\DocumentMetadata{
    lang = en-US,
    pdfversion = 2.0,
    pdfstandard = ua-2,
    testphase = latest
}

\documentclass[11pt]{report}

\usepackage{amsmath}
\usepackage{amssymb}
\usepackage{mathtools}

\usepackage[utf8]{inputenc}
\usepackage[english]{babel}
\usepackage{hyperref}
\hypersetup{
    hidelinks,
    pdfauthor={Kia Karbasi},
    pdftitle={On Applications of Natural Language Processing and Artificial Intelligence in Intelligent Tutoring Systems},
    pdfsubject={Doctoral dissertation},
    pdfkeywords={intelligent tutoring systems, natural language processing, artificial intelligence, knowledge tracing},
    pdflang={en-US},
    pdfdisplaydoctitle=true,
}
 
\DeclarePairedDelimiter\abs{\lvert}{\rvert}%
\DeclarePairedDelimiter\norm{\lVert}{\rVert}%

\makeatletter
\let\oldabs\abs
\def\abs{\@ifstar{\oldabs}{\oldabs*}}
\let\oldnorm\norm
\def\norm{\@ifstar{\oldnorm}{\oldnorm*}}
\makeatother

\usepackage[letterpaper,
    left=1in,
    right=1in,
    top=1in,
    bottom=1in]{geometry}

\usepackage{listings}
\usepackage{color}

\definecolor{dkgreen}{rgb}{0,0.6,0}
\definecolor{gray}{rgb}{0.5,0.5,0.5}
\definecolor{mauve}{rgb}{0.58,0,0.82}
\definecolor{promptframe}{RGB}{220,220,220}

\usepackage{array, booktabs}
\usepackage{graphicx}
\usepackage[section]{placeins}
\usepackage[x11names,table]{xcolor}
\usepackage{caption}
\usepackage{subcaption}
\DeclareCaptionFont{blue}{\color{LightSteelBlue3}}
\providecommand{\Description}[1]{}
\newcommand{\DataTableHeaderRows}[1]{\tagpdfsetup{table/tagging=true,table/header-rows={#1},table/header-columns={}}}
\newcommand{\LayoutTableTagging}{\tagpdfsetup{table/tagging=presentation}}

\usepackage{tikz}
	
\tikzstyle{state}=[shape=circle,draw=blue!50,fill=blue!20]
\tikzstyle{observation}=[shape=rectangle,draw=orange!50,fill=orange!20]
\tikzstyle{lightedge}=[<-,dotted]
\tikzstyle{mainstate}=[state,thick]
\tikzstyle{mainedge}=[<-,thick]

\makeatletter
\newcounter{subsubsubsection}[subsection]

\newcommand\subsubsubsection{\@startsection{subsubsubsection}{4}{\z@}
  {-3.25ex\@plus -1ex \@minus -.2ex}
  {1.5ex \@plus .2ex}
  {\normalfont\normalsize\bfseries}}
\newcommand\l@subsubsubsection{\@dottedtocline{4}{10em}{4.1em}}
\makeatother
\usepackage{setspace}
\graphicspath{ {./images/} }

\newcommand{\thesistitle}{On Applications of Natural Language Processing and Artificial Intelligence in Intelligent Tutoring Systems}
\newcommand{\thesisauthor}{Kia Karbasi}
\newcommand{\thesisdegree}{Doctor of Philosophy in Electrical and Computer Engineering}
\newcommand{\thesisyear}{2026}

\title{\thesistitle}
\author{\thesisauthor}
\date{\thesisyear}

\renewcommand{\contentsname}{TABLE OF CONTENTS}
\renewcommand{\listfigurename}{LIST OF FIGURES}
\renewcommand{\listtablename}{LIST OF TABLES}
\renewcommand{\bibname}{References}

\begin{document}

\renewcommand{\contentsname}{TABLE OF CONTENTS}
\renewcommand{\listfigurename}{LIST OF FIGURES}
\renewcommand{\listtablename}{LIST OF TABLES}

\setlength{\parskip}{0pt}

\begin{titlepage}
\begin{center}
\begin{singlespace}
\vspace*{0.55in}
UNIVERSITY OF CALIFORNIA\\[1em]
Los Angeles

\vspace{1.55in}
\thesistitle

\vspace{1.55in}
A dissertation submitted in partial satisfaction\\
of the requirements for the degree\\
\thesisdegree

\vspace{0.55in}
by

\vspace{0.55in}
\thesisauthor

\vfill
\thesisyear
\end{singlespace}
\end{center}
\end{titlepage}

\clearpage
\thispagestyle{empty}
\begin{center}
\begin{doublespace}
\vspace*{\fill}
\copyright{} Copyright by\\
\thesisauthor\\
\thesisyear
\end{doublespace}
\end{center}
\clearpage

\pagenumbering{roman}
\setcounter{page}{2}

\thispagestyle{plain}
\begin{center}
\begin{singlespace}
ABSTRACT OF THE DISSERTATION

\vspace{1.5em}
\thesistitle

\vspace{1.5em}
by

\vspace{1.5em}
\thesisauthor

\vspace{1.5em}
\thesisdegree\\
University of California, Los Angeles, \thesisyear\\
Professor Gregory J. Pottie, Chair
\end{singlespace}
\end{center}

\vspace{1.5em}

Intelligent tutoring systems adapt instruction by estimating what a student understands, selecting what the student should encounter next, and delivering interventions that support learning. Natural language processing and artificial intelligence offer new ways to improve this loop because tutoring systems are built around linguistic and educational artifacts: questions, hints, explanations, standards, knowledge components, and student histories. This dissertation studies where those techniques help, where they fail, and how they should be connected to specific tutoring functions.

The dissertation examines five research directions across intervention generation and student modeling. First, it shows that structured multi-agent collaboration can improve the pedagogical quality of automatically generated mathematics questions. Second, it analyzes Deep Knowledge Tracing and argues that the relations learned by the model are better interpreted as prerequisite-like dependencies than as generic exercise associations. Third, it tests whether pretrained semantic representations can improve knowledge tracing by initializing knowledge-component inputs, finding that semantic similarity alone does not reliably improve performance. Fourth, it shows that text embeddings can be more useful when they reorganize the knowledge-component space itself, improving downstream knowledge tracing under the right data conditions. Finally, it introduces ReasonKT as a preliminary framework for reasoning-based knowledge tracing and evaluates a prompt-only instantiation that produces above-chance ranking signal and inspectable reasoning traces.

Across these studies, the dissertation argues that NLP and AI techniques are useful in intelligent tutoring systems when their representational, generative, or reasoning capabilities are tied to a concrete educational role. Language is not uniformly beneficial as an added feature; it becomes useful when it helps model student knowledge, interpret educational structure, generate instructional interventions, or connect these components into a more coherent adaptive system.

\clearpage

\thispagestyle{plain}
\begin{center}
\begin{singlespace}
\vspace*{1.0in}
The dissertation of \thesisauthor{} is approved.

\vspace{1.0in}
Jonathan Kao\\[1.5em]
Lieven Vandenberghe\\[1.5em]
Xiang Chen\\[1.5em]
Gregory J. Pottie, Committee Chair

\vfill
University of California, Los Angeles\\
\thesisyear
\end{singlespace}
\end{center}
\clearpage

\thispagestyle{plain}
\vspace*{\fill}
\begin{center}
\itshape
To my mother, Fereshteh,\\
my father, Bahram,\\
and my brother, Kaveh,\\
for their love, support, and encouragement throughout this journey.
\end{center}
\vspace*{\fill}
\clearpage

\tableofcontents

\clearpage

\listoffigures

\clearpage

\listoftables

\clearpage

\chapter*{ACKNOWLEDGMENTS}

I would first like to thank my advisor, Professor Gregory Pottie, for his guidance, support, and mentorship throughout my graduate studies. His advice helped shape both this dissertation and my development as a researcher.

I am also grateful to my dissertation committee members, Professor Jonathan Kao, Professor Lieven Vandenberghe, and Professor Xiang Anthony Chen, for their time, thoughtful feedback, and encouragement.

I would like to thank my research collaborators, Kevin Hong, Kaveh Karbasi, Mohammad Amin Samadi, and Mahdi Saiedi, for their contributions, discussions, and support throughout the projects that contributed to this dissertation. I am also thankful to my labmates, Yunzheng Zhu, Omead Pooladzandi, and Sunay Bhat, for their friendship, feedback, and many helpful conversations.

Portions of Chapter~3 are based on the following coauthored work: Kia Karbasi, Kevin Hong, Mohammad Amin Samadi, and Gregory J. Pottie. ``Multi-Agent Collaborative Framework For Math Problem Generation.'' In \textit{Proceedings of the 18th International Conference on Educational Data Mining}, pp. 613--618, 2025. I created the agents and workflows, ran all experiments, and performed most of the analysis. Kevin Hong helped set up AutoGen and performed initial testing. Mohammad Amin Samadi helped with analysis, paper formatting, and plots. Gregory J. Pottie served as advisor and principal investigator.

Portions of Chapter~4 are based on the following coauthored work: Kevin Hong, Kia Karbasi, and Gregory J. Pottie. ``Extracting Causal Relations in Deep Knowledge Tracing.'' In \textit{Proceedings of the 18th International Conference on Educational Data Mining}, pp. 586--591, 2025. I conceived the project, developed the code, directed the experimental work, and performed the primary analysis. Kevin Hong helped conduct the experiments and helped write the original paper. Gregory J. Pottie served as advisor and principal investigator.

This work was supported throughout my graduate studies by the UCLA Graduate Division, the Network Systems Department at The Aerospace Corporation, and the Office of Naval Research (ONR). I also acknowledge Google Education for providing a Google Cloud research grant.

I am grateful to the graduate staff at UCLA, especially Deeona Columbia and Tatyana Vigil, as well as the department administrators and staff, for their help and support. I also thank the department chair at Worcester Polytechnic Institute and Professor Neil Heffernan's lab for providing access to data that supported this research.

AI-based writing tools were used to assist with copyediting and language refinement; all research, analysis, interpretation, and conclusions are the author's own.

Finally, I thank my family for their love, patience, and encouragement: my mother, Fereshteh Rafieimoghadam; my father, Bahram Karbasi; and my brother, Kaveh Karbasi. Their support has been a constant foundation throughout this journey.

\clearpage

\chapter*{VITA}

\begin{singlespace}
\noindent
\textbf{Education}

\vspace{0.75em}

\noindent
{\LayoutTableTagging
\begin{tabular}{>{\raggedright\arraybackslash}p{1.05in}>{\raggedright\arraybackslash}p{5.25in}}
2014 & B.S., Electrical Engineering, University of Maryland, College Park.\\[0.75em]
2018 & M.S., Electrical and Computer Engineering, University of California, Los Angeles.\\
\end{tabular}
}

\vspace{1.5em}

\noindent\textbf{Professional Experience}

\vspace{0.75em}

\noindent
{\LayoutTableTagging
\begin{tabular}{>{\raggedright\arraybackslash}p{1.05in}>{\raggedright\arraybackslash}p{5.25in}}
2014--2015 & Software Engineer, Visa Inc.\\[0.75em]
2019--2026 & Senior Member of Technical Staff, The Aerospace Corporation.\\
\end{tabular}
}

\vspace{1.5em}

\noindent\textbf{Publications}

\begin{itemize}
    \item Kia Karbasi, Kevin Hong, Mohammad Amin Samadi, and Gregory J. Pottie. ``Multi-Agent Collaborative Framework For Math Problem Generation.'' In \textit{Proceedings of the 18th International Conference on Educational Data Mining}, pp. 613--618, 2025.
    \item Kevin Hong, Kia Karbasi, and Gregory J. Pottie. ``Extracting Causal Relations in Deep Knowledge Tracing.'' In \textit{Proceedings of the 18th International Conference on Educational Data Mining}, pp. 586--591, 2025.
    \item Ehsan Keramat, Nicholas Kauffroath, Kia Karbasi, Heng Zhao, Babak Daneshrad, and Gregory J. Pottie. ``Experimental results for low overhead frequency offset estimation in MANETs with concurrent transmission.'' In \textit{Proceedings of the IEEE Military Communications Conference}, pp. 69--72, 2017.
\end{itemize}
\end{singlespace}

\clearpage
\pagenumbering{arabic}

\chapter{Introduction}

An intelligent tutoring system (ITS) is an adaptive educational system that manages a student's learning trajectory by coordinating three core functions. It estimates what the student currently understands, selects what the student should encounter next, and delivers an instructional intervention that can improve future learning~\cite{fengSystematicReviewLiterature2021, kulikEffectivenessIntelligentTutoring2016}. These functions are linked: a generated question becomes evidence for the next student-model update, and a student model is useful only insofar as it supports better instructional action.

This dissertation studies how natural language processing and AI techniques can improve different parts of that adaptive loop: student modeling and intervention generation. The motivation is not simply that tutoring systems contain text. Questions, explanations, hints, standards labels, and knowledge-component descriptions are linguistic artifacts, but their value depends on how they enter the tutoring system. If they are reduced too early to anonymous identifiers, the system may lose information about conceptual similarity, prerequisite structure, difficulty, and pedagogical intent. If they are used too naively, however, language representations can add surface similarity without improving the underlying tutoring decision.

The central question is therefore not whether language is useful in general, but where and how it changes the tutoring problem. A language representation might help prediction by capturing similarity across questions. It might help interpretation by making learned exercise relations easier to analyze. It might help intervention by supporting the generation of new practice problems rather than only selecting from existing ones. It might also fail when the dominant signal is behavioral, temporal, psychometric, or curriculum-structural rather than semantic. This dissertation tests those possibilities across student modeling and intervention generation instead of treating ``adding language'' as a uniform modeling improvement.

On the student-modeling side, the thesis studies knowledge tracing (KT), the task of estimating the probability that a student will answer the next item correctly from prior interactions~\cite{corbett_knowledge_1995, piech_deep_2015}. KT has moved from interpretable probabilistic formulations to powerful neural sequence models, but the sources of those models' gains are often hard to localize: performance may reflect temporal signal, concept structure, question identity, prerequisite relations, or some combination of all four~\cite{pandey_self-attentive_2019, zhang_dynamic_2017, sonkar_qdkt_2020}. On the intervention side, the thesis treats automatic question generation as an ITS capability rather than a generic text-generation task, because a generated question is the instructional action that the system uses to collect the next piece of evidence~\cite{heilman_good_2010, zhang_review_2022}.

The research chapters follow this adaptive loop. Chapter~3 addresses intervention generation and shows that structured multi-agent collaboration can improve the pedagogical quality of automatically generated mathematics questions. Chapter~4 returns to student modeling and argues that the exercise-relation structure learned by DKT is better understood as prerequisite-like dependency than as generic association, making the learned structure more educationally interpretable.

Chapters~5 and~6 then test two different ways of bringing language into KT. Chapter~5 shows that, in the evaluated DKT setting, the tested semantic initializations do not reliably capture the structure needed by the student model: inserting pretrained semantic representations into the inherited knowledge-component space does not reliably improve DKT. Chapter~6 changes the representational target: instead of using text only to initialize model inputs, it uses text embeddings to reorganize the knowledge-component space itself. Under the right data conditions, that reorganization improves downstream DKT. This contrast gives the dissertation its clearest KT finding: language is more useful when it reshapes the educational structure over which the model learns than when it is added only as a surface semantic prior.

Chapter~7 extends this line of work from structural language use to explicit reasoning. It introduces \emph{ReasonKT} as a framework for reasoning over student history, question content, and concept organization, while evaluating only a prompt-only preliminary instantiation of that broader framework. The prompt-only study is preliminary, but it shows above-chance ranking signal and inspectable reasoning traces in a KT-style next-step prediction setting. Chapter~8 synthesizes these results and uses them to motivate future work on reasoning-based student modeling and more unified ITS architectures.

Taken together, the dissertation argues that NLP and AI techniques are useful in tutoring systems when their representational or generative strengths are tied to a specific tutoring function: modeling student knowledge, interpreting exercise structure, generating interventions, or connecting those pieces into a more coherent adaptive system.

\section{Main Contributions}

Across its research chapters, this dissertation makes five main contributions:
\begin{itemize}
    \item We show that collaborative multi-agent workflows improve the pedagogical quality of automatically generated mathematics questions, especially on relevance, answerability, and difficulty matching.
    \item We demonstrate that the dependency structure learned by Deep Knowledge Tracing is better understood in causal or prerequisite-like terms than as generic exercise association, and we introduce a more stable method for extracting those relations.
    \item We show that, in the evaluated DKT setting, multiple text-informed initialization pipelines do not consistently improve over standard sparse baselines, revealing a gap between surface semantic representations and the structure used by the student model.
    \item We demonstrate that text embeddings can improve downstream knowledge tracing when they are used to redefine the knowledge-component space, with positive results on ASSISTments 2012 and ASSISTments 2017.
    \item We present preliminary evidence for reasoning-based student modeling and use that evidence to frame a broader ITS research agenda connecting knowledge tracing, intervention selection, and intervention generation.
\end{itemize}

The rest of this dissertation is structured as follows. Chapter~2 provides the shared technical background; Chapters~3--7 present the five research studies; Chapter~8 synthesizes the dissertation-level findings and outlines future directions; and the appendices collect supplementary chapter-specific results and diagnostics.

\chapter{Background}

In this chapter, we review neural representations, sequence models, transformers and prompting, structure-discovery methods, reinforcement learning, and evaluation concepts.

\section{Neural Representations and Embeddings}

\subsection{Sparse and Dense Representations}

Many learning problems begin with a representational question. Before a model can estimate, compare, or generate anything, the objects of interest must be encoded numerically. A standard approach is to use a sparse symbolic representation such as a one-hot vector, in which each category is assigned its own coordinate and exactly one coordinate is active. This encoding is easy to specify and preserves identity perfectly, but it treats all categories as equally unrelated. Two tokens that are near-synonyms and two tokens that share no semantic relation at all are separated in essentially the same way.

\begin{figure}[!htbp]
    \centering
    \includegraphics[width=0.72\linewidth]{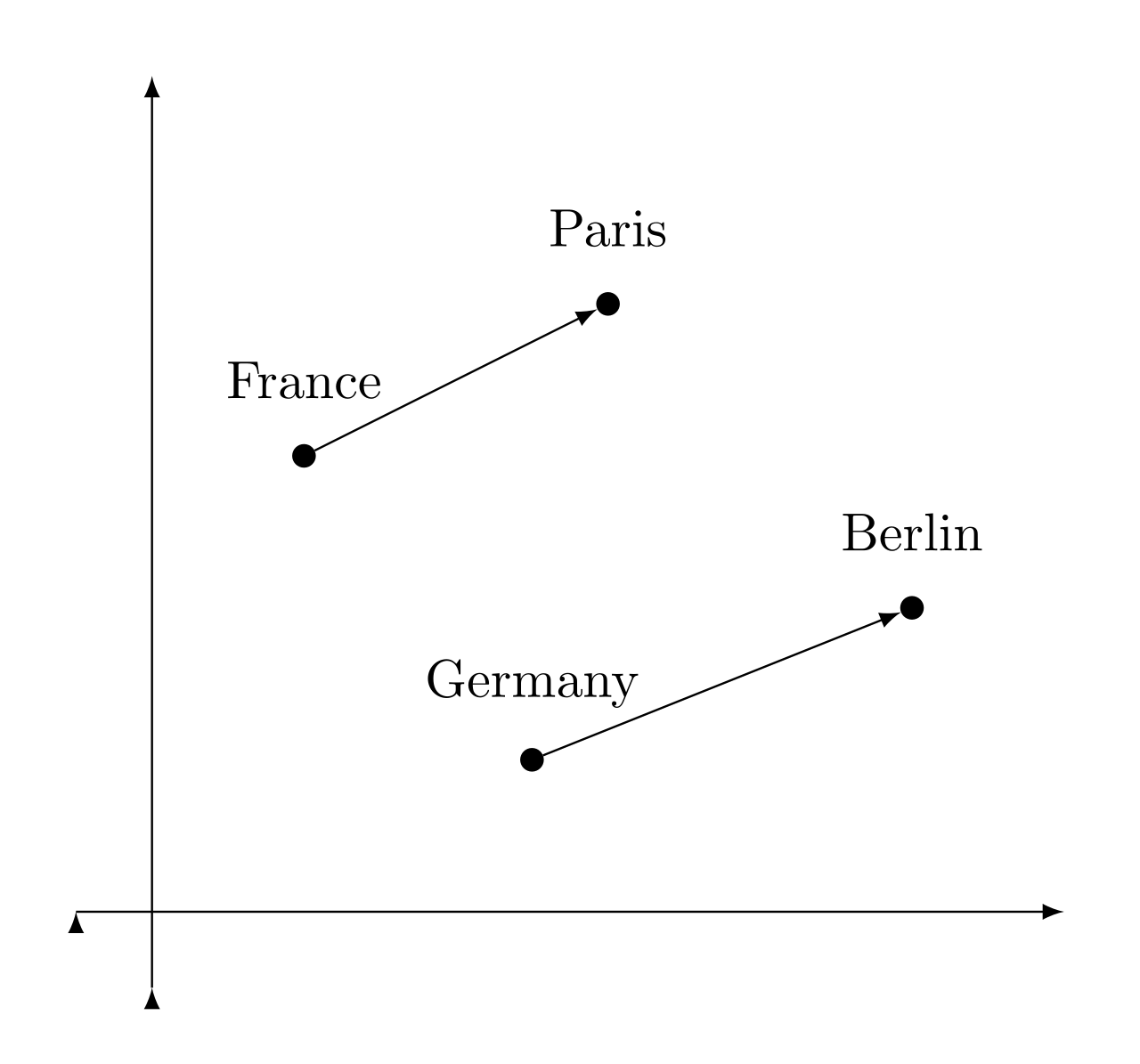}
    \caption{A schematic embedding space. Related words occupy nearby regions rather than being represented as isolated one-hot symbols.}
    \label{fig:background_embeddings}
    \Description{Two-dimensional illustration of an embedding space in which semantically related words are closer together.}
    \caption*{\footnotesize Source note: Wikimedia Commons, File \texttt{Word\_embedding\_illustration.svg}, CC BY-SA 4.0, \url{https://commons.wikimedia.org/wiki/File:Word_embedding_illustration.svg}.}
\end{figure}

\emph{Embeddings} address this limitation by mapping discrete objects into dense vectors in a continuous space~\cite{mikolov_distributed_2013, pennington_glove_2014, bojanowski_enriching_2017}. In such a space, objects that behave similarly in data can be assigned nearby representations, while dissimilar objects can be placed farther apart. The value of an embedding is therefore not that it provides a human-interpretable coordinate system in which each dimension has a simple semantic meaning. Rather, it provides a learned geometry in which proximity, direction, and relative arrangement encode regularities that are useful for prediction. Figure~\ref{fig:background_embeddings} illustrates this geometric intuition at a schematic level.

The geometry in Figure~\ref{fig:background_embeddings} should not be read too literally. Real embedding spaces are high-dimensional, and their usefulness arises from distributed structure across many coordinates at once. Nevertheless, the figure captures the central idea: a learned representation can support transfer because information acquired about one object can generalize to nearby objects that occupy similar regions of the space.

\subsection{Similarity and Context}

In language applications, embeddings can be learned at several levels of granularity. Earlier approaches typically assigned one vector to each word type, regardless of context~\cite{mikolov_distributed_2013, pennington_glove_2014}. More recent models instead produce \emph{contextual} representations, meaning that the representation of a token depends on the surrounding sequence~\cite{devlin_bert_2019}. This distinction is consequential because linguistic meaning is often context-sensitive. The token ``bank'' in ``river bank'' and the same token in ``bank account'' should not, in general, be represented identically if the model is expected to reason effectively about meaning.

Representations can also be defined above the token level. Sentence encoders and related models map entire sentences or short passages into single vectors intended to summarize their semantic content at a coarser scale~\cite{reimers_sentence-bert_2019, gao_simcse_2021}. Such representations are particularly appropriate when the object of interest is not an isolated word but a larger unit such as a question, explanation, or paragraph.

A common way to compare two embeddings is cosine similarity:
\[
\cos(u,v) = \frac{u^\top v}{\|u\| \, \|v\|}.
\]
Cosine similarity measures the angle between two vectors rather than their raw Euclidean separation. Two vectors can therefore be judged similar if they point in nearly the same direction even when their magnitudes differ. In many text-representation settings, that directional information is more informative than absolute scale because it captures relational structure in the representation space.

\section{Sequence Models for Time-Ordered Data}

\subsection{Recurrent Neural Networks}

Many machine-learning problems concern ordered observations rather than independent examples. Natural language is sequential, sensor streams are sequential, and longitudinal student histories are sequential. In such settings, the model must update its internal representation as new observations arrive while retaining relevant information from prior steps. A \emph{recurrent neural network} (RNN) addresses this requirement by maintaining a hidden state that is updated recursively over time. At step \(t\), the model combines the current input with the hidden state from step \(t-1\) to produce a new hidden state that summarizes the sequence prefix observed so far.

\begin{figure}[!htbp]
    \centering
    \begin{subfigure}{0.48\linewidth}
        \centering
        \includegraphics[width=\linewidth]{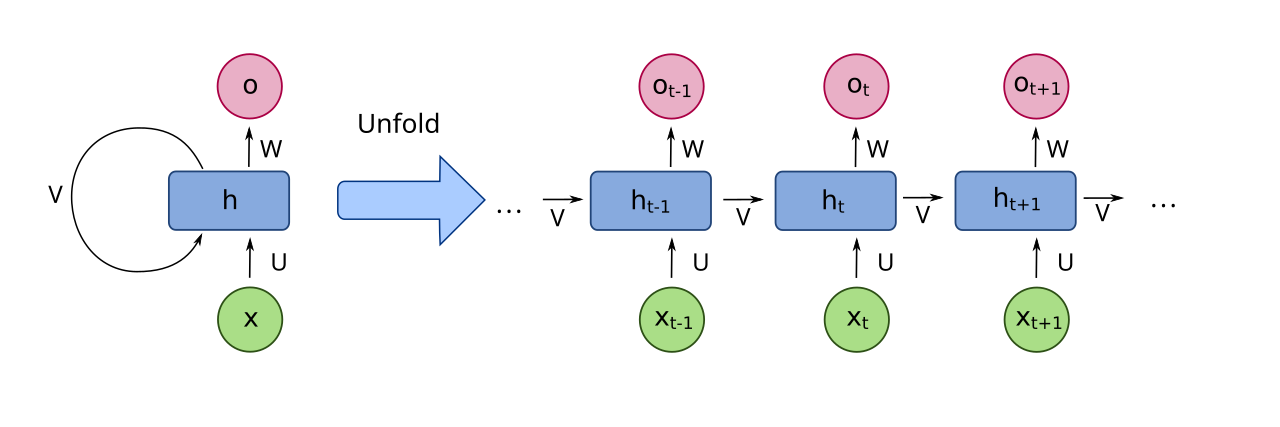}
        \caption{An unfolded RNN.}
        \Description{A recurrent neural network shown in compact and unfolded forms across time steps.}
    \end{subfigure}\hfill
    \begin{subfigure}{0.48\linewidth}
        \centering
        \includegraphics[width=\linewidth]{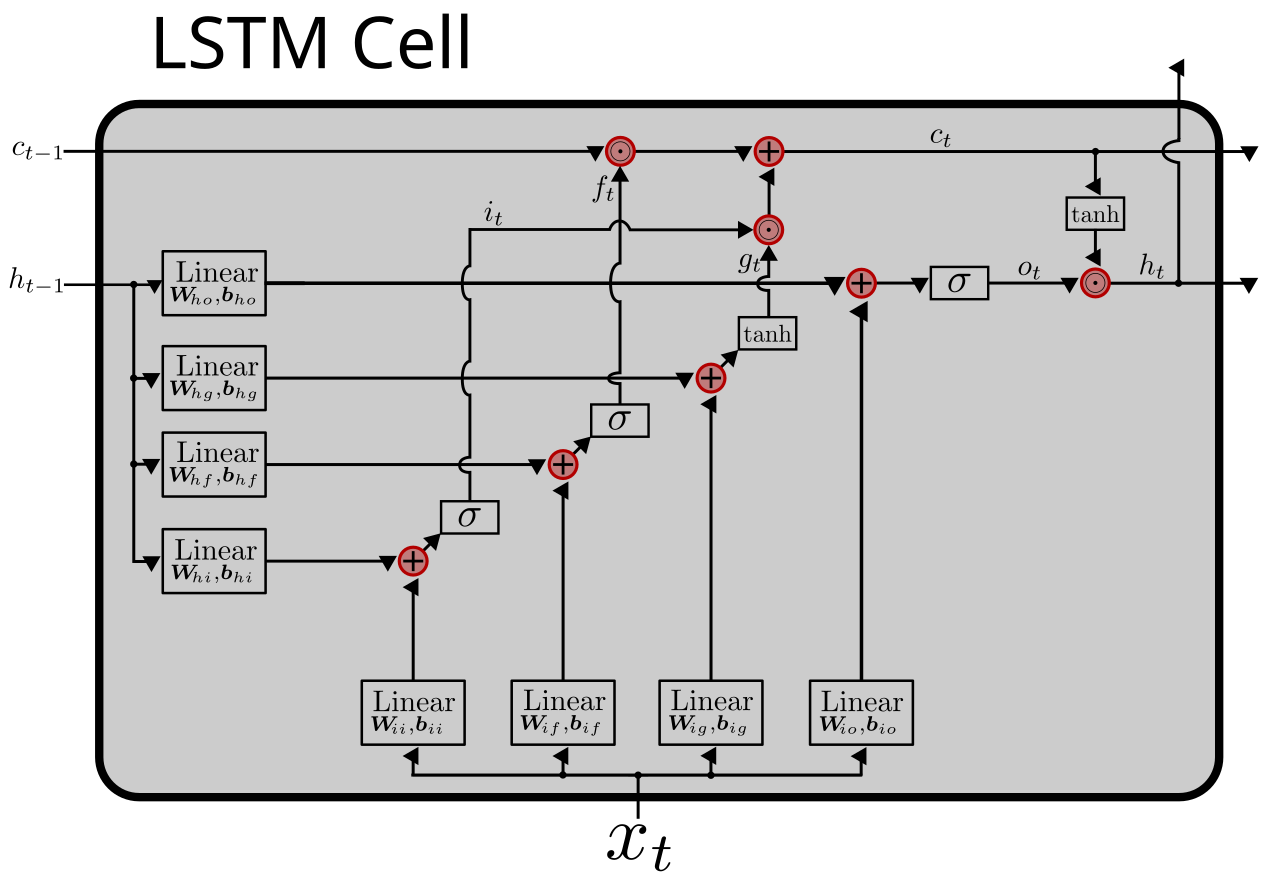}
        \caption{An LSTM cell.}
        \Description{A long short-term memory cell showing gates and memory flow.}
    \end{subfigure}
    \caption{Two views of sequence modeling. The left panel shows recurrence as repeated application of a shared transition, and the right panel shows how an LSTM regulates memory flow through learned gates.}
    \label{fig:background_sequence_models}
    \caption*{\footnotesize Source notes: left, Wikimedia Commons, File \texttt{Recurrent\_neural\_network\_unfold.svg}, CC BY-SA 4.0, \url{https://commons.wikimedia.org/wiki/File:Recurrent_neural_network_unfold.svg}; right, Wikimedia Commons, File \texttt{NN\_LSTM-Cell.svg}, CC0, \url{https://commons.wikimedia.org/wiki/File:NN_LSTM-Cell.svg}.}
\end{figure}

An RNN can be drawn either as a compact recurrent block or as an unfolded chain in which the same computation is repeated across time steps. The unfolded view is especially useful because it makes explicit that the parameters are shared across positions in the sequence. Figure~\ref{fig:background_sequence_models} uses this perspective to contrast a plain recurrent architecture with the more structured LSTM update.

The principal limitation of plain RNNs is not expressive inadequacy in principle, but difficulty in optimization. When training requires information from distant time steps to influence present predictions, gradients must propagate backward through many recurrent updates. These gradients may decay toward zero or grow uncontrollably, making long-range dependencies difficult to learn reliably. In practical terms, the model either loses earlier information too quickly or becomes numerically unstable during training.

\subsection{Long Short-Term Memory Networks}

\emph{Long short-term memory} (LSTM) networks were introduced to mitigate this problem. An LSTM augments the recurrent update with a memory cell and a collection of learned gates that regulate what information is forgotten, what new information is written, and what portion of the internal state is exposed at the output. As Figure~\ref{fig:background_sequence_models} shows, this architecture imposes more structure on the update than a plain RNN and thereby gives the model greater control over information flow across long sequences.

The practical significance of LSTMs is not that they preserve all prior information indefinitely. Rather, they make it easier for the model to preserve \emph{relevant} information over longer spans when the task requires it. For that reason, LSTMs became a standard sequence-modeling architecture in many domains prior to the rise of transformer-based language models. The same principle matters whenever data arrive as temporally ordered interactions: the model must represent not only the current observation but also the evolving state induced by the preceding sequence.

\section{Transformers, Large Language Models, and Generative Tasks}

\subsection{Tokens, Self-Attention, and Transformers}

Modern language models do not process raw text as uninterpreted character streams. They first convert text into a sequence of discrete \emph{tokens}, which may correspond to whole words, subword fragments, punctuation marks, or special symbols. Tokenization is therefore a design choice that mediates between linguistic structure and computational efficiency. It defines the units over which the model learns statistical regularities.

The transformer architecture fundamentally changed language modeling by replacing recurrence with \emph{self-attention}~\cite{vaswani_attention_2017}. In self-attention, each position in a sequence computes weighted interactions with other positions, allowing the model to represent dependencies without compressing the entire past into a single recurrent state. This design has two important consequences. First, it permits more direct modeling of long-range relationships. Second, it allows highly parallel computation during training because many token-to-token interactions can be evaluated simultaneously.

\begin{figure}[!htbp]
    \centering
    \includegraphics[width=0.65\linewidth]{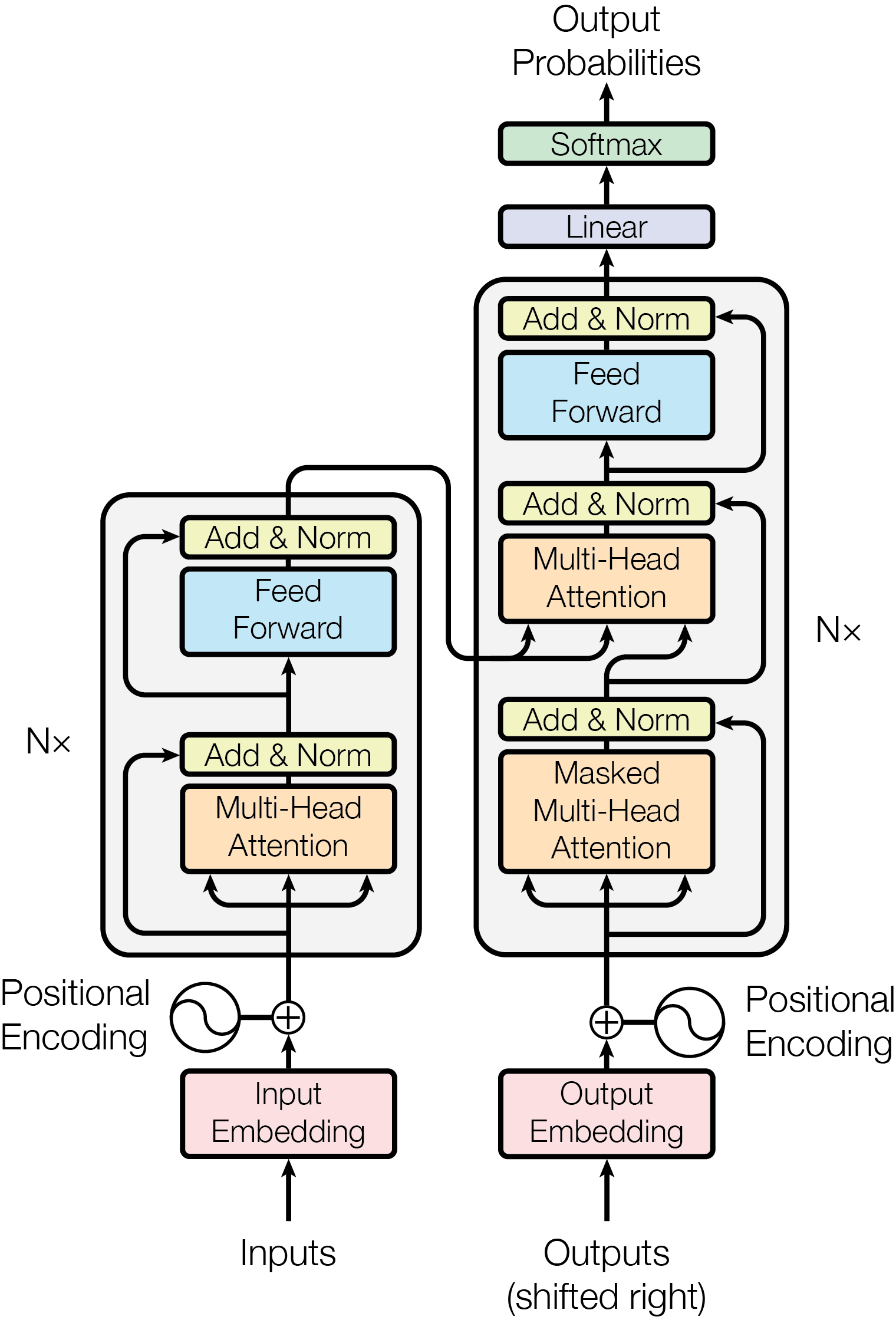}
    \caption{A standard transformer encoder--decoder architecture. Repeated attention and feedforward blocks replace recurrent state transitions as the primary computational mechanism.}
    \label{fig:background_transformer}
    \Description{Diagram of a transformer encoder-decoder architecture with stacked attention and feedforward blocks.}
    \caption*{\footnotesize Source note: Wikimedia Commons, File \texttt{Attention\_Is\_All\_You\_Need\_-\_Encoder-decoder\_Architecture.png}, CC BY-SA 4.0, \url{https://commons.wikimedia.org/wiki/File:Attention_Is_All_You_Need_-_Encoder-decoder_Architecture.png}.}
\end{figure}

Figure~\ref{fig:background_transformer} shows the canonical encoder--decoder transformer architecture. Although many modern language models adopt decoder-only or otherwise modified variants, the figure captures the central idea: language understanding and generation are driven by repeated attention-based transformations of token representations rather than by a single recurrent state passed forward step by step.

Each transformer layer alternates between attention and feedforward computation, progressively refining token representations in light of the full context. Stacking many such layers yields a deep model capable of representing rich syntactic, semantic, and discourse-level structure. This attention-based design now underlies most contemporary large language models.

\subsection{Large Language Models and Generative Tasks}

A \emph{large language model} (LLM) is typically a transformer-based model trained on very large text corpora using a generative objective, often followed by additional post-training stages that improve instruction following, dialogue quality, or other downstream behavior~\cite{brown_language_2020, ouyang_instructgpt_2022, guo_deepseekr1_2025}. The importance of scale is not simply that larger models contain more parameters. Large-scale pretraining allows a single model to acquire broad statistical regularities about language that can then be adapted across a wide range of tasks.

The standard generative objective is next-token prediction: given a prefix of tokens, estimate the conditional probability distribution over the next token. Despite its apparent simplicity, this objective is remarkably general. If the prefix is the beginning of a paragraph, the model performs text continuation. If the prefix is an instruction, it attempts to generate a compliant response. If the prefix frames a question, the continuation may take the form of an answer. Many tasks that appear distinct at the application level can therefore be cast as conditional generation under a common predictive mechanism.

This unification is why a single LLM can be used for text continuation, summarization, question answering, question generation, dialogue, translation, and code generation without changing the underlying architecture. The task changes because the conditioning context changes, not because the core predictive machinery does. That flexibility matters here because the same generative formalism can support question generation, structured prompting, and other language-model uses without changing the underlying predictive mechanism.

\section{Decoding, Sampling, and Prompting}

\subsection{Decoding and Temperature}

Once a language model has assigned probabilities to candidate next tokens, a second question arises: how should the system convert that distribution into an actual output sequence? This stage is known as \emph{decoding}. The simplest strategy is \emph{greedy decoding}, which selects the single most probable token at each step. Greedy decoding is deterministic and often produces stable outputs, but it can also be overly conservative and may amplify repetition because it never explores lower-probability alternatives.

Sampling-based decoding instead draws from the predicted distribution. This introduces stochasticity and can increase diversity, but it also exposes the model to the risk of low-quality continuations if the sampling procedure is too permissive. The \emph{temperature} parameter controls this trade-off by sharpening or flattening the distribution before sampling. Low temperatures concentrate probability mass on the most likely tokens, producing safer and more predictable outputs. High temperatures distribute probability more broadly, increasing diversity at the cost of greater variance and a higher risk of incoherence. Temperature therefore does not change what the model knows; it changes how aggressively the decoding procedure explores the distribution already induced by the model.

Two further controls are commonly used in practice. \emph{Top-\(k\)} sampling restricts selection to the \(k\) most probable next tokens, while \emph{top-\(p\)} or nucleus sampling restricts selection to the smallest token set whose cumulative probability exceeds a chosen threshold. Both mechanisms attempt to preserve diversity without sampling too heavily from the extreme tail of the distribution.

\subsection{Prompting}

Prompting refers to the use of input text to specify task structure, desired behavior, or reasoning format at inference time. In \emph{zero-shot} prompting, the model receives only a task description or instruction. In \emph{few-shot} prompting, the prompt additionally includes a small number of exemplars that demonstrate the intended input--output relation~\cite{kojima_zero-shot_2022}. In either case, the model parameters remain fixed; behavior changes because the prompt alters the context in which generation takes place.

\emph{Chain-of-thought} prompting goes further by encouraging the model to produce intermediate reasoning steps before the final answer~\cite{wei_cot_2022, wang_self-consistency_2023, zhou_least-to-most_2023}. This can improve performance on tasks that benefit from decomposition or explicit intermediate structure. At the same time, such generated rationales should not automatically be treated as transparent windows into the model's internal computation. They are outputs of the model, not guaranteed faithful explanations of its hidden-state dynamics. Table~\ref{tab:background_decoding_prompting} summarizes the main decoding and prompting choices.

\begin{table}[!htbp]
\centering
\small
\DataTableHeaderRows{1}
\begin{tabular}{p{0.21\linewidth} p{0.2\linewidth} p{0.25\linewidth} p{0.22\linewidth}}
\toprule
Method & Category & Main effect & Typical trade-off \\
\midrule
Greedy decoding & Decoding & Chooses the most probable next token at each step & Stable, but can be repetitive or conservative \\
Sampling & Decoding & Draws from the next-token distribution & More diverse, but less predictable \\
Temperature & Decoding control & Sharpens or flattens the distribution before sampling & Lower values increase reliability; higher values increase variety \\
Top-\(k\) / top-\(p\) & Decoding control & Restricts sampling to a plausible candidate set & Reduces low-probability errors while preserving diversity \\
Zero-shot prompting & Prompting & Relies only on an instruction & Fast, but sensitive to wording \\
Few-shot prompting & Prompting & Uses examples in the prompt & Better task steering, but consumes context length \\
Chain-of-thought & Prompting & Encourages intermediate reasoning steps & Often helps difficult tasks, but increases cost and does not guarantee faithfulness \\
\bottomrule
\end{tabular}
\caption{Common decoding and prompting choices for language models.}
\label{tab:background_decoding_prompting}
\end{table}

\section{Graphs, Clustering, and Structure Discovery}

\subsection{Graphs and Directed Acyclic Graphs}

A \emph{graph} represents entities as nodes and relations among them as edges. If the edges have directions, the graph is \emph{directed}; if they also carry numerical strengths, it is a weighted directed graph. This framework is appropriate when the central object of study is relational structure rather than isolated instances. In educational and machine-learning settings, graphs are often used to represent similarity, dependency, information flow, or possible prerequisite relations among concepts.

A \emph{cycle} occurs when a directed path eventually returns to its starting node. A \emph{directed acyclic graph} (DAG) is a directed graph with no such cycles. DAGs support a partial order: some nodes can precede others without requiring every pair of nodes to be directly comparable. When a directed graph is interpreted as a prerequisite structure, acyclicity is required for a clean ordering interpretation. A cycle between concepts \(A\) and \(B\) may still indicate association, but it is difficult to interpret as a clean ordering in which one concept should be mastered before the other.

\subsection{Dimensionality Reduction and Clustering}

\begin{figure}[!htbp]
    \centering
    \includegraphics[width=0.62\linewidth]{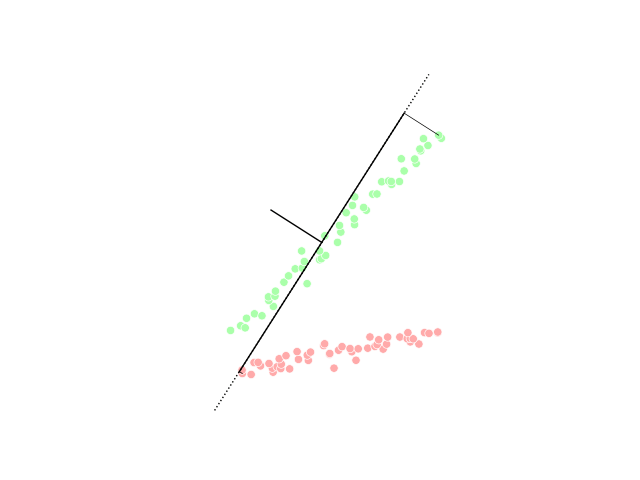}
    \caption{A PCA projection illustration. High-dimensional variation is approximated by projection onto a smaller number of principal directions.}
    \label{fig:background_pca}
    \Description{An illustration showing data points projected onto a principal component axis.}
    \caption*{\footnotesize Source note: Wikimedia Commons, File \texttt{PCA\_Projection\_Illustration.gif}, CC BY-SA 4.0, \url{https://commons.wikimedia.org/wiki/File:PCA_Projection_Illustration.gif}. Static frame reproduced from the original animation.}
\end{figure}

When observations are represented as high-dimensional vectors, dimensionality reduction is often applied before visualization or structural analysis. \emph{Principal component analysis} (PCA) finds new orthogonal axes that capture variance in descending order~\cite{jolliffePrincipalComponentAnalysis2002}. Retaining only the leading principal components yields a lower-dimensional approximation that preserves as much variation as possible subject to the chosen dimension. Figure~\ref{fig:background_pca} illustrates this projection idea geometrically.

PCA does not itself discover semantic groups; it is a linear compression method. \emph{Clustering}, by contrast, asks whether points in a representation space can be partitioned into internally similar groups. A standard example is \(k\)-means clustering~\cite{macqueenSomeMethodsClassification1967, arthurKmeansPlusPlus2007}, which alternates between assigning each point to its nearest centroid and updating the centroids to match the current assignments. The procedure is conceptually simple, but the choice of the number of clusters \(k\) is substantive rather than merely technical. Small values of \(k\) impose coarse structure; larger values permit finer distinctions. Cluster granularity is therefore a modeling assumption about how much structure the analyst wishes to resolve.

\section{Reinforcement Learning and GRPO}

\subsection{Reinforcement Learning Basics}

In \emph{reinforcement learning} (RL), an agent interacts with an environment by taking actions and receiving evaluative feedback in the form of rewards~\cite{sutton_barto_2018}. The agent's behavior is specified by a \emph{policy}, often written as \(\pi_\theta(a \mid s)\), which denotes the probability of taking action \(a\) in state \(s\) under parameters \(\theta\). Unlike supervised learning, RL does not typically provide a target output for every step. Instead, the agent must infer which behaviors are desirable from the rewards generated by its interactions.

The usual objective is to maximize expected cumulative reward:
\[
\max_{\theta} \; \mathbb{E}_{\tau \sim \pi_{\theta}} [R(\tau)],
\]
where \(\tau\) denotes a trajectory and \(R(\tau)\) its total reward. The central difficulty is that reward is often delayed. A successful or unsuccessful final outcome may depend on a long sequence of earlier choices, yet the learner is not told directly which choice deserves credit or blame. This is the classical \emph{credit-assignment problem}. In contemporary reasoning-oriented systems, a related notion is that of a \emph{verifier}: an external procedure that can determine whether a generated solution satisfies a correctness criterion. Verifiers matter because they convert task success into a more concrete reward signal.

\subsection{PPO and GRPO}

\emph{Proximal policy optimization} (PPO) is a widely used RL algorithm that seeks to improve a policy while constraining each update to remain reasonably close to the previous one~\cite{schulman_ppo_2017}. In large-model applications, PPO is often coupled with a learned value function that estimates expected return and helps reduce the variance of policy-gradient updates. This framework is effective, but it introduces additional components that must themselves be trained and stabilized.

\begin{figure}[!htbp]
    \centering
    \includegraphics[width=0.9\linewidth]{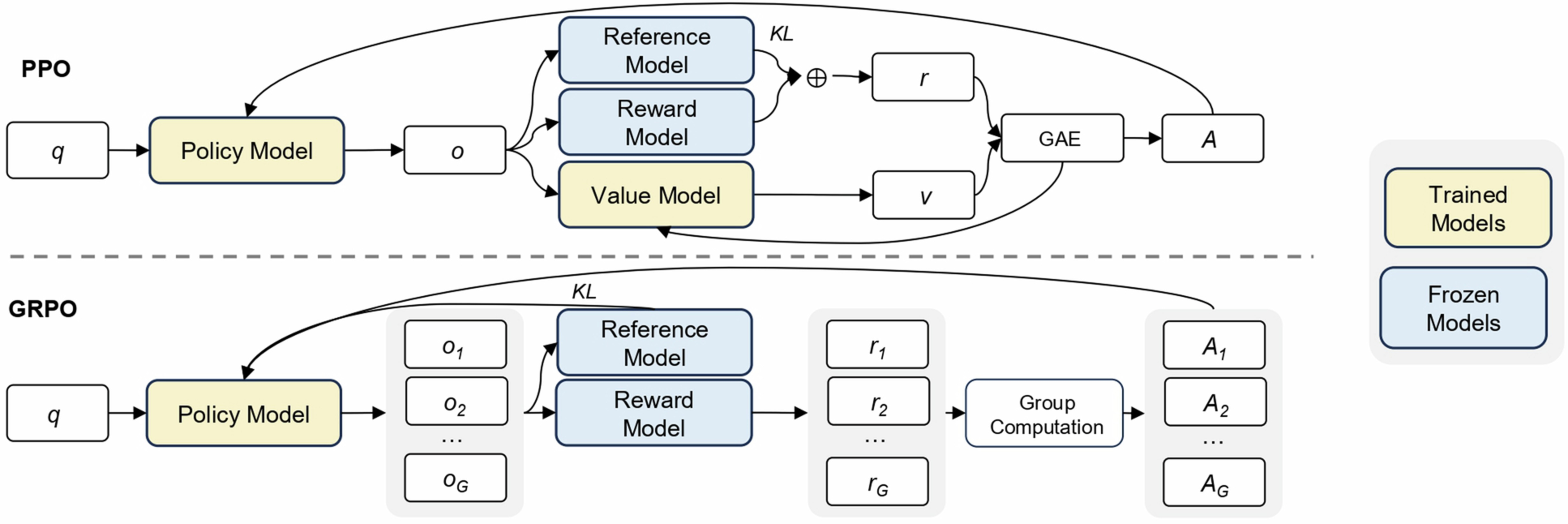}
    \caption{PPO and GRPO compared at a high level. GRPO removes the separate value model and instead computes relative advantages across a sampled group of responses for the same query.}
    \label{fig:background_grpo}
    \Description{Diagram comparing PPO and GRPO pipelines for reinforcement-learning-based language-model training.}
    \caption*{\footnotesize Source note: Guo et al., ``DeepSeek-R1 incentivizes reasoning in LLMs through reinforcement learning,'' Extended Data Fig. 2, open-access Nature article, CC BY 4.0, \url{https://www.nature.com/articles/s41586-025-09422-z/figures/4}.}
\end{figure}

\emph{Group Relative Policy Optimization} (GRPO) simplifies that setup and has become relevant in recent reasoning-focused language-model work~\cite{shao_deepseekmath_2024, guo_deepseekr1_2025}. Instead of learning a separate value model, GRPO samples multiple responses for the same prompt and evaluates them relative to one another under a reward model or verifier. Figure~\ref{fig:background_grpo} shows this contrast directly: PPO relies on an explicit value model, whereas GRPO computes relative advantages within a sampled response group.

This group-relative formulation is well suited to settings where several candidate outputs can be generated for the same problem and scored against a common criterion. Some candidates may be more accurate, more efficient, or otherwise better aligned with the task than others. GRPO uses these relative differences to shape the policy without requiring a separate learned value baseline. It does not remove the fundamental challenges of RL, especially credit assignment, but it reduces architectural complexity in a way that is well matched to modern reasoning-model training.

\section{Core Model Development and Evaluation Concepts}

\subsection{Training, Validation, and Generalization}

Machine-learning systems are ordinarily developed using separate training, validation, and test sets. The \emph{training set} is used to fit model parameters. The \emph{validation set} is used to select hyperparameters, choose stopping criteria, or compare design variants. The \emph{test set} is reserved for final evaluation and should not influence model development directly. This separation is methodologically essential because a model can appear to improve simply by adapting too closely to the data used for tuning.

\emph{Overfitting} occurs when a model captures idiosyncratic patterns in the observed data that do not generalize to new examples. \emph{Generalization} is the complementary objective: strong performance on data not seen during training. A model's \emph{inductive bias} is the set of structural assumptions built into the model class or training procedure that makes some solutions easier to learn than others. Such bias is unavoidable and, in practice, often beneficial. The relevant question is not whether bias exists, but whether it is well aligned with the structure of the task.

\subsection{Common Evaluation Measures}

Different evaluation metrics quantify different properties of model behavior, so no single summary number is universally adequate. \emph{Accuracy} measures the proportion of correct decisions once predicted probabilities have been thresholded into labels. \emph{ROC curves} characterize the trade-off between true-positive and false-positive rates across decision thresholds, and the \emph{area under the ROC curve} (AUC) summarizes ranking quality over that threshold range~\cite{pelanek_details_2017}. \emph{Calibration} asks a distinct question: whether predicted probabilities correspond to empirical frequencies, so that a prediction of 0.8 is borne out approximately 80\% of the time~\cite{guo_calibration_2017}.

Other measures compare patterns rather than binary decisions. \emph{Pearson correlation} measures linear association between two numeric quantities, whereas \emph{Spearman correlation} measures whether rank order is preserved. In clustering analysis, \emph{adjusted Rand index} (ARI) quantifies agreement between two partitions while correcting for agreement expected by chance~\cite{hubert_comparing_1985}. \emph{Normalized mutual information} (NMI) measures how much information one partition provides about the other~\cite{vinh_information_2010}. \emph{Homogeneity} asks whether each cluster predominantly contains members of a single reference class, whereas \emph{completeness} asks whether members of a reference class are assigned predominantly to the same cluster~\cite{rosenberg_v-measure_2007}. Table~\ref{tab:background_metrics} summarizes these distinctions.

\begin{table}[!htbp]
\centering
\small
\DataTableHeaderRows{1}
\begin{tabular}{p{0.2\linewidth} p{0.31\linewidth} p{0.35\linewidth}}
\toprule
Measure & What it captures & Useful intuition \\
\midrule
Accuracy & Fraction of decisions that are correct & Sensitive to the threshold used to turn probabilities into labels \\
ROC / AUC & Ranking quality across thresholds & A high AUC means positives tend to receive higher scores than negatives \\
Calibration & Agreement between predicted probabilities and observed frequencies & A confident model is not necessarily a well-calibrated one \\
Pearson correlation & Linear association between two numeric patterns & Best when relative magnitudes matter linearly \\
Spearman correlation & Rank-order association & Useful when order matters more than exact spacing \\
ARI & Partition agreement corrected for chance & High ARI means two clusterings make similar pairwise grouping decisions \\
NMI & Shared information between partitions & High NMI means one partition tells you a great deal about the other \\
Homogeneity / Completeness & Purity of clusters vs integrity of reference classes & High homogeneity avoids mixed clusters; high completeness avoids splitting one class across many clusters \\
\bottomrule
\end{tabular}
\caption{Common development and evaluation concepts.}
\label{tab:background_metrics}
\end{table}

Together, these concepts define a shared foundation for representation learning, sequential prediction, reasoning-based optimization, and model evaluation.

\chapter{Multi-Agent Collaborative Framework For Math Problem Generation}

\section{Introduction and Related Work}

Dynamically generating mathematics practice exercises and homework problems is a longstanding challenge for intelligent tutoring systems (ITS) and educators~\cite{heilman_good_2010}. Typically, educators and ITS either rely on existing problems or create new variations through minor modifications. Automatic Question Generation (AQG) aims to address this challenge but faced limited early success~\cite{heilman_good_2010,zhang_review_2022}. Over time, diverse AQG methods have emerged, including the fine-tuning of pretrained language models for topic-controlled question generation~\cite{li_novel_2025,vaswani_attention_2017}, often leveraging contrastive examples (examples of similar questions that differ in topic) to help the model distinguish and generate questions suited to specific concepts. More recently, prompt engineering guided by Bloom’s Taxonomy~\cite{blobstein_angel_2024}, few-shot prompting with chain-of-thought reasoning~\cite{scaria_automated_2024,blobstein_angel_2024}, and contextualized math word problem generation with equation constraints~\cite{hwang_contextualized_2024,wang_math_2021} have further advanced the field. Modern ITS require robust AQG capabilities to dynamically provide personalized practice problems, effectively intervening in students' learning trajectories~\cite{piech_deep_2015}.

\begin{figure}
    \centering
    \includegraphics[width=0.82\linewidth]{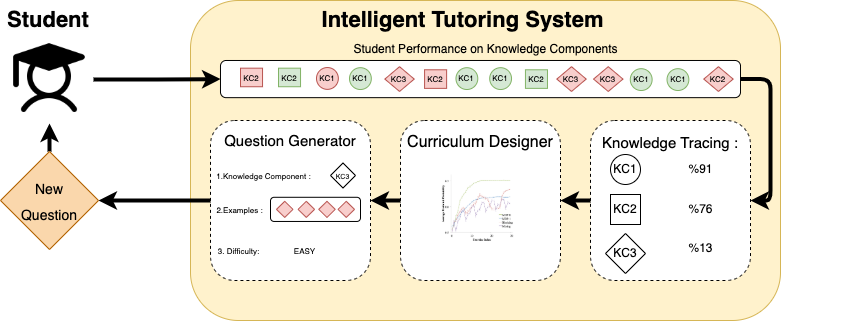}
    \caption{Typical workflow of an ITS.}
    \label{fig:ITS}
    \Description{Workflow of an ITS.}
\end{figure}

An ITS can be conceptualized as managing a student's knowledge state. Figure~\ref{fig:ITS} illustrates a typical ITS architecture where a knowledge-tracing module continuously monitors student comprehension across various knowledge components (KCs). At each interaction, the ITS selects a KC based on the student's knowledge state and, using a Curriculum Designer module commonly implemented via reinforcement learning (RL) models or search algorithms like expectiminimax~\cite{piech_deep_2015}, presents an appropriate problem as an intervention. This adaptive process supports personalized curricula aimed at continuously enhancing student mastery of KCs. Consequently, ITS must be capable of dynamically generating practice problems for any KC, leveraging heuristics from example problems and incorporating difficulty assignment mechanisms.

The AQG task can vary significantly based on system requirements. For instance,~\cite{blobstein_angel_2024} utilizes Large Language Models (LLMs) prompted explicitly by textbook chapters and related question sets, while other systems may rely solely on textbook chapters without additional prompts. Another common variation is concept-based generation, where AQG systems create questions derived directly from explicitly defined concepts or knowledge graphs. In this chapter, we approach AQG from an ITS perspective, defining inputs as: (1) the KC name per Common Core State Standards definitions, (2) a set of example questions for that KC, and (3) the required difficulty level (easy, medium, or hard). The system outputs a generated question and its answer.

AQG aligns with broader natural language generation (NLG) challenges, particularly the lack of definitive ground truths and the subjective nature of human evaluations. Because of this, evaluating NLG has been a key area of recent research, giving rise to frameworks like G-Eval~\cite{liu_g-eval_2023}, which leverages GPT-4 as a reference-free evaluator for tasks such as summarization, demonstrating improved alignment with human judgments. However, biases may persist in these LLM-based evaluation methods, as highlighted by~\cite{koo_benchmarking_2024,wang_dhp_2025}. Task-specific initiatives like QGEval~\cite{fu_qgeval_2024} reveal continued difficulties in ensuring strong alignment between automated metrics and human ratings in question generation, underscoring the broader challenges posed by subjectivity and the absence of definitive ground truth. Another major ongoing challenge in AQG is ensuring high-quality questions that are coherent, relevant, and accurately aligned with the intended difficulty level.

To address these issues, we explore recent advancements in LLMs, particularly in Inference Time Computation (ITC) techniques~\cite{snell_scaling_2024}. ITC generally refers to methods that extend or refine a model’s outputs during deployment without additional training, by orchestrating extra computational steps, retrieving context, or allowing models to interact with each other. Within this paradigm, we focus on multi-agent collaborative systems, also known as agentic workflows~\cite{wu_autogen_2023}, which improve LLM performance by facilitating interactions between multiple agents. This allows them to debate, complement each other’s strengths, and incorporate diverse perspectives. Of all the agent collaboration techniques, debate emerges as a particularly effective approach, inspired by the concept of the Society of Mind~\cite{minsky_society_1988} to harness collective knowledge. Studies show that multi-agent debate can enhance factuality and reasoning~\cite{du_improving_2023}, foster divergent thinking~\cite{liang_encouraging_2024}, and even achieve state-of-the-art performance in mathematical reasoning~\cite{wu_autogen_2023}. Other lines of work focus on mechanisms such as hierarchical or roleplaying-based collaboration~\cite{zhang_review_2022,chen_oscars_2025} and competitive settings~\cite{zeng_autodefense_2024}, while recent frameworks like AutoGen~\cite{wu_autogen_2023}, Camel~\cite{li_camel_2023}, and MetaGPT~\cite{hong_metagpt_2024} facilitate a variety of multi-agent collaboration structures. By enabling agents to engage in structured interactions, these workflows enhance factual accuracy, reasoning capabilities, and the overall quality of generated questions.

In this chapter, we investigate the potential of integrating ITC techniques through collaborative multi-agent frameworks into AQG. We evaluate the outputs based on clarity, relevance, importance, answerability, and difficulty matching. Our contributions are three-fold. First, we propose two novel collaborative frameworks specifically designed for AQG tasks. Second, we develop a self-curation method for AQG guided by cognitive demand criteria from Bloom's taxonomy. Third, we introduce an automated evaluation framework to assess automatically generated math questions, considering clarity, relevance, importance, answerability, and difficulty control.

Our research explores two key questions:
\begin{enumerate}
    \item Does difficulty matching depend on the proposed difficulty?
    \item Do agentic workflows improve question generation compared to baseline models?
\end{enumerate}

\section{Method and Experimental Setup}

We propose two multi-agent systems that generate a question and answer based on a given difficulty level, set of examples, and the KC name for use in the ITS. In this section, we explain the data, agents, multi-agent workflows, prompting strategies, curation method, and evaluation metrics of the system.

\subsection{Data}

Our experiments utilize the Problem Bodies dataset (\url{https://sites.google.com/site/assistmentsdata/home/assistments-problems}). The dataset is an extension of the ASSISTments dataset, a widely used benchmark from ITS. Problem Bodies contains middle-school math questions accompanied by empirical student performance metrics. Each question in this dataset includes a ``percent correct'' attribute, representing the percentage of students who answered correctly. This allows us to categorize the questions into easy, medium, and hard difficulty levels, providing an empirically grounded measure of difficulty aligned with real-world student performance.

\subsection{Agents}

In our experiments, we use the following agents and roles:
\begin{description}
    \item[Teacher:] Generates math questions and answers for a specific KC at a set difficulty level. It returns concise responses while adjusting based on conversation context.
    \item[Generic Critic:] Provides objective, high-level feedback on clarity, relevance, and difficulty alignment without introducing new content. It references the overall discussion for consistent evaluation.
    \item[Consensus CEO:] Serves as the final decision-maker by reviewing conversation history and selecting the best question-answer pair from multiple agents. If a consensus exists, it reports that; otherwise, it chooses the option aligning best with the KC, difficulty requirements, and sample solutions.
    \item[Versatile Agent:] Dynamically participates in collaborative discussions, making one of three decisions based on chat history: (1) generate a new Q\&A pair, (2) revise an existing pair, or (3) endorse a peer’s pair while providing constructive feedback. This agent uses prior messages and examples to ensure its contributions align with the KC, intended difficulty, and the collective goal of reaching consensus.
\end{description}

\subsection{Prompt Design}

The full implementation prompts for the Teacher-Critic Cycle agents, the Collective Consensus agents, and the evaluation module are reported in Appendix~\ref{sec:appendix_ch3_prompt_templates}. In the main text, we focus on the role structure rather than reproducing the full prompt text. The Teacher prompt is responsible for question generation and revision, the Critic prompt for structured feedback, the Versatile Agent prompt for proposal, revision, or endorsement within the collective workflow, and the Consensus CEO prompt for final selection.

\subsection{Workflows and Experimental Setup}

\begin{figure}
    \centering
    \includegraphics[width=0.8\linewidth]{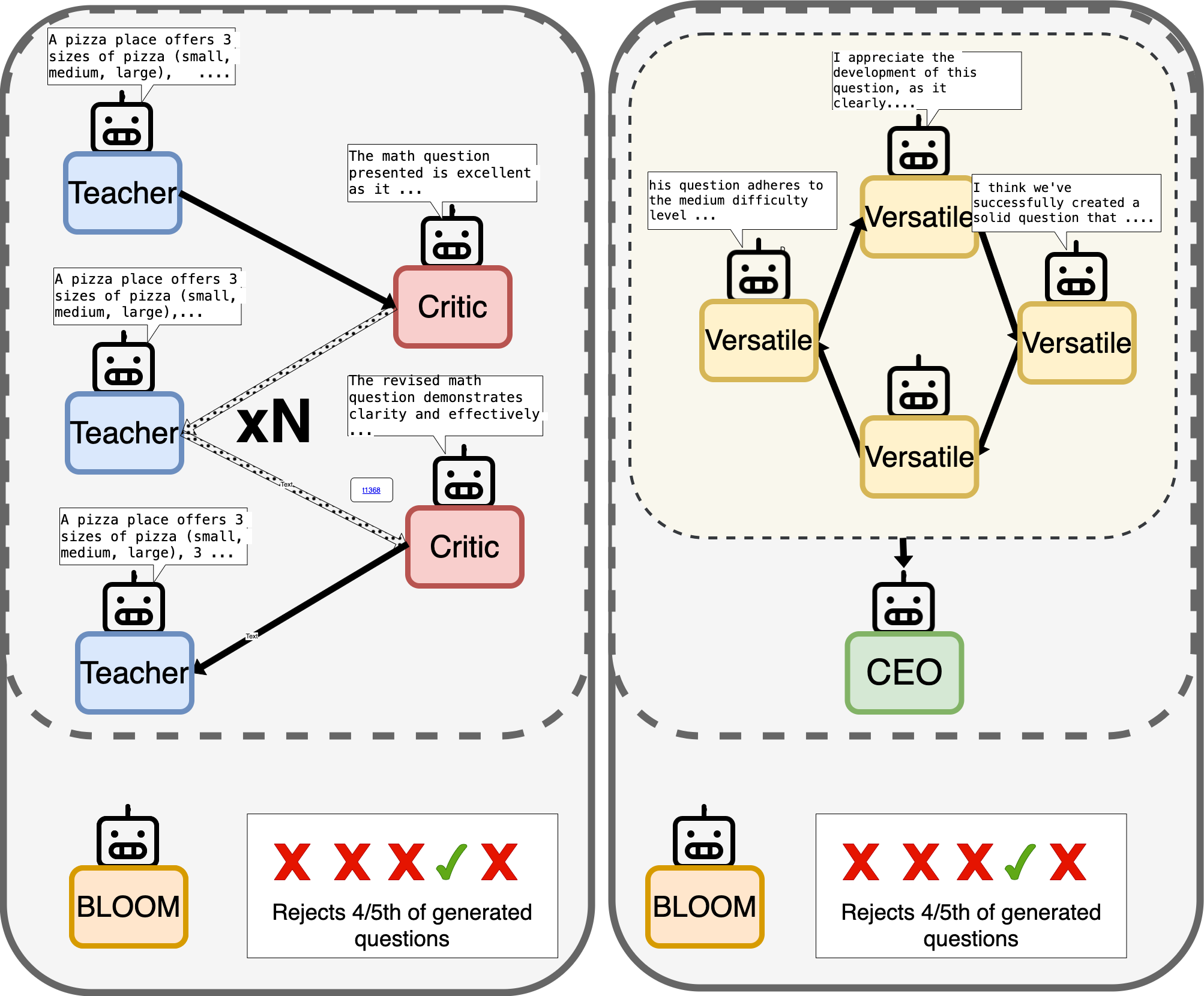}
    \caption{Agentic workflows. Left: Teacher Critic Cycle. Right: Collective Consensus.}
    \label{fig:agentic_workflows}
    \Description{Agentic workflows used in this chapter.}
\end{figure}

We systematically evaluated four distinct approaches, including two baseline scenarios and two agentic workflows (see Figure~\ref{fig:agentic_workflows}) designed to generate educational math question-answer pairs, each with varying complexity and collaborative structures:
\begin{description}
    \item[Baseline Teacher Zero-Shot:] This workflow uses a single teacher agent with a zero-shot prompt approach to generate self-contained math questions and answers without relying on prior examples or conversation history.
    \item[Baseline Teacher Few-Shot:] In this variant, the teacher agent is guided by a few example question-answer pairs, aligning it more closely with other few-shot-based workflows.
    \item[Teacher-Critic Cycle (TCC):] TCC involves two agents: a teacher generates an initial math question and answer, followed by iterative critique from a generic critic agent. The critic assesses clarity, relevance, difficulty, and pedagogical appropriateness, with the teacher refining the content across multiple feedback rounds. The full implementation prompts for these two roles are reported in Appendix~\ref{sec:appendix_ch3_prompt_templates}. Experimentally, we varied the number of interaction rounds between two to five, and evaluated the effectiveness of two prompt engineering techniques: Auto Chain-of-Thought (AutoCoT) and explicit Solution Generation, by enabling and disabling them. In Solution Generation, the teacher (or versatile agent) tries to output only the final answer to a math problem, while in AutoCoT, the teacher (or versatile agent) uses a step-by-step reasoning process before arriving at the final answer.
    \item[Collective Consensus (CC):] CC is a multi-agent workflow involving a collaborative conversation to reach consensus on math question-answer pairs. Initially, one versatile agent generates a question-answer pair, after which two to four versatile agents sequentially contribute by either creating a new question, revising an existing question with feedback, or explicitly agreeing with feedback. The corresponding collaboration and selection prompts are also reported in Appendix~\ref{sec:appendix_ch3_prompt_templates}. Decoding parameters, including sampling seed and temperature, are randomized per agent to encourage diverse perspectives. Following iterative discussion (ranging from two to five rounds), a consensus CEO agent reviews the conversation. If consensus is achieved, the CEO selects the agreed-upon final pair; otherwise, it chooses the best candidate based on collective judgment and educational alignment.
\end{description}

Across all workflows, we tested three prompting strategies, which we detail in Section~\ref{sec:difficulty_prompting_strategies}: Empirical Difficulty Prompting, Empirical Prompting, and Simple Prompting. These strategies help assess whether providing explicit difficulty-level context enhances question quality relative to random or difficulty-specific examples.

\subsection{Difficulty Prompting Strategies}
\label{sec:difficulty_prompting_strategies}

\begin{description}
    \item[Empirical:] All but the baseline use a few-shot approach. Math questions are labeled ``easy'', ``medium'', or ``hard'' using real student performance data (higher ``percent correct'' indicates easier). These labeled examples guide the model in generating questions matching the requested difficulty.
    \item[Prompting Empirical:] This variant also relies on empirical difficulty data but only presents examples matching the requested difficulty. For instance, if the model is asked to produce an ``easy'' question, only empirically identified easy examples are shown.
    \item[Prompting Simple (Random Examples):] Here, the model receives randomly selected examples from all difficulty tiers. It must rely solely on the requested difficulty (``easy'', ``medium'', or ``hard'') without explicit guidance from example labels, placing more responsibility on general instructions.
\end{description}

\subsection{Self-Curation}

We design a \emph{Bloom Agent} that assigns each output question a Bloom evaluation score (1--5) based on its cognitive demands, ensuring alignment with the three Bloom tiers—lower (Remembering, Understanding), middle (Applying, Analyzing), and upper (Evaluating, Creating). Because generating many candidate questions is computationally inexpensive, we filter out those that fail the expected cognitive challenge and retain only those demonstrating strong Bloom alignment. We also compare this curation method against Random Curation (RC) of our methods for comparison. Unlike conventional curation, which may rely on syntactic features, our approach prioritizes pedagogical rigor over surface textual similarity, ensuring that even a surplus of suboptimal questions yields a subset with greater alignment to Bloom-level demands.

\subsection{Evaluation Metrics}

To evaluate the generated questions, we use prompt engineering to quantify the quality of each instance based on the following criteria:
\begin{description}
    \item[Relevance:] Measures how well the generated questions align with example questions and the targeted KC. Scored on a scale of 1 (not relevant) to 5 (highly relevant).
    \item[Importance:] Assesses whether the question emphasizes key conceptual components of the targeted KC. Rated from 1 (least important) to 5 (essential).
    \item[Clarity:] Evaluates structural coherence and linguistic precision of a question to ensure it is easily understood by middle-school students. Scored from 1 (unclear) to 5 (exceptionally clear).
    \item[Difficulty Matching:] Determines how well a question’s complexity aligns with the specified difficulty level (easy, medium, hard). Cognitive difficulty ranges from basic factual recall (easy) to moderate conceptual reasoning (medium) and advanced analytical skills (hard). Scored from 1 (no alignment) to 5 (perfect alignment).
    \item[Answerability:] Evaluates whether a middle-school student can reasonably understand and respond to the question based on the provided information. Rated from 1 (unanswerable) to 5 (clearly answerable).
\end{description}

\subsubsection{Evaluation Prompt Templates}

The full evaluator prompts for clarity, relevance, importance, difficulty matching, and answerability are reported in Appendix~\ref{sec:appendix_ch3_prompt_templates}. Figure~\ref{fig:eval-bloom-prompt} remains in the main chapter because the Bloom-alignment prompt is central to the self-curation mechanism.

\begin{figure}[htbp]
\centering
\setlength{\fboxsep}{8pt}
\fcolorbox{promptframe}{white}{%
\parbox{0.94\linewidth}{%
\ttfamily\footnotesize\raggedright
You are an experienced mathematics educator. Your task is to evaluate whether a generated mathematics question aligns with the intended cognitive challenge level defined by Bloom's Taxonomy.\par
\par
Bloom's Taxonomy includes six levels: Remembering, Understanding, Applying, Analyzing, Evaluating, and Creating.\par
\par
The target mapping is:\par
\par
- Easy: Remembering and Understanding\par
- Medium: Applying and Analyzing\par
- Hard: Evaluating and Creating\par
\par
You will be given:\par
\par
- [Question]\par
- [Target knowledge component]\par
- [Intended difficulty level]\par
\par
Analyze the cognitive demands of the question, compare them with the intended Bloom levels, assign a five-point Bloom-alignment score, and explain your reasoning step by step.%
}}
\caption{Bloom-alignment evaluation prompt.}
\Description{Bloom-alignment evaluation prompt.}
\label{fig:eval-bloom-prompt}
\end{figure}

\section{Results and Discussion}

\begin{figure}
    \centering
    \includegraphics[width=0.7\linewidth]{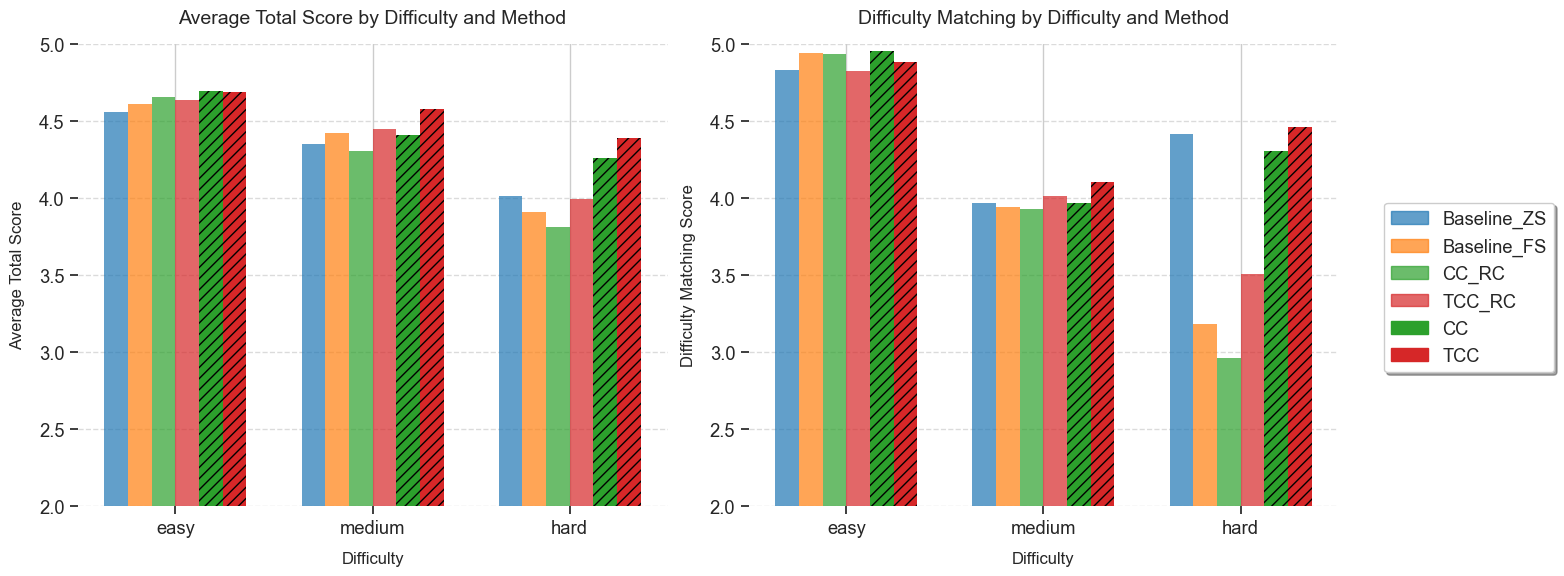}
    \Description{Bar graph showing average score and difficulty matching by difficulty and methods.}
    \caption{Average score and difficulty matching by difficulty and methods.}
    \label{fig:res_bargraph}
\end{figure}

\begin{table}
    \centering
    \small
    \renewcommand{\arraystretch}{0.7}
    \setlength{\tabcolsep}{4pt}
    \caption{Evaluation results of different question generation workflows.}
    \label{tab:results}
    \DataTableHeaderRows{1}
    \begin{tabular}{lcccccc}
        \toprule
        \textbf{Method} & \textbf{Clarity} & \textbf{Relevance} & \textbf{Importance} & \textbf{Difficulty} & \textbf{Answerability} & \textbf{Avg.} \\
        \midrule
        \textbf{Baseline\_ZS}  & 3.66 & 4.61 & 4.67 & 4.41 & 4.65 & 4.40 \\
        \textbf{Baseline\_FS}  & 3.70 & 4.93 & \textbf{4.73} & 4.02 & 4.71 & 4.42 \\
        \textbf{CC\_RC}        & 3.50 & \textbf{4.95} & 4.71 & 3.94 & 4.61 & 4.34 \\
        \textbf{TCC-RC}        & \textbf{3.72} & 4.90 & \textbf{4.73} & 4.11 & 4.79 & 4.45 \\
        \textbf{CC}            & 3.60 & \textbf{4.99} & \textbf{4.76} & \textbf{4.96} & \textbf{4.94} & \textbf{4.65} \\
        \textbf{TCC}           & \textbf{3.75} & 4.92 & 4.70 & \textbf{4.88} & \textbf{4.94} & \textbf{4.64} \\
        \bottomrule
    \end{tabular}
\end{table}

Table~\ref{tab:results} shows the evaluation results of our methods, TCC and CC, compared to both the baseline and non-curated agentic methods. While TCC and CC outperform both the baselines and their non-curated counterparts across all evaluation metrics, the improvements are incremental rather than drastic. The greatest gains are observed in Difficulty Matching and Relevance, which suggests that incorporating iterative critique and collective refinement enhances alignment with the intended cognitive challenge and pedagogical quality. These findings directly address the second research question. Additionally, the fact that non-curated agentic methods underperform the baseline suggests that agent-based generation alone is insufficient without structured selection or refinement. Agent responses exhibit greater variability, sometimes failing to meet standards of clarity, difficulty alignment, or answerability. This highlights the importance of iterative curation mechanisms to ensure consistency and reliability in multi-agent AQG.

Figure~\ref{fig:res_bargraph} presents the Difficulty Matching and Average Score across different difficulty levels (easy, medium, hard) for each method. The trends observed align with the overall evaluation results. Our curated methods (TCC and CC) outperform the baselines and non-curated methods, though the improvements remain incremental. A general pattern emerges where both Difficulty Matching and Average Score decrease as question difficulty increases. This finding addresses the first research question. The observed trend suggests that difficulty matching is indeed more challenging for harder questions, indicating that LLM-based generation struggles to maintain appropriate cognitive complexity as difficulty increases. Notably, the baseline models and non-curated agentic methods perform significantly worse on hard questions. This reinforces the need for structured prompting and iterative refinement to produce well-calibrated high-difficulty questions.

Unexpectedly, the Baseline Zero-Shot method achieves higher Difficulty Matching scores than both the Baseline Few-Shot model and several agentic methods (including curated and non-curated variants), which suggests that few-shot examples may introduce biases or inconsistencies in aligning questions with intended difficulty. These findings underscore the intricate interplay between prompting strategies and difficulty alignment, prompting further study of how example-based prompts shape model outputs.

Taken together, these trends suggest that optimizing inference-time computation should be treated as a tuning problem rather than a simple scaling rule where more interaction always produces better questions. They also indicate that some of the remaining performance differences among stronger methods may be difficult to interpret cleanly without more discriminative evaluation. This motivates a more careful treatment of both workflow design and evaluation reliability, especially when improvements are modest but consistent.

\begin{figure}
    \centering
    \captionsetup{font=small}
    \begin{subfigure}{0.32\linewidth}
        \centering
        \includegraphics[width=\linewidth]{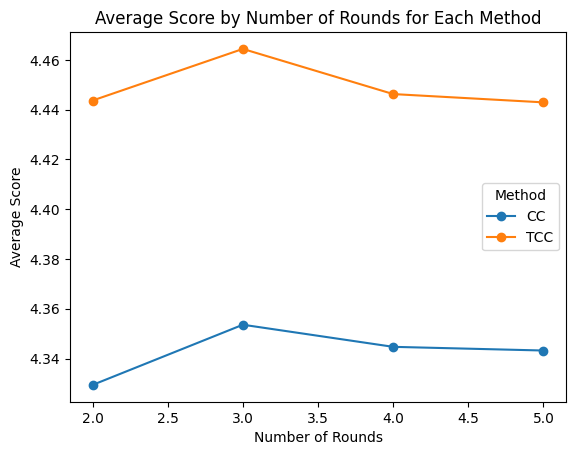}
        \caption{}
        \label{fig:rounds}
        \Description{Average Score by Number of Rounds for Each Method.}
    \end{subfigure}\hfill
    \begin{subfigure}{0.32\linewidth}
        \centering
        \includegraphics[width=\linewidth]{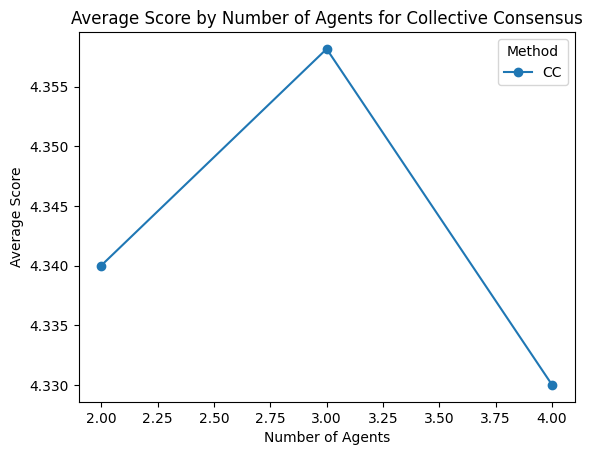}
        \caption{}
        \label{fig:agents}
        \Description{Average Score by Number of Agents for Collective Consensus.}
    \end{subfigure}\hfill
    \begin{subfigure}{0.32\linewidth}
        \centering
        \includegraphics[width=\linewidth]{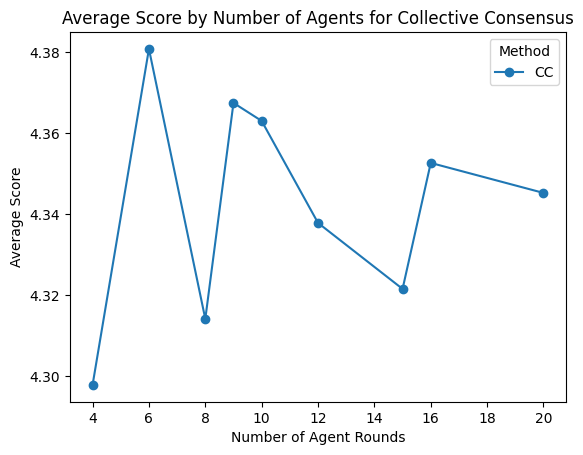}
        \caption{}
        \label{fig:n_agent_round}
        \Description{Average Score by Number of Agent Rounds for Collective Consensus.}
    \end{subfigure}\hfill
    \caption{Average score plotted against: (a) number of agentic discussion rounds for both TCC and CC methods, (b) number of agents in the CC workflow, and (c) number of agentic rounds (number of agents multiplied by number of rounds).}
    \label{fig:combined_plot}
\end{figure}

A key question involves how much inference computation is optimal. Since these are iterative ITC methods, merely increasing the number of rounds or agents does not guarantee performance gains and can even yield diminishing returns, as shown in Figure~\ref{fig:combined_plot}. The number-of-rounds trend is shown in Figure~\ref{fig:rounds}, where both TCC and CC improve from two to three rounds, after which the gains flatten or slightly reverse. This suggests that a small amount of iterative refinement is useful, but extended back-and-forth does not continue to improve outputs at the same rate. The agent-count pattern appears in Figure~\ref{fig:agents}, where CC reaches its best average score with three agents and declines when a fourth agent is added, indicating that additional participants may introduce coordination costs or redundant feedback rather than steadily improving the final question.

The total agent-round budget further reinforces this interpretation. As shown in Figure~\ref{fig:n_agent_round}, performance varies non-monotonically as the amount of computation increases, with some of the strongest settings appearing at small-to-moderate budgets rather than at the highest values. This is important from a systems perspective because it suggests that the marginal value of extra inference-time collaboration depends on how that computation is organized, not only on how much of it is used. For an ITS deployment, this makes workflow configuration a practical optimization problem: the goal is not to maximize discussion length, but to identify the point where additional deliberation no longer improves pedagogical quality enough to justify the extra computational cost.

Furthermore, our study finds that the effectiveness of prompting strategies for enabling the in-context learning ability of LLMs is minimal. As shown in Table~\ref{tab:difficulty_method}, different few-shot prompting techniques had little to no impact on performance. This suggests that current few-shot learning strategies may be suboptimal for the AQG task. Future research could explore alternative prompt engineering and adaptive few-shot learning techniques to better leverage in-context learning for question generation.

\begin{figure}
    \centering
    \includegraphics[width=0.7\linewidth]{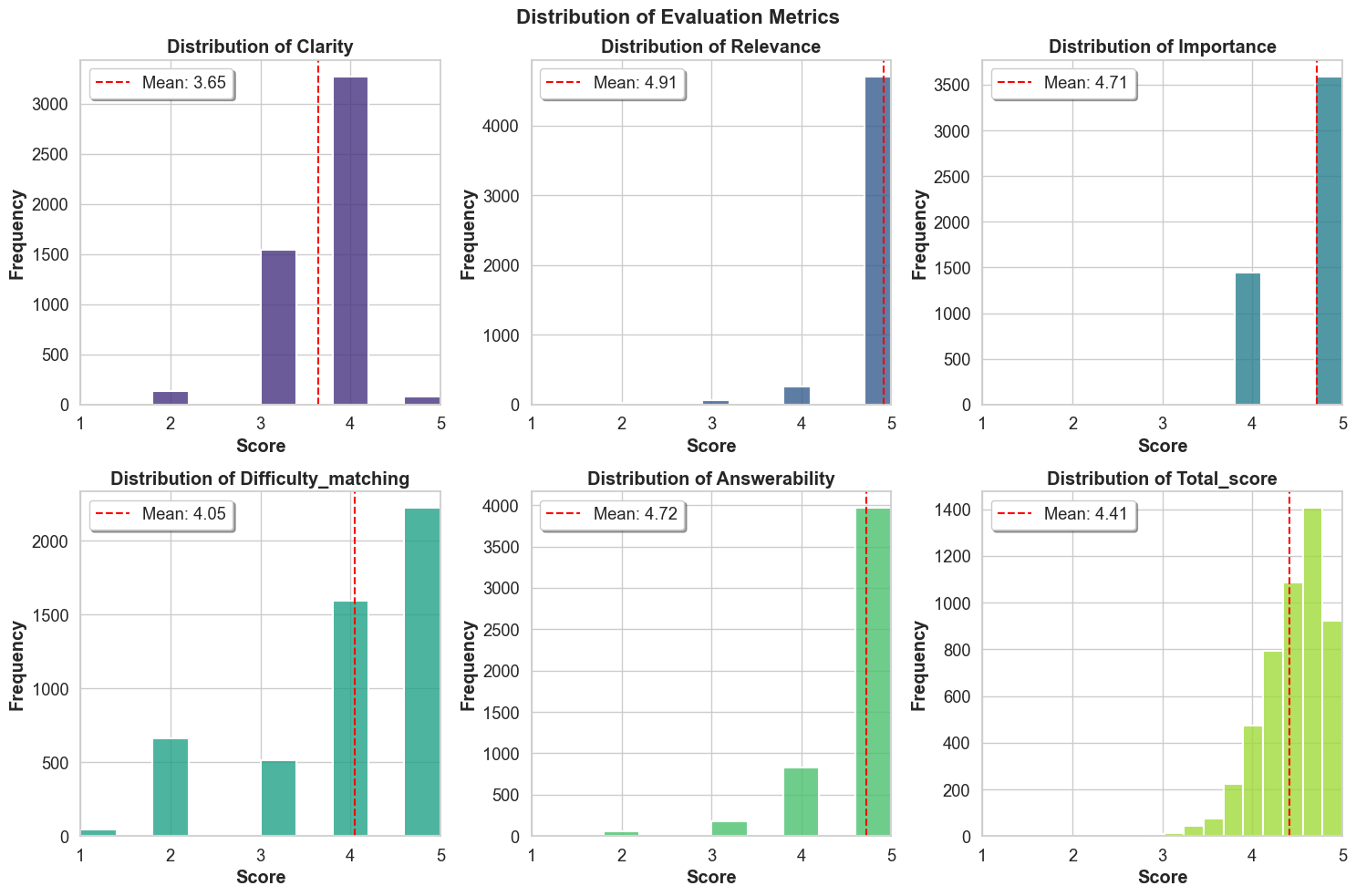}
    \caption{Histogram of evaluation metrics.}
    \label{fig:histogram}
    \Description{Distribution of evaluation metrics.}
\end{figure}

Finally, a critical area for future work is the quality of evaluation itself. All our findings rely on automated evaluations conducted by GPT-4 and based on state-of-the-art NLG evaluation frameworks~\cite{liu_g-eval_2023,fu_qgeval_2024}, which raises concerns about potential biases and limitations in LLM-based self-evaluation. In particular, we observed a ceiling effect in the resulting evaluation scores, similar to what the G-Eval authors reported, as shown in Figure~\ref{fig:histogram}. The distributions for Relevance, Importance, and Answerability are heavily concentrated near the upper end of the scale, which indicates limited spread among strong outputs and therefore reduced ability to separate good generations from very good ones. By contrast, Clarity is somewhat lower and more dispersed, suggesting that wording, phrasing, and question formulation remain more variable than the other evaluated dimensions.

Difficulty Matching shows a broader spread and a lower mean than the strongest metrics, which is consistent with the broader pattern observed in the bar chart and supports the conclusion that calibrated difficulty remains one of the hardest aspects of AQG. The Total Score distribution is also compressed into a relatively high range, reinforcing the idea that small numerical differences between strong systems should be interpreted cautiously. If the evaluation system exhibits misalignment with human judgment, then even consistent ranking differences could reflect evaluator bias rather than meaningful pedagogical improvement. Future validation should collect human evaluation data and use it to calibrate the automated evaluation module against human judgments. This is especially important here, where subtle differences between high-performing workflows need stronger validation than automated scoring alone can provide.

Overall, while our study demonstrates that the success of difficulty matching is dependent on the proposed difficulty level and that agentic workflows can improve question generation compared to the baseline, these improvements come with important caveats. Optimization of inference computation, refinement of few-shot learning strategies, and enhancing evaluation reliability are key directions for future research to advance AI-driven educational content generation.

\begin{table}
    \centering
    \caption{Difficulty prompting strategies results.}
    \label{tab:difficulty_method}
    \DataTableHeaderRows{1}
    \begin{tabular}{llcc}
        \toprule
        \textbf{Method} & \textbf{Strategy} & \textbf{Difficulty} & \textbf{Avg. Score} \\
        \midrule
        CC  & empirical           & \textbf{4.71} & 4.60 \\
        CC  & prompting empirical & 4.68          & 4.64 \\
        CC  & prompting simple    & \textbf{4.70} & 4.61 \\
        TCC & empirical           & 4.66          & \textbf{4.66} \\
        TCC & prompting empirical & 4.61          & 4.64 \\
        TCC & prompting simple    & 4.67          & \textbf{4.65} \\
        \bottomrule
    \end{tabular}
\end{table}

\chapter{Extracting Causal Relations in Deep Knowledge Tracing}

\section{Introduction}

Computer-assisted educational technology, such as intelligent tutoring systems~\cite{fengSystematicReviewLiterature2021, kulikEffectivenessIntelligentTutoring2016}, enables personalizing activities to suit individuals with varying levels of proficiency. A key component of these adaptive systems is knowledge tracing (KT), which aims to estimate students' mastery of several exercise concepts, known as knowledge components (KCs), as they interact with the corresponding exercises~\cite{explainable_kt, chenPrerequisiteDrivenDeepKnowledge2018}. Formally, \(x_i = \{e_t, a_t\}\) represents a student's answer pair, where \(e_t\) denotes the exercise ID and \(a_t\) denotes whether the student answered correctly. Given a series of past interactions, \(X = \{x_1, x_2, \dots, x_t\}\), and the next concept exercise, \(e_{t+1}\), the task of the KT model is to estimate the likelihood of the student answering correctly, \(a_{t+1}\)~\cite{corbett_knowledge_1995}. Traditionally, Bayesian Knowledge Tracing (BKT)~\cite{corbett_knowledge_1995} was used to perform KT. Despite its popularity, BKT is often critiqued for its binary representation of knowledge states, which oversimplifies the learning process. Several models were later proposed to address these limitations. Learning Factors Analysis~\cite{cenLearningFactorsAnalysis2006} improved upon BKT by representing learning as a continuous process influenced by multiple exercise interactions and accumulated practice. Performance Factors Analysis~\cite{jrPerformanceFactorsAnalysis} further captured the complexity of learning by tracking the effects of correct and incorrect prior exercise attempts on performance. These developments laid the groundwork for Deep Knowledge Tracing (DKT)~\cite{piech_deep_2015}, which popularized a key extension to KT: the ability to implicitly infer relationships between exercises.

In this chapter, we focus on explainability within DKT. Prior work has often attributed DKT's gains to its ability to model bidirectional relationships among KCs, allowing the system to infer understanding of one concept from performance on another. Here, we argue for a different interpretation: DKT's strength lies more in its implicit representation of prerequisite structure as a causal graph. By pruning exercise-relation graphs into directed acyclic graphs (DAGs) and training DKT on causal subsets of the ASSISTments datasets, we show that DKT's predictive behavior aligns strongly with these causal structures. We also introduce a modified method for extracting exercise-relation DAGs from DKT's learned representations and provide empirical evidence that it yields more stable and informative dependency patterns. In this chapter, we therefore reframe DKT's interpretability in terms of causal dependencies rather than generic bidirectional relations, and we make the supporting code publicly available for follow-on work.\footnote{\url{https://github.com/kevinhongca/dkt-causal-relations}}

The relationships between exercises can take the form of prerequisite dependencies, where understanding one exercise improves performance on another, or corequisites, where exercises depend on each other. Researchers have demonstrated that exercise relationships can be mapped into an exercise graph by analyzing student learning data. An exercise graph represents the dependencies between different concept exercises based on student learning patterns. Creating accurate exercise graphs allows educators to better sequence lessons and ensure that students master foundational concepts before progressing to more advanced ones. KT models are preferred for this task because they learn temporal patterns in student data to capture how mastery of one exercise affects success on another~\cite{chenPrerequisiteDrivenDeepKnowledge2018, piech_deep_2015, tong_hgkt_2022}. This enables KT models to uncover learning dependencies that content-based or expert-defined methods may overlook. Furthermore, automating the discovery of exercise relations through KT models offers a scalable alternative to manually defining these relationships.

The original DKT work made a significant contribution to exercise-relation discovery~\cite{piech_deep_2015}. Its method assigns an influence score \(J_{ij}\) to every directed pair of exercises \(i\) and \(j\), defined as the conditional probability of correctly answering exercise \(j\) after correctly answering exercise \(i\) in the previous timestep, normalized by the sum of such conditional probabilities. The influence score from \(i\) to \(j\) is interpreted as evidence that performance on exercise \(i\) informs predicted success on exercise \(j\), so \(i\) may function as a prerequisite for \(j\). We refer to this procedure as the DKT method. Many later studies incorporated exercise relations into KT models to improve performance and interpretability. For example, Hierarchical Graph Knowledge Tracing (HGKT)~\cite{tong_hgkt_2022}, Structure-based Knowledge Tracing (SKT)~\cite{tongStructureBasedKnowledgeTracing2020}, and Deep Knowledge Tracing with Multiple Relations (DKTMR)~\cite{duanMoreAccurateInterpretable2024} track prerequisite and corequisite relationships between exercises. Graph Attention Knowledge Tracing (GAKT)~\cite{zhaoResearchDeepKnowledge2022} uses a graph attention network to uncover the underlying structure between exercises, learning these relationships using the method introduced in DKT. Graph-based Knowledge Tracing (GKT)~\cite{nakagawaGraphbasedKnowledgeTracing2019} explores multiple statistics-based and learning-based methods to infer a latent graph structure and use it for KT. Among the methods evaluated in GKT, the DKT procedure performed best on ASSISTments 2009~\cite{fengAddressingAssessmentChallenge2009}.

\section{Methodology}

We conduct our study using the ASSISTments datasets, which are among the largest publicly available KT datasets and are widely used as benchmarks for KT models~\cite{abdelrahmanKnowledgeTracingSurvey2023, fengAddressingAssessmentChallenge2009, priharExploringCommonTrends2022}. These datasets capture student interactions over extended periods and across a wide range of grade-school mathematics exercises. We utilize three datasets: ASSISTments 2009\footnote{\url{https://sites.google.com/site/assistmentsdata/home/2009-2010-assistment-data/skill-builder-data-2009-2010}} (\textit{skill builder data corrected collapsed}), ASSISTments 2012\footnote{\url{https://sites.google.com/site/assistmentsdata/2012-13-school-data-with-affect}} (\textit{2012-2013 data with predictions 4 final}), and ASSISTments 2017\footnote{\url{https://sites.google.com/view/assistmentsdatamining/dataset}} (\textit{anonymized full release competition dataset}). The 2009 dataset contains 346,860 exercise attempts from 4,217 students across 123 exercises. The 2012 dataset includes 6,123,270 attempts from 46,674 students across 265 exercises. The 2017 dataset consists of 942,816 attempts from 1,709 students across 102 exercises. The datasets do not contain personal information.

We use DKT, implemented via the pyKT library~\cite{liuPyKTPythonLibrary2023}, to learn latent exercise relations and generate the graphs used throughout the analysis. We first train a DKT model on each of the three ASSISTments datasets. We then switch the models to evaluation mode and apply the original DKT relation-extraction method. To ensure the resulting graph represents a causal structure, we apply a minimum threshold to the influence scores. This filtering step removes weaker edges that could introduce cycles, allowing us to construct a DAG. Since the distribution of influence scores varies across datasets, we select the minimum dataset-specific thresholds that enforce acyclicity. We use thresholds of 0.0107 for ASSISTments 2009, 0.0051 for ASSISTments 2012, and 0.0139 for ASSISTments 2017.

Using the exercise-relation DAG, we create a causal subset of each ASSISTments dataset by filtering interactions to include only exercises with at least one incoming or outgoing edge. This ensures that the subset consists exclusively of exercises with learned causal relationships, which allows us to isolate and evaluate the influence of these structures. We then retrain DKT on this subset and evaluate its predictive accuracy. To isolate the impact of causal structure, we also generate five random subsets with bidirectional relations. Each random subset contains the same number of exercise concepts as its corresponding DAG-based subset, but the exercises are selected randomly rather than by causal structure. The numbers of exercise concepts in these subsets are 60 for ASSISTments 2009, 83 for ASSISTments 2012, and 17 for ASSISTments 2017. We train a separate DKT model on each random subset and compute the mean and standard deviation of their AUC scores. Finally, we compute a \(z\)-score to compare the AUC of the DAG-based subset against the distribution of AUC scores from the random subsets across the three ASSISTments datasets.

Recognizing a limitation in DKT's standard relation-extraction method, where influence scores are computed from a single correct response per exercise, we propose a modified approach that simulates a student repeatedly answering the same type of exercise correctly until the model's estimated knowledge of that exercise stabilizes. Rather than relying on a one-time response, we feed multiple consecutive correct answers for a given exercise, allowing the model to iteratively update its estimate of the student's mastery over all exercises.

Let \(\hat{K}_t\) represent the student's estimated knowledge level of an exercise after the student has answered correctly \(t\) times. We continue feeding correct responses until the estimated mastery stabilizes according to the following criterion:

\[
\hat{K}_t = \hat{K}_{t-1}, \quad \forall t \in [t_0 - T, t_0] \text{ s.t. } t_0 \geq T,
\]

where, if the model's estimate does not change for \(T\) consecutive iterations, we stop feeding additional responses and take \(\hat{K}_t\) as the final knowledge estimate. In our experiments, we set \(T = 100\), chosen heuristically as a reasonable value to allow the knowledge estimates to stabilize. Then, inspired by the method proposed by Piech et al., we use the modified influence score

\begin{equation}
\label{eq:causal_jij}
J_{ij} = \frac{z(j|i)}{\sum_k z(k|i)}.
\end{equation}

Here, \(z(j|i)\) is the correctness probability assigned to exercise \(j\) on the next timestep given that the student answered exercise \(i\) correctly \(t_0\) times on the previous \(t_0\) timesteps. This iterative process provides a more stable estimate of concept relationships by reducing the influence of single-response variations. We evaluate the effectiveness of this modified relation-extraction method by repeating the earlier methodology using Equation~\ref{eq:causal_jij} to generate newly constructed causal graphs: we extract the new DAG-causal subset, retrain DKT on this subset, and compare its AUC scores against those of randomly selected subsets. When extracting the new DAG-causal subset, we again apply dataset-specific thresholds to the influence scores: 0.0129 for ASSISTments 2009, 0.0067 for ASSISTments 2012, and 0.0167 for ASSISTments 2017.

These thresholds are chosen as the minimum values that make the extracted exercise graph acyclic. Let
\[
G_{\tau} = (V, E_{\tau}), \qquad E_{\tau} = \{(i,j) : J_{ij} \ge \tau\},
\]
where \(V\) is the set of exercises and \(J_{ij}\) is the extracted influence score. As \(\tau\) increases, weak edges are removed from \(E_{\tau}\), so the graph becomes sparser. We select
\[
\tau^{*} = \inf \{\tau : G_{\tau}\ \text{is a directed acyclic graph}\},
\]
that is, the smallest threshold for which all directed cycles disappear. Intuitively, if \(\tau\) is too low, the graph contains reciprocal or longer cyclic dependencies that are inconsistent with a prerequisite interpretation. Raising \(\tau\) removes the weakest candidate relations first, and \(\tau^{*}\) is the point at which the remaining edges define a DAG. We use dataset-specific values because the empirical distribution of influence scores differs across ASSISTments 2009, 2012, and 2017, so the minimum acyclic threshold is not the same across datasets.

\section{Results}

\begin{table}
\centering
\renewcommand{\arraystretch}{1.2}
\caption{AUC results for the DAG-based causal subsets (Causal) and matched random subsets generated using the original DKT relation-extraction method.}
\DataTableHeaderRows{1}
\begin{tabular}{cccc}
\hline
Dataset & \textit{AUC} & Exercises (\#) & \(z\)-score \\
\hline
2009 Causal & $0.905$ & 60 & 3.50 \\
2009 Random & $0.859 \pm 0.013$ & 60 & -- \\
\hline
2012 Causal & $0.727$ & 83 & 1.50 \\
2012 Random & $0.721 \pm 0.004$ & 83 & -- \\
\hline
2017 Causal & $0.718$ & 17 & 1.02 \\
2017 Random & $0.704 \pm 0.011$ & 17 & -- \\
\hline
\end{tabular}
\label{tab:causal_dkt_method}
\end{table}

Across all three ASSISTments datasets, the causal subsets derived from the DKT method~\cite{piech_deep_2015} consistently achieve higher AUC scores than the mean AUC of randomly selected subsets, as shown in Table~\ref{tab:causal_dkt_method}. This suggests that well-defined causal knowledge dependencies improve a KT model's ability to trace student learning, and that KT models appear to learn causal relationships more readily than bidirectional ones. These findings also introduce a potential evaluation criterion for knowledge-structure graphs: graphs that more accurately capture prerequisite relationships may yield higher AUC scores when their corresponding concepts are used for training.

\begin{table}
\centering
\renewcommand{\arraystretch}{1.2}
\caption{AUC results for the DAG-based modified causal subsets (MC) and matched random subsets generated using the stabilized influence score from Equation~\ref{eq:causal_jij}.}
\DataTableHeaderRows{1}
\begin{tabular}{cccc}
\hline
Dataset & \textit{AUC} & Exercises (\#) & \(z\)-score \\
\hline
2009 MC & $0.875$ & 68 & 1.67 \\
2009 Random & $0.851 \pm 0.014$ & 68 & -- \\
\hline
2012 MC & $0.758$ & 90 & 5.93 \\
2012 Random & $0.721 \pm 0.004$ & 90 & -- \\
\hline
2017 MC & $0.712$ & 59 & 1.85 \\
2017 Random & $0.703 \pm 0.005$ & 59 & -- \\
\hline
\end{tabular}
\label{tab:causal_modified_method}
\end{table}

Applying the modified relation-extraction method yields the same qualitative pattern: the new causal subsets consistently outperform the mean AUC of the random subsets. Table~\ref{tab:causal_modified_method} also shows that the modified approach produces higher \(z\)-scores than the original DKT method on two of the three datasets, with a substantially higher average \(z\)-score overall. To quantify this comparison, we define

\[
z = \frac{\text{AUC}_{\text{causal}} - \mu_{\text{random}}}{\sigma_{\text{random}}},
\]

where \(\text{AUC}_{\text{causal}}\) is the AUC of the causal subset and \(\mu_{\text{random}}\) and \(\sigma_{\text{random}}\) are the mean and standard deviation of the AUCs obtained from the random subsets. These results suggest that the modified probing procedure may yield more accurate representations of the underlying knowledge structure.

\begin{figure}
    \centering
    \includegraphics[width=0.92\linewidth]{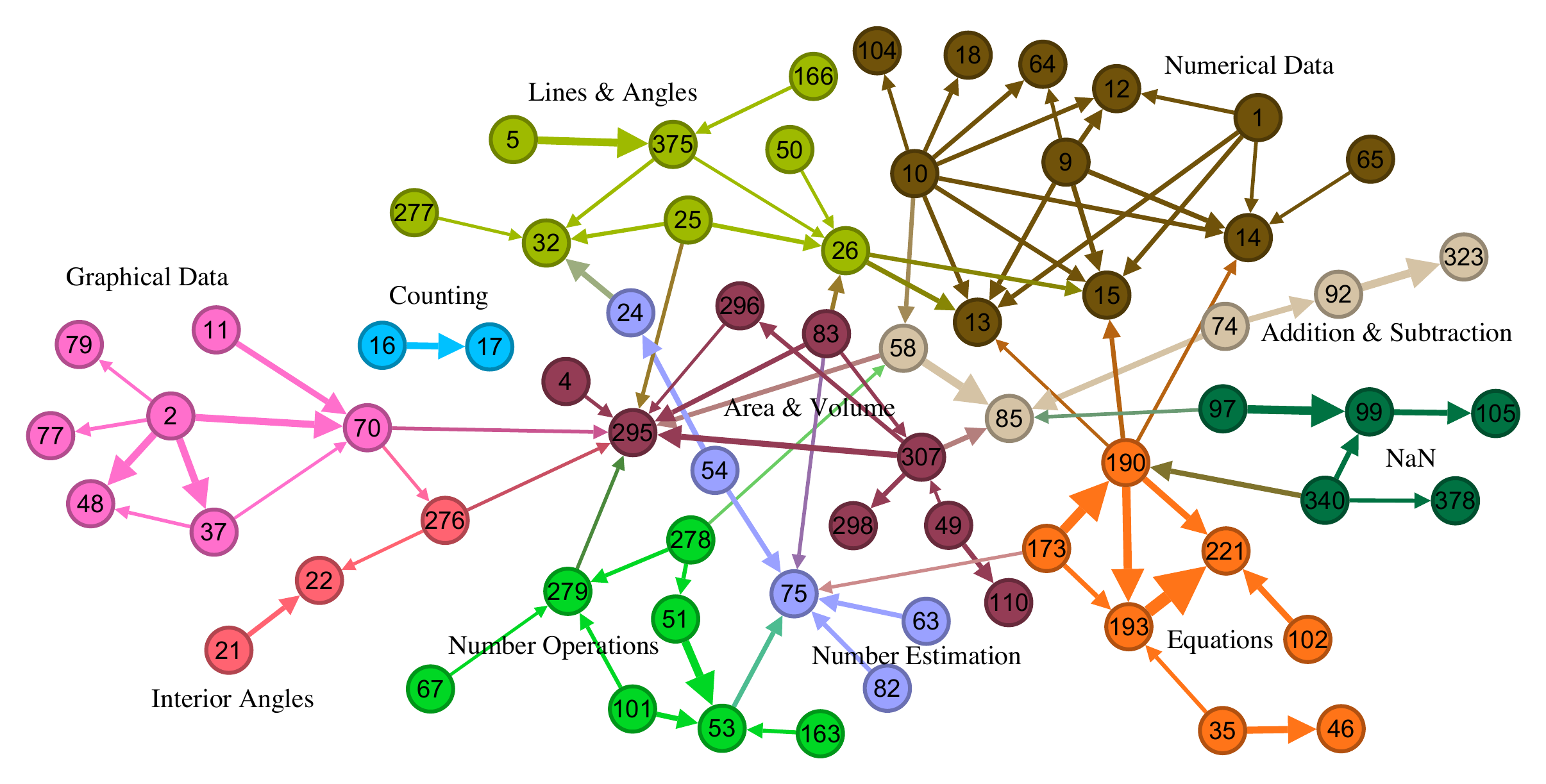}
    \Description{Exercise relation graph for ASSISTments 2009 using the modified influence-score method.}

    \vspace{0.75em}

    \includegraphics[width=0.92\linewidth]{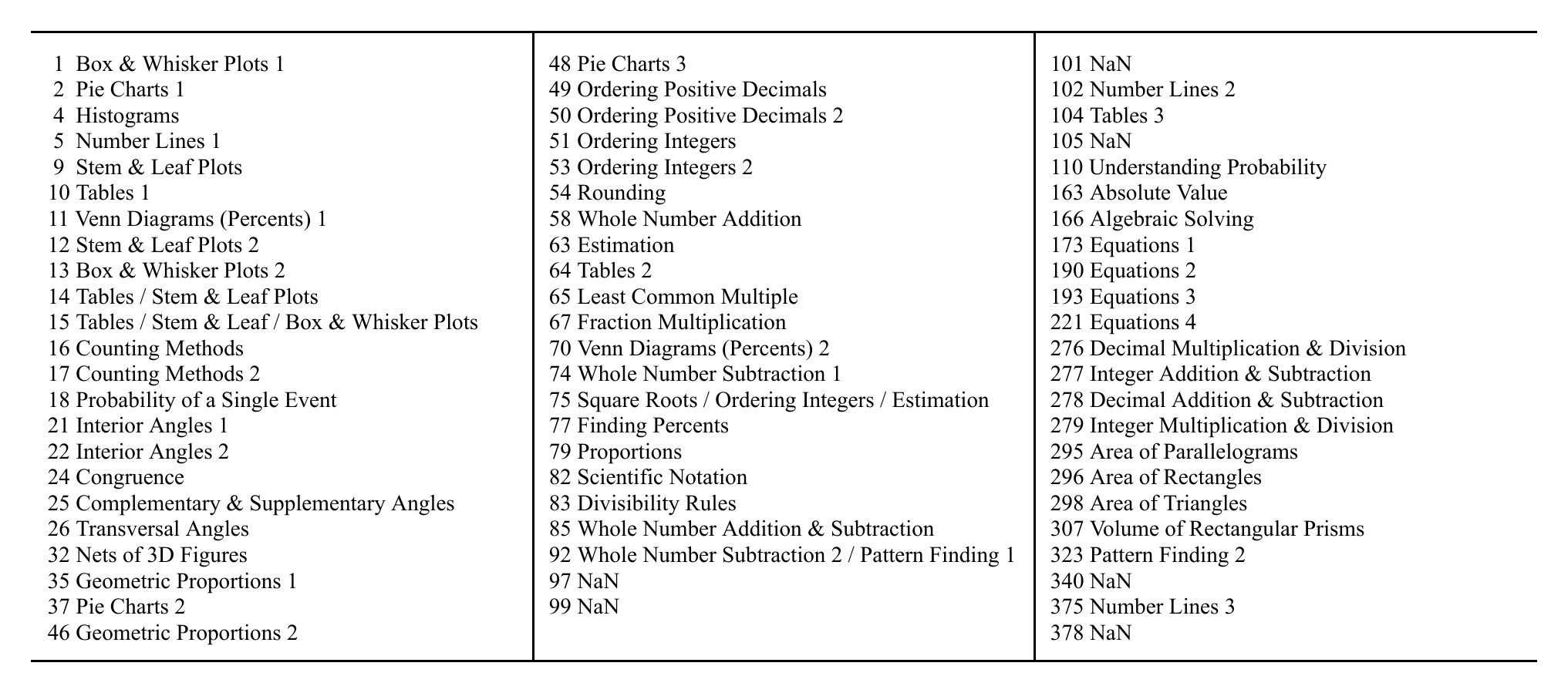}
    \Description{Exercise IDs and their corresponding names for ASSISTments 2009.}
    \caption{ASSISTments 2009 DAG graph of exercise relations produced using Equation~\ref{eq:causal_jij}. Arrow weight indicates prerequisite connection strength, and the lower panel provides the exercise ID lookup used to interpret the graph.}
    \label{fig:causal_graph}
\end{figure}

To interpret the results further, we use Equation~\ref{eq:causal_jij} to generate and analyze the directed exercise-relation graphs learned by the models. Figure~\ref{fig:causal_graph} presents the graph for ASSISTments 2009, where node numbers correspond to exercise IDs. To better identify topics across exercise groups, we apply the algorithms described in~\cite{blondelFastUnfoldingCommunities2008, lambiotteLaplacianDynamicsMultiscale2014} to partition the exercises into topics and use color coding to visually distinguish them. The resulting graph reveals a meaningful causal structure. Focusing on graphical-data topics, Exercise 2 (Pie Charts 1) serves as a prerequisite for multiple exercises. Since pie charts rely heavily on percentage calculations, it follows that Exercise 2 also contributes to learning related topics, such as Venn diagrams represented as percentages (Exercise 70) and proportion calculations (Exercise 79). The model also captures increasingly advanced topic structure: Exercise 2 (Pie Charts 1) is a prerequisite for Exercise 37 (Pie Charts 2), and both Exercises 2 and 37 serve as prerequisites for Exercise 48 (Pie Charts 3). Not every directed edge necessarily indicates a true prerequisite relationship, since some connections may still result from indirect correlations rather than direct dependencies, but the overall structure remains interpretable and coherent.

\section{Discussion}
\label{sec:causal_discussion}

This section provides intuition for the modified approach to extracting directed relations. We use a DKT model trained on the full ASSISTments 2009 dataset and set it to evaluation mode to predict the knowledge state of a new simulated student. Our goal is to identify the student's top three highest-ranked exercise concepts, sorted by estimated knowledge mastery, after correctly completing concept 278 (Decimal Addition \& Subtraction). Using the original method, we determine the top three predicted KCs immediately after answering concept 278 correctly. In the modified approach, we allow the student's understanding of concept 278 to stabilize through repeated correct responses before extracting the top three concept estimates. In other words, we seek three values of \(j\) that maximize \(y(j|i)\) and three that maximize \(z(j|i)\), given \(i = 278\) and \(i \neq j\).

\begin{table}
    \centering
    \small
    \setlength{\tabcolsep}{4pt}
    \renewcommand{\arraystretch}{1.2}
    \caption{Comparison of the top three KC mastery estimates produced by the original DKT method and the modified approach after a correct response to concept 278.}
    \DataTableHeaderRows{1,2}
    \begin{tabular}{c >{\raggedright\arraybackslash}p{2.9cm} c c >{\raggedright\arraybackslash}p{2.9cm} c}
        \toprule
        \multicolumn{3}{c}{\textit{DKT Method}} & \multicolumn{3}{c}{\textit{Modified Approach}} \\
        \cmidrule(lr){1-3} \cmidrule(lr){4-6}
        Exercise ID & Exercise Name & Mastery & Exercise ID & Exercise Name & Mastery \\
        \midrule
        24 & Congruence & 0.677 & 51 & Ordering Integers & 0.842 \\
        307 & Volume of Rectangular Prisms & 0.673 & 279 & Integer Multiplication \& Division & 0.839 \\
        26 & Transversal Angles & 0.667 & 58 & Whole Number Addition & 0.815 \\
        \bottomrule
    \end{tabular}
    \label{tab:causal_top_kc}
\end{table}

As shown in Table~\ref{tab:causal_top_kc}, the top three predicted KC masteries for concept 278 differ substantially between the DKT method and the modified approach. The highest-ranked concepts from the original DKT method are primarily geometry-related, whereas the modified method identifies exercises that are more directly relevant to concept 278, all involving integer operations. In this example, it takes only three consecutive correct responses on exercise 278 for the top-three ordering to stabilize. Because \(y(j|i)\) and \(z(j|i)\) correspond to the numerators of the original DKT method and the modified method, respectively, these values directly affect how exercise relations are constructed. The graph in Figure~\ref{fig:causal_graph} is consistent with this interpretation.

The modified method nevertheless has a practical limitation: students typically do not answer more than a few questions per exercise. The assumption that a student can be prompted up to 100 times on a given exercise is unlikely to reflect real learning scenarios. Future work should therefore explore more practical probing budgets, such as five responses, or convergence-based stopping criteria that terminate the iterations when the difference between successive knowledge estimates falls below a threshold.

Another limitation concerns the evaluation setup used to compare DAG-based causal subsets and random subsets. Although both subset types contain the same number of exercises, they may still differ in concept coverage, exercise difficulty, or the types of students engaging with the exercises. Some portion of the observed performance gains may therefore be attributable to differences in exercise or student characteristics rather than graph structure alone. Future work should explore more controlled subset construction, such as concept-matched sampling, in which random subsets reflect the distribution of concept types found in the DAG-based subset.

\section{Conclusion}

In this chapter, we show that DKT achieves better predictive performance when trained on DAG exercise subsets, which suggests that it effectively learns causal concept dependencies. We also introduce a modified method for calculating influence scores that stabilizes knowledge estimates and helps construct more accurate exercise-relation graphs. Together, these results support a causal interpretation of DKT's exercise dependencies while also highlighting the practical limitations of repeated probing and the need for more controlled experimental comparisons.

\chapter{Semantic Initialization for Deep Knowledge Tracing}

\section{Introduction}

In this chapter, we evaluate the hypothesis that question text carries information about what a student is being asked to do and that semantic embeddings might therefore improve DKT. We test that hypothesis under controlled conditions rather than assume in advance that semantic information should help.

Educational questions are not arbitrary identifiers. They contain instructions, mathematical language, and cues about the skill being assessed. Modern text representation methods are designed to place semantically similar texts near one another in vector space~\cite{devlin_bert_2019, reimers_sentence-bert_2019, gao_simcse_2021}. One hypothesis is that these semantic representations could provide a better inductive bias than sparse ID-based inputs.

There is also methodological precedent for testing this hypothesis directly. Prior work has augmented KT models with exercise-side information, built exercise-aware architectures, proposed specialized question representations, and pretrained question embeddings for educational prediction~\cite{su_exercise-enhanced_2018, liu_ekt_2019, yin_quesnet_2019, wang_seep_2022}. At the same time, work on KC imputation suggests that text is only part of the story and that surrounding interaction context can be just as important~\cite{pardos_imputing_2017}. That tension makes a controlled test necessary rather than assuming in advance that text should help.

In this study, we test a deliberately constrained version of the intervention. Instead of redesigning the whole KT model, we replace the default input embedding table in pyKT's DKT implementation with pretrained embeddings derived from question text through nine construction pipelines. The goal is to isolate one claim: whether semantic information in question text, when mapped into KC-response representations, improves DKT performance.

The final result is negative. Across the pipelines studied here, pretrained text-derived representations do not produce a consistent improvement over the standard DKT baseline. This negative result constrains what semantic similarity can plausibly explain in KT and suggests that the dominant signal in benchmark DKT may be behavioral, temporal, structural, or psychometric rather than purely linguistic. That interpretation is consistent with later KT work that couples question information to additional structural or contextual signals rather than relying on text alone~\cite{ghosh_context-aware_2020, pardos_imputing_2017, wang_seep_2022}.

In this chapter, we make four contributions. First, we formulate a direct semantic hypothesis for DKT and test it under a controlled representation replacement setting. Second, we implement and compare nine pipelines for constructing KC-response embeddings from question text. Third, we evaluate those embeddings both under the standard training setup and under reduced-data training regimes. Fourth, we conduct a post hoc analysis of whether DKT-derived KC graphs align with pretrained KC embedding representations.

The research questions investigated in this chapter are as follows:
\begin{itemize}
    \item \textbf{RQ1.} Do pretrained semantic embeddings improve DKT performance relative to the standard DKT baseline with randomly initialized interaction embeddings?
    \item \textbf{RQ2.} Do semantic embeddings improve data efficiency when the amount of training data is reduced?
    \item \textbf{RQ3.} To what extent do semantic KC representations align with the KC graphs extracted from DKT?
\end{itemize}
The first two questions are about predictive performance. \textbf{RQ3} asks whether semantic KC representations recover the relation structure that DKT appears to use.

The rest of this chapter is structured as follows. We first review the background most relevant to the study, then describe the methodology, report the results, and discuss what those results imply for KT.

\section{Background and Related Work}

\subsection{Knowledge Tracing and DKT}

KT models estimate the probability that a student will answer the next question correctly from the student's prior interaction history~\cite{corbett_knowledge_1995, piech_deep_2015}. In datasets such as ASSISTments, each interaction is linked to a question and one or more KCs, and the model must infer latent mastery over time. BKT does this with an explicit hidden-state probabilistic model, whereas DKT updates a shared latent state from sequences of KC-response interactions with recurrent networks. Because DKT replaces manually specified transition structure with learned sequence representations, it became a natural setting for asking what kinds of relations among questions and KCs are actually useful for prediction.

\begin{figure}[!htbp]
    \centering
    \includegraphics[width=0.95\linewidth,alt={Visualization from the original DKT paper showing predicted probabilities over multiple skills across a student's sequence of exercise attempts.}]{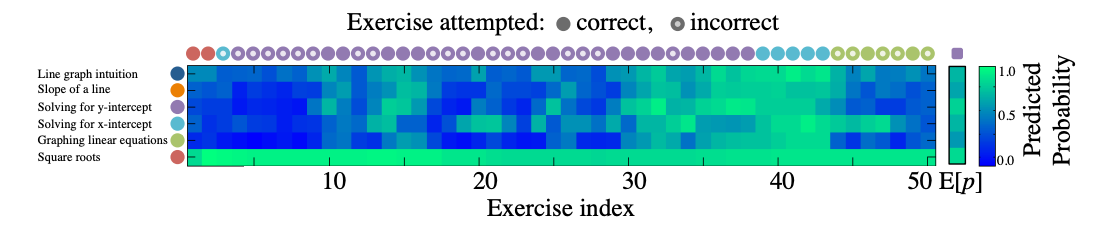}
    \caption{Visualization from the original DKT paper showing predicted probabilities over multiple skills across a student's sequence of exercise attempts~\cite{piech_deep_2015}.}
    \label{fig:chapter5_dkt_original_screengrab}
    \Description{Visualization from the original DKT paper showing predicted probabilities over multiple skills across a student's sequence of exercise attempts.}
    \caption*{\footnotesize Source note and fair-use rationale: Reproduced from Piech et al., \emph{Deep Knowledge Tracing}, Figure~1, arXiv:1506.05908, \url{https://arxiv.org/abs/1506.05908}. The arXiv record links the paper to CC BY-NC-SA 3.0; this dissertation uses the graphic as a fair-use excerpt: one attributed figure for noncommercial scholarly analysis of the DKT model, resized only for layout.}
\end{figure}

The original DKT paper also made this sequence-modeling view visually explicit. Piech et al.~\cite{piech_deep_2015} plotted the model's predicted probabilities over multiple skills across a student's exercise sequence, showing how each interaction updates not only the predicted probability for the current skill but also the broader skill-state profile represented in the recurrent hidden state. Figure~\ref{fig:chapter5_dkt_original_screengrab} reproduces that visualization. It is included here for a narrow scholarly purpose: the chapter is explaining and evaluating the representational claim made by DKT, and the figure is the canonical visual example of that claim. The reuse is limited to one attributed figure from the open-access arXiv version of the paper, is noncommercial and educational, and does not substitute for the original article. The horizontal axis indexes successive exercise attempts, the markers above the heatmap indicate whether each attempted exercise was answered correctly or incorrectly, and each row tracks the model's predicted probability for a different skill over time. The figure shows why DKT became such an influential benchmark: it framed knowledge tracing as the continuous updating of a shared latent state over a sequence of interactions rather than as independent skill-wise updates.

However, later KT research suggests that those useful relationships are not obviously semantic. DKT's gains were followed by critiques showing that model behavior can be unstable or hard to interpret, and that preprocessing and benchmark details can materially affect conclusions~\cite{khajah_deep_2016, xiong_going_2016, yeung_addressing_2018}. More recent architectures often improve by modeling memory, recency, attention, question difficulty, or question-level structure more explicitly~\cite{zhang_dynamic_2017, pandey_self-attentive_2019, sonkar_qdkt_2020, ghosh_context-aware_2020}. That broader literature motivates the present chapter: if semantic structure is truly central to KT, then injecting semantics directly into the input representation should help even without a more elaborate architecture.

\subsection{Pretrained Embeddings as KT Inputs}

The motivation for semantic initialization comes from modern representation learning. Chapter~2 introduced the general embedding background; here the focus is on how those representations enter DKT. Distributional and contextual embedding methods map text into dense vector spaces in which lexical, topical, and semantic relations are partially preserved~\cite{mikolov_distributed_2013, bojanowski_enriching_2017, pennington_glove_2014, devlin_bert_2019, reimers_sentence-bert_2019, gao_simcse_2021}. Sentence-level embedding methods are especially relevant here because the study compares full questions and groups of questions rather than isolated tokens. The representation flow is conceptually simple:
\[
\text{question text} \rightarrow \text{question embedding} \rightarrow \text{KC representation} \rightarrow \text{KT input}.
\]

If semantic structure were aligned with the dependencies useful for KT, then these dense representations could improve generalization, reduce sparsity, and help the model learn effectively with less data. That expectation is consistent with several neighboring strands of work. Exercise-enhanced and exercise-aware KT models incorporate richer representations of questions rather than treating them as anonymous IDs~\cite{su_exercise-enhanced_2018, liu_ekt_2019}. Question-centric KT models and related architectures emphasize that question representation itself is a major axis of model design~\cite{sonkar_qdkt_2020, ghosh_context-aware_2020}. Outside KT proper, educational question representation work such as QuesNet argues that test questions are heterogeneous objects containing both linguistic content and domain logic, which may not be well served by generic off-the-shelf embeddings alone~\cite{yin_quesnet_2019}. More directly, pretraining question embeddings for KT with structural constraints has been shown to help in settings where question--concept relations and difficulty are modeled explicitly~\cite{wang_seep_2022}.

At the same time, the literature also suggests reasons this hypothesis might fail. Pardos and Dadu show that for KC imputation, contextual information from neighboring problems can be as predictive as the problem text itself~\cite{pardos_imputing_2017}. The semantics relevant to educational prediction may therefore be distributed across wording, curriculum structure, and usage context rather than residing only in the text. In ASSISTments-style data, operational KC labels, student behavior, and dataset construction may not align perfectly with pure linguistic similarity.

For that reason, in this chapter, we use semantic initialization as a controlled probe of DKT rather than as a general claim that text should always help. If replacing random KC-response embeddings with pretrained semantic ones improves DKT, that would support the idea that semantic structure contributes directly to the benchmark. If it does not, the result would indicate that semantic initialization alone is insufficient under the present setup.

\section{Methodology}

\subsection{Data Sources}

In this project, we use the ASSISTments 2009, 2012, and 2017 KT benchmarks~\cite{abdelrahmanKnowledgeTracingSurvey2023, fengAddressingAssessmentChallenge2009, priharExploringCommonTrends2022}. Because the question here is whether question text improves DKT, the experiments also rely on the Problem Bodies resource, which provides text for only a subset of ASSISTments items. That partial linkage is the main practical constraint on the study: only questions that can be matched to Problem Bodies text can receive text-derived representations.

Table~\ref{tab:chapter5_pb_data} reports the number of unique questions and KCs in each interaction log together with the corresponding counts in the matched Problem Bodies subset. The key methodological point is that the text-linked subset is much smaller than the full interaction log, especially at the question level, and that filtering constraint shapes every text-based experiment in the chapter.

\begin{table}[!htbp]
    \centering
    \small
    \setlength{\tabcolsep}{6pt}
    \renewcommand{\arraystretch}{1.1}
    \DataTableHeaderRows{1}
    \begin{tabular}{lcccc}
        \toprule
        Dataset & \# Questions & \# Questions in PB & \# KCs & \# KCs in PB \\
        \midrule
        ASSISTments 2009 & 26,688 & 7,836 & 123 & 76 \\
        ASSISTments 2012 & 179,999 & 32,792 & 265 & 162 \\
        ASSISTments 2017 & 3,162 & 638 & 102 & 68 \\
        \bottomrule
    \end{tabular}
    \caption{Question and KC counts in the ASSISTments logs and in the matched Problem Bodies subset used for text-based experiments in Chapter 5.}
    \label{tab:chapter5_pb_data}
\end{table}

This table defines the coverage constraint that applies to every embedding pipeline evaluated here, because each pipeline depends on question text.

\subsection{How Pretrained Embeddings Are Used in DKT}

The baseline we use in this chapter is the DKT implementation provided by the pyKT library~\cite{liuPyKTPythonLibrary2023}. In pyKT, each student interaction is represented as a discrete KC-response state, so the model does not consume a literal one-hot vector directly at the recurrent layer. Instead, it uses an embedding lookup table indexed by the interaction state. If there are \(N\) KCs, then there are \(2N\) possible interaction states corresponding to \((KC,\mathrm{incorrect})\) and \((KC,\mathrm{correct})\). Let \(c_t \in \{1,\dots,N\}\) denote the KC at time \(t\), and let \(r_t \in \{0,1\}\) denote whether the response is incorrect or correct. The interaction index can be written as
\[
i_t = 2(c_t - 1) + r_t,
\]
or equivalently as any indexing scheme that maps each KC-response pair to one of \(2N\) rows in an embedding matrix.

In the standard pyKT DKT setup, this interaction embedding matrix is randomly initialized and learned during training. If \(E \in \mathbb{R}^{2N \times d}\) denotes the embedding table and \(e_{i_t}\) its \(i_t\)-th row, then the recurrent model receives
\[
x_t = e_{i_t}, \qquad E \sim \text{random initialization},
\]
where \(d\) is the embedding dimension. In practice, pyKT realizes this through a learnable embedding lookup table with \(2N\) rows, one for each KC-response state. So the baseline comparison in this chapter is not between a fully hand-engineered one-hot representation and a pretrained representation, but between a randomly initialized learned interaction-embedding table and a pretrained interaction-embedding table with the same role in the model.

Our modification replaces that randomly initialized lookup table with pretrained interaction embeddings constructed by the evaluated pipeline variants. Let \(\tilde{E} \in \mathbb{R}^{2N \times d'}\) denote the pretrained matrix whose rows correspond to the response-specific KC embeddings generated from question text. The modified model then uses
\[
x_t = \tilde{e}_{i_t}, \qquad \tilde{E} = 
\begin{bmatrix}
\tilde{e}_{(1,\mathrm{incorrect})} \\
\tilde{e}_{(1,\mathrm{correct})} \\
\vdots \\
\tilde{e}_{(N,\mathrm{incorrect})} \\
\tilde{e}_{(N,\mathrm{correct})}
\end{bmatrix},
\]
where each row is supplied by one of the nine embedding pipelines described later in the methodology. In the modified implementation, pyKT loads a pretrained matrix from disk when one is provided, adjusts the embedding dimensionality to match that matrix, and initializes the interaction lookup table from those pretrained weights rather than from random values.

\paragraph{Pretrained Embedding Initialization}

Under pretrained embedding initialization, the implementation requires two final vectors for every KC: one corresponding to \((KC,\mathrm{correct})\) and one corresponding to \((KC,\mathrm{incorrect})\). A practical consequence is that the representation pipeline cannot stop at a KC-level semantic vector. It has to map from the text associated with a KC's questions to a pair of response-specific interaction embeddings that can populate the \(2N\)-row DKT interaction table.

\begin{figure}[!htbp]
    \centering
    \begin{subfigure}{0.48\linewidth}
        \centering
        \includegraphics[width=\linewidth]{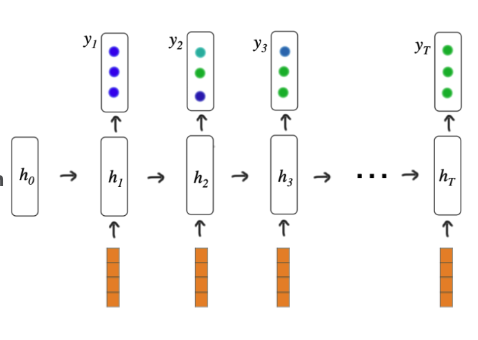}
        \caption{Pretrained embedding initialization}
        \Description{Architecture figure showing the Chapter 5 intervention of replacing the default DKT interaction embeddings with pretrained text-derived KC-response embeddings.}
    \end{subfigure}\hfill
    \begin{subfigure}{0.48\linewidth}
        \centering
        \includegraphics[width=\linewidth]{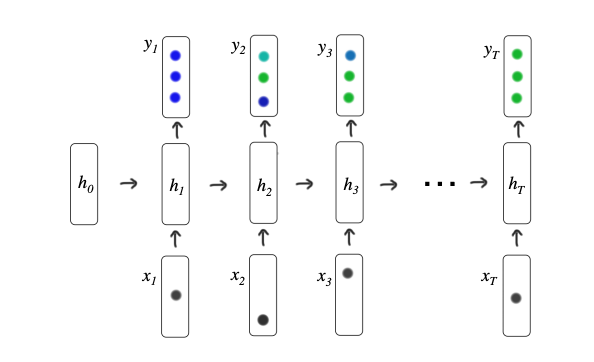}
        \caption{Baseline DKT with randomly initialized interaction embeddings}
        \Description{Architecture figure showing the baseline DKT input setup based on randomly initialized interaction embeddings rather than pretrained text-derived embeddings.}
    \end{subfigure}
    \caption{Pretrained embedding initialization. The panels compare the Chapter~5 text-derived KC-response initialization with the baseline DKT interaction representation based on randomly initialized embeddings.}
    \label{fig:chapter5_pretrained_embedding_initialization}
\end{figure}

If \(i_t\) denotes the KC-response index at time \(t\), then the baseline uses \(x_t = e_{i_t}\) with \(E\) learned from random initialization, whereas the pretrained-initialization variant uses
\[
x_t = \tilde{e}_{i_t}, \qquad \tilde{E} \in \mathbb{R}^{2N \times d},
\]
where \(\tilde{E}\) is populated by one of the nine text-derived pipelines described later in this section. This representation replacement is the shared starting point for all text-informed variants evaluated in the chapter, and the architecture is depicted in Figure~\ref{fig:chapter5_pretrained_embedding_initialization}.

Evaluation follows standard KT practice by focusing on predictive performance metrics such as area under the ROC curve and accuracy.

\subsection{Constructing Text-Informed KC Representations}

Having defined the target object required by the modified DKT implementation, the next question is how those response-specific embeddings are constructed from question text. The central methodological problem is therefore not simply to embed questions, but to map from the text of a KC's associated questions to two final vectors representing \((KC,\mathrm{correct})\) and \((KC,\mathrm{incorrect})\).

\begin{figure}[!htbp]
    \centering
    \includegraphics[width=0.72\linewidth]{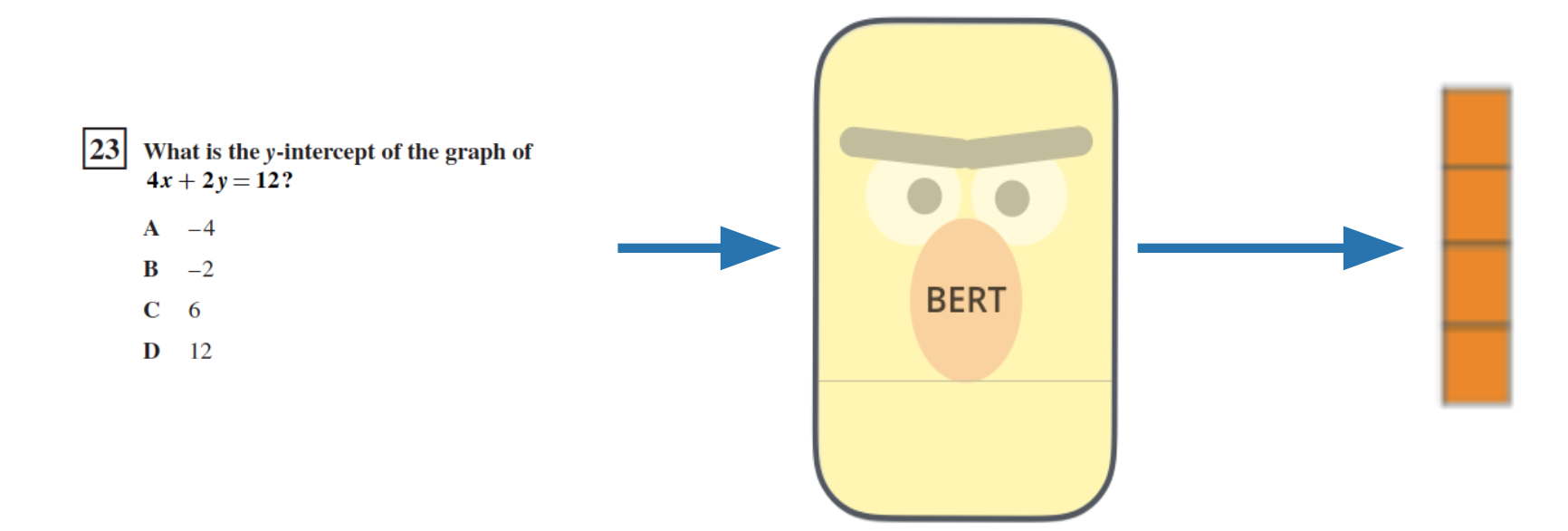}
    \caption{Schematic illustrating the question representation step of the Chapter 5 methodology.}
    \label{fig:text_informed_question_embeddings}
    \Description{Schematic of question embeddings for text-informed knowledge tracing.}
\end{figure}

This construction can be described in terms of three representation steps. The first is the \emph{question representation step}, in which raw question text is mapped into a dense vector representation. The second is the \emph{KC construction step}, in which one or more question-level representations are transformed into a single KC-level vector. The third is the \emph{response-state construction step}, in which the KC representation is converted into the two response-specific interaction embeddings required by DKT. The initial question-to-embedding step is illustrated in Figure~\ref{fig:text_informed_question_embeddings}.

The implementation explores multiple embedding families, including sentence-transformer models, commercial embedding APIs, and dataset-specific fine-tuned BERT variants. The evaluated pipeline implementation instantiates nine concrete pipelines. All nine pipelines produce the two response-specific vectors required by DKT, but they differ in how the question representation step, KC construction step, and response-state construction step are instantiated.

\subsubsection{Implemented Embedding Models}

\begin{table}[!htbp]
    \centering
    \small
    \setlength{\tabcolsep}{5pt}
    \renewcommand{\arraystretch}{1.1}
    \caption{Embedding models implemented in the Chapter~5 text-informed embedding pipeline.}
    \label{tab:chapter5_embedding_models}
    \DataTableHeaderRows{1}
    \begin{tabular}{>{\raggedright\arraybackslash}p{0.23\linewidth}>{\raggedright\arraybackslash}p{0.31\linewidth}>{\raggedright\arraybackslash}p{0.19\linewidth}>{\raggedright\arraybackslash}p{0.13\linewidth}}
        \toprule
        Family & Model & Dimensionality & Dataset scope \\
        \midrule
        OpenAI & \texttt{text-embedding-3-small} & 1536 & all three datasets \\
        OpenAI & \texttt{text-embedding-3-large} & 3072 & all three datasets \\
        OpenAI & \texttt{text-embedding-ada-002} & 1536 & all three datasets \\
        SentenceTransformer & \texttt{all-MiniLM-L6-v2} & 384 & all three datasets \\
        SentenceTransformer & \texttt{all-mpnet-base-v2} & 768 & all three datasets \\
        SentenceTransformer & \texttt{multi-qa-mpnet-base-dot-v1} & 768 & all three datasets \\
        SentenceTransformer & \texttt{all-MiniLM-L12-v2} & 384 & all three datasets \\
        SentenceTransformer & \texttt{paraphrase-MiniLM-L6-v2} & 384 & all three datasets \\
        Cohere & \texttt{embed-v4.0} & 1536 in the experimental configuration & all three datasets \\
        Cohere & \texttt{embed-english-v3.0} & 1024 in the experimental configuration & all three datasets \\
        Cohere & \texttt{embed-multilingual-v3.0} & 384 in the experimental configuration & all three datasets \\
        Fine-tuned BERT & Assist2009 KC classifier & dataset-specific & Assist2009 only \\
        Fine-tuned BERT & Assist2012 KC classifier & dataset-specific & Assist2012 only \\
        \bottomrule
    \end{tabular}
\end{table}

The experimental pipeline implementation is structured so that the same nine pipeline constructions can be evaluated across different embedding backbones, with model-dependent differences such as embedding dimensionality and context-window limits handled inside the pipeline utilities.

\subsubsection{Fine-Tuned BERT Models}

In addition to the off-the-shelf embedding models, we include two dataset-specific fine-tuned BERT models in this chapter: one trained on ASSISTments 2009 and one trained on ASSISTments 2012. These models differ conceptually from the general-purpose encoders in Table~\ref{tab:chapter5_embedding_models} because they are explicitly trained to predict KC labels from problem text rather than relying on general semantic similarity alone.

Fine-tuning is formulated as a multiclass text-classification problem in which the input is the raw question text and the target is the numeric KC index associated with that question. For each dataset, question--KC pairs are extracted from the corresponding \texttt{questions\_with\_kc.csv} mapping file, rows with missing text or missing skill identifiers are dropped, and KC classes with fewer than two examples are removed so that each valid class can appear in both training and validation. The remaining examples are split independently for each dataset into stratified train, validation, and test partitions with a 70/15/15 ratio and random seed 42.

All fine-tuning experiments start from \texttt{bert-base-uncased}~\cite{devlin_bert_2019}. The model is instantiated as \texttt{BertForSequenceClassification} with the number of output labels set to the number of valid KC classes in the dataset and with hidden-state outputs enabled so that the encoder can later be reused for embedding extraction. Training uses AdamW with learning rate \(2 \times 10^{-5}\), weight decay \(0.01\), warmup over the first \(10\%\) of optimization steps, training batch size 16, evaluation batch size 32, maximum sequence length 128, and up to 10 epochs with early stopping based on validation loss using patience 3. Because the KC distributions are highly imbalanced, the loss is class-weighted cross-entropy with weights inversely proportional to class frequency in the training split. Gradient norms are clipped to 1.0, and the checkpoint with the lowest validation loss is kept.

After fine-tuning, the classification head is discarded and embeddings are extracted from the final-layer \texttt{[CLS]} representation, yielding a 768-dimensional vector for each question text. These per-question vectors are then passed through the same nine KC/response construction pipelines used for the off-the-shelf providers, so the BERT-based experiments remain comparable to the rest of the chapter. For ASSISTments 2009 and 2012, KC positions are resolved against each dataset's \texttt{keyid2idx.json} index, and KCs missing from the text mapping are assigned random embeddings drawn from the empirical distribution of the mapped vectors.

The resulting tensors are loaded into DKT as pretrained interaction embeddings. Unlike the frozen semantic-input experiments described earlier, the fine-tuned BERT embedding tables remain trainable during KT optimization, allowing the domain-adapted initialization to continue adapting to the sequential prediction objective. An optional projection layer can still be applied before the recurrent model, which makes the fine-tuned BERT family comparable to the projection-based architecture modifications described later in the chapter. These BERT fine-tuning experiments are limited to ASSISTments 2009 and 2012, and no cross-dataset transfer is performed.

\subsubsection{Implemented Pipeline Families}

The nine implemented pipelines instantiate four experimental design families. In this notation, student action (SA) denotes the response state attached to a KC, either correct or incorrect:
\begin{enumerate}
    \item \textbf{Question $\rightarrow$ KC $\rightarrow$ SA pipelines (Pipelines 1--4):} raw question text is embedded first, KC embeddings are then created either by averaging all questions in a KC or by sampling a single representative question, and finally correctness is added using either a mirror transformation or a concatenated antonym embedding.
    \item \textbf{KC $\rightarrow$ Question $\rightarrow$ SA pipelines (Pipelines 5--6):} each KC is represented directly by concatenating as many of its question texts as will fit in a model-dependent context window and embedding that single stuffed text. Correctness is then added using either the mirror or stack-antonym transformation.
    \item \textbf{SA $\rightarrow$ Question $\rightarrow$ KC pipelines (Pipelines 7--8):} correctness is injected before embedding by prefixing each question text with repeated \texttt{CORRECT} or \texttt{INCORRECT} tokens. The modified questions are then embedded and aggregated into separate correct and incorrect KC vectors using either averaging or single-question sampling.
    \item \textbf{SA $\rightarrow$ KC $\rightarrow$ Question pipeline (Pipeline 9):} correctness is again injected into question text first, but the KC is then built using separate stuffed windows for the correct and incorrect versions of each KC before the modified questions are embedded individually.
\end{enumerate}

Two implementation details matter for interpreting these pipelines. First, the \texttt{mirror} strategy keeps the incorrect KC vector as the base embedding \(e\) and maps the correct version to \(-e\). Second, the \texttt{stack\_antonym} strategy concatenates the KC vector with a separately embedded \texttt{CORRECT} or \texttt{INCORRECT} token, doubling the dimensionality of the final response-specific input. By contrast, the \texttt{add\_words} pipelines do not apply a post-hoc correctness transformation at the KC level; correctness is already embedded in the modified question text itself. In the experiments reported here, those prefixes are repeated twice before embedding.

\begin{table}[!htbp]
    \centering
    \small
    \renewcommand{\arraystretch}{1.15}
    \caption{Nine implemented pipelines for generating response-specific KC embeddings.}
    \label{tab:text_informed_pipelines}
    \DataTableHeaderRows{1}
    \resizebox{\textwidth}{!}{%
    \begin{tabular}{llllll}
        \toprule
        Pipeline & Execution order & Question-text operation & Question representation step & KC construction step & Response-state construction step \\
        \midrule
        P1 & question $\rightarrow$ KC $\rightarrow$ SA & raw question text & embed raw question & average & mirror \\
        P2 & question $\rightarrow$ KC $\rightarrow$ SA & raw question text & embed raw question & average & stack antonym \\
        P3 & question $\rightarrow$ KC $\rightarrow$ SA & raw question text & embed raw question & sample single question & mirror \\
        P4 & question $\rightarrow$ KC $\rightarrow$ SA & raw question text & embed raw question & sample single question & stack antonym \\
        P5 & KC $\rightarrow$ question $\rightarrow$ SA & raw question text & embed stuffed KC text & stuff window & mirror \\
        P6 & KC $\rightarrow$ question $\rightarrow$ SA & raw question text & embed stuffed KC text & stuff window & stack antonym \\
        P7 & SA $\rightarrow$ question $\rightarrow$ KC & add words & embed modified question & average & encoded in text \\
        P8 & SA $\rightarrow$ question $\rightarrow$ KC & add words & embed modified question & sample single question & encoded in text \\
        P9 & SA $\rightarrow$ KC $\rightarrow$ question & add words & embed stuffed modified KC text & stuff window & encoded in text \\
        \bottomrule
    \end{tabular}
    }
\end{table}

The implementation also adapts the context window used by the \texttt{stuff\_window} pipelines to the embedding model. In these experiments, OpenAI embedding models use a larger window than the sentence-transformer variants. This means that the stuffed-window pipelines are not merely averaging more question embeddings; they are re-embedding a single concatenated document intended to represent the full textual scope of a KC within the available context budget. The nine pipelines combine these reusable operations in different ways, as summarized in Table~\ref{tab:text_informed_pipelines}.

\begin{figure}[!htbp]
    \centering
    \begin{subfigure}[t]{0.3\linewidth}
        \centering
        \includegraphics[width=\linewidth,height=0.16\textheight,keepaspectratio]{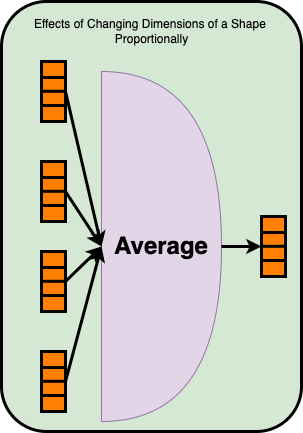}
        \caption{Average}
        \label{fig:concept_average}
    \end{subfigure}\hfill
    \begin{subfigure}[t]{0.3\linewidth}
        \centering
        \includegraphics[width=\linewidth,height=0.16\textheight,keepaspectratio]{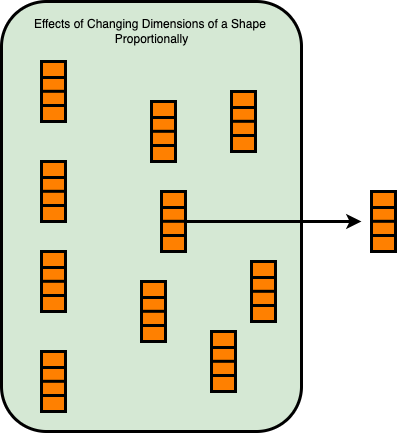}
        \caption{Sample one question}
        \label{fig:concept_sample_single_question}
    \end{subfigure}

    \vspace{0.75em}

    \begin{subfigure}[t]{0.44\linewidth}
        \centering
        \includegraphics[width=\linewidth,height=0.18\textheight,keepaspectratio]{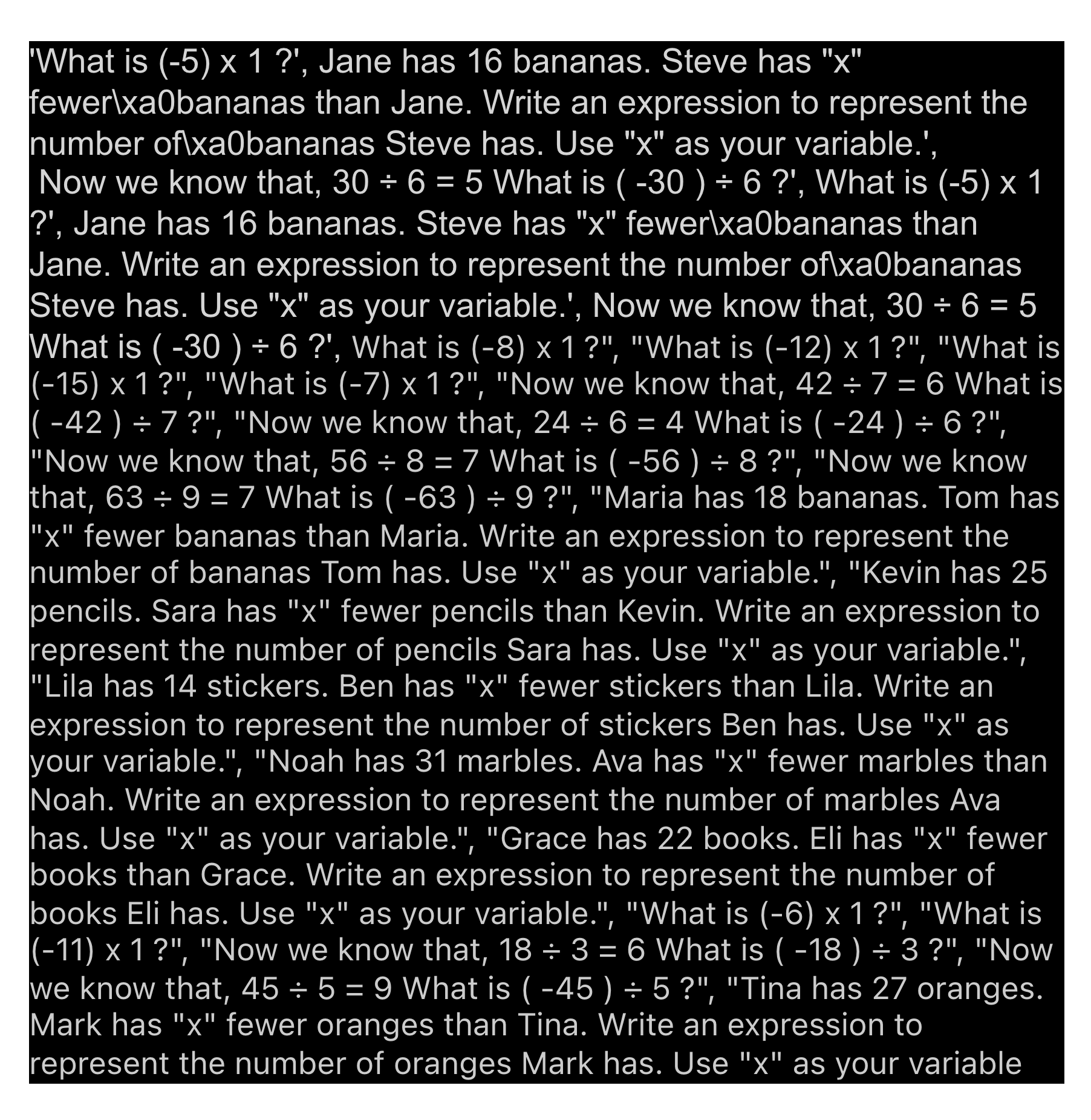}
        \caption{Stuff window}
        \label{fig:concept_stuff_window}
    \end{subfigure}
    \caption{KC-construction operations used in the Chapter~5 text-informed embedding pipelines.}
    \label{fig:chapter5_kc_construction_concepts}
    \Description{Grouped concept figure showing three KC-construction operations: averaging question embeddings, sampling a single representative question, and stuffing multiple question texts into one embedding window.}
\end{figure}

Figure~\ref{fig:concept_average} illustrates averaging, one of the KC-construction operations used in the pipeline family. Under this strategy, multiple question embeddings associated with the same KC are pooled into a single KC representation by arithmetic mean.

Figure~\ref{fig:concept_sample_single_question} illustrates single-question sampling. Instead of pooling all question embeddings within a KC, this strategy selects one representative question and uses its embedding as the KC representation.

Figure~\ref{fig:concept_stuff_window} illustrates the stuff-window strategy. In this case, multiple question texts associated with a KC are concatenated into a single text window and then embedded as one document, subject to the context limits of the embedding model.

\begin{figure}[!htbp]
    \centering
    \begin{subfigure}[t]{0.3\linewidth}
        \centering
        \includegraphics[width=\linewidth,height=0.16\textheight,keepaspectratio]{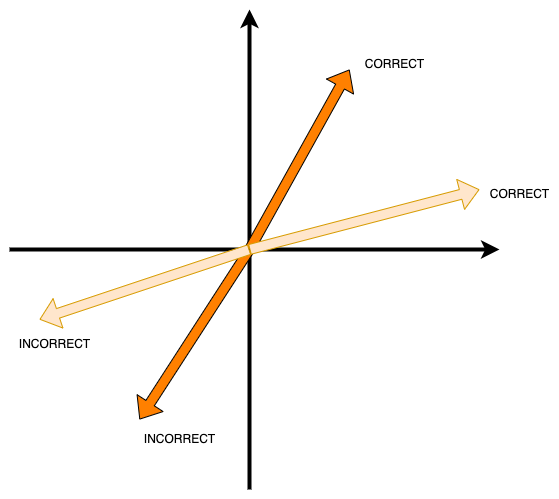}
        \caption{Mirror}
        \label{fig:concept_mirror}
    \end{subfigure}\hfill
    \begin{subfigure}[t]{0.3\linewidth}
        \centering
        \includegraphics[width=\linewidth,height=0.16\textheight,keepaspectratio]{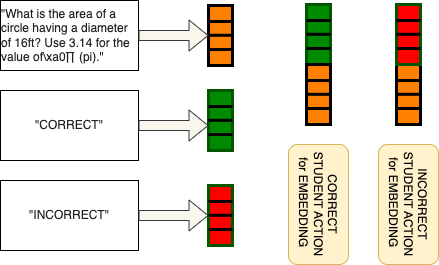}
        \caption{Stack antonym}
        \label{fig:concept_stack_antonym}
    \end{subfigure}

    \vspace{0.75em}

    \begin{subfigure}[t]{0.72\linewidth}
        \centering
        \includegraphics[width=\linewidth,height=0.17\textheight,keepaspectratio]{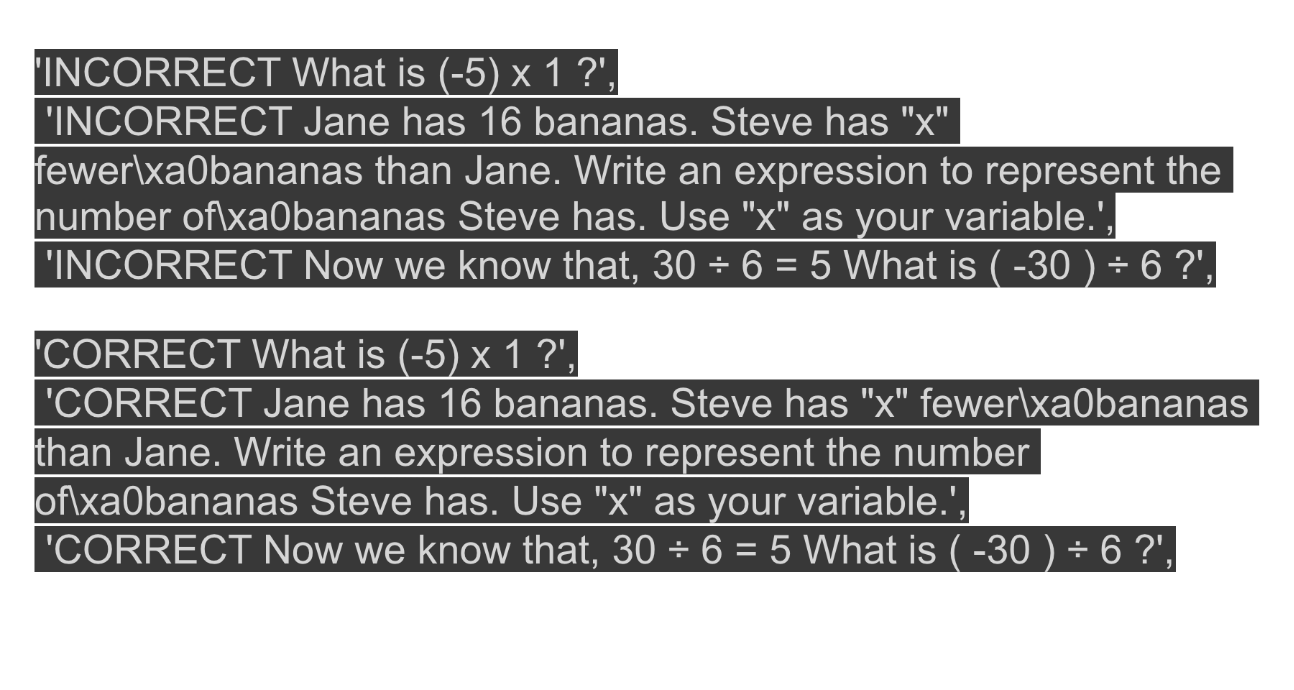}
        \caption{Add words}
        \label{fig:concept_add_words}
    \end{subfigure}
    \caption{Response-state construction operations used in the Chapter~5 text-informed embedding pipelines.}
    \label{fig:chapter5_sa_construction_concepts}
    \Description{Grouped concept figure showing three response-state construction operations: mirroring a KC embedding, concatenating a correctness-specific antonym embedding, and injecting CORRECT or INCORRECT directly into question text.}
\end{figure}

Figure~\ref{fig:concept_mirror} illustrates mirroring. Under this response-state strategy, the incorrect state remains anchored at the base KC representation, while the correct state is mapped to its reflected counterpart.

Figure~\ref{fig:concept_stack_antonym} illustrates stack antonym. In this strategy, a KC embedding is concatenated with a separately embedded correctness marker, thereby producing distinct representations for correct and incorrect interaction states.

Figure~\ref{fig:concept_add_words} illustrates the add-words strategy. Here, correctness is injected into the text itself before any embedding is computed, so the semantic distinction between correct and incorrect states is introduced at the text level rather than as a later vector-space transformation.

The nine pipelines test three distinct hypotheses about how correctness should interact with semantic representations. The \emph{mirror} pipelines assume that correct and incorrect responses should occupy opposite positions around the same KC center. The \emph{stack-antonym} pipelines assume that correctness should be represented by attaching an explicit semantic token to the KC vector. The \emph{add-words} pipelines assume that correctness should change the textual object being embedded before any KC aggregation occurs. This framing clarifies why the negative result reported here is stronger than a failure of any one pipeline family.

\FloatBarrier

\subsection{Architecture Modifications}

Beyond the base replacement of pyKT's randomly initialized interaction embeddings with pretrained text-derived KC-response embeddings, the implemented study evaluated several additional architecture modifications motivated by the same semantic hypothesis. These modifications are reported separately because they change different parts of the DKT computation while keeping pretrained embedding initialization fixed.

\paragraph{Projection Bottleneck}

\begin{figure}[!htbp]
    \centering
    \includegraphics[width=0.72\linewidth]{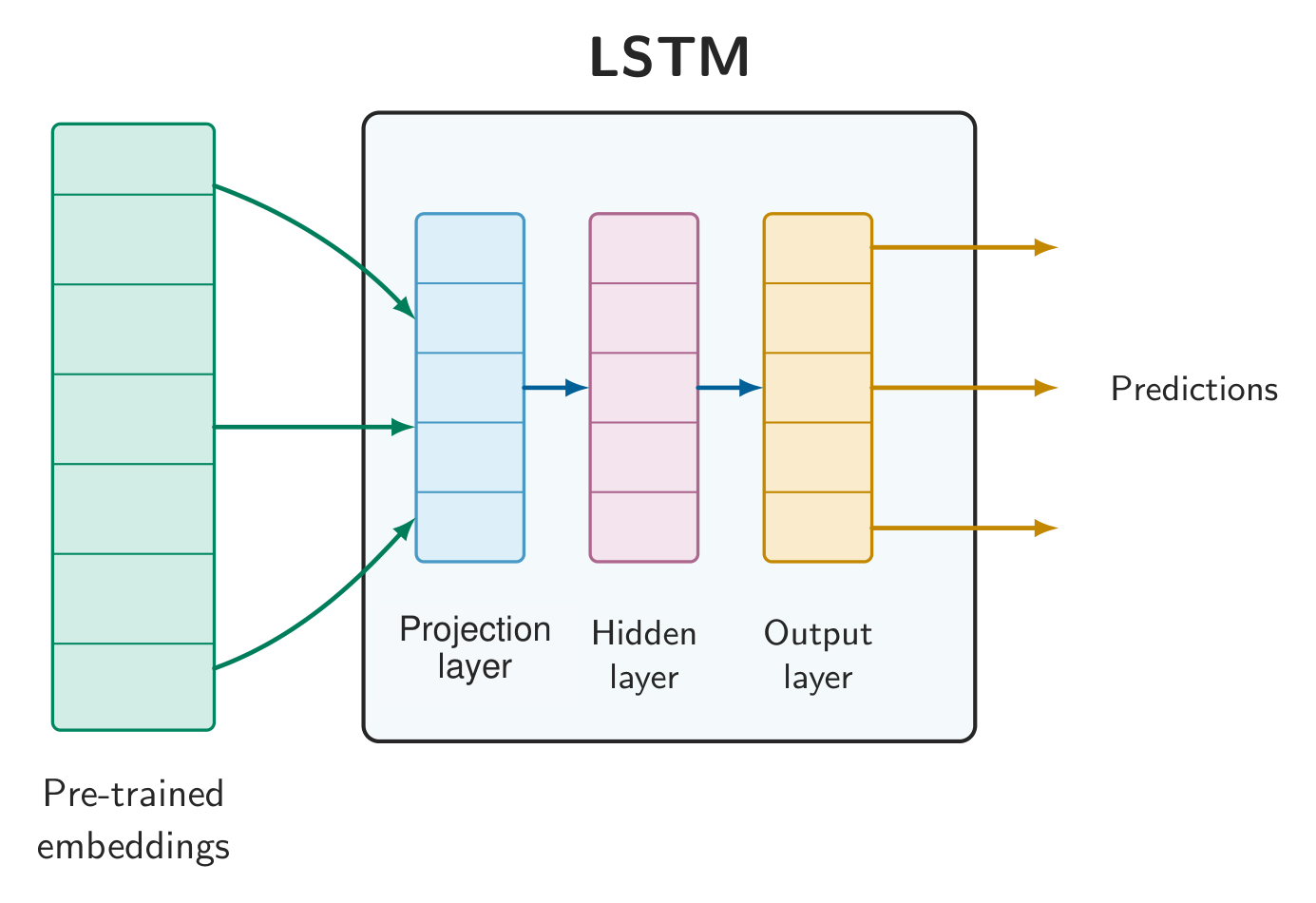}
    \caption{Projection bottleneck applied to the pretrained-input DKT model.}
    \Description{Diagram of an LSTM-based DKT model with pretrained embeddings passed through a learned projection layer before the recurrent hidden and output layers.}
    \label{fig:chapter5_projection_bottleneck}
\end{figure}

The second architecture modification inserts a learned projection layer before the recurrent model. As shown in Figure~\ref{fig:chapter5_projection_bottleneck}, this adds a learned compression stage between the pretrained interaction embedding and the recurrent computation. If \(x_t \in \mathbb{R}^{d}\) denotes the pretrained interaction embedding, the projected representation is
\[
z_t = W_{\mathrm{proj}} x_t + b_{\mathrm{proj}},
\]
where \(W_{\mathrm{proj}} \in \mathbb{R}^{d' \times d}\). In the bottleneck setting, \(d' < d\), so the projection compresses the pretrained representation before sequence modeling. This serves two purposes. First, it adapts high-dimensional pretrained embeddings to the dimensionality expected by the recurrent model. Second, it forces the model to compress the semantic representation into a smaller task-facing subspace, which may help filter irrelevant semantic variation.

\paragraph{Projection Bottleneck with Regularization}

The projection bottleneck was also evaluated together with explicit L2 regularization. This variant keeps the same forward architecture and changes only the training objective. Let \(r_{t+1} \in \{0,1\}\) denote the observed next response and let \(\hat{p}_{t+1}\) denote the model's predicted probability of correctness for that response. The regularized objective is
\[
\mathcal{L}_{\mathrm{reg}}
=
-\sum_{t=1}^{T-1} \Bigl[r_{t+1}\log \hat{p}_{t+1} + (1-r_{t+1})\log(1-\hat{p}_{t+1})\Bigr]
+ \lambda \lVert \Theta \rVert_2^2,
\]
where \(\Theta\) denotes the trainable parameters and \(\lambda\) controls the strength of the L2 penalty. In this chapter, the point of the regularized bottleneck setting is methodological rather than architectural novelty: it tests whether constraining parameter magnitude improves stability or generalization when the model is driven by high-dimensional pretrained inputs.

\paragraph{Adaptive Recurrent Capacity}

\begin{figure}[!htbp]
    \centering
    \includegraphics[width=0.72\linewidth]{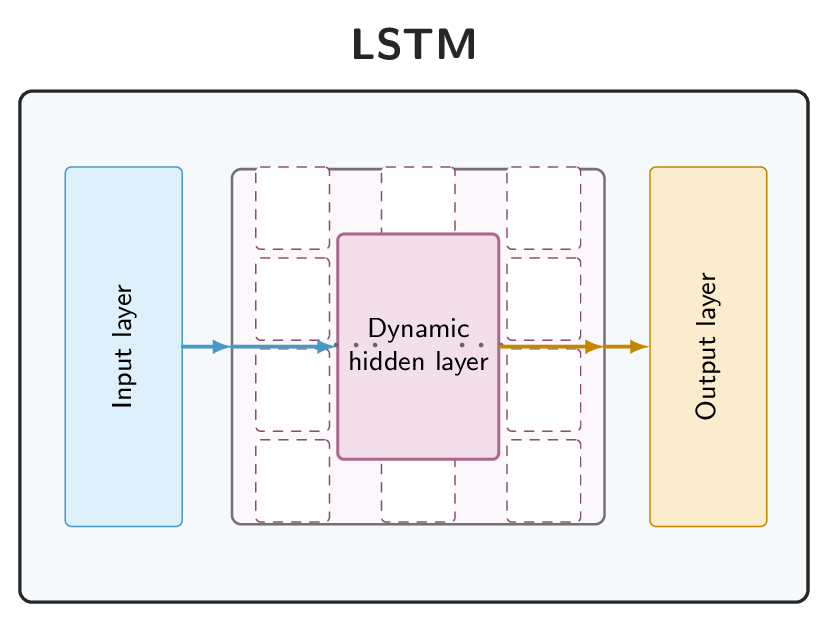}
    \caption{Adaptive recurrent capacity in the pretrained-input DKT model.}
    \Description{Diagram of an LSTM-based DKT model with a dynamic hidden layer size inside the recurrent block.}
    \label{fig:chapter5_adaptive_recurrent_capacity}
\end{figure}

The third architecture modification is adaptive recurrent capacity. Figure~\ref{fig:chapter5_adaptive_recurrent_capacity} highlights that this variant changes the recurrent block itself rather than the semantic input construction. Rather than assuming that the architecture used for sparse ID embeddings remains optimal for pretrained semantic inputs, the experiments allow hidden size, number of recurrent layers, and dropout to vary. In abstract form, the recurrent update becomes
\[
h_t = \mathrm{LSTM}_{\theta}(u_t, h_{t-1}),
\]
where \(u_t = x_t\) in the direct-input case and \(u_t = z_t\) in the projected-input case, while \(\theta\) is selected from a family of recurrent architectures with different hidden widths, depths, and dropout settings. This tests whether semantic inputs require a different recurrent capacity in order to become useful.

\paragraph{Semantic Output Layer}

\begin{figure}[!htbp]
    \centering
    \includegraphics[width=0.72\linewidth]{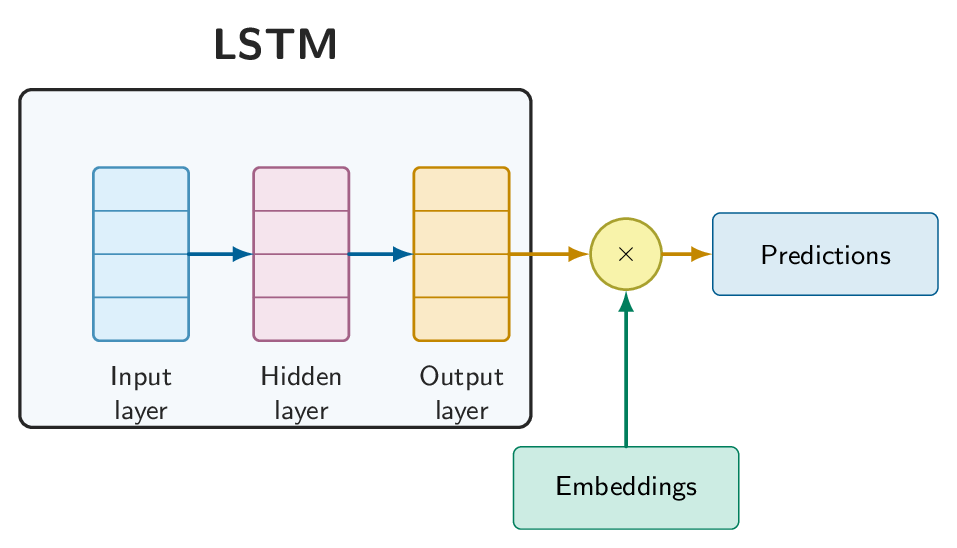}
    \caption{Semantic output layer in the pretrained-input DKT model.}
    \Description{Diagram of an LSTM-based DKT model whose prediction head combines the recurrent output with semantic embeddings before producing predictions.}
    \label{fig:chapter5_semantic_output_layer}
\end{figure}

The fourth architecture modification is the \emph{Semantic Output Layer}. As illustrated in Figure~\ref{fig:chapter5_semantic_output_layer}, this variant changes the prediction head so that the hidden state is scored against semantic KC representations instead of being sent directly to a conventional classifier. In the standard DKT setup, the hidden state is mapped directly to output logits over KCs by a learned linear layer. In the semantic-output variant, the hidden state is first projected into the same space as the projected pretrained KC or question embeddings, and prediction is then based on a dot product between the projected hidden state and those stored semantic representations. Writing \(h_t\) for the recurrent hidden state, one implementation-matched form is
\[
\hat{h}_t = W_h h_t + b_h, \qquad \bar{e}_k = W_{\mathrm{proj}} e_k, \qquad s_{t,k} = \hat{h}_t^\top \bar{e}_k,
\]
where \(e_k\) is the stored pretrained representation associated with KC \(k\), \(W_{\mathrm{proj}}\) is the same projection family used on the input side, and the score is implemented as a dot product. In the implementation variants used here, the stored semantic targets are built from the pretrained interaction table and then projected once into output space; one version averages the correct and incorrect state embeddings for each KC, while another uses a single branch directly. Conceptually, this changes the prediction problem from assigning score mass to a discrete KC index to measuring how well the current student state aligns with a semantic KC representation. This idea is inspired by \emph{weight tying} in NLP language models, where the geometry of the input embedding space is reused at the output layer rather than discarded in favor of an unrelated classifier matrix.

\FloatBarrier

\subsection{Data-Efficiency Protocol}

In addition to the full-data comparison, the methodology includes a reduced-data evaluation designed to test whether semantic representations provide a stronger inductive bias when the training signal is weaker. This part of the study asks whether pretrained text-derived embeddings improve sample efficiency even when they do not improve performance in the full-data setting.

The data-efficiency analysis is carried out on ASSISTments 2009, 2012, and 2017 by creating nested subsamples of each base dataset at \(10\%, 20\%, \dots, 90\%\) of the original row count. The sampling procedure is designed to preserve skill coverage: for each dataset, a mandatory set of rows is first selected so that every KC appearing in the full dataset appears at least once in the sampled dataset, and the remaining rows are then randomly permuted with a fixed seed. The resulting samples are nested, so the \(10\%\) sample is a subset of the \(20\%\) sample, the \(20\%\) sample is a subset of the \(30\%\) sample, and so on. This design makes performance trends across sample size easier to interpret because the larger samples extend the smaller ones rather than replacing them with unrelated draws.

Two conditions are compared on these sampled datasets. The first is the baseline pyKT DKT model without pretrained semantic embeddings. The second is the pretrained condition, in which the model is initialized with a fixed pretrained interaction embedding matrix loaded from one of the pipeline output files. For ASSISTments 2009, all nine pipelines are evaluated in the pretrained condition; for ASSISTments 2012 and 2017, the reduced-data experiments are restricted to the final pipeline variant used in the analysis setup. In all cases, the pretrained matrices remain fixed during training so that the comparison continues to isolate the effect of semantic initialization rather than joint end-to-end adaptation.

Each sampled condition is optimized with a dataset-specific W\&B hyperparameter sweep. Final reporting follows the same validation-selection rule used in the full-data benchmark runs: for each condition, the run with the highest validation AUC is selected, and the selected run's test AUC and test accuracy are reported. For ASSISTments 2009, pipeline selection is nested inside that procedure, so the best pretrained result at each sample size is the pipeline/run combination with the highest validation AUC among the available semantic alternatives. This protocol ensures that model selection is carried out on validation performance rather than on the test set.

\section{Results}

\noindent Table~\ref{tab:chapter5_results_combined} summarizes the strongest validation-selected configuration for each experimental condition in the chapter. Each row reports the associated test AUC and test accuracy, and the grouping labels follow the methodological variants introduced earlier.

\subsection{Main Outcome}

The principal empirical finding of this chapter is negative: pretrained text-derived embeddings do not yield a consistent improvement over the baseline DKT configuration. Although a small number of configurations produce localized gains, those improvements are neither stable across datasets nor robust across pipeline and architecture choices. Accordingly, the evidence does not support the claim that replacing the default interaction embedding table with semantic text representations reliably improves KT performance.

This finding directly answers the first research question in the negative. Under the experimental conditions examined here, pretrained semantic embeddings do not outperform the standard DKT baseline in any general or conclusive sense. This result should not be interpreted as evidence that language is irrelevant to KT. Rather, it indicates that the particular form of semantic signal encoded by these pipelines is insufficient, on its own, to capture the dependencies that most strongly govern DKT predictions.

\begin{table}[!htbp]
    \centering
    \small
    \setlength{\tabcolsep}{3pt}
    \renewcommand{\arraystretch}{1.1}
    \DataTableHeaderRows{1}
    \resizebox{\textwidth}{!}{%
    \begin{tabular}{lp{3.5cm}p{2.6cm}cccccc}
        \toprule
        Dataset & Variant & Embedding & Pipe. & Proj. & SemOut & Test AUC & $\Delta$AUC & Test Acc. \\
        \midrule
        \multicolumn{9}{l}{\textbf{ASSISTments 2009}} \\
        2009 & Baseline DKT & -- & -- & -- & No & \textbf{0.8582} & \textbf{+0.0000} & 0.7851 \\
        2009 & Pretrained Embedding Initialization & text-embedding-3-small & P1 & -- & No & 0.8550 & -0.0031 & 0.7851 \\
        2009 & Pretrained Embedding Initialization + Regularization & text-embedding-3-small & P1 & -- & No & 0.8567 & -0.0015 & 0.7865 \\
        2009 & Projection Bottleneck & text-embedding-3-large & P4 & 256 & No & 0.8552 & -0.0030 & \textbf{0.7870} \\
        2009 & Adaptive recurrent capacity & embed-v4.0 & P6 & 256 & No & 0.8342 & -0.0240 & 0.7708 \\
        2009 & Semantic Output Layer & text-embedding-3-small & P1 & 256 & Yes & 0.8522 & -0.0060 & 0.7811 \\
        2009 & Semantic Output Layer + Projection Bottleneck & bert\_ft\_assist2009 & P6 & 256 & Yes & 0.8349 & -0.0233 & 0.7727 \\
        \midrule
        \multicolumn{9}{l}{\textbf{ASSISTments 2012}} \\
        2012 & Baseline DKT & -- & -- & -- & No & \textbf{0.7370} & \textbf{+0.0000} & 0.7437 \\
        2012 & Pretrained Embedding Initialization & text-embedding-3-large & P2 & 512 & No & 0.7265 & -0.0105 & 0.7344 \\
        2012 & Projection Bottleneck & text-embedding-3-large & P6 & 512 & No & 0.7351 & -0.0019 & 0.7373 \\
        2012 & Projection Bottleneck & embed-v4.0 & P6 & 512 & No & 0.7353 & -0.0016 & 0.7367 \\
        2012 & Adaptive recurrent capacity & text-embedding-3-large & P1 & 48 & No & 0.7317 & -0.0053 & 0.7365 \\
        2012 & Semantic Output Layer & text-embedding-3-large & P1 & 192 & Yes & 0.7356 & -0.0014 & \textbf{0.7376} \\
        2012 & Semantic Output Layer + Projection Bottleneck & text-embedding-3-large & P6 & 256 & Yes & 0.7358 & -0.0011 & 0.7371 \\
        \midrule
        \multicolumn{9}{l}{\textbf{ASSISTments 2017}} \\
        2017 & Baseline DKT & -- & -- & -- & No & \textbf{0.7246} & \textbf{+0.0000} & \textbf{0.6889} \\
        2017 & Pretrained Embedding Initialization & text-embedding-3-small & P3 & -- & No & 0.7207 & -0.0039 & 0.6811 \\
        2017 & Projection Bottleneck & text-embedding-3-small & P4 & 256 & No & 0.7196 & -0.0049 & 0.6808 \\
        2017 & Projection Bottleneck with Regularization & text-embedding-3-small & P6 & 96 & No & 0.7129 & -0.0116 & 0.6758 \\
        2017 & Adaptive recurrent capacity & text-embedding-3-small & P6 & 96 & No & 0.7197 & -0.0049 & 0.6803 \\
        2017 & Semantic Output Layer & text-embedding-3-large & P1 & 192 & Yes & 0.7195 & -0.0051 & 0.6785 \\
        2017 & Semantic Output Layer & text-embedding-3-large & P1 & 256 & Yes & 0.7193 & -0.0053 & 0.6798 \\
        \bottomrule
    \end{tabular}
    }
\caption{Combined Chapter 5 results across datasets. For each experimental condition, the reported row is the validation-selected configuration, together with its test AUC and test accuracy. Boldface indicates the highest value in each dataset block at the reported precision. Across all semantic variants, none exceeded the baseline test AUC on any dataset.}
\label{tab:chapter5_results_combined}
\end{table}

\FloatBarrier

\subsection{Pipeline Comparison}

Appendix Tables~\ref{tab:app_ch5_pipeline_2009}--\ref{tab:app_ch5_pipeline_2017} compare the semantic variants against one another, while the DKT baseline with randomly initialized interaction embeddings remains the external reference point.

Across datasets, no single pipeline emerges as a universal winner. On ASSISTments 2009, Pipeline~1 achieved the highest semantic test AUC (\(0.8567\)) while Pipeline~4 achieved the highest semantic test accuracy (\(0.7870\)). On ASSISTments 2012, Pipelines~2, 3, and~6 tie for the highest reported semantic test AUC (\(0.7358\)), while Pipeline~7 achieved the highest semantic test accuracy (\(0.7382\)). On ASSISTments 2017, Pipeline~3 was the strongest semantic pipeline on both test AUC (\(0.7207\)) and test accuracy (\(0.6811\)). These detailed comparisons sharpen the same conclusion as Table~\ref{tab:chapter5_results_combined}: pipeline choice affects the margins, but none of the semantic pipelines displaces the baseline DKT test AUC in any dataset.

\subsection{Data-Efficiency Results}

Reduced-data evaluation serves as a secondary empirical test of whether semantic initialization helps when the training signal is weaker. Model selection follows the protocol described earlier: for each sampled dataset, the representative run is selected by validation AUC, and the selected run's test AUC and test accuracy are reported. For ASSISTments 2009, this includes a pipeline-selection step across all nine pipelines at each sample size; for ASSISTments 2012 and 2017, the final sampled-data setup uses Pipeline 3 only.

\begin{table}[!htbp]
    \centering
    \small
    \setlength{\tabcolsep}{4pt}
    \renewcommand{\arraystretch}{1.1}
    \DataTableHeaderRows{1}
    \begin{tabular}{lcccccc}
        \toprule
        Dataset & Mean $\Delta$AUC & Mean $\Delta$Acc. & Positive AUC splits & Best split & Best pipeline & Worst split \\
        \midrule
        ASSISTments 2009 & +0.0004 & +0.0000 & 5 / 9 & 40\% & P4 & 20\% \\
        ASSISTments 2012 & +0.0002 & -0.0028 & 6 / 9 & 80\% & P3 & 40\% \\
        ASSISTments 2017 & +0.0096 & +0.0020 & 8 / 9 & 30\% & P3 & 20\% \\
        \bottomrule
    \end{tabular}
    \caption{Dataset-level summary of the data-efficiency experiments. Mean deltas are computed as pretrained minus baseline over the nine sampled splits from \(10\%\) to \(90\%\). Positive AUC splits counts how often the selected pretrained run exceeds the selected baseline run on test AUC.}
    \label{tab:chapter5_data_efficiency_summary}
\end{table}

\begin{figure}[!htbp]
    \centering
    \includegraphics[width=\linewidth]{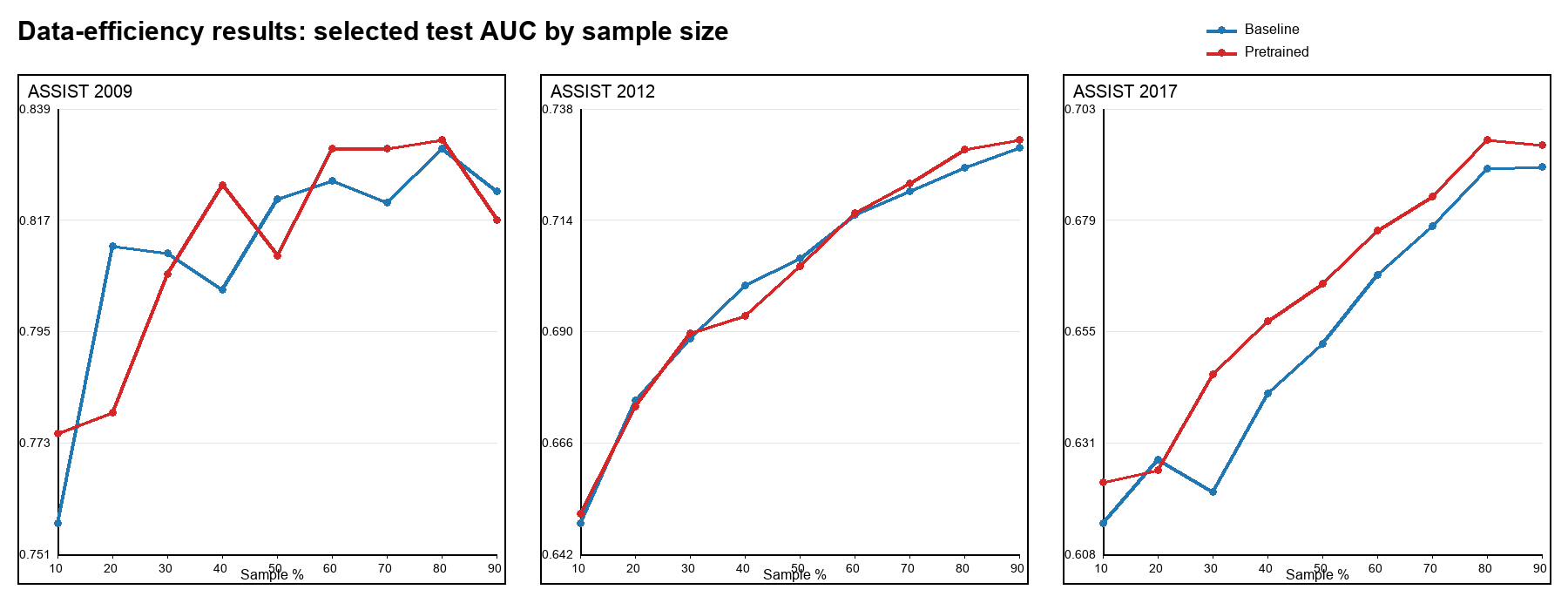}
\caption{Selected test AUC by sample size for the baseline and pretrained data-efficiency experiments. Each curve uses the validation-selected run at each sampled split.}
\label{fig:chapter5_data_efficiency_auc}
\Description{Three-panel figure showing baseline and pretrained selected test AUC curves across sample sizes from 10 percent to 90 percent for ASSISTments 2009, 2012, and 2017.}
\end{figure}

Figure~\ref{fig:chapter5_data_efficiency_auc} reports the selected test AUC curves for the baseline and pretrained settings across sample size. The overall pattern is mixed. ASSISTments 2009 is unstable: the pretrained condition exceeds the baseline on five of the nine sampled splits, but the mean AUC gain is effectively zero and the winning pipeline changes with the sampling level. ASSISTments 2012 is similarly weak, with near-zero mean AUC change and a small decline in mean test accuracy. The most favorable dataset is ASSISTments 2017, for which the pretrained condition improves test AUC on eight of nine sampled splits and yields a mean gain of approximately 0.01 AUC.

\FloatBarrier

ASSISTments 2009 is the only reduced-data setting in which all nine pipelines were evaluated at each sample size. The detailed pipeline-by-pipeline view is provided in Appendix Figure~\ref{fig:app_ch5_data_efficiency_2009_pipelines}. No single pipeline dominates the reduced-data regime across all splits: the best selected pipeline switches among P3, P4, P6, and P8, which reinforces the broader conclusion that the semantic benefit is not robust to reasonable design choices. By contrast, the reduced-data experiments for ASSISTments 2012 and 2017 use Pipeline~3 throughout, so those trends isolate the effect of semantic initialization more narrowly but do not test pipeline diversity in the same way.

These findings provide limited positive evidence, especially for ASSISTments 2017, but they remain insufficient to support a conclusive sample-efficiency claim. They do not justify the broader conclusion that pretrained semantic representations reliably improve DKT when training data are scarce. The ASSISTments 2017 trend indicates that semantic initialization may offer some benefit in lower-data regimes for particular datasets, but that effect is not sufficiently stable across datasets or pipeline choices to overturn the overall negative result of this study.

\subsection{Fine-Tuned BERT Results}

The fine-tuned BERT experiments test a stronger semantic initialization strategy than the generic embedding providers used in the rest of the chapter. Instead of relying on general-purpose text similarity, these embeddings are derived from \texttt{bert-base-uncased} models fine-tuned to predict KC labels directly from question text. If the encoder learns KC-discriminative structure, then the resulting embedding space should in principle be better aligned with the curriculum than off-the-shelf text embeddings. The downstream question is whether that stronger alignment translates into better KT performance once the embeddings are injected into DKT.

The fine-tuning results confirm that the encoder learns dataset-specific KC distinctions with highly variable success across datasets. Table~\ref{tab:chapter5_bert_classifier_results} shows that the classifier performs extremely well on ASSISTments 2009 and 2012, but fails to learn a comparably strong decision boundary on ASSISTments 2017. This performance gap reflects the much more data-constrained 2017 classification setting. As Table~\ref{tab:chapter5_pb_data} already shows, ASSISTments 2017 is the most text-constrained dataset in the chapter, and the filtered classifier split used here is smaller still, with 448 training examples and 96 test examples over 51 KC classes. Under those conditions, weak classifier performance on 2017 is unsurprising.

\begin{table}[!htbp]
    \centering
    \small
    \setlength{\tabcolsep}{5pt}
    \renewcommand{\arraystretch}{1.1}
    \DataTableHeaderRows{1}
    \begin{tabular}{l r r r r r r}
        \toprule
        Dataset & \#KCs & Train & Test & Test Acc. & Macro F1 & Weighted F1 \\
        \midrule
        ASSISTments 2009 & 65 & 4297 & 921 & \textbf{0.9967} & \textbf{0.9665} & \textbf{0.9963} \\
        ASSISTments 2012 & 130 & 13894 & 2978 & 0.9795 & 0.8434 & 0.9772 \\
        ASSISTments 2017 & 51 & 448 & 96 & 0.1354 & 0.1095 & 0.0860 \\
        \bottomrule
    \end{tabular}
\caption{KC-classification performance of the fine-tuned \texttt{bert-base-uncased} models used to generate the Chapter 5 BERT embedding family.}
\label{tab:chapter5_bert_classifier_results}
\end{table}

The ASSISTments 2017 classifier is included in Table~\ref{tab:chapter5_bert_classifier_results} as a diagnostic of this data constraint rather than as a downstream initialization condition. Its small filtered training set and weak held-out performance made it unsuitable for the fine-tuned BERT DKT comparison, so the downstream results in Table~\ref{tab:chapter5_bert_ft_results} are limited to ASSISTments 2009 and 2012.

\begin{table}[!htbp]
    \centering
    \small
    \setlength{\tabcolsep}{5pt}
    \renewcommand{\arraystretch}{1.1}
    \DataTableHeaderRows{1}
    \begin{tabular}{l l l c c c c}
        \toprule
        Dataset & Condition & Winning pipeline & Test AUC & $\Delta$AUC & Test Acc. & $\Delta$Acc. \\
        \midrule
        ASSISTments 2009 & Matched baseline & -- & 0.8552 & -- & \textbf{0.7875} & -- \\
        ASSISTments 2009 & Fine-tuned BERT & P4 & \textbf{0.8581} & +0.0029 & \textbf{0.7875} & +0.0000 \\
        ASSISTments 2012 & Matched baseline & -- & 0.7363 & -- & \textbf{0.7384} & -- \\
        ASSISTments 2012 & Fine-tuned BERT & P4 & \textbf{0.7364} & +0.0001 & 0.7371 & -0.0013 \\
        \bottomrule
    \end{tabular}
\caption{Matched downstream DKT comparison for the fine-tuned BERT embedding family. For each dataset, the baseline and fine-tuned rows are selected by validation AUC within their respective comparison conditions.}
\label{tab:chapter5_bert_ft_results}
\end{table}

\FloatBarrier

Table~\ref{tab:chapter5_bert_ft_results} indicates that the fine-tuned BERT family is more competitive than the generic text embedding providers, but still not strong enough to alter the main conclusion of this study. On ASSISTments 2009, where Table~\ref{tab:chapter5_bert_classifier_results} shows near-perfect KC classification, the best fine-tuned BERT configuration improves test AUC by approximately 0.0029 over its matched baseline while leaving test accuracy unchanged. On ASSISTments 2012, the best fine-tuned BERT run produces only a negligible AUC gain of approximately 0.0001 and a small decrease in test accuracy despite similarly strong classifier performance. In both datasets, the best downstream configuration uses Pipeline 4, which suggests that, within these sweeps, the most effective use of the fine-tuned encoder was to construct KC embeddings from a single representative question and then inject response state through the stack-antonym scheme.

These results support a more nuanced interpretation than the generic embedding experiments alone. Dataset-specific supervision appears to make the semantic initialization more competitive, especially on ASSISTments 2009. However, the gains remain small and inconsistent, and they do not establish a decisive advantage over standard DKT. Thus, improving KC discriminability at the text-encoder level may be beneficial to some degree, but it is not sufficient by itself to produce robust KT gains. The 2017 classifier results reinforce this point in a complementary way: when the text-linked dataset is very small, the model struggles even to learn the KC classification problem itself, which in turn limits the quality of the embeddings available for downstream KT.

\subsection{Post Hoc Analysis: Alignment Between DKT KC Graphs and Pretrained KC Representations}

We return here to \textbf{RQ3}: if semantic initialization does not reliably improve DKT, do the semantic KC representations at least recover the same relation structure that DKT appears to use? The post hoc analysis addresses this question by comparing two structures over the same set of KCs: the KC graph extracted from DKT and the geometry induced by the pretrained KC embeddings. If semantic initialization were beneficial because it encoded the same inter-concept structure that DKT exploits, then the relation graph induced by the embeddings would be expected to align closely with the DKT-derived KC graph.

This analysis follows the relation-discovery motivation introduced in the original DKT work~\cite{piech_deep_2015}, but it uses the KC-graph extraction procedure developed in Chapter 4 rather than the original single-step variant. For each dataset, the extracted KC graph is thresholded upward until it becomes acyclic, so the comparison target is the minimum-threshold DAG recovered from DKT. On the embedding side, the analysis uses the top half of each pretrained interaction tensor and treats those rows as the KC representations. Cosine similarity over these KC vectors produces an embedding-induced relation matrix, which is then compared with the DKT-derived graph using Pearson correlation, Spearman correlation, Mantel correlation, top-\(k\) edge overlap, neighborhood overlap, and node-strength agreement.

Table~\ref{tab:chapter5_posthoc_best_by_dataset} summarizes the strongest dataset-level results in this analysis. The overall pattern is consistent across the post hoc comparisons: alignment exists, but it remains weak across all three datasets. ASSISTments 2012 yields the highest values in the sweep, where \texttt{embed\_english\_v3.0} with Pipeline~3 reaches Pearson correlation \(0.0894\) and top-10\% edge overlap \(0.1571\). ASSISTments 2009 is weaker, with its best run coming from \texttt{bert\_ft\_assist2009} and Pipeline~5 at Pearson \(0.0573\). ASSISTments 2017 is weakest, where the best run, \texttt{all\_MiniLM\_L6\_v2} with Pipeline~5, reaches only Pearson \(0.0247\). These values are all positive, but none of them approach a regime that would justify describing the embedding-derived graphs as strong recoveries of the DKT graph.

The weak alignment in the post hoc comparisons provides a direct explanation for the negative benchmark outcome. If pretrained text-semantic embeddings improved DKT by encoding the same relational structure that DKT exploits, then the embedding-derived KC graphs would have aligned much more closely with the DKT-derived KC graphs. They do not. The tested embeddings therefore do not appear to encode the same KC relation structure that makes DKT useful.

\begin{table}[!htbp]
    \centering
    \small
    \setlength{\tabcolsep}{5pt}
    \renewcommand{\arraystretch}{1.1}
    \DataTableHeaderRows{1}
    \begin{tabular}{l l c c c c}
        \toprule
        Dataset & Best model family & Pipeline & Pearson & Spearman & Top-10\% overlap \\
        \midrule
        ASSISTments 2009 & \texttt{bert\_ft\_assist2009} & 5 & 0.0573 & 0.0276 & 0.1151 \\
        ASSISTments 2012 & \texttt{embed\_english\_v3.0} & 3 & \textbf{0.0894} & \textbf{0.0782} & \textbf{0.1571} \\
        ASSISTments 2017 & \texttt{all\_MiniLM\_L6\_v2} & 5 & 0.0247 & 0.0114 & 0.1404 \\
        \bottomrule
    \end{tabular}
    \caption{Best dataset-level alignment results between semantic KC representations and DKT-derived KC graphs. Each row reports the highest-Pearson result for that dataset under the post hoc analysis setting.}
    \label{tab:chapter5_posthoc_best_by_dataset}
\end{table}

The three DKT-derived target graphs are included in Appendix Figure~\ref{fig:app_ch5_posthoc_kc_graph_triplet}. Their visual differences across datasets already suggest that a single semantic prior would have difficulty reproducing all three graphs uniformly. The ASSISTments 2012 graph is also the one for which the embedding families achieve the strongest, albeit still weak, agreement.

Model comparison shows that some embedding families are systematically less misaligned than others, but only at the margin. The detailed family-level ranking is reported in Appendix Table~\ref{tab:app_ch5_posthoc_family_summary}. The global leader is \texttt{bert\_ft\_assist2012}, followed closely by \texttt{text-embedding-3-large}. This ranking is informative, but its absolute values remain small: the best global family reaches mean Pearson \(0.0321\), which is still far from strong graph agreement.

At the model--dataset level, no family dominates all three datasets. ASSISTments 2009 is led by \texttt{bert\_ft\_assist2009}, ASSISTments 2012 is led by API-style embeddings such as \texttt{embed\_english\_v3.0}, and ASSISTments 2017 is led by a sentence-transformer model. This instability across datasets argues against a single universal semantic encoder explanation for the relation structure used by DKT.

Pipeline comparison leads to a similar conclusion. Appendix Table~\ref{tab:app_ch5_posthoc_pipeline_summary} shows that Pipeline~5 is the strongest global pipeline, followed closely by Pipelines~9 and~3. The gap between pipelines is modest. Pipeline design changes the results incrementally; it does not create a setting in which semantic KC graphs become strong proxies for the DKT-derived graph.

At the dataset--pipeline level, the pattern is also unstable. Pipeline~5 is best for ASSISTments 2009 and ASSISTments 2017 in the broad sweep, whereas ASSISTments 2012 is best under Pipeline~3 in the broad sweep and under Pipeline~4 in the selected-comparison combination summary. This reinforces the same conclusion: pipeline effects are real, but they are dataset-dependent and not strong enough to define a universally successful semantic recipe. The larger pipeline-by-dataset heatmap comparison is provided in Appendix Figure~\ref{fig:app_ch5_posthoc_pipeline_grid}, where the same point becomes visible directly: changing the pipeline alters the cosine-similarity structure, but the resulting relation maps do not converge toward a single DKT-like organization.

The interaction between models and pipelines is most clearly visible in ASSISTments 2012, which is the strongest dataset according to Table~\ref{tab:chapter5_posthoc_best_by_dataset}. ASSISTments 2012 contains the numerically strongest broad-sweep case and the densest cluster of high-performing combinations. It is also the dataset with the largest text-linked question pool in Table~\ref{tab:chapter5_pb_data}, which plausibly gives semantic representations more data from which to form stable KC neighborhoods. Even so, the 2012 combinations still reflect only partial alignment.

Appendix Table~\ref{tab:app_ch5_posthoc_combinations} collects the strongest dataset--embedding-model--pipeline combinations from the selected comparison subset together with the strongest fine-tuned BERT alignment results. The strongest selected-subset combination is ASSISTments 2012 with \texttt{text-embedding-3-large} and Pipeline~4. The BERT follow-up sharpens the interpretation rather than overturning it. On ASSISTments 2009, \texttt{bert\_ft\_assist2009} is the best family-level result in the ranking table, which shows that KC-supervised semantics can help in one dataset. On ASSISTments 2012, however, \texttt{bert\_ft\_assist2012} ranks only 12th among families despite being the strongest family in the selected global average. This contrast shows that stronger supervision can improve semantic alignment modestly while still failing to produce a robust, dataset-general recovery of the DKT graph.

\begin{figure}[!htbp]
    \centering
    \begin{subfigure}{0.48\linewidth}
        \centering
        \includegraphics[width=\linewidth]{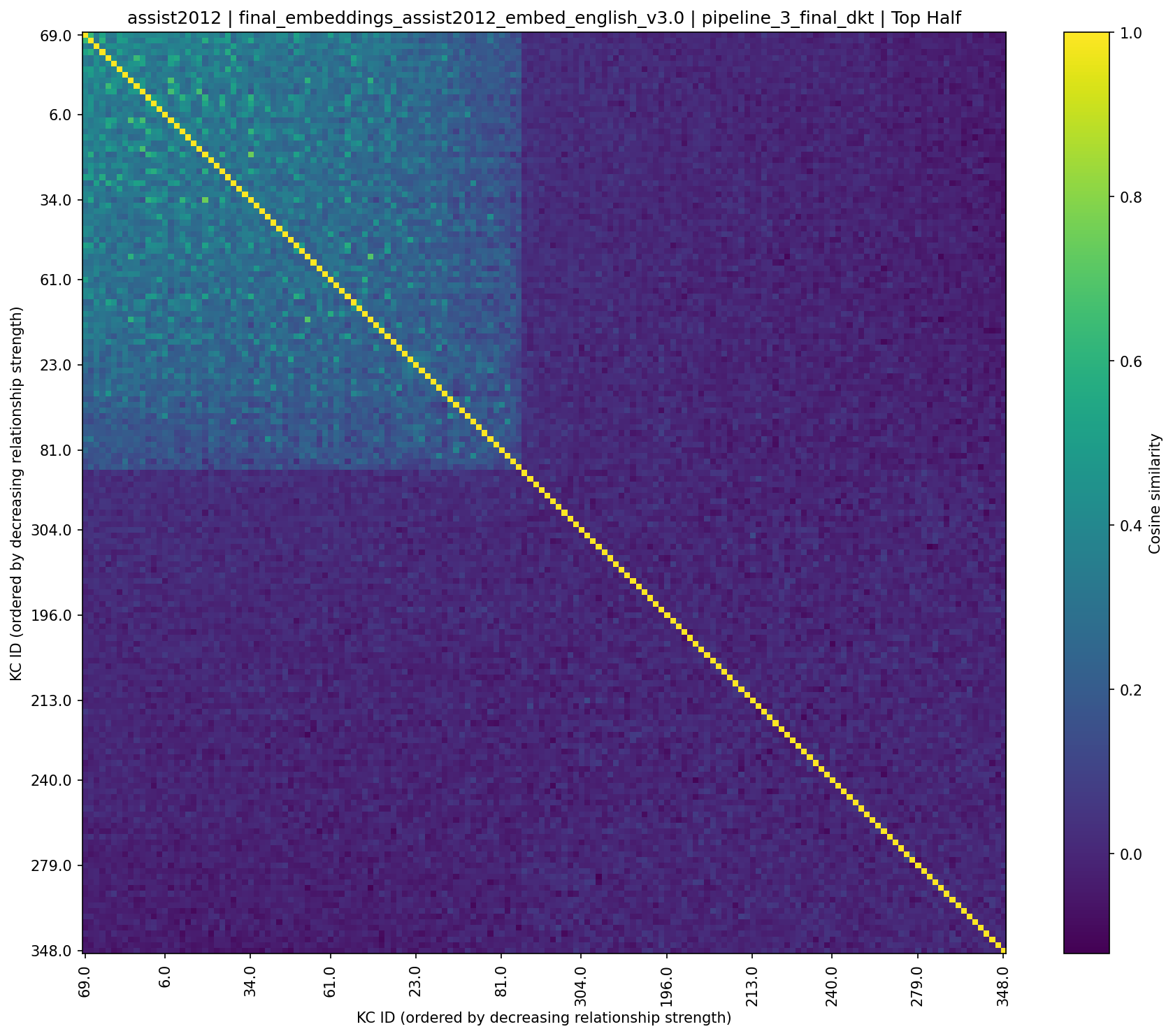}
        \caption{Embedding cosine-similarity matrix}
    \end{subfigure}\hfill
    \begin{subfigure}{0.48\linewidth}
        \centering
        \includegraphics[width=\linewidth]{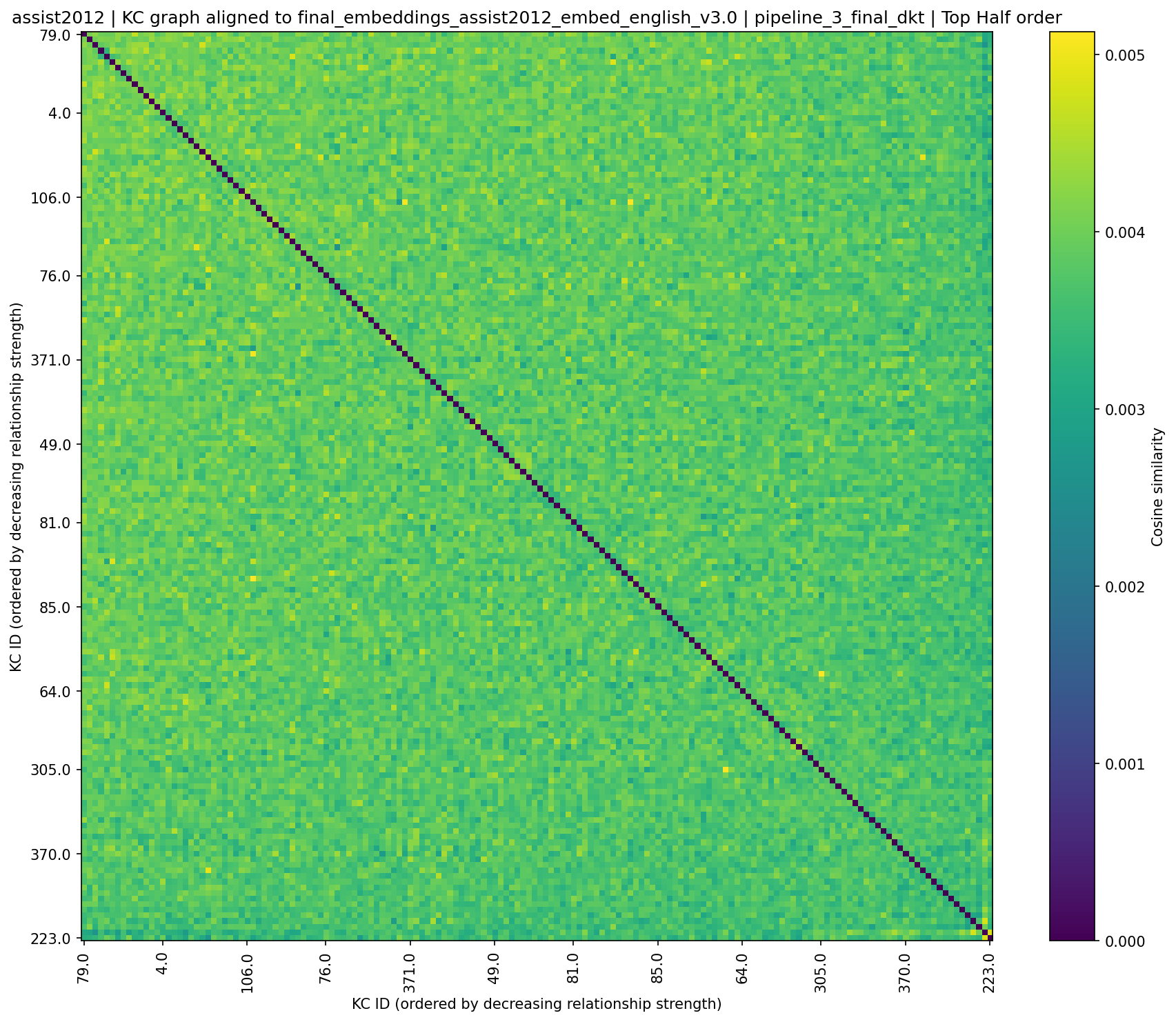}
        \caption{Aligned DKT KC graph}
    \end{subfigure}
    \caption{Best-case heatmap comparison for the strongest broad-sweep alignment result: ASSISTments 2012 with \texttt{embed\_english\_v3.0} and Pipeline~3.}
    \label{fig:chapter5_posthoc_bestcase_heatmaps}
    \Description{Two-panel figure showing the embedding cosine-similarity heatmap and the aligned DKT KC graph heatmap for the strongest broad-sweep post hoc alignment case.}
\end{figure}

Figure~\ref{fig:chapter5_posthoc_bestcase_heatmaps} shows the strongest broad-sweep case directly as two aligned heatmaps: the cosine-similarity matrix induced by the semantic KC representations and the DKT-derived KC graph reordered to the same KC order. This provides a visual comparison at the level of the embedding relations themselves. The strongest case remains ASSISTments 2012 with \texttt{embed\_english\_v3.0} and Pipeline~3, because Table~\ref{tab:chapter5_posthoc_best_by_dataset} identifies it as the numerically strongest broad-sweep result. The semantic heatmap does exhibit structure, but that structure does not closely reproduce the target DKT graph.

\begin{figure}[!htbp]
    \centering
    \begin{subfigure}{0.48\linewidth}
        \centering
        \includegraphics[width=\linewidth]{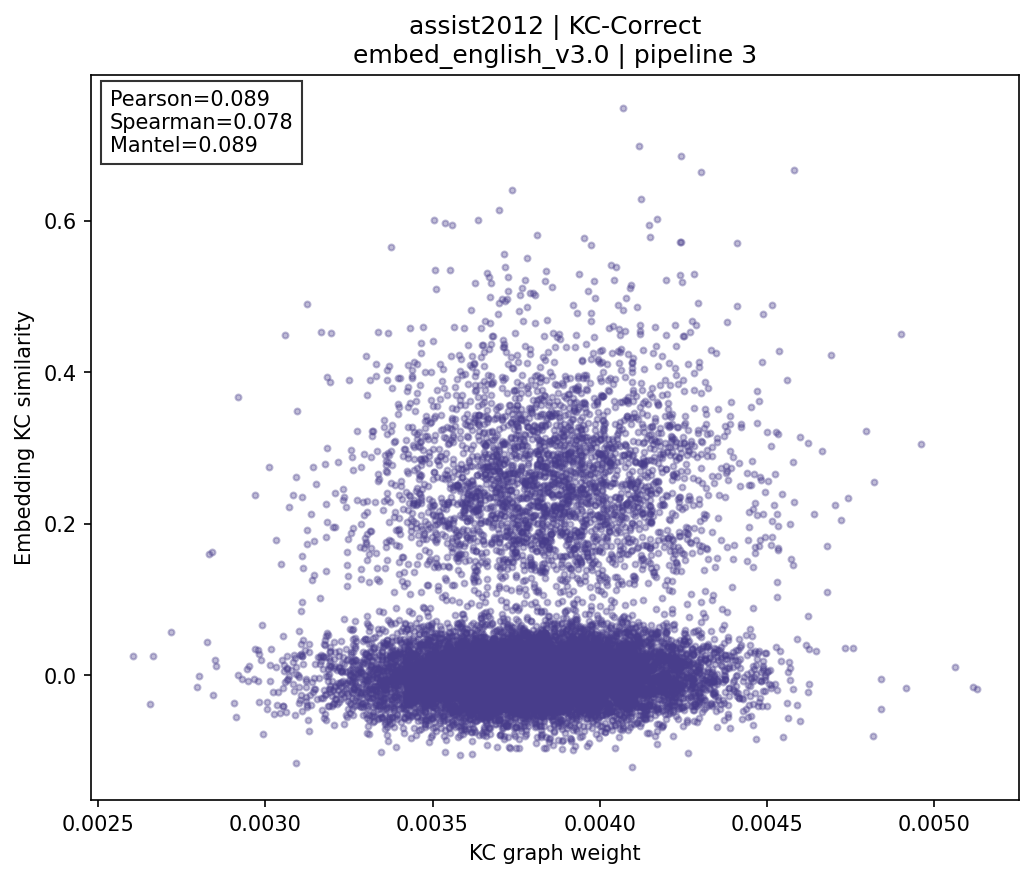}
        \caption{Pairwise edge agreement}
    \end{subfigure}\hfill
    \begin{subfigure}{0.48\linewidth}
        \centering
        \includegraphics[width=\linewidth]{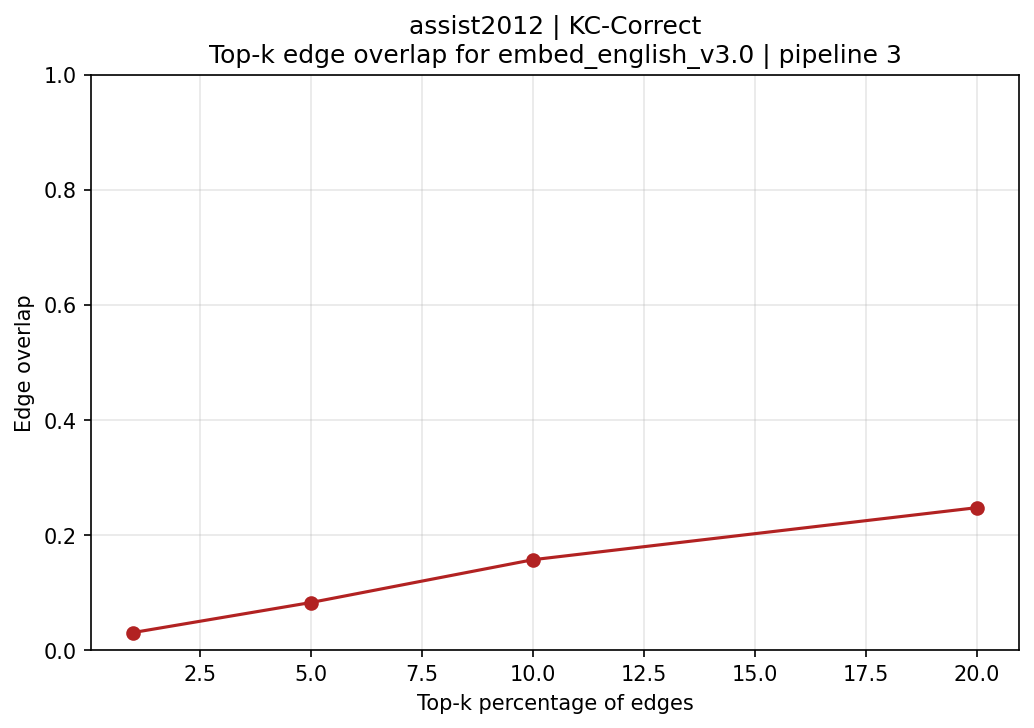}
        \caption{Top-\(k\) edge overlap}
    \end{subfigure}
\caption{Direct agreement diagnostics for the strongest broad-sweep alignment case: ASSISTments 2012 with \texttt{embed\_english\_v3.0} and Pipeline~3.}
\label{fig:chapter5_posthoc_bestcase_alignment}
\Description{Two-panel figure showing scatter-based and top-k-overlap-based agreement diagnostics for the strongest broad-sweep post hoc alignment case.}
\end{figure}

Figure~\ref{fig:chapter5_posthoc_bestcase_alignment} shows the corresponding direct agreement diagnostics for the same strongest broad-sweep case. The scatter plot and top-\(k\) overlap curve make the failure mode visible quantitatively: the trend is positive, but it is diffuse, and the strongest DKT edges are only partially recovered by the semantic graph. This indicates weak or partial alignment rather than strong relational equivalence.

\FloatBarrier

These comparisons answer RQ3 directly. Semantic KC representations align with DKT-derived KC graphs only weakly and inconsistently. ASSISTments 2012 is the strongest case, but still only partially aligned. ASSISTments 2017 is the weakest case, consistent with the tighter text-coverage constraint already shown in Table~\ref{tab:chapter5_pb_data}. ASSISTments 2009 falls between these extremes and shows that moderate text coverage does not guarantee strong semantic recovery. Model choice, pipeline choice, and their interaction all matter, but only at the margin. The overall result therefore supports the interpretation that semantic similarity does not strongly recover the relational structure that DKT uses.

\section{Discussion}

\textbf{RQ1.} Pretrained semantic embeddings do not improve DKT performance in any robust or generalizable sense. Across the full-data benchmark experiments, none of the semantic variants consistently exceeds the baseline DKT model across datasets. Small gains are observed in isolated settings, but they do not survive comparison across datasets, pipelines, and architecture modifications.

\textbf{RQ2.} The reduced-data results provide more limited evidence than the full-data benchmark results, but they are still inconclusive. The signal is strongest on ASSISTments 2017, yet it remains too dataset-dependent and too sensitive to pipeline choice to support a general statement that semantic initialization improves sample efficiency in KT.

\textbf{RQ3.} Semantic KC representations align with the KC graphs extracted from DKT only weakly and inconsistently. A small positive signal is present, especially on ASSISTments 2012, but it remains far too limited to support the claim that semantic KC geometry strongly recovers the graph structure used by DKT.

\subsection{Interpreting the Main Result}

The negative benchmark result is inconsistent with a primarily semantic explanation for DKT performance. If the relations exploited by DKT were primarily semantic, then replacing random interaction embeddings with semantically meaningful text-derived representations should have produced a more reliable benefit. The fact that this does not occur suggests that the dominant structure used by DKT is not well characterized by semantic proximity alone.

This also clarifies the limits of direct transfer from mainstream NLP to educational modeling. In many NLP tasks, semantic similarity is a strong prior. In KT, the relevant structure may depend on response behavior, temporal order, latent prerequisite dependencies, psychometric difficulty, and dataset-specific instructional organization. These factors may correlate with semantics, but they need not be reducible to semantics. The present results therefore suggest that KC relations in KT may occupy a latent learning space: a representation of how different KCs are organized in the learner's knowledge state, only weakly aligned with the semantic geometry induced by text embeddings.

\subsection{What the Limited Positive Signals Mean}

The reduced-data experiments refine this interpretation rather than overturning it. If pretrained semantic embeddings provided a strong inductive bias for KT, their advantage would be expected to emerge more clearly when the training signal is weak. That pattern is not observed consistently. ASSISTments 2009 is unstable, with different pipelines winning at different sample sizes and with an average AUC gain that is effectively zero. ASSISTments 2012 provides even weaker evidence for a semantic advantage, since its mean AUC change is negligible and its mean test accuracy decreases slightly under pretrained initialization.

ASSISTments 2017 is the one dataset that shows a more favorable reduced-data trend, with positive AUC deltas on most sampled splits and a modest mean AUC improvement. This pattern provides limited support for a low-data semantic effect, but it remains insufficient to make the claim conclusive. ASSISTments 2017 is also the most text-constrained setting in Table~\ref{tab:chapter5_pb_data}, so any useful prior carried by the embeddings has more opportunity to matter. ASSISTments 2012, by contrast, has much richer text coverage but also a larger and potentially more heterogeneous KC space, which may dilute the value of a text-only prior.

ASSISTments 2009 falls between these extremes. That intermediate position is consistent with semantic initialization being competitive in isolated settings, including the fine-tuned BERT experiments, yet still failing to produce a decisive full-data advantage. The results therefore indicate that semantic initialization may help in specific low-data regimes while remaining too unstable to support a general data-efficiency claim.

\subsection{Why the Post Hoc Analysis Matters}

The post hoc analysis provides the clearest explanation for why the semantic-input variants do not reliably improve DKT. If semantic initialization were helping because it encoded the same relational structure that DKT exploits, then the semantic KC graphs would be expected to align strongly with the DKT-derived KC graphs. The analysis does not support that expectation. Instead, it finds only weak alignment, even in the strongest cases.

The dataset pattern indicates the same conclusion. ASSISTments 2012 is the strongest case, but still only weakly positive. ASSISTments 2009 is weaker. ASSISTments 2017 is weakest and close to zero by the standards of graph recovery. Model family and pipeline choice can improve the margins, but they do not change the overall conclusion. The best settings improve alignment only modestly; they do not produce the kind of graph agreement that would explain reliable downstream gains.

This dataset ordering is also plausible given Table~\ref{tab:chapter5_pb_data}. ASSISTments 2012 has the richest text coverage, ASSISTments 2017 the weakest, and ASSISTments 2009 falls between them. That pattern does not prove that data volume alone determines graph alignment, but it does suggest that available question text constrains how well semantic KC relations can be estimated.

Dataset size alone is still not a sufficient explanation. Even the strongest runs on ASSISTments 2012 remain only partially aligned with the DKT graph, so the explanation appears structural rather than purely data-limited: semantic relation graphs and DKT-derived KC graphs encode overlapping but genuinely different kinds of information.

This interpretation is reinforced by the model and pipeline comparisons. Model-family and pipeline differences are present, but they improve alignment only modestly rather than producing strong graph recovery. Even the strongest BERT-based cases do not change that basic pattern. The same point helps interpret the fine-tuned BERT experiments: stronger semantic supervision can make the initialization more competitive, but it still does not recover the structure DKT seems to use.

The pipeline grid comparison supports the same conclusion visually. Different pipelines induce visibly different cosine-similarity patterns, so the construction choice is not trivial. However, the grid does not reveal any pipeline whose heatmaps consistently reorganize the KC space into something that resembles the DKT-derived graphs across datasets. The differences are real, but they remain variations within a weak-alignment regime rather than evidence of a successful semantic graph construction strategy.

\subsection{Broader Implications for Knowledge Tracing}

These findings have implications beyond the experiments reported here. The present study sets a boundary condition for attempts to use text more structurally in KT. It shows that simply substituting semantic embeddings into the KT input representation is insufficient.

The benchmark results, the limited low-data signal, and the post hoc graph-alignment analysis all support the same broader lesson: semantics alone is too weak a proxy for the relations that govern DKT performance. A stronger use of text would have to act on the structure of the concept space itself rather than only on the input representation.

\subsection{Limitations}

This study is subject to several limitations that bound the interpretation of the results. First, text coverage is incomplete and uneven across datasets, as shown earlier in Table~\ref{tab:chapter5_pb_data}. The semantic pipelines can be applied only to the text-linked subset, so the effective sample differs from the full interaction log and is especially restrictive on ASSISTments 2017.

Second, the reported comparisons are necessarily conditional on the set of embedding families, pipeline constructions, and architecture modifications evaluated here. We cover a broad space of semantic initialization choices, including fine-tuned BERT variants and several recurrent-model modifications, but we do not exhaust all possible ways of integrating language into KT. In particular, the findings should not be read as ruling out approaches that combine semantic information with richer structural, curricular, or interaction-based signals.

Third, the data-efficiency results, although weakly positive in some settings, remain too dataset-dependent to support a general claim. The positive signal on ASSISTments 2017 is suggestive, but the corresponding effects are weak or unstable on ASSISTments 2009 and 2012. These reduced-data experiments therefore motivate further investigation rather than establishing a settled conclusion.

Fourth, the post hoc graph-alignment analysis depends on the DKT-derived KC graphs produced by the extraction procedure introduced in Chapter 4 and then thresholded to the minimum acyclic DAG. That choice keeps the comparison methodologically consistent, but it also means that the alignment findings should be interpreted relative to that specific graph-construction procedure rather than as statements about every possible notion of KT structure.

These limitations constrain the scope of the chapter's conclusions, but they do not overturn its central result. Under the benchmark conditions studied here, semantic initialization does not reliably improve DKT, and semantic KC representations align with DKT-derived KC graphs only weakly.

\section{Conclusion}

In this chapter, we evaluated a clear hypothesis: whether semantic representations derived from question text improve Deep Knowledge Tracing. The answer to the first research question is negative. Across the benchmark experiments, text-informed KC representations do not reliably outperform the sparse baseline. The answer to the second research question is more tentative: the reduced-data results are positive in some settings, especially ASSISTments 2017, but they remain insufficiently stable to support a conclusive claim that semantic initialization improves data efficiency in KT. The answer to the third research question is that semantic KC representations align with DKT-derived KC graphs only weakly and inconsistently. Even the strongest dataset shows only partial alignment.

This negative result constrains the role that language-based representations are likely to play in KT. Rather than treating semantics as a general solution to sparse educational identifiers, the evidence presented here supports a more precise interpretation: the relations exploited by DKT appear to reside in a learning-relevant latent space that is only weakly aligned with semantic proximity among questions and KCs. The post hoc graph-alignment analysis strengthens this interpretation by showing that the relation graphs induced by the embeddings do not strongly recover the KC graph extracted from DKT itself.

\chapter{Reorganizing Knowledge Components with Text Embeddings}

\section{Introduction}

KT models are usually built on a predefined mapping from questions to knowledge components (KCs)~\cite{corbett_knowledge_1995, piech_deep_2015, abdelrahmanKnowledgeTracingSurvey2023, zhang_dynamic_2017, pandey_self-attentive_2019, ghosh_context-aware_2020, sonkar_qdkt_2020}. That mapping is often treated as part of the dataset rather than part of the model. In practice, however, it shapes what the model is allowed to learn. If the KC partition is too coarse, too fragmented, or poorly aligned with the actual structure of the questions, even a strong KT architecture may be learning over the wrong concept space.

In this chapter, we build directly on the result in Chapter~5. There, we found that semantic initialization did not reliably improve DKT inside the inherited KC space. Here, we ask whether text embeddings are more useful for changing the concept partition than for enriching the original one. The relabeling procedure is simple: represent questions with text embeddings, reduce and cluster those representations, treat each cluster as an embedding-derived KC grouping, relabel the interaction data, and then compare DKT under the revised mapping against DKT under the original manual labels.

This changes the role of text in KT. Instead of using language as an auxiliary feature, we use it to reorganize the question space itself. The goal is to test whether the original KC taxonomy is the right supervision structure in the first place. The study also examines how the embedding-derived groupings relate to the benchmark's CCSS-style labels rather than evaluating the method only through AUC.

This question also matters for ITS decision-making. In an ITS, the model is used to decide what the student should see next: which question to assign, which topic to revisit, or which instructional action to deliver. Those decisions are based on knowledge estimates defined over the KC partition itself, so changing that partition changes the state over which student understanding is aggregated and acted upon. In the original DKT paper, exercise selection is framed as a Markov decision problem, and the reported objective is to choose sequences that increase predicted future knowledge across the course rather than to optimize one isolated KC at a time~\cite{piech_deep_2015}. If a different KC partition changes downstream KT performance materially, then the way the course is decomposed into KCs shapes the state representation used by downstream tutoring decisions rather than serving only as an administrative labeling choice.

We find that embedding-derived KC regrouping can improve downstream KT, but not uniformly across datasets. The mixed pattern appears to depend on dataset properties, preprocessing sensitivity, and the constraints of mapping questions to cluster-derived KCs. In this chapter, we therefore study not only whether the new grouping improves prediction, but also how it agrees with, splits, merges, or recombines the original CCSS-style KCs.

The ASSISTments datasets also differ in size, sequence structure, and preprocessing sensitivity~\cite{fengAddressingAssessmentChallenge2009, priharExploringCommonTrends2022, xiong_going_2016}. In addition, some datasets contain multi-KC structure that does not map cleanly onto one-cluster-per-question relabeling. For that reason, the main issue we study in this chapter is not simply whether clustering can raise AUC, but when KC redefinition helps and what kinds of datasets benefit from it.

The research questions investigated in this chapter are as follows:
\begin{itemize}
    \item \textbf{RQ1.} To what extent can redefining the knowledge-component space through embedding-derived question groupings improve downstream predictive performance in knowledge tracing?
    \item \textbf{RQ2.} To what extent do embedding-derived knowledge-component groupings correspond to, diverge from, or reorganize the human-authored CCSS-based knowledge-component taxonomy?
\end{itemize}

The rest of this chapter is structured as follows. We first review the background most relevant to the study, then describe the methodology, report the results, and discuss what those results imply for KT and ITS.

\section{Background and Related Work}

\subsection{Knowledge Tracing}

Across Bayesian, recurrent, attention-based, graph-based, and question-centric models, KT usually assumes that the question-to-KC mapping supplied with the dataset is adequate to define the concept space over which mastery should be tracked~\cite{corbett_knowledge_1995, piech_deep_2015, zhang_dynamic_2017, pandey_self-attentive_2019, ghosh_context-aware_2020, sonkar_qdkt_2020}. That assumption is easy to miss because it is built into the benchmark, but it is still a modeling choice.

This becomes more important once benchmark sensitivity is taken seriously. Follow-up critiques of DKT showed that results can depend strongly on preprocessing, formatting decisions, and consistency constraints~\cite{khajah_deep_2016, xiong_going_2016, yeung_addressing_2018}. If those choices can change conclusions, then the KC taxonomy should not be treated as a neutral background condition. It is part of what determines what the model can and cannot learn.

\subsection{Knowledge Component Taxonomy}

The idea that concept structure should be tested rather than adopted without further evaluation is already well established in educational data mining. Learning Factors Analysis treats cognitive model improvement as a search problem and uses predictive fit to refine the skill model itself~\cite{cenLearningFactorsAnalysis2006}. Performance Factors Analysis similarly shows that different ways of structuring skills change what the learner model can represent~\cite{jrPerformanceFactorsAnalysis}. Later work such as Knowledge Tracing Machines continues this line by allowing multiple KCs and richer side information, again making it clear that KC representation is part of the model and not merely an administrative labeling choice~\cite{vie_knowledge_2018}.

That literature matters here because it makes KC redefinition a legitimate modeling question. The issue is not only whether a KT architecture is adequate, but whether the concept partition it inherits is too coarse, too fine, or misaligned with the structure that supports prediction. In that sense, KC discovery, KC split/merge, and taxonomy refinement provide the main background for the present study.

This question is especially important when the inherited KCs come from standards-based taxonomies. Common Core State Standards and related curriculum frameworks were designed to define grade-level expectations, support alignment, and coordinate curriculum and assessment systems across jurisdictions~\cite{ccssDevelopmentProcess2010, ccssValidationReport2010, porterCommonCoreStandards2011}. That gives them clear educational value, but it does not mean they were designed to be optimal KT taxonomies. A standards label can be useful for curriculum design and still be a poor fit for sequential prediction.

The broader CCSS literature reinforces that point. Research on implementation repeatedly shows that standards do not operate in isolation; their practical effects depend on curriculum materials, professional learning, assessment systems, and local implementation quality~\cite{heffernanASSISTmentsEcosystem2014, priharExploringCommonTrends2022}. For this chapter, the main implication is that CCSS-style KCs should be treated as one meaningful partition of educational content rather than as unquestioned ground truth about latent concept structure.

Granularity is another central issue. Broad manual KCs can collapse heterogeneous questions into the same label, while very fine-grained representations move toward question-centric KT~\cite{sonkar_qdkt_2020}. The approach we use in this chapter sits between these two extremes. By changing cluster count, it moves from coarse regroupings toward finer partitions of the question space. In the limit, increasing the number of clusters approaches a question-as-KC setting, so cluster count is not just a technical hyperparameter. It is a modeling decision about concept granularity.

\subsection{Text-Aware Educational Modeling and Question Representations}

There is already substantial work showing that question text can carry useful signal for educational modeling. Exercise-enhanced recurrent models use item text to improve prediction and to address cold-start settings where sparse IDs are not enough~\cite{su_exercise-enhanced_2018}. Exercise-aware KT argues more generally that question content should be part of the sequential model rather than discarded~\cite{liu_ekt_2019}. QuesNet pushes this further by treating educational questions as structured objects rather than interchangeable symbols~\cite{yin_quesnet_2019}. More recently, semantic-enhanced pretraining methods have shown that question embeddings can be improved by bringing in side information and structural constraints before downstream KT training~\cite{wang_seep_2022}.

These studies support the idea that question text matters, but most of them still preserve the original KC taxonomy and use text to enrich item or interaction representations. This work uses text differently. The goal here is not just to improve question representations inside an unchanged concept space, but to use question text to change the concept partition itself.

The boundary between content and context still matters. Pardos and Dadu show that for KC imputation, surrounding problem context can be as informative as the problem text and that combining the two improves inference~\cite{pardos_imputing_2017}. That is an important caution for the present chapter. If text-derived KCs fail to align cleanly with the manual taxonomy or help only on some datasets, one plausible reason is that text captures only part of the structure that matters for KT.

\subsection{Text Embedding Clustering}

Chapter~2 introduced the general background on embeddings, PCA, and clustering. Here the focus is only on the parts of that toolkit that matter for regrouping educational questions. Modern text representations range from static embeddings such as word2vec and GloVe to contextual transformer encoders~\cite{mikolov_distributed_2013, pennington_glove_2014, bojanowski_enriching_2017, devlin_bert_2019}. Sentence-level models such as Sentence-BERT and SimCSE are especially relevant when the object being compared is the full question rather than individual words~\cite{reimers_sentence-bert_2019, gao_simcse_2021}. These models make it plausible that semantically related questions will lie near one another in embedding space.

In this chapter, those embeddings are not fed directly into KT. Instead, they are converted into discrete groupings through dimensionality reduction and clustering. This step tests whether an embedding-induced partition of the question space can serve as a better KC taxonomy for DKT.

This use of clustering is consistent with prior work that uses question text and latent structure to organize educational items and concept relations~\cite{pardos_imputing_2017, yin_quesnet_2019, wang_seep_2022}. In machine learning more broadly, clustering is a standard way to convert continuous representation spaces into discrete groups for downstream analysis~\cite{macqueenSomeMethodsClassification1967, lloydLeastSquaresQuantization1982, arthurKmeansPlusPlus2007}. These precedents do not prove that a clustered KC taxonomy will be best, but they support the methodological premise tested here.

The pipeline used here combines principal component analysis with \(k\)-means clustering. PCA provides a transparent reduction of high-dimensional embedding space, and \(k\)-means provides a simple partitioning mechanism that can be studied through cluster-count sweeps and standard diagnostics. This simple design isolates the effect of the concept partition rather than shifting attention to a more elaborate clustering objective.

At the same time, the resulting groups should not be treated as recovered ground truth. They depend on the embedding model, the reduction step, the number of clusters, and the clustering objective. For that reason, this work treats embedding-derived KCs as candidate partitions to be compared against the inherited manual taxonomy rather than as the one correct ontology of mathematical concepts.

\section{Methodology}

\subsection{Data Setting}

In this project, we use three ASSISTments benchmarks that are widely used in KT research: ASSISTments 2009, ASSISTments 2012, and ASSISTments 2017~\cite{fengAddressingAssessmentChallenge2009, priharExploringCommonTrends2022, abdelrahmanKnowledgeTracingSurvey2023}. As in Chapter~5, the analysis is constrained by the availability of question text through the Problem Bodies resource, so the Chapter 6 pipeline operates on the subset of questions for which textual content can be linked reliably to interaction data.

This constraint determines both how many questions can be clustered and how much relabeled supervision is available for DKT. In this study, interactions for questions without linked text are removed rather than kept under a hybrid mapping. The study therefore operates on a text-filtered subset of each benchmark.

The preprocessing code also cleans the question text before embedding. Problem Bodies HTML is parsed with \texttt{BeautifulSoup}, image tags are replaced with their \texttt{alt} text when available, script and style blocks are removed, and the remaining content is converted into normalized plain text. The text-linkage and cleaning step is performed separately for each dataset.

\begin{figure}[!htbp]
    \centering
    \includegraphics[width=0.92\linewidth]{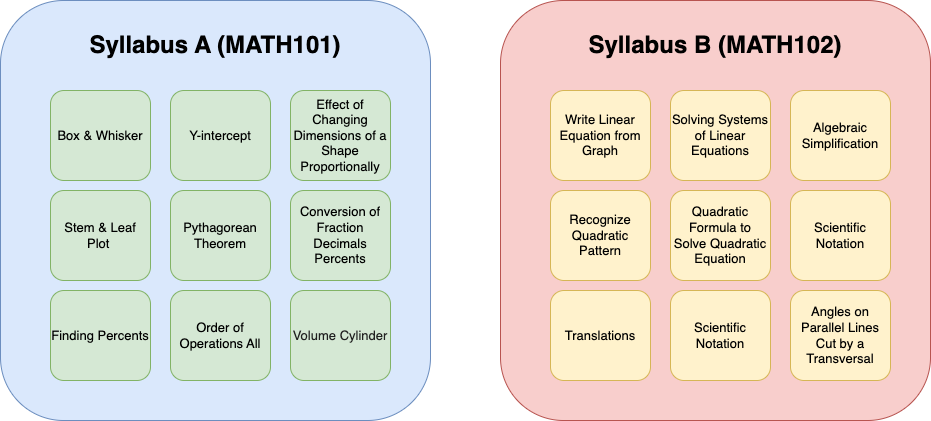}
    \caption{Illustrative example of an inherited syllabus-level organization in which course content is partitioned into manually specified knowledge components.}
    \label{fig:chapter6_manual_kc_structure}
    \Description{Two example syllabi, labeled Math101 and Math102, each containing sets of manually defined knowledge components represented as topic boxes.}
\end{figure}

Figure~\ref{fig:chapter6_manual_kc_structure} gives a simplified picture of the kind of structure that motivates this study. Educational content is usually organized into syllabus- or curriculum-level units, and those units are then operationalized as KCs. That organization is educationally meaningful, but it also means that a manual partition is imposed on the question space before the KT model sees the data.

\subsection{Question Embeddings and Dimensionality Reduction}

Questions in the text-linked subset are represented with OpenAI's \texttt{text-embedding-3-large} model. We fix this model to keep the main study focused on the KC-redefinition pipeline rather than introducing embedding-family choice as a second major experimental variable.

Because these vectors are high-dimensional, we apply principal component analysis (PCA) before clustering. This step is primarily practical. It compresses the embedding space, may reduce redundancy before clustering, and makes centroid-based clustering more manageable in a lower-dimensional representation. The number of PCA dimensions is treated as a search parameter rather than fixed in advance. In the clustering code, PCA is fit separately for each dataset with \texttt{random\_state=42}.

Let \(e_i \in \mathbb{R}^{d}\) denote the embedding of question \(q_i\). After dimensionality reduction, each question is represented as
\[
z_i = P(e_i), \qquad z_i \in \mathbb{R}^{m},
\]
where \(P(\cdot)\) is the PCA projection and \(m\) is the chosen dimensionality.

\subsection{Question Clustering and Embedding-Derived KC Groupings}

The core methodological pipeline is:

\begin{enumerate}
    \item collect question text for the text-linked subset of each dataset;
    \item encode each question with \texttt{text-embedding-3-large};
    \item apply dimensionality reduction using principal component analysis (PCA);
    \item cluster the reduced question representations using \(k\)-means;
    \item treat each cluster as a replacement KC label for the corresponding questions;
    \item relabel the interaction data with these embedding-derived groupings;
    \item train and evaluate DKT on the relabeled datasets.
\end{enumerate}

Clustering is performed with \(k\)-means. Here clustering is run separately for each dataset using \texttt{random\_state=42} and \texttt{n\_init=10}. The number of clusters is treated as a modeling choice, so we evaluate multiple candidate partitions of the question space rather than assuming a single correct KC granularity in advance. As the number of clusters increases, the representation moves toward a more question-centric regime in which each question or small set of questions becomes its own concept-like unit.

Given a chosen cluster count \(K\), the clustering objective is
\[
\min_{\{\mu_k\}_{k=1}^{K}} \sum_{i=1}^{N} \left\lVert z_i - \mu_{c_i} \right\rVert_2^2,
\]
where \(c_i \in \{1,\dots,K\}\) is the cluster assignment for question \(q_i\) and \(\mu_k\) is the centroid of cluster \(k\). Each cluster then becomes an embedding-derived KC grouping.

This design treats the KC taxonomy itself as a variable rather than as a fixed benchmark artifact. The resulting labels are not meant to replace educational interpretation outright. Instead, they define a data-driven concept space that can be compared to the inherited CCSS-style taxonomy both structurally and through downstream KT performance.

\begin{figure}[!htbp]
    \centering
    \includegraphics[width=0.98\linewidth]{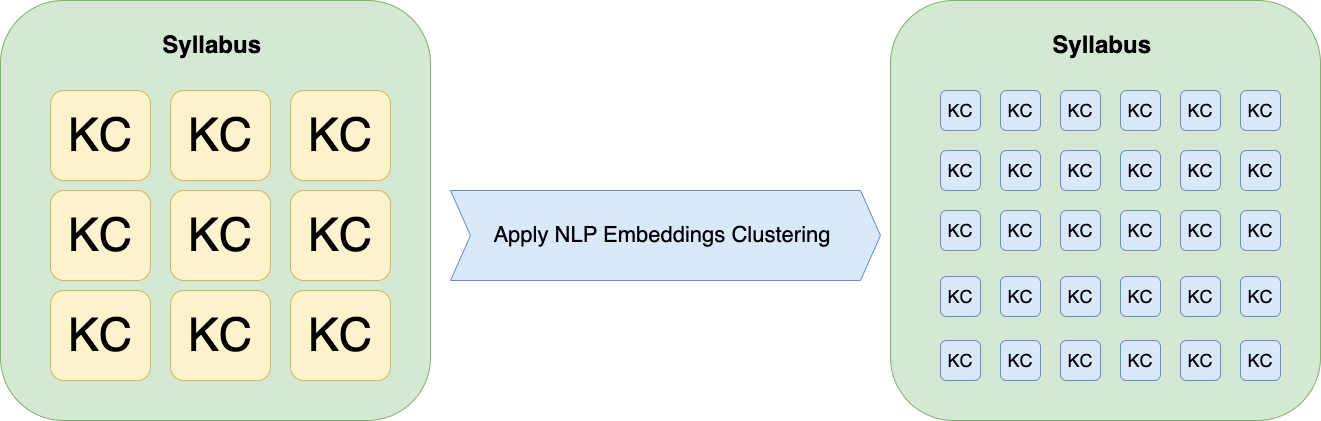}
    \caption{Conceptual schematic of the Chapter 6 pipeline: an inherited manual KC structure is transformed into a finer embedding-derived grouping through text embedding and clustering.}
    \label{fig:chapter6_embedding_clustering_concept}
    \Description{A schematic showing a syllabus with a small number of manual KC blocks on the left, an arrow labeled apply NLP embeddings clustering in the middle, and a syllabus with a larger set of regrouped KC blocks on the right.}
\end{figure}

Figure~\ref{fig:chapter6_embedding_clustering_concept} shows the main methodological change in this study. The goal is not to add text-derived features to the existing KC labels, but to regroup questions by embedding similarity and use those new groupings as an alternative supervision structure for DKT.

\subsection{Relabeling the Interaction Data}

After clustering, each question is assigned the KC label of its cluster. The original CCSS-based KC label is then replaced with this embedding-derived label in the interaction data. The student sequences therefore preserve the same question order, but they no longer preserve the original manual concept partition. Since DKT is trained over KC-response sequences, changing the KC labels changes the supervision structure substantially even when the underlying problem sequence remains the same.

Let a student's original interaction sequence be written as
\[
\mathcal{S}
= \bigl((q_1, s_1, r_1), (q_2, s_2, r_2), \dots, (q_T, s_T, r_T)\bigr),
\]
where \(q_t\) is the question at time \(t\), \(s_t\) is the original CCSS-based KC label attached to that question, and \(r_t \in \{0,1\}\) is the correctness label. In the original benchmark representation, the KC sequence seen by DKT is therefore
\[
(s_1, s_2, \dots, s_T),
\]
or, more precisely in pyKT's default encoding, the corresponding KC-response state sequence
\[
(x_1, x_2, \dots, x_T),
\]
where each \(x_t\) is the discrete interaction state derived from the pair \((s_t, r_t)\).

After clustering, each text-linked question \(q_i\) is assigned a cluster identity \(c_i \in \{1,\dots,K\}\), where \(K\) is the chosen number of clusters. The relabeling map then replaces the original KC label with the cluster-derived label. If an interaction at time \(t\) is associated with question \(q_t\), and if \(s_t\) denotes the original manual KC label, then the relabeling is
\[
s_t^{\prime} = c(q_t),
\]
where \(c(q_t)\) is the cluster assignment of the question and \(s_t^{\prime}\) is the embedding-derived KC label used in the relabeled dataset. The relabeling is therefore not a change to the order of interactions, but a change to the concept identity attached to those interactions.

The relabeled student sequence is therefore
\[
\mathcal{S}^{\prime}
= \bigl((q_1, s_1^{\prime}, r_1), (q_2, s_2^{\prime}, r_2), \dots, (q_T, s_T^{\prime}, r_T)\bigr),
\]
with the same question order and the same response sequence as before, but with a different KC sequence,
\[
(s_1^{\prime}, s_2^{\prime}, \dots, s_T^{\prime}).
\]
Under pyKT's default DKT implementation, this relabeled sequence is again converted into a discrete KC-response state sequence
\[
(x_1^{\prime}, x_2^{\prime}, \dots, x_T^{\prime}),
\]
where each \(x_t^{\prime}\) is now derived from the relabeled pair \((s_t^{\prime}, r_t)\). The temporal order of student interactions is unchanged, but the concept identity attached to each interaction is replaced by an embedding-derived concept assignment.

This relabeling step is the core KC-redefinition operation. The comparison is not between two different KT architectures, but between two different ways of defining what the model should treat as a concept.

\subsection{DKT Training and Evaluation}

The primary comparison we make in this chapter is between two supervision regimes:

\begin{itemize}
    \item the original manual KC labels supplied with the benchmark data;
    \item the embedding-derived KC groupings obtained through clustering.
\end{itemize}

As in Chapters~4 and~5, the downstream KT model is standard DKT implemented through pyKT on top of PyTorch. The experiments were run locally on Apple Silicon hardware using PyTorch's MPS backend. The evaluation reports AUC and accuracy for each dataset and clustering setting.

Unlike Chapter~5, in this chapter, we do not change the input representation or architecture of DKT. The model returns to the default pyKT DKT setup, and the methodological change is only in the KC labels used to construct the interaction sequences. pyKT implements the standard discrete KC-response encoding through a learned embedding lookup table rather than feeding a literal one-hot vector directly into the recurrent layer. That embedding table is randomly initialized and trained with the model. Here, we keep that default architecture and change only the question-to-KC mapping.

The comparison is intentionally simple. The goal is to isolate the effect of changing the concept partition rather than combine KC redefinition with several other architectural changes. DKT is therefore used as the main downstream evaluation model so that the study can answer a focused question: does the revised KC space provide a better basis for sequential prediction than the inherited manual taxonomy?

\subsection{Structural Comparison with Common Core State Standards}

In addition to the downstream DKT experiments, we include a separate analysis comparing the machine-derived KC groupings against the original CCSS-assigned KCs. This comparison supports the second research question by asking whether the discovered grouping resembles the manual taxonomy, diverges from it, or reorganizes it in a qualitatively different way.

The exact structural metrics used in that comparison are reported together with the analysis results. This analysis asks not only whether the relabeled concept space improves KT, but also how that new concept space relates to the human-authored Common Core State Standards taxonomy it replaces.

\section{Results}

\subsection{Main Benchmark Results}

The main empirical question we study in this chapter is whether relabeling questions with embedding-derived KC groupings improves DKT relative to the original manual KC labels. Table~\ref{tab:chapter6_main_results} gives the final benchmark-sweep results, while the structural comparison against CCSS appears in the subsequent structural analysis.

\begin{table}[!htbp]
\centering
\small
\DataTableHeaderRows{1}
\begin{tabular}{lccccc}
\toprule
Dataset & Baseline AUC & Best Clustered AUC & $\Delta$AUC & PCA Dim & Clusters \\
\midrule
ASSISTments 2009 & \textbf{0.8545} & 0.7723 & -0.0822 & 10 & 150 \\
ASSISTments 2012 & 0.7367 & \textbf{0.7841} & +0.0474 & 100 & 300 \\
ASSISTments 2017 & 0.7131 & \textbf{0.7403} & +0.0272 & 10 & 100 \\
\bottomrule
\end{tabular}
\caption{Best DKT results by dataset. The clustered condition replaces the original CCSS-style KC mapping with embedding-derived cluster labels.}
\label{tab:chapter6_main_results}
\end{table}

Table~\ref{tab:chapter6_main_results} summarizes the main cross-dataset result. ASSISTments 2012 shows the largest positive change, with the best clustered configuration improving AUC by 0.0474 over the original-KC baseline. ASSISTments 2017 is a second positive case, with a gain of 0.0272. ASSISTments 2009 remains a negative case: no clustered setting in the evaluated sweep matches the baseline, and the best clustered result remains 0.0822 AUC below the original KC representation.

The dataset-level analyses show whether each gain is broad across the evaluated grid or concentrated in a narrow hyperparameter region. The figures in the main chapter focus on the benchmark sweep and the structural analyses that are central to the argument. Appendix~\ref{sec:appendix_ch6_benchmark} provides additional significance, robustness, convergence, and cluster-granularity diagnostics that support the same overall reading.

\subsection{Hyperparameter Effects}

\paragraph{ASSISTments 2012.}
The best downstream configuration in the main benchmark occurs at PCA dimension 100 and 300 clusters. Every clustered setting in the evaluated grid improves on the original-KC baseline, and the strongest gains come from increasing the number of clusters. Across the sweep, variation in downstream performance is larger along cluster count than along PCA dimension. Performance improves from 50 to 300 clusters across the tested PCA values, while PCA mainly determines whether enough semantic structure survives dimensionality reduction.

\begin{figure}[!htbp]
    \centering
    \includegraphics[width=0.9\linewidth]{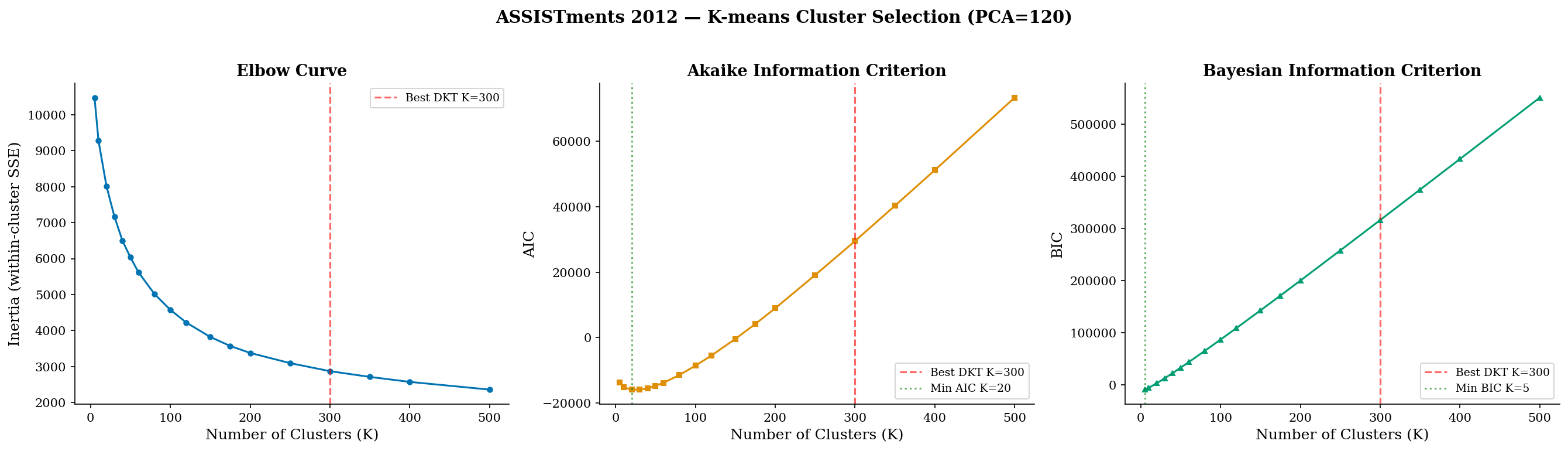}
    \caption{Clustering diagnostics for ASSISTments 2012. The elbow and information-criterion curves mark where geometric compression begins to flatten, but they do not identify the best downstream KT setting on their own.}
    \Description{Clustering diagnostics for ASSISTments 2012. The elbow and information-criterion curves mark where geometric compression begins to flatten, but they do not identify the best downstream KT setting on their own.}
    \label{fig:chapter6_assist2012_elbow}
\end{figure}

Figure~\ref{fig:chapter6_assist2012_elbow} shows that the information-criterion minima occur at smaller \(K\) values than the best downstream DKT setting, so unsupervised compactness alone does not select the most effective KT partition.

\begin{figure}[!htbp]
    \centering
    \includegraphics[width=0.86\linewidth]{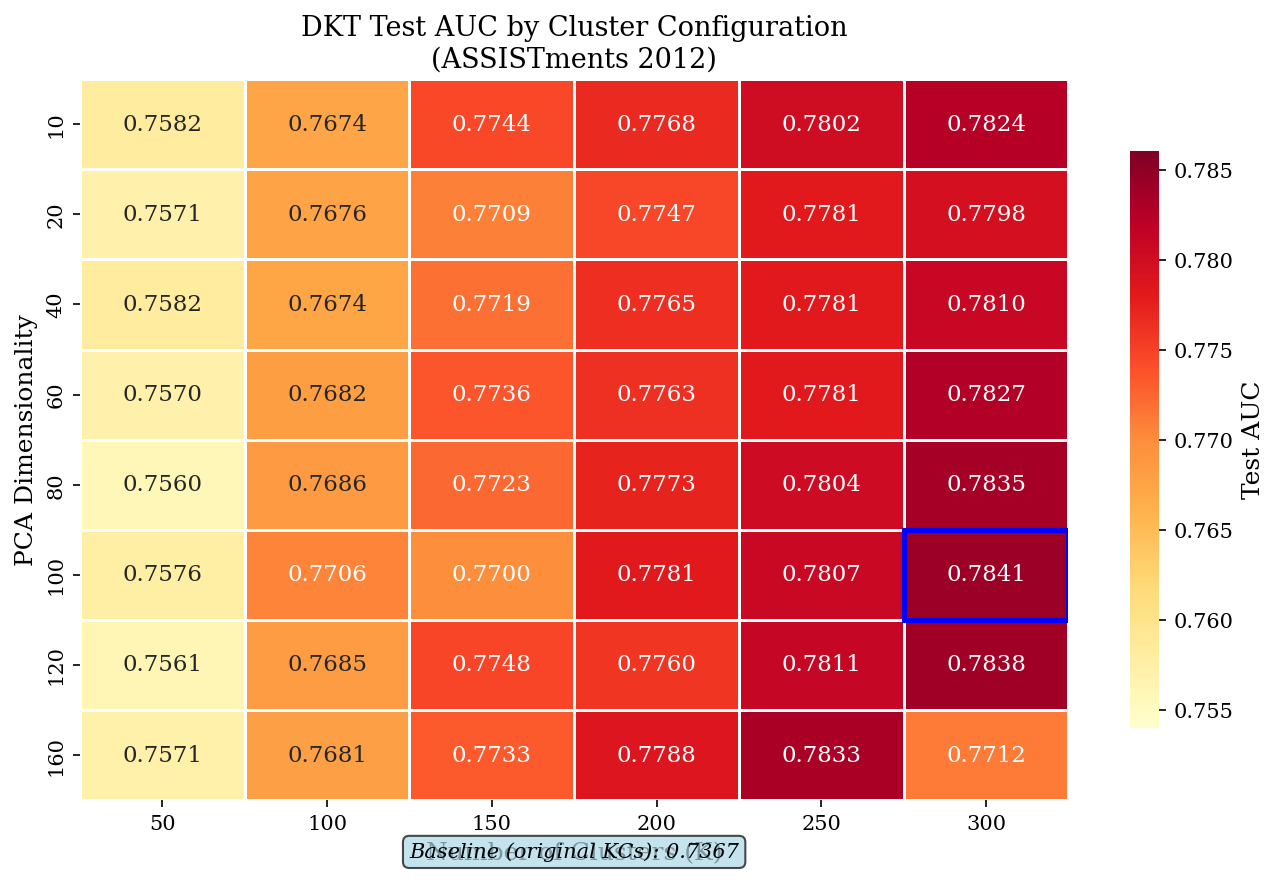}
    \caption{Validation-AUC heatmap for ASSISTments 2012 across PCA dimensionality and number of clusters. The gains remain positive across the tested grid and are largest at higher cluster counts.}
    \Description{Validation-AUC heatmap for ASSISTments 2012 across PCA dimensionality and number of clusters. The gains remain positive across the tested grid and are largest at higher cluster counts.}
    \label{fig:chapter6_assist2012_auc_heatmap}
\end{figure}

Figure~\ref{fig:chapter6_assist2012_auc_heatmap} shows that all tested clustered settings exceed the original-KC baseline and that the best region spans multiple PCA settings at higher cluster counts.

\begin{figure}[!htbp]
    \centering
    \includegraphics[width=0.86\linewidth]{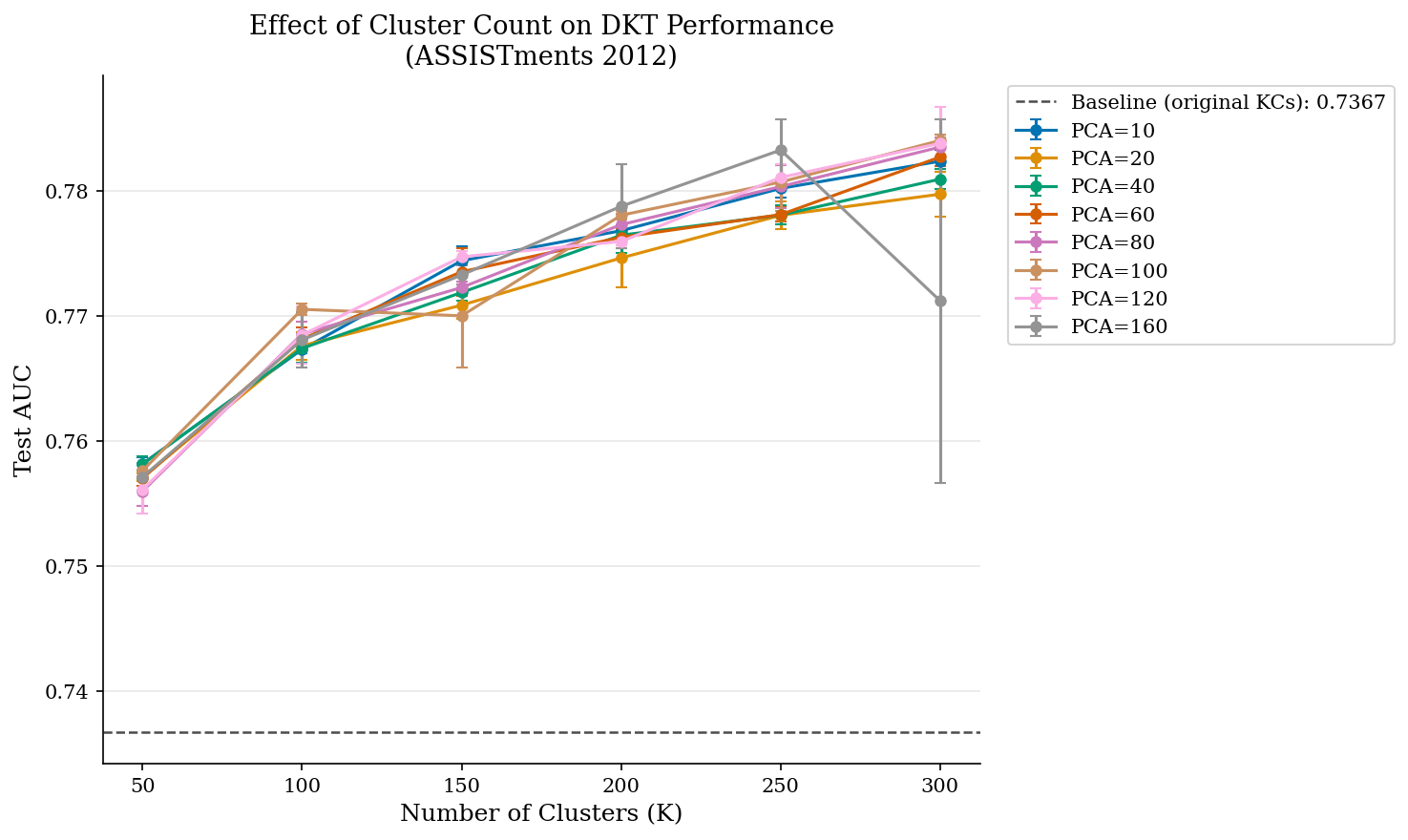}
    \caption{Effect of number of clusters on ASSISTments 2012. Increasing cluster count consistently improves downstream performance over the tested range, and the best results occur at the highest tested cluster counts.}
    \Description{Effect of number of clusters on ASSISTments 2012. Increasing cluster count consistently improves downstream performance over the tested range, and the best results occur at the highest tested cluster counts.}
    \label{fig:chapter6_assist2012_effect_k}
\end{figure}

Figure~\ref{fig:chapter6_assist2012_effect_k} makes the dominant role of cluster count explicit: the curves rise as \(K\) increases across the tested range.

\begin{figure}[!htbp]
    \centering
    \includegraphics[width=0.86\linewidth]{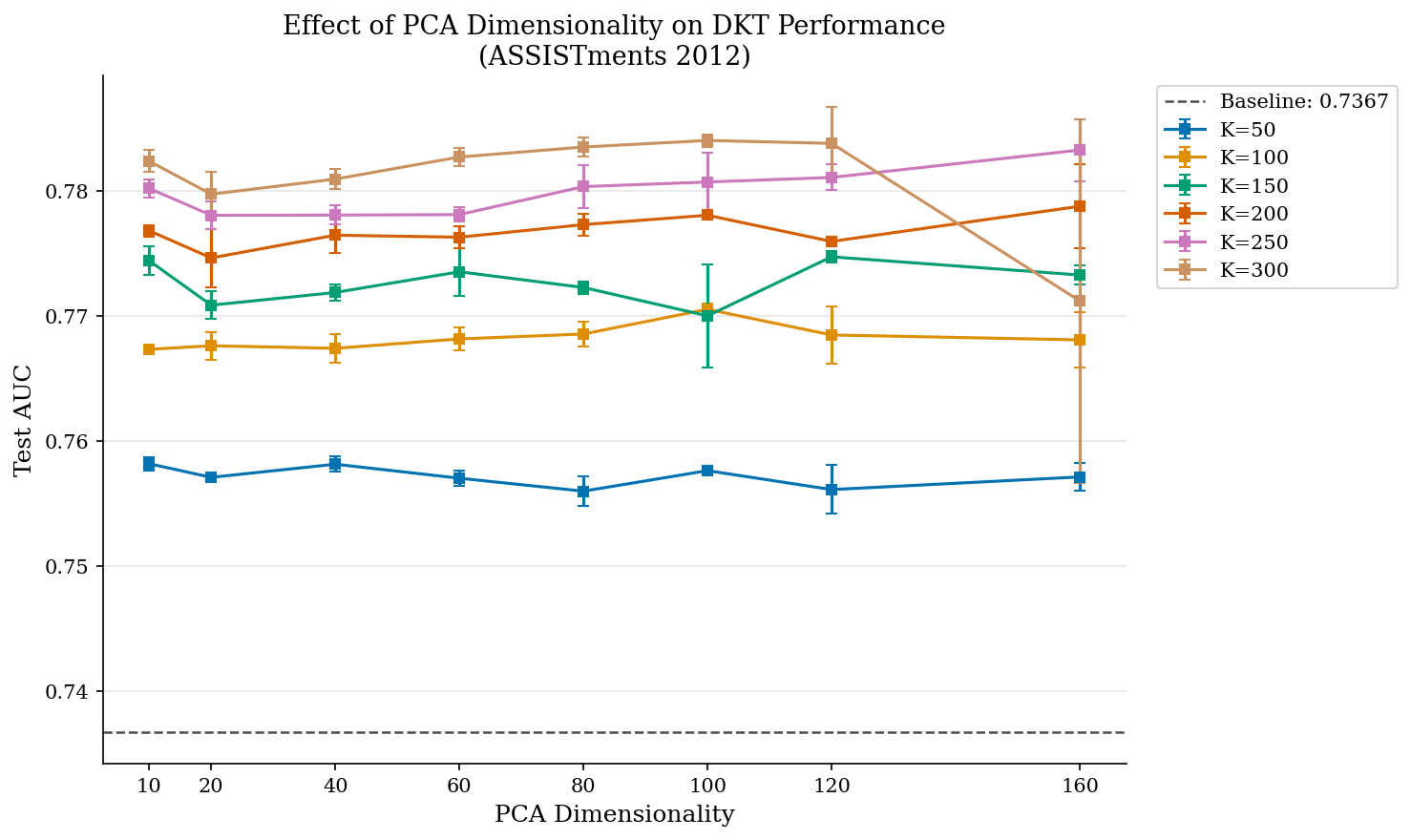}
    \caption{Effect of PCA dimensionality on ASSISTments 2012. Very low-dimensional projections remain competitive, but the strongest results occur in a moderate-dimensional range rather than at the smallest tested projection.}
    \Description{Effect of PCA dimensionality on ASSISTments 2012. Very low-dimensional projections remain competitive, but the strongest results occur in a moderate-dimensional range rather than at the smallest tested projection.}
    \label{fig:chapter6_assist2012_effect_pca}
\end{figure}

Figure~\ref{fig:chapter6_assist2012_effect_pca} shows that PCA dimensionality matters less than cluster count and that the strongest region lies in a moderate-dimensional band rather than at the smallest tested projection.

\begin{figure}[!htbp]
    \centering
    \includegraphics[width=0.86\linewidth]{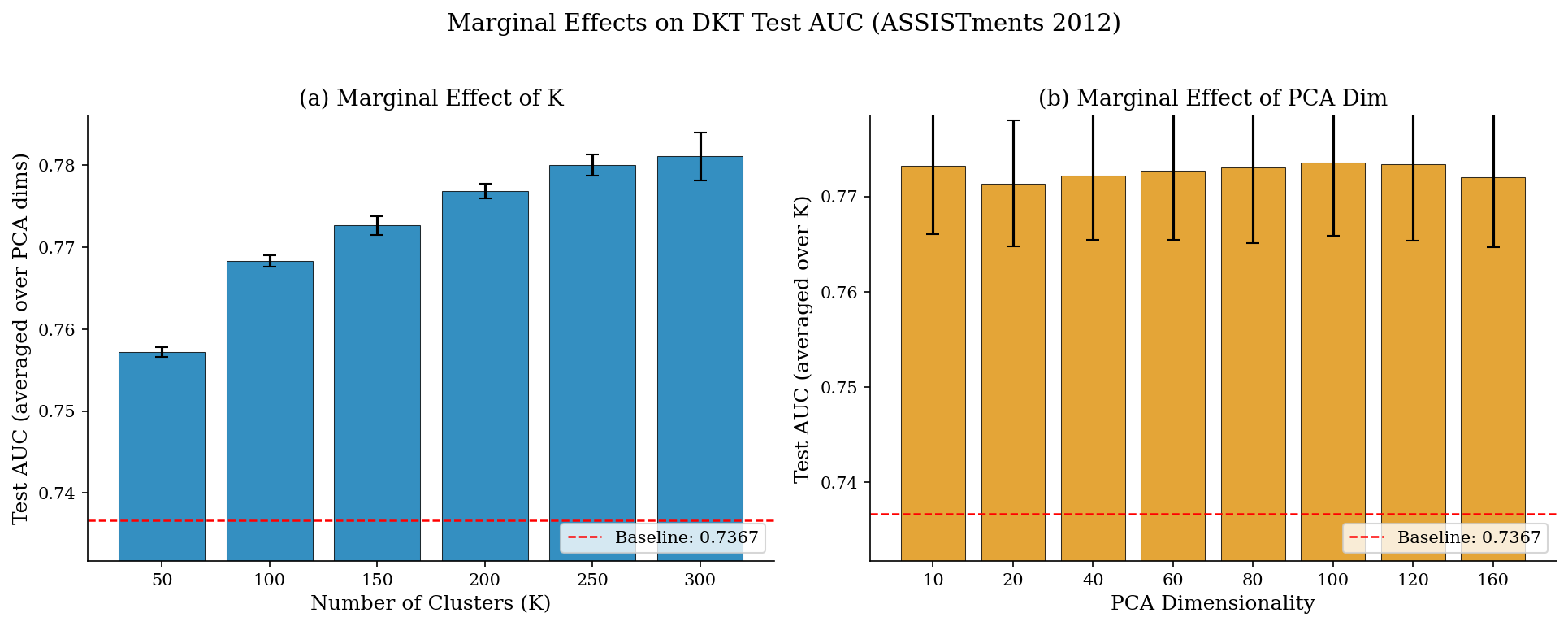}
    \caption{Marginal effects on ASSISTments 2012. Averaged across the other variable, cluster count has the larger effect on downstream performance, while PCA dimensionality mainly changes performance within a narrower band.}
    \Description{Marginal effects on ASSISTments 2012. Averaged across the other variable, cluster count has the larger effect on downstream performance, while PCA dimensionality mainly changes performance within a narrower band.}
    \label{fig:chapter6_assist2012_marginal_effects}
\end{figure}

\begin{figure}[!htbp]
    \centering
    \includegraphics[width=0.86\linewidth]{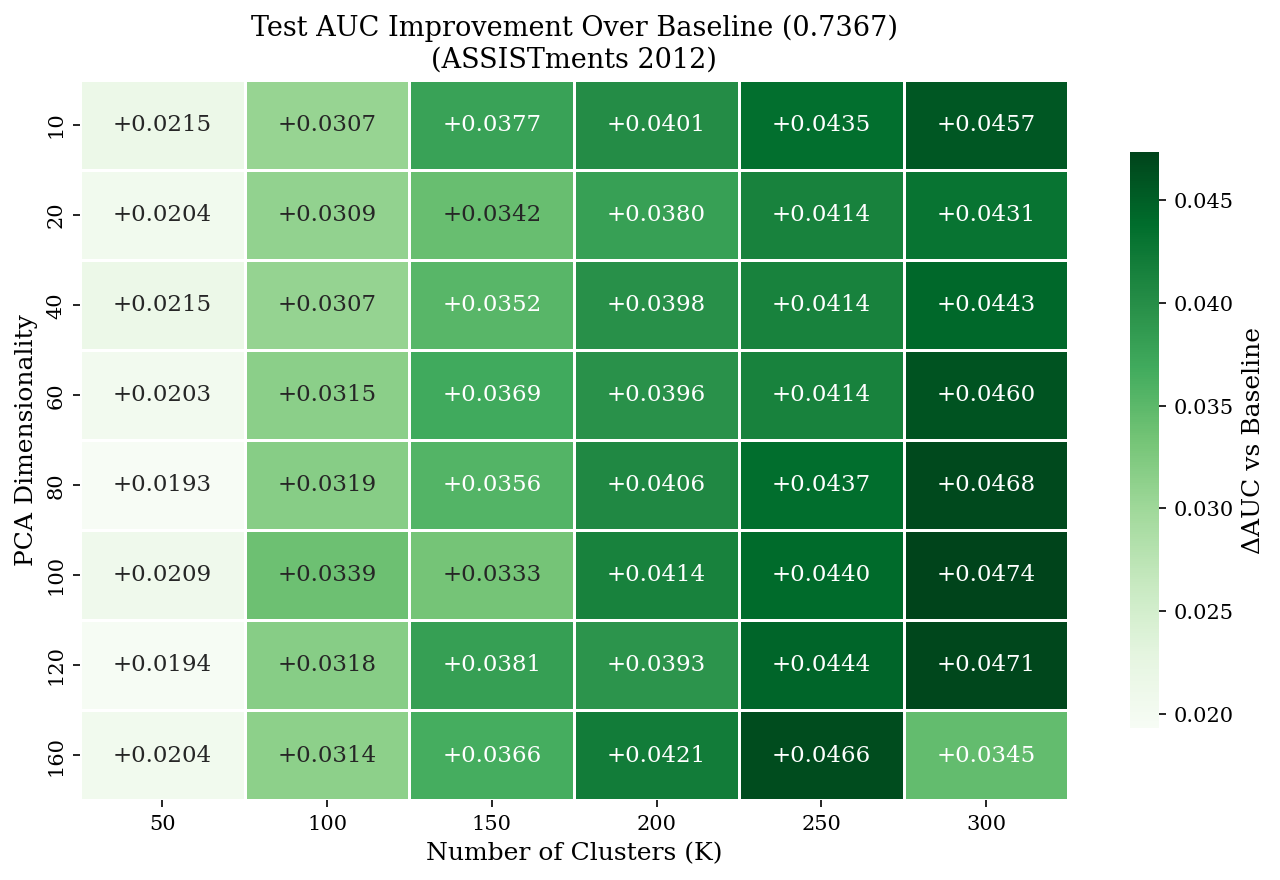}
\caption{Improvement over the original-KC baseline for ASSISTments 2012. The gains are positive across the tested grid and are largest at higher cluster counts.}
\Description{Improvement over the original-KC baseline for ASSISTments 2012. The gains are positive across the tested grid and are largest at higher cluster counts.}
\label{fig:chapter6_assist2012_improvement_heatmap}
\end{figure}

Taken together, Figures~\ref{fig:chapter6_assist2012_elbow}--\ref{fig:chapter6_assist2012_improvement_heatmap} show that the 2012 improvement is distributed across a broad region of the sweep and that cluster count has the larger effect on downstream performance over the tested range.

\FloatBarrier

\paragraph{ASSISTments 2017.}
The best downstream configuration occurs at PCA dimension 10 and 100 clusters. The optimum is tighter than on 2012. Low PCA dimensions dominate, and performance remains strong at 150 clusters for the best low-dimensional projections, but the gains flatten quickly as \(K\) increases and as higher-dimensional projections are kept.

\begin{figure}[!htbp]
    \centering
    \includegraphics[width=0.9\linewidth]{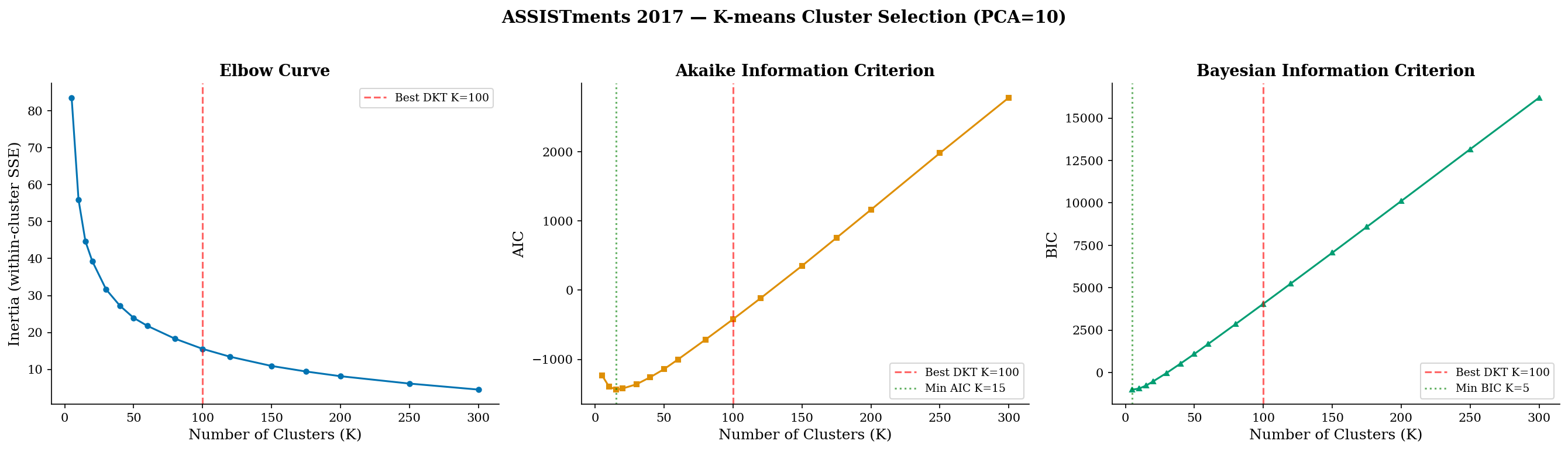}
    \caption{Clustering diagnostics for ASSISTments 2017. The best downstream KT setting occurs at a larger cluster count than the information-criterion minima, which reinforces that unsupervised clustering diagnostics alone do not select the best educational partition.}
    \Description{Clustering diagnostics for ASSISTments 2017. The best downstream KT setting occurs at a larger cluster count than the information-criterion minima, which reinforces that unsupervised clustering diagnostics alone do not select the best educational partition.}
    \label{fig:chapter6_assist2017_elbow}
\end{figure}

Figure~\ref{fig:chapter6_assist2017_elbow} again places the information-criterion minima below the best downstream setting, so the KT-optimal partition is not the same as the geometry-optimal clustering under those criteria.

\begin{figure}[!htbp]
    \centering
    \includegraphics[width=0.86\linewidth]{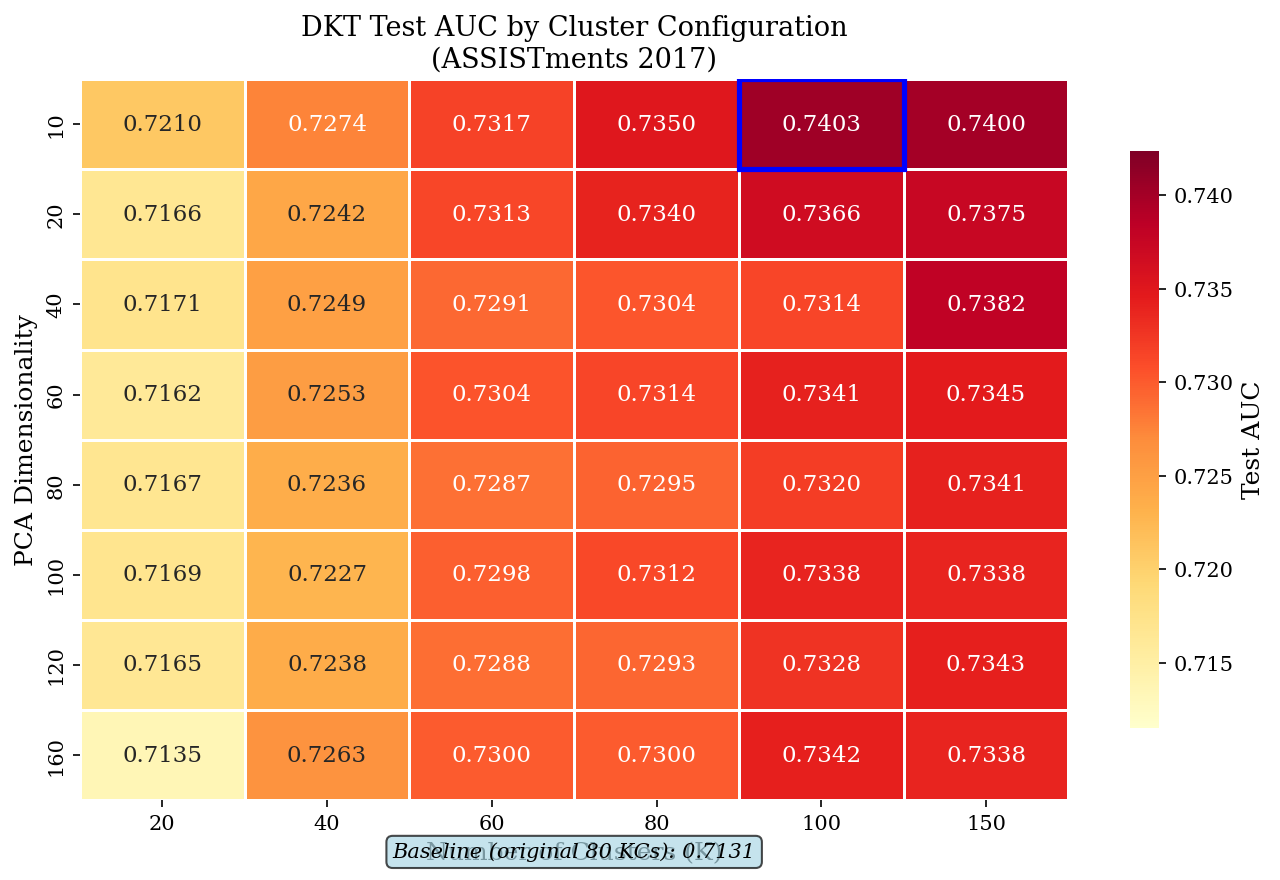}
    \caption{AUC heatmap for ASSISTments 2017 across PCA dimensionality and number of clusters. The improvement region is narrower than on 2012, but all tested clustered settings still outperform the original-KC baseline.}
    \Description{AUC heatmap for ASSISTments 2017 across PCA dimensionality and number of clusters. The improvement region is narrower than on 2012, but all tested clustered settings still outperform the original-KC baseline.}
    \label{fig:chapter6_assist2017_auc_heatmap}
\end{figure}

Figure~\ref{fig:chapter6_assist2017_auc_heatmap} shows that the improvement region is concentrated in the low-dimensional part of the grid and is visibly narrower than the corresponding region on ASSISTments 2012.

\begin{figure}[!htbp]
    \centering
    \includegraphics[width=0.86\linewidth]{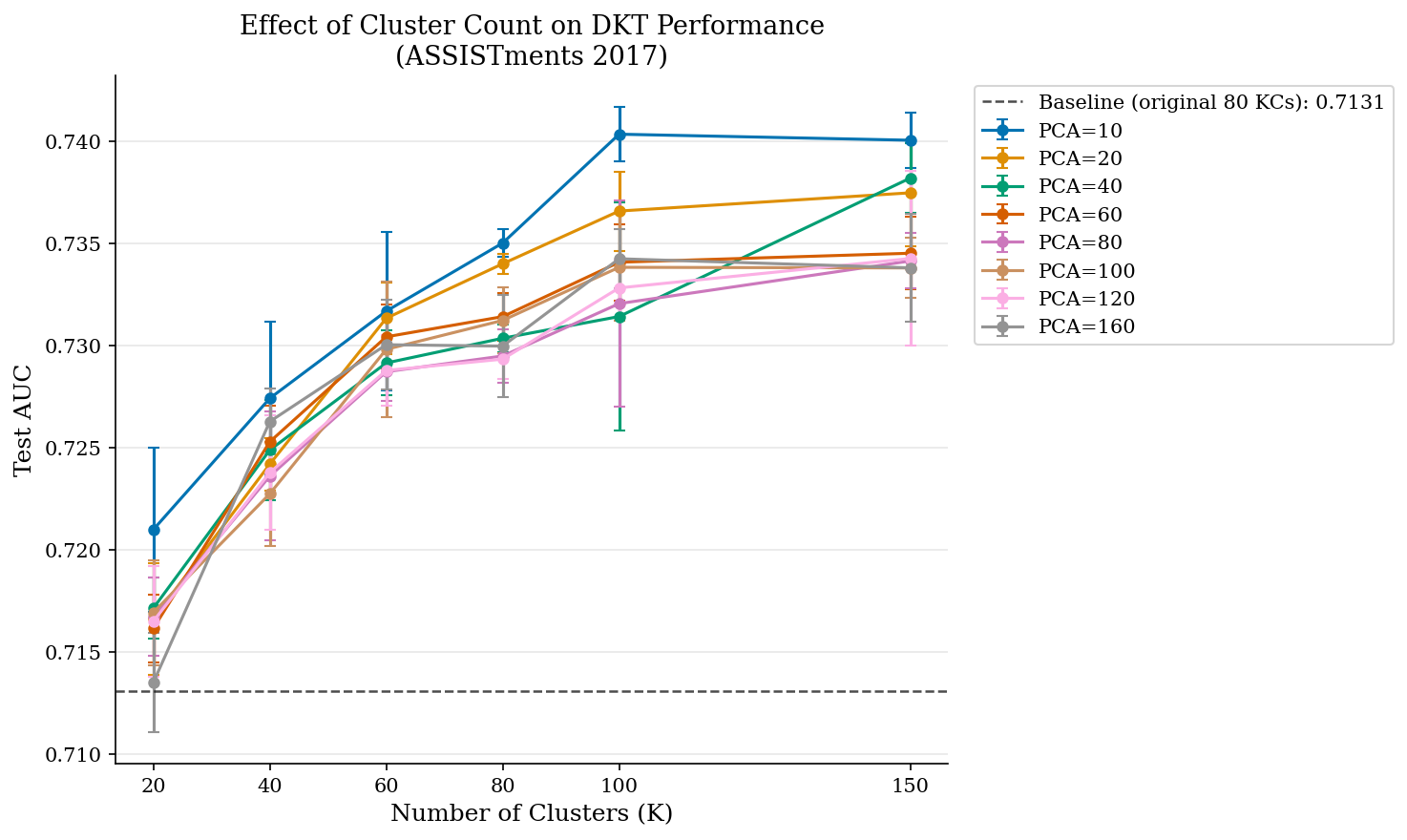}
    \caption{Effect of number of clusters on ASSISTments 2017. The best region lies around \(K=100\) to \(K=150\), and the gains flatten beyond that range.}
    \Description{Effect of number of clusters on ASSISTments 2017. The best region lies around \(K=100\) to \(K=150\), and the gains flatten beyond that range.}
    \label{fig:chapter6_assist2017_effect_k}
\end{figure}

Figure~\ref{fig:chapter6_assist2017_effect_k} places the best downstream region around \(K=100\) to \(K=150\), after which the curves flatten.

\begin{figure}[!htbp]
    \centering
    \includegraphics[width=0.86\linewidth]{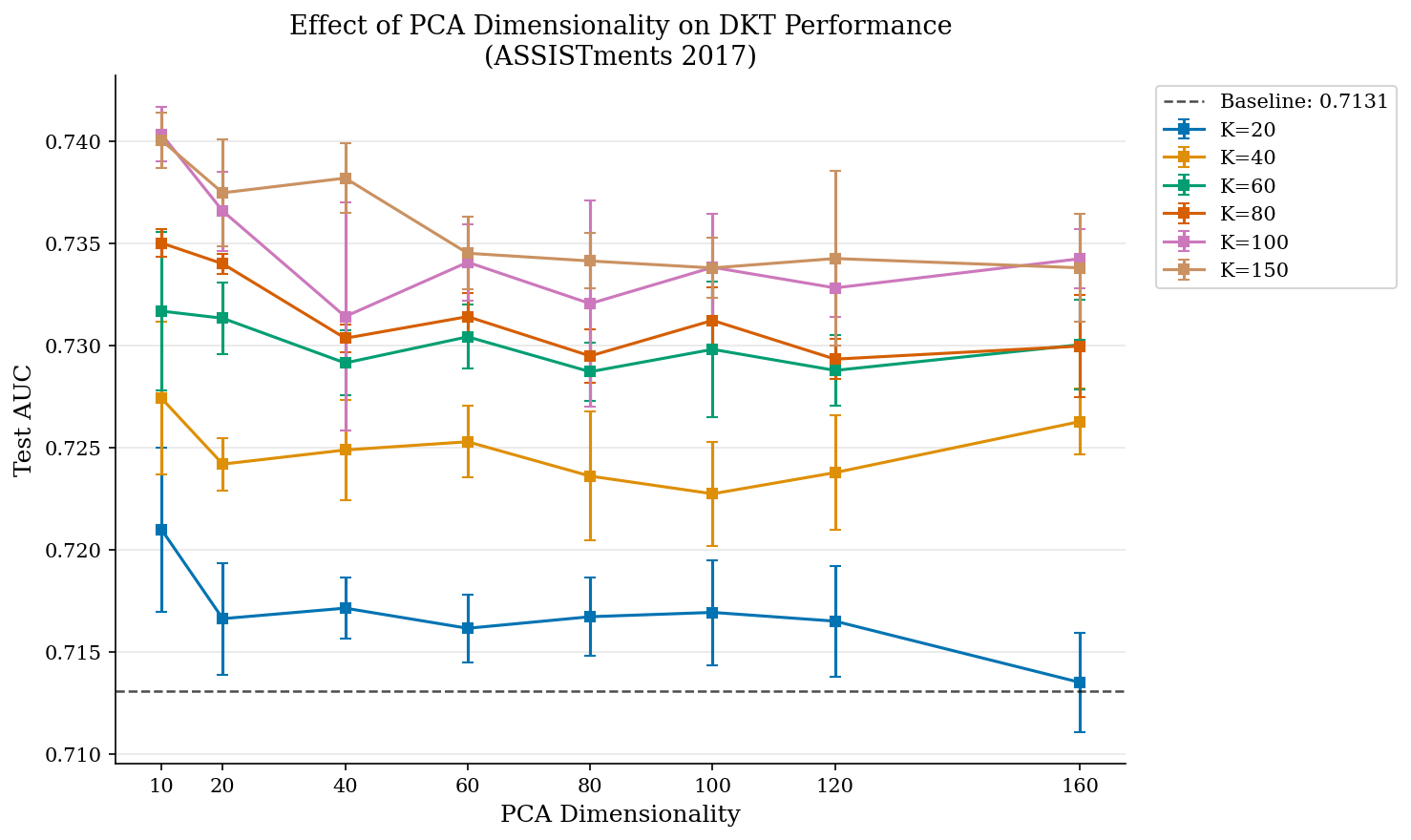}
\caption{Effect of PCA dimensionality on ASSISTments 2017. The best results occur at very low PCA dimensions, indicating that a small number of principal components retain most of the useful clustering signal for this dataset.}
\Description{Effect of PCA dimensionality on ASSISTments 2017. The best results occur at very low PCA dimensions, indicating that a small number of principal components retain most of the useful clustering signal for this dataset.}
\label{fig:chapter6_assist2017_effect_pca}
\end{figure}

Figures~\ref{fig:chapter6_assist2017_elbow}--\ref{fig:chapter6_assist2017_effect_pca} therefore show a tighter but still favorable region on 2017. Useful refinement arrives earlier and saturates faster than on 2012, low-dimensional projections dominate, and the best cluster counts stay in the moderate range.

\FloatBarrier

\paragraph{ASSISTments 2009.}
The same pipeline behaves very differently on ASSISTments 2009. No clustered setting in the evaluated grid beats the original-KC baseline, and the mean gap versus baseline remains strongly negative. Cluster count still matters more than PCA in shaping the clustered results, but here that does not translate into a win over the manual taxonomy. The best 2009 setting is again at 150 clusters, yet the baseline with the original 76 KCs remains much stronger.

\begin{figure}[!htbp]
    \centering
    \includegraphics[width=0.9\linewidth]{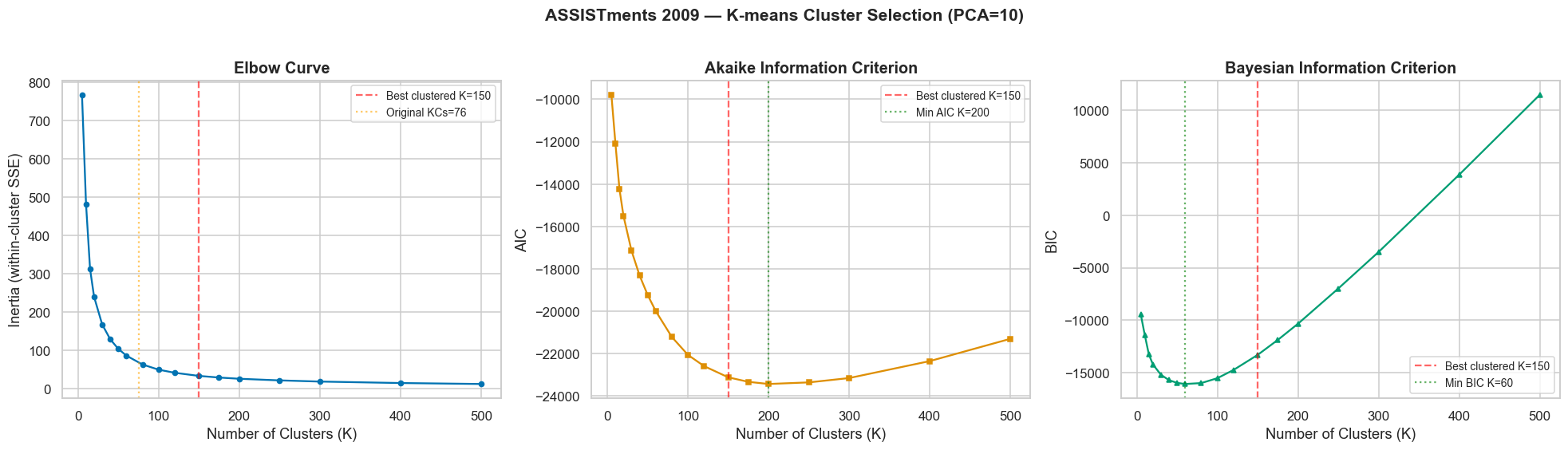}
    \caption{Clustering diagnostics for ASSISTments 2009. As on the other datasets, these curves characterize geometric clustering behavior, but they do not recover a clustered setting that matches the original-KC baseline.}
    \Description{Clustering diagnostics for ASSISTments 2009. As on the other datasets, these curves characterize geometric clustering behavior, but they do not recover a clustered setting that matches the original-KC baseline.}
    \label{fig:chapter6_assist2009_elbow}
\end{figure}

Figure~\ref{fig:chapter6_assist2009_elbow} shows the same disconnect between geometric clustering criteria and downstream KT performance, but here no clustered configuration matches the original-KC baseline.

\begin{figure}[!htbp]
    \centering
    \includegraphics[width=0.86\linewidth]{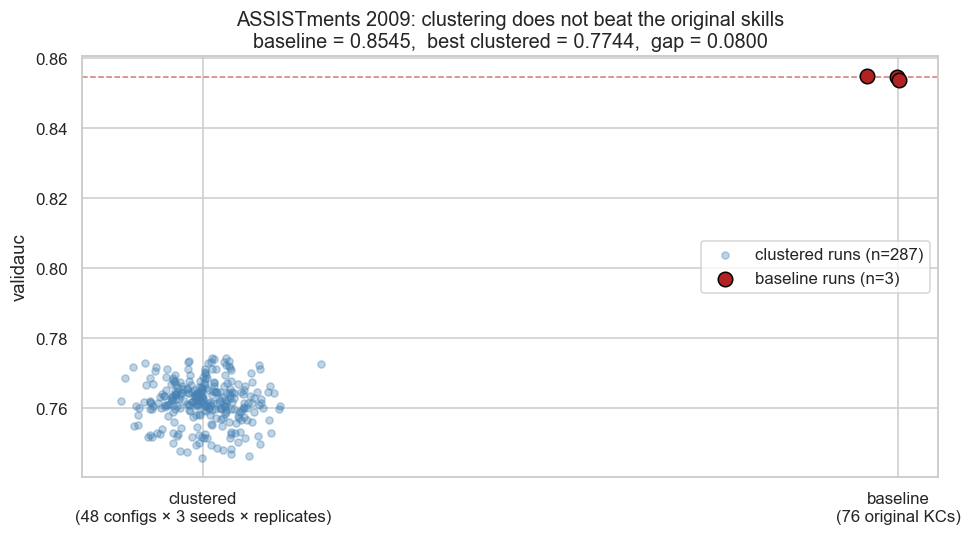}
    \caption{Headline comparison for ASSISTments 2009. The full family of clustered configurations sits well below the original-KC baseline, so the negative comparison is visible before the hyperparameter analysis is added.}
    \Description{Headline comparison for ASSISTments 2009. The full family of clustered configurations sits well below the original-KC baseline, so the negative comparison is visible before the hyperparameter analysis is added.}
    \label{fig:chapter6_assist2009_headline}
\end{figure}

Figure~\ref{fig:chapter6_assist2009_headline} shows that the negative gap is dataset-wide rather than confined to a small subset of hyperparameter settings.

\begin{figure}[!htbp]
    \centering
    \includegraphics[width=0.86\linewidth]{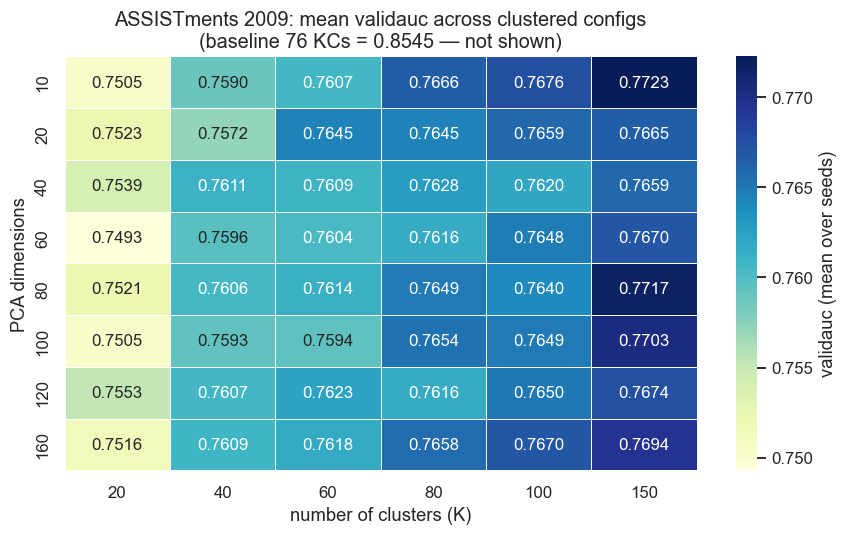}
    \caption{Validation-AUC heatmap for ASSISTments 2009 across PCA dimensionality and number of clusters. The clustered settings do show internal variation, but none of them exceed the original-KC baseline.}
    \Description{Validation-AUC heatmap for ASSISTments 2009 across PCA dimensionality and number of clusters. The clustered settings do show internal variation, but none of them exceed the original-KC baseline.}
    \label{fig:chapter6_assist2009_auc_heatmap}
\end{figure}

Figure~\ref{fig:chapter6_assist2009_auc_heatmap} shows internal variation within the clustered family, but every configuration remains below the manual-KC baseline.

\begin{figure}[!htbp]
    \centering
    \includegraphics[width=0.86\linewidth]{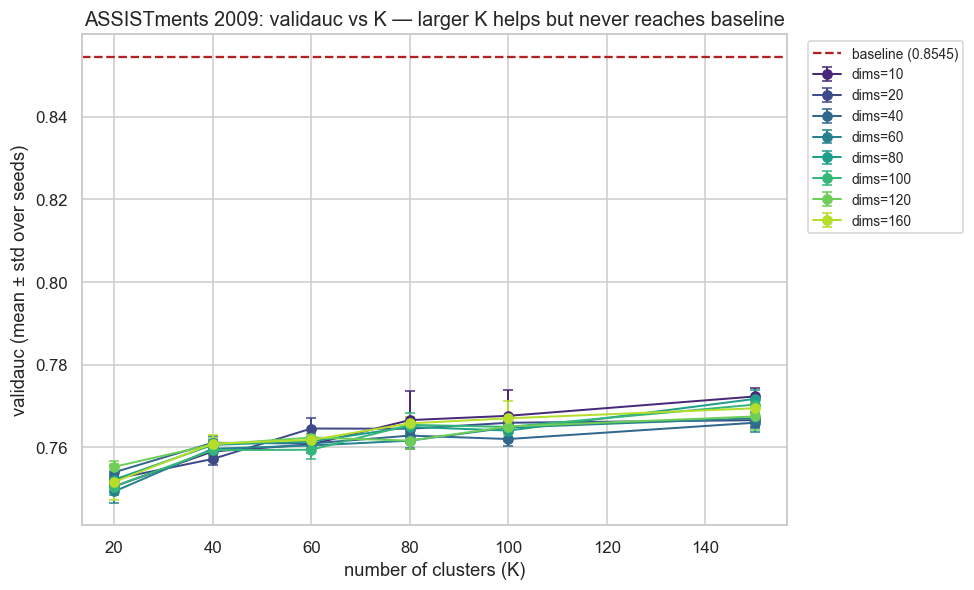}
    \caption{Effect of number of clusters on ASSISTments 2009. The best clustered results occur around 150 clusters, but increasing cluster count also increases the train--validation gap, which reinforces the negative comparison against the original-KC baseline.}
    \Description{Effect of number of clusters on ASSISTments 2009. The best clustered results occur around 150 clusters, but increasing cluster count also increases the train--validation gap, which reinforces the negative comparison against the original-KC baseline.}
    \label{fig:chapter6_assist2009_effect_k}
\end{figure}

Figure~\ref{fig:chapter6_assist2009_effect_k} shows that larger \(K\) improves the clustered family relative to itself, but not enough to reach the original-KC baseline.

\begin{figure}[!htbp]
    \centering
    \includegraphics[width=0.86\linewidth]{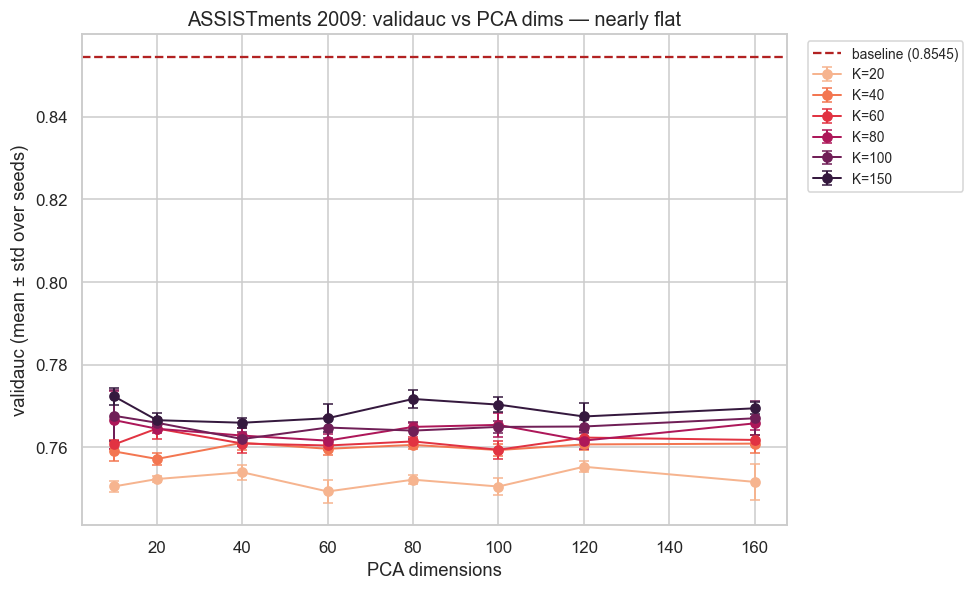}
    \caption{Effect of PCA dimensionality on ASSISTments 2009. PCA dimensionality matters less than cluster count, and moderate dimensions do not rescue the clustered representation enough to surpass the manual baseline.}
    \Description{Effect of PCA dimensionality on ASSISTments 2009. PCA dimensionality matters less than cluster count, and moderate dimensions do not rescue the clustered representation enough to surpass the manual baseline.}
    \label{fig:chapter6_assist2009_effect_pca}
\end{figure}

\begin{figure}[!htbp]
    \centering
    \includegraphics[width=0.86\linewidth]{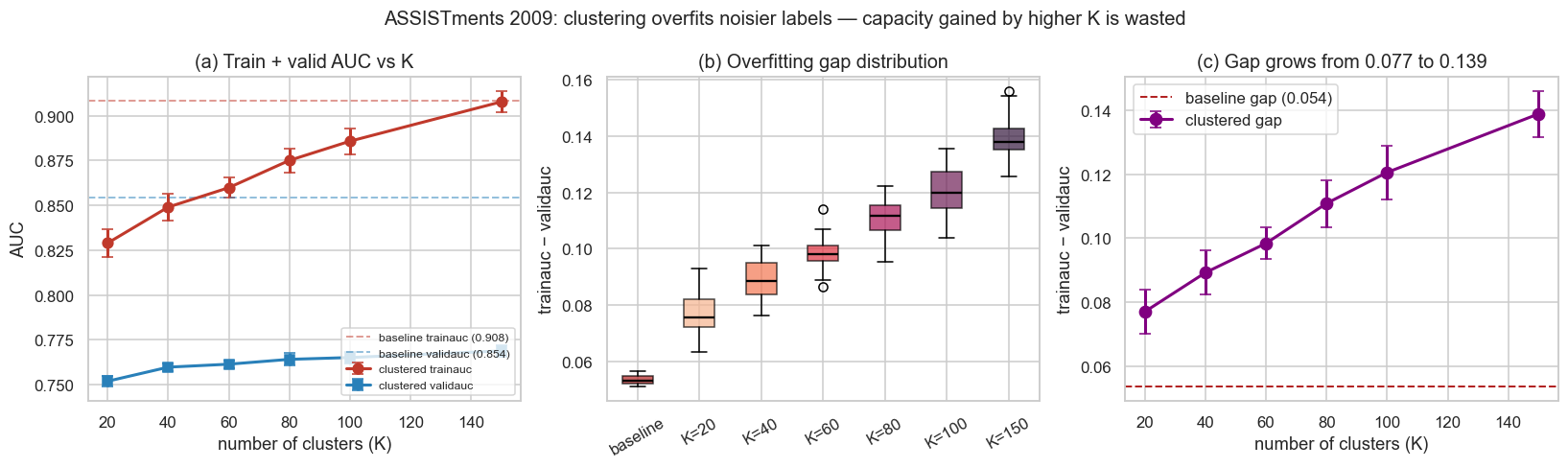}
\caption{Overfitting gap for ASSISTments 2009. As cluster count increases, the train--validation gap expands substantially, indicating that finer partitions add capacity without producing a better benchmark result.}
\Description{Overfitting gap for ASSISTments 2009. As cluster count increases, the train--validation gap expands substantially, indicating that finer partitions add capacity without producing a better benchmark result.}
\label{fig:chapter6_assist2009_overfitting_gap}
\end{figure}

Figures~\ref{fig:chapter6_assist2009_elbow}--\ref{fig:chapter6_assist2009_overfitting_gap} show that the clustered family is not flat or degenerate, but all of its useful internal variation occurs below the original-KC baseline. Larger \(K\) helps the clustered family relative to itself, yet the widening train--validation gap shows that this extra granularity does not translate into a better benchmark result.

There are therefore at least three different optimization targets in play here: downstream DKT performance, external agreement with the manual taxonomy, and internal cluster geometry. The sweep figures show that these do not collapse to the same setting. On ASSISTments 2017, for example, the best downstream result occurs at \(K=100\), while the information-criterion minima occur at smaller \(K\) values.

\FloatBarrier

\subsection{Structural Comparison with Common Core State Standards}

We therefore analyze the structural relation between the embedding-derived grouping and the original CCSS-style taxonomy separately from the benchmark sweep. This part of the chapter addresses the second research question by characterizing where the new grouping agrees with the manual labels, where it splits broad manual KCs into several clusters, where it merges multiple manual KCs into one cluster, and where it produces more mixed or overlapping reorganizations. Additional stability, internal-validity, and purity diagnostics are collected in Appendix~\ref{sec:appendix_ch6_structural}.

For ASSISTments 2012, the structural deep-dive is centered on PCA dimension 120 and 300 clusters. Across the 2012 structural grid, normalized mutual information remains high and peaks near 0.838, while adjusted Rand index reaches about 0.495. This means the embedding-derived grouping preserves a substantial amount of the manual taxonomy even before any downstream KT evaluation is considered.

For ASSISTments 2017, the structural deep-dive is centered on PCA dimension 10 and 100 clusters, which is also the best downstream DKT setting. In this filtered structural-analysis subset, the original manual baseline contains 80 KCs. External-agreement scores are lower than on 2012 and 2009 in absolute terms, with the best ARI at about 0.128 and the best NMI at about 0.671. The structural match is therefore weaker, but still strong enough to suggest that the learned grouping is refining rather than replacing the manual taxonomy.

For ASSISTments 2009, the structural analysis still shows strong correspondence with the manual taxonomy under metrics such as NMI and ARI, even though the downstream DKT comparison remains negative. The learned partition is therefore not random or degenerate. The failure is not that clustering cannot find structure at all, but that the recovered structure does not yield a better KT supervision scheme on this dataset.

\begin{table}[!htbp]
\centering
\small
\DataTableHeaderRows{1}
\begin{tabular}{lcc}
\toprule
Dataset & Best ARI configuration & Best NMI configuration \\
\midrule
ASSISTments 2012 & PCA 120, $K=50$, ARI $\approx 0.495$ & PCA 160, $K=150$, NMI $\approx 0.838$ \\
ASSISTments 2017 & PCA 60, $K=20$, ARI $\approx 0.128$ & PCA 120, $K=150$, NMI $\approx 0.671$ \\
ASSISTments 2009 & PCA 160, $K=60$, ARI $\approx 0.659$ & PCA 100, $K=60$, NMI $\approx 0.900$ \\
\bottomrule
\end{tabular}
\caption{Top structural configurations from the deep cluster-analysis. These are not the same objects as the best downstream DKT settings, which shows that structural agreement and predictive performance do not collapse to the same optimum.}
\label{tab:chapter6_structural_top_configs}
\end{table}

\begin{figure}[!htbp]
    \centering
    \includegraphics[width=0.92\linewidth]{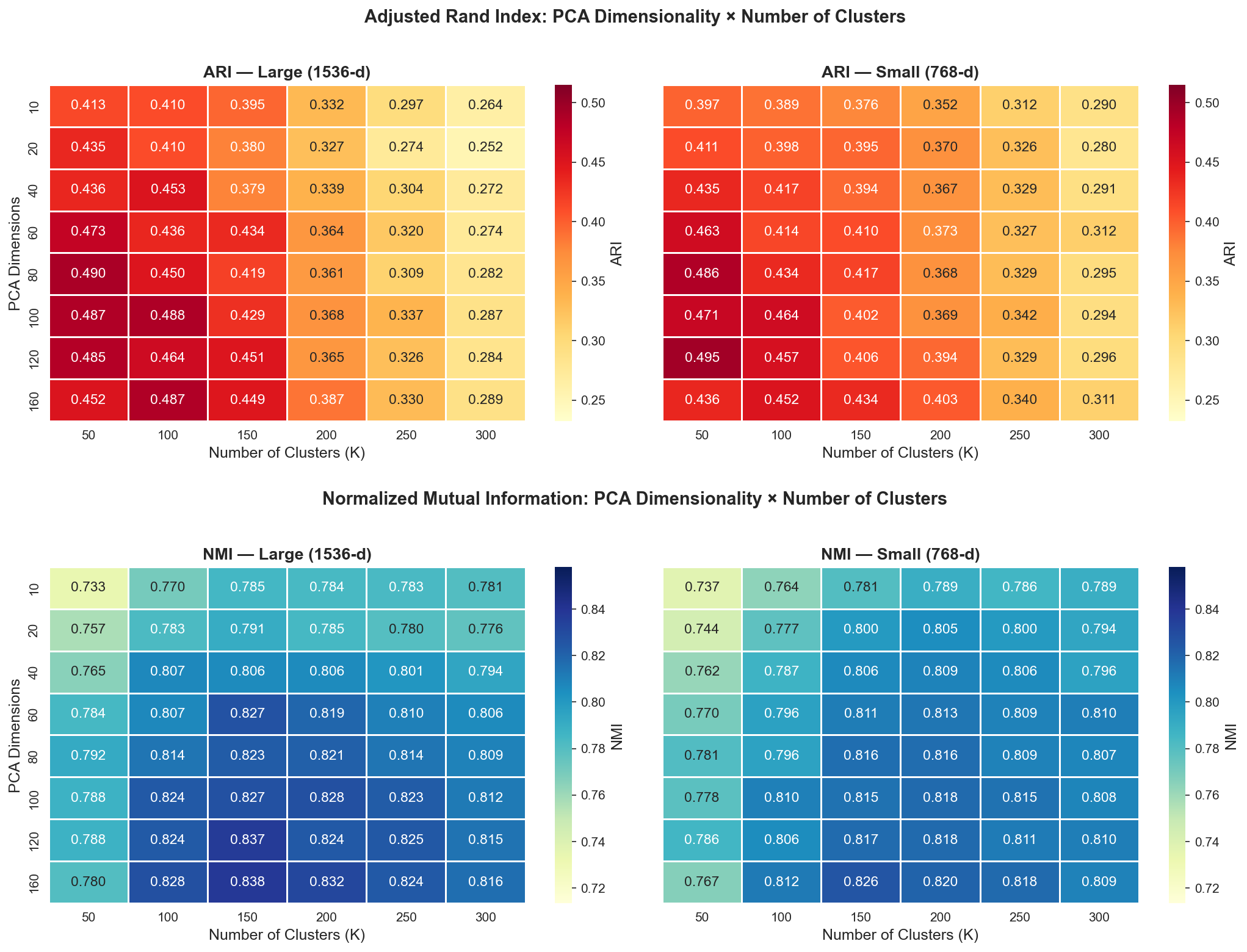}
    \caption{Structural-agreement heatmaps for ASSISTments 2012. The strongest CCSS correspondence appears in a moderate range of PCA dimensions and cluster counts, but the settings that maximize structural agreement do not exactly match the best downstream DKT setting.}
    \Description{Structural-agreement heatmaps for ASSISTments 2012. The strongest CCSS correspondence appears in a moderate range of PCA dimensions and cluster counts, but the settings that maximize structural agreement do not exactly match the best downstream DKT setting.}
    \label{fig:chapter6_cluster2012_metric_heatmaps}
\end{figure}

\begin{figure}[!htbp]
    \centering
    \includegraphics[width=0.82\linewidth]{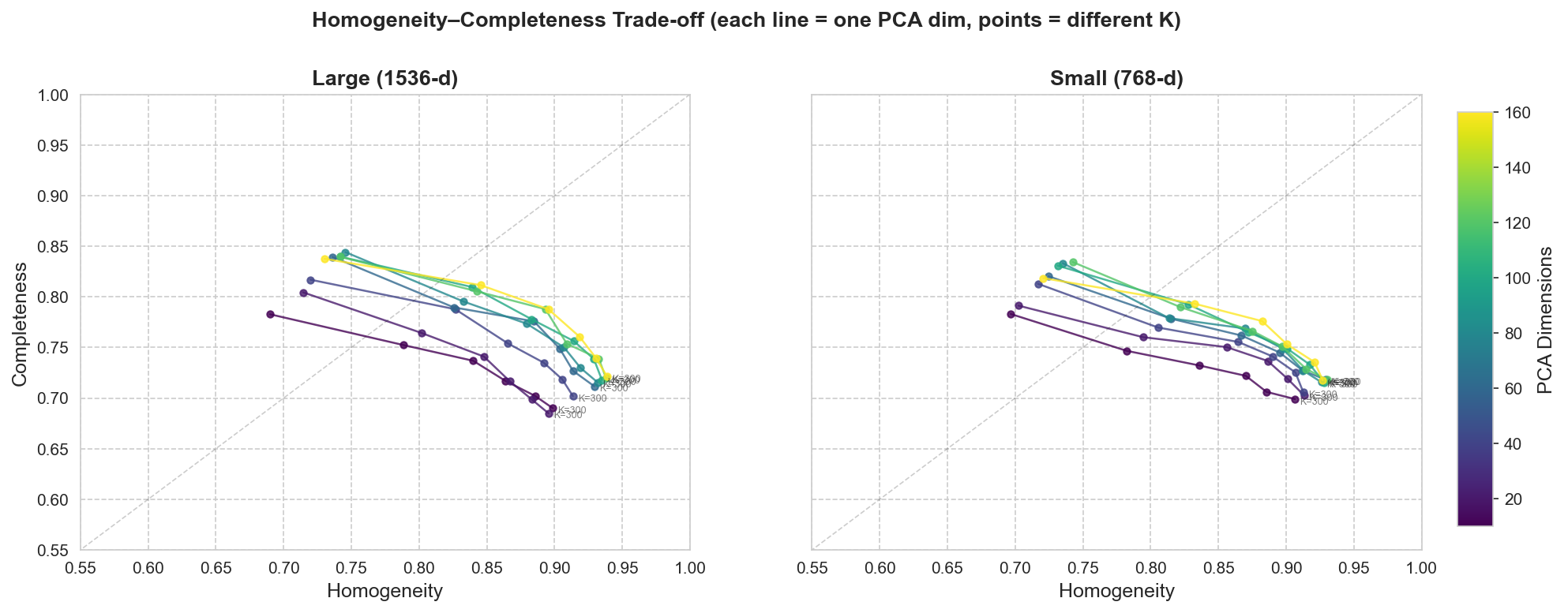}
    \caption{Homogeneity--completeness trade-off for ASSISTments 2012. Increasing the number of clusters yields purer clusters, but it also fragments the original manual KCs. The best DKT setting does not coincide exactly with the best taxonomy-matching setting.}
    \Description{Homogeneity--completeness trade-off for ASSISTments 2012. Increasing the number of clusters yields purer clusters, but it also fragments the original manual KCs. The best DKT setting does not coincide exactly with the best taxonomy-matching setting.}
    \label{fig:chapter6_cluster2012_hc}
\end{figure}

Figures~\ref{fig:chapter6_cluster2012_metric_heatmaps} and~\ref{fig:chapter6_cluster2012_hc} show that high agreement with CCSS persists across multiple PCA and cluster-count settings and that increasing \(K\) raises homogeneity while fragmenting the original manual KCs.

\begin{figure}[!htbp]
    \centering
    \includegraphics[width=0.92\linewidth]{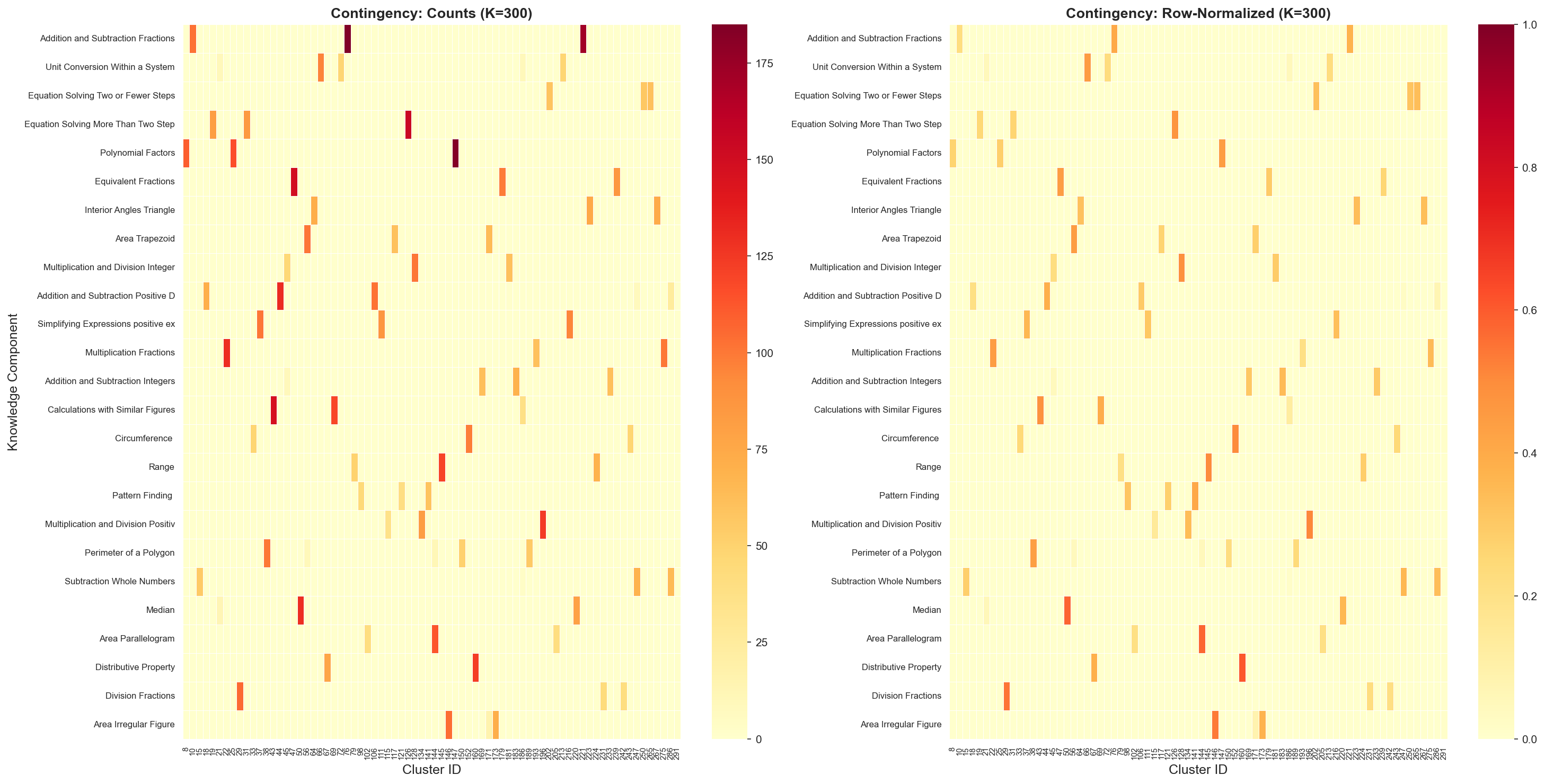}
    \caption{Contingency heatmap for the analyzed ASSISTments 2012 deep-dive configuration. Off-diagonal structure shows that the learned grouping both preserves large parts of the original taxonomy and reorganizes some broad manual KCs into multiple cleaner clusters.}
    \Description{Contingency heatmap for the analyzed ASSISTments 2012 deep-dive configuration. Off-diagonal structure shows that the learned grouping both preserves large parts of the original taxonomy and reorganizes some broad manual KCs into multiple cleaner clusters.}
    \label{fig:chapter6_cluster2012_contingency}
\end{figure}

\begin{figure}[!htbp]
    \centering
    \includegraphics[width=0.92\linewidth]{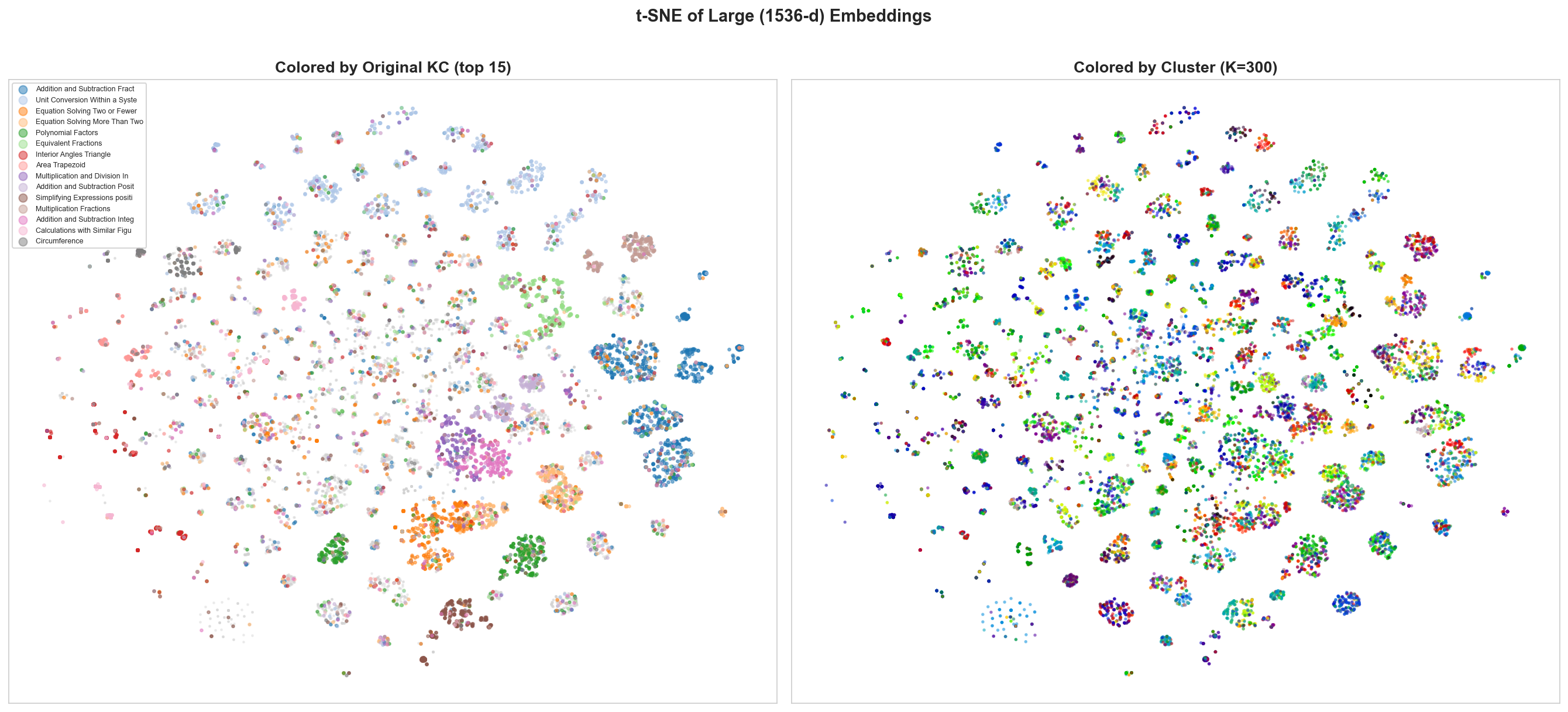}
\caption{Embedding-space visualization for ASSISTments 2012. The similarity between the manual-label coloring and the cluster-label coloring shows that the embedding-derived grouping is not arbitrary, while the visible boundary differences show that it is not simply reproducing CCSS verbatim.}
\Description{Embedding-space visualization for ASSISTments 2012. The similarity between the manual-label coloring and the cluster-label coloring shows that the embedding-derived grouping is not arbitrary, while the visible boundary differences show that it is not simply reproducing CCSS verbatim.}
\label{fig:chapter6_cluster2012_embedding_space}
\end{figure}

Figures~\ref{fig:chapter6_cluster2012_contingency} and~\ref{fig:chapter6_cluster2012_embedding_space} show that the analyzed partition preserves much of the manual taxonomy while splitting several broad KCs into more localized clusters.

\FloatBarrier

\begin{figure}[!htbp]
    \centering
    \includegraphics[width=0.92\linewidth]{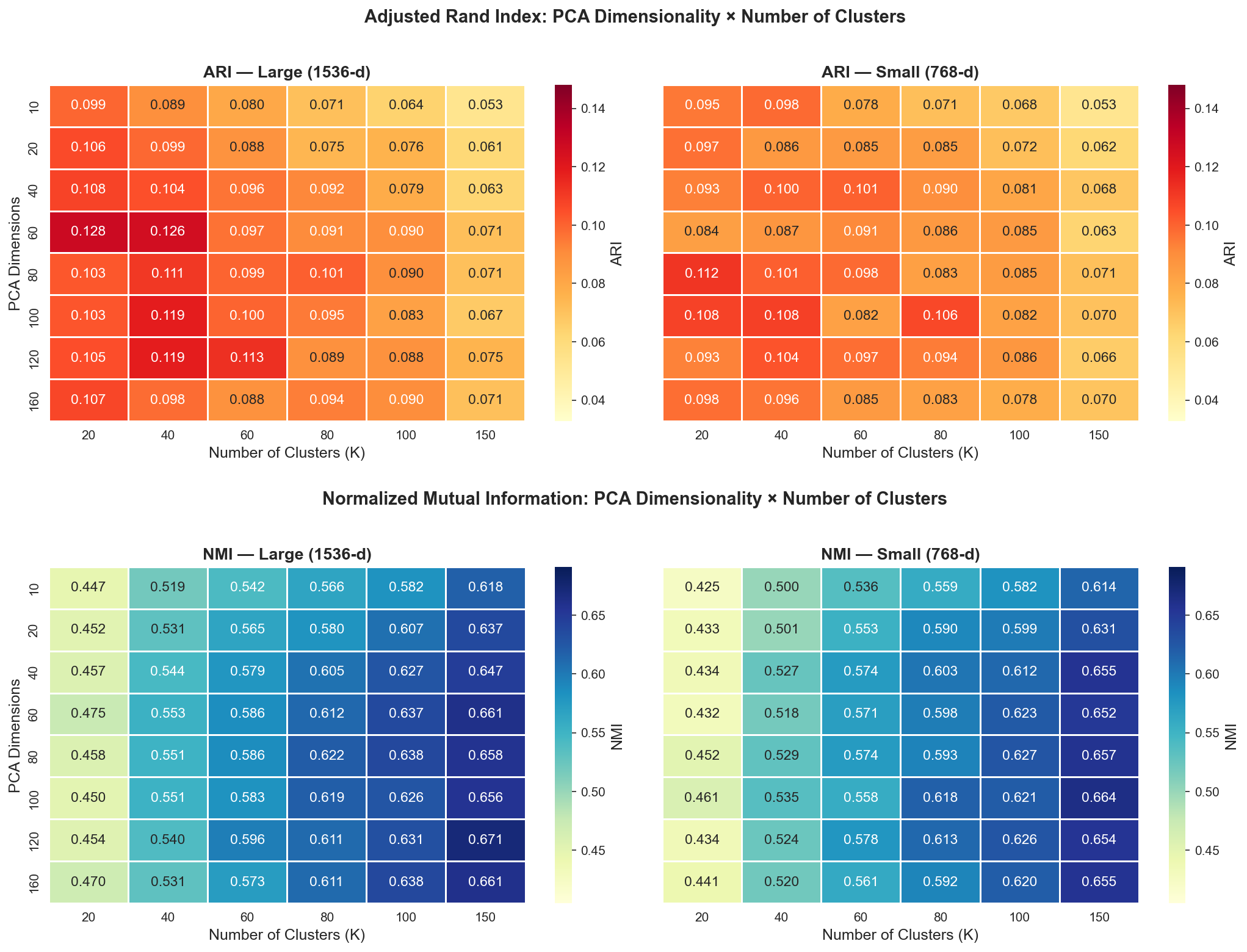}
    \caption{Structural-agreement heatmaps for ASSISTments 2017. Agreement is weaker than on 2012 and 2009 in absolute terms, but the heatmaps still show a structured relation to the original taxonomy.}
    \Description{Structural-agreement heatmaps for ASSISTments 2017. Agreement is weaker than on 2012 and 2009 in absolute terms, but the heatmaps still show a structured relation to the original taxonomy.}
    \label{fig:chapter6_cluster2017_metric_heatmaps}
\end{figure}

\begin{figure}[!htbp]
    \centering
    \includegraphics[width=0.82\linewidth]{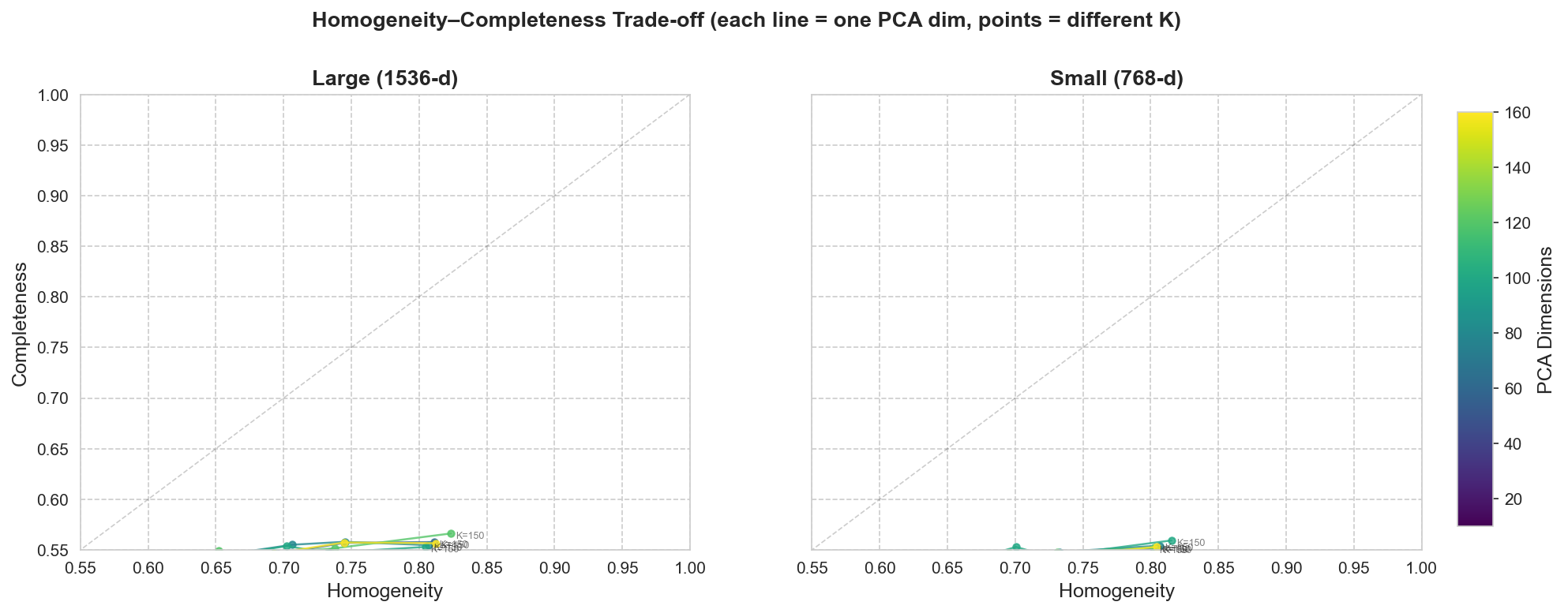}
    \caption{Homogeneity--completeness trade-off for ASSISTments 2017. The main refinement region appears around \(K=100\), where clusters are cleaner than the original labels without yet fragmenting the small problem bank too severely.}
    \Description{Homogeneity--completeness trade-off for ASSISTments 2017. The main refinement region appears around \(K=100\), where clusters are cleaner than the original labels without yet fragmenting the small problem bank too severely.}
    \label{fig:chapter6_cluster2017_hc}
\end{figure}

Figures~\ref{fig:chapter6_cluster2017_metric_heatmaps} and~\ref{fig:chapter6_cluster2017_hc} show that the 2017 agreement region is narrower than the 2012 region and centered on lower-dimensional embeddings.

\begin{figure}[!htbp]
    \centering
    \includegraphics[width=0.92\linewidth]{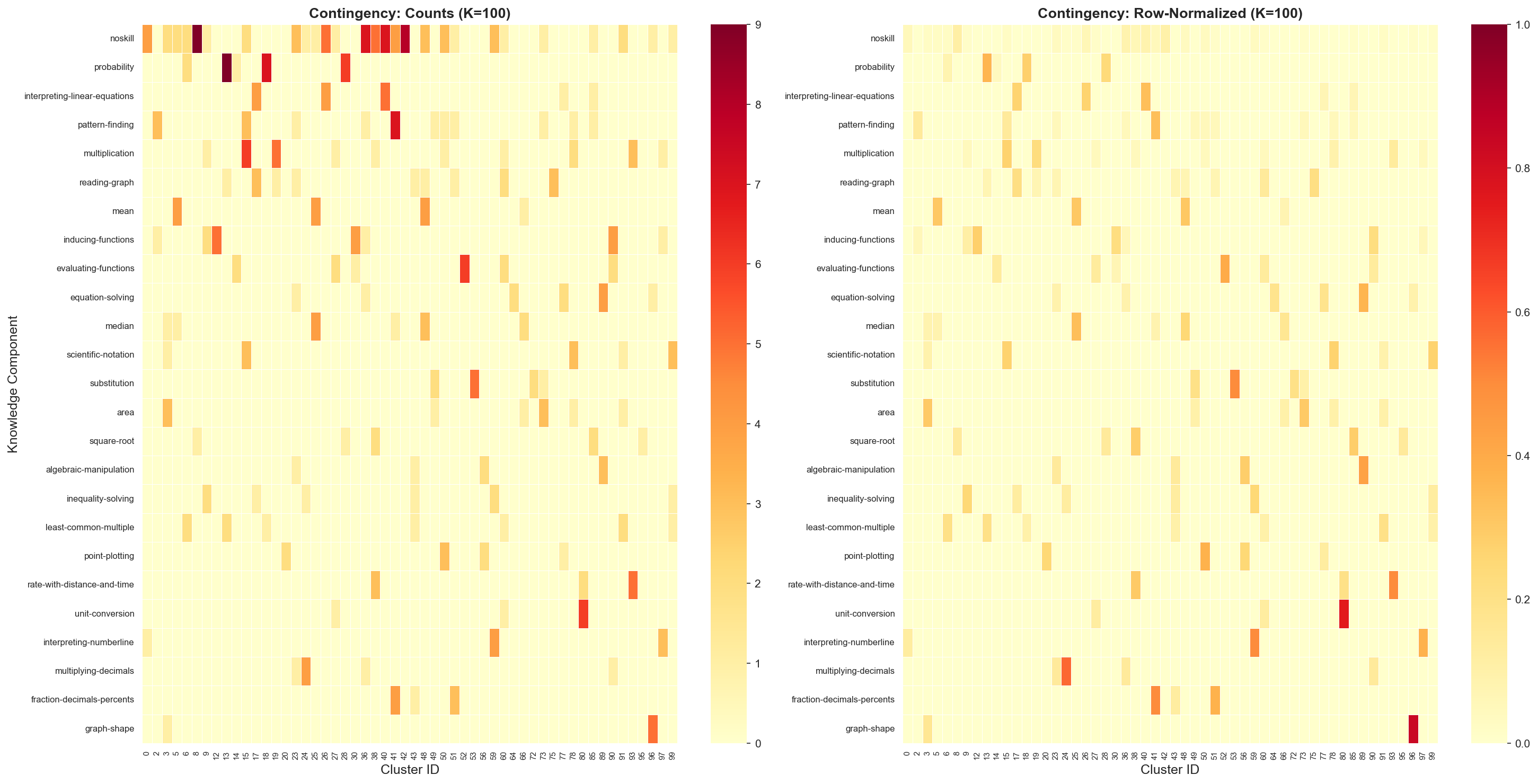}
    \caption{Contingency heatmap for the analyzed ASSISTments 2017 deep-dive configuration. The learned grouping partly preserves the manual taxonomy and partly recombines it, especially in larger and more heterogeneous KCs.}
    \Description{Contingency heatmap for the analyzed ASSISTments 2017 deep-dive configuration. The learned grouping partly preserves the manual taxonomy and partly recombines it, especially in larger and more heterogeneous KCs.}
    \label{fig:chapter6_cluster2017_contingency}
\end{figure}

\begin{figure}[!htbp]
    \centering
    \includegraphics[width=0.92\linewidth]{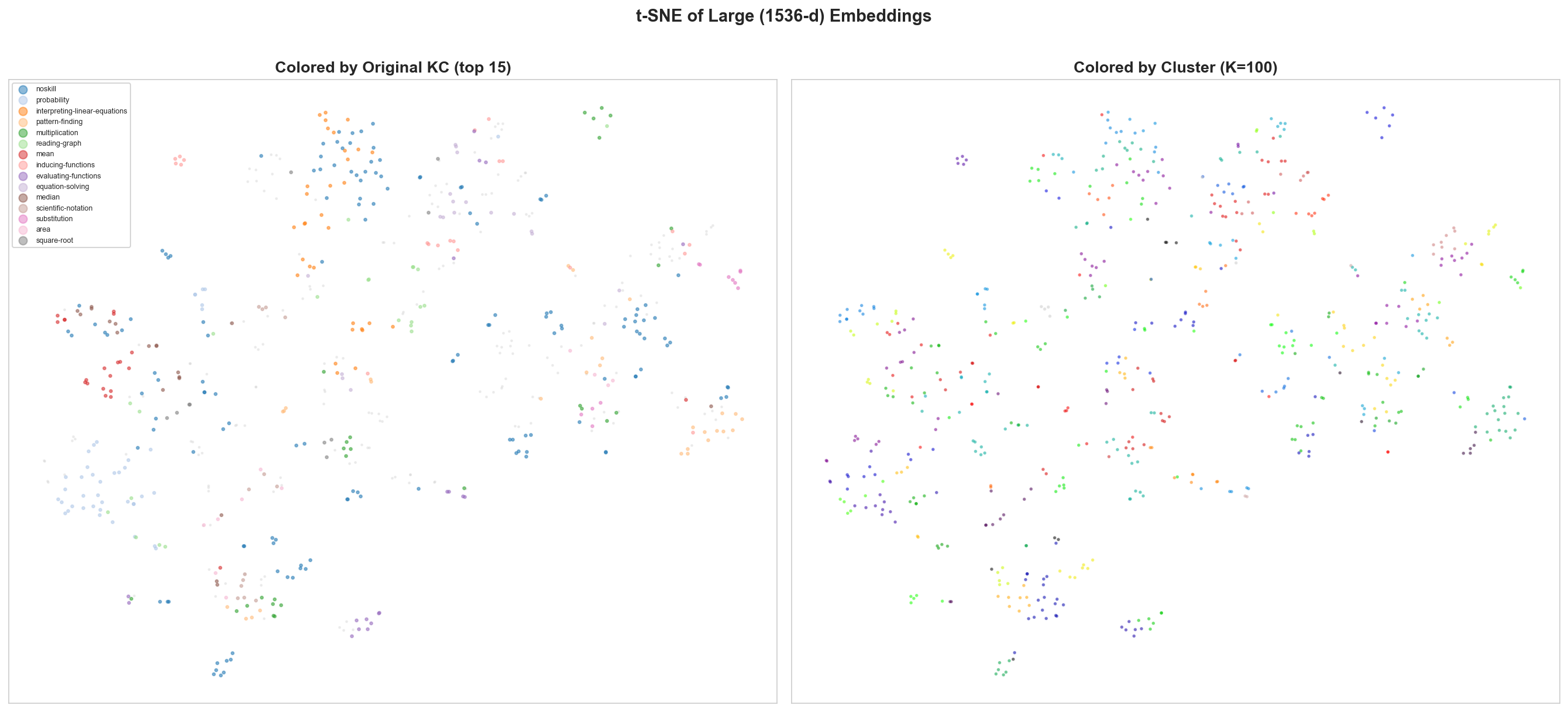}
\caption{Embedding-space visualization for ASSISTments 2017. The cluster boundaries track visible semantic structure even in a much smaller question bank, consistent with a moderate downstream KT gain.}
\Description{Embedding-space visualization for ASSISTments 2017. The cluster boundaries track visible semantic structure even in a much smaller question bank, consistent with a moderate downstream KT gain.}
\label{fig:chapter6_cluster2017_embedding_space}
\end{figure}

Figures~\ref{fig:chapter6_cluster2017_contingency} and~\ref{fig:chapter6_cluster2017_embedding_space} show moderate reorganization rather than wholesale replacement of the manual taxonomy, which matches the smaller downstream gain.

\FloatBarrier

\begin{figure}[!htbp]
    \centering
    \includegraphics[width=0.92\linewidth]{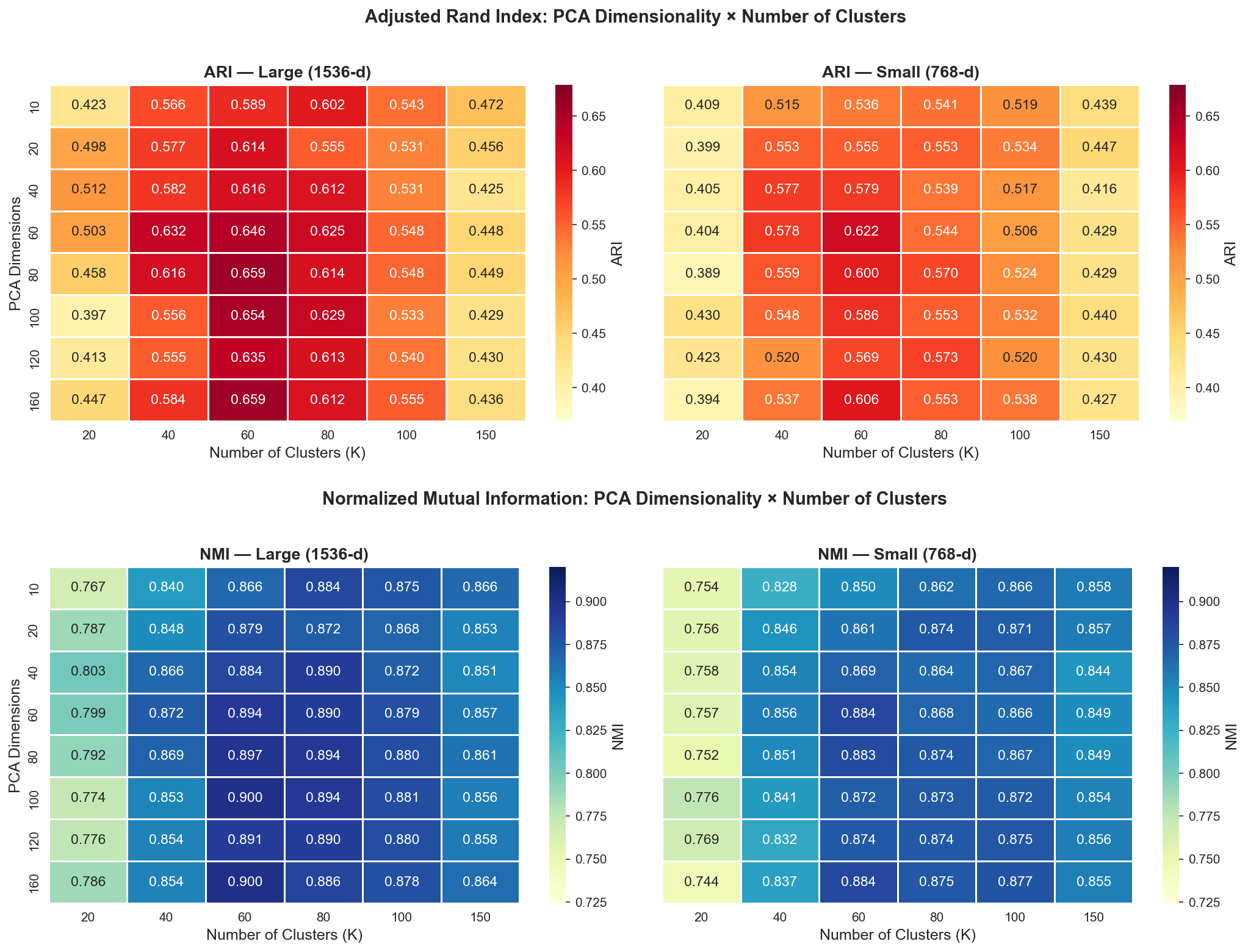}
    \caption{Structural-agreement heatmaps for ASSISTments 2009. The clusters can still align strongly with CCSS-like structure even though the downstream DKT comparison remains negative. Taxonomy agreement and predictive performance therefore peak at different settings.}
    \Description{Structural-agreement heatmaps for ASSISTments 2009. The clusters can still align strongly with CCSS-like structure even though the downstream DKT comparison remains negative. Taxonomy agreement and predictive performance therefore peak at different settings.}
    \label{fig:chapter6_cluster2009_metric_heatmaps}
\end{figure}

\begin{figure}[!htbp]
    \centering
    \includegraphics[width=0.82\linewidth]{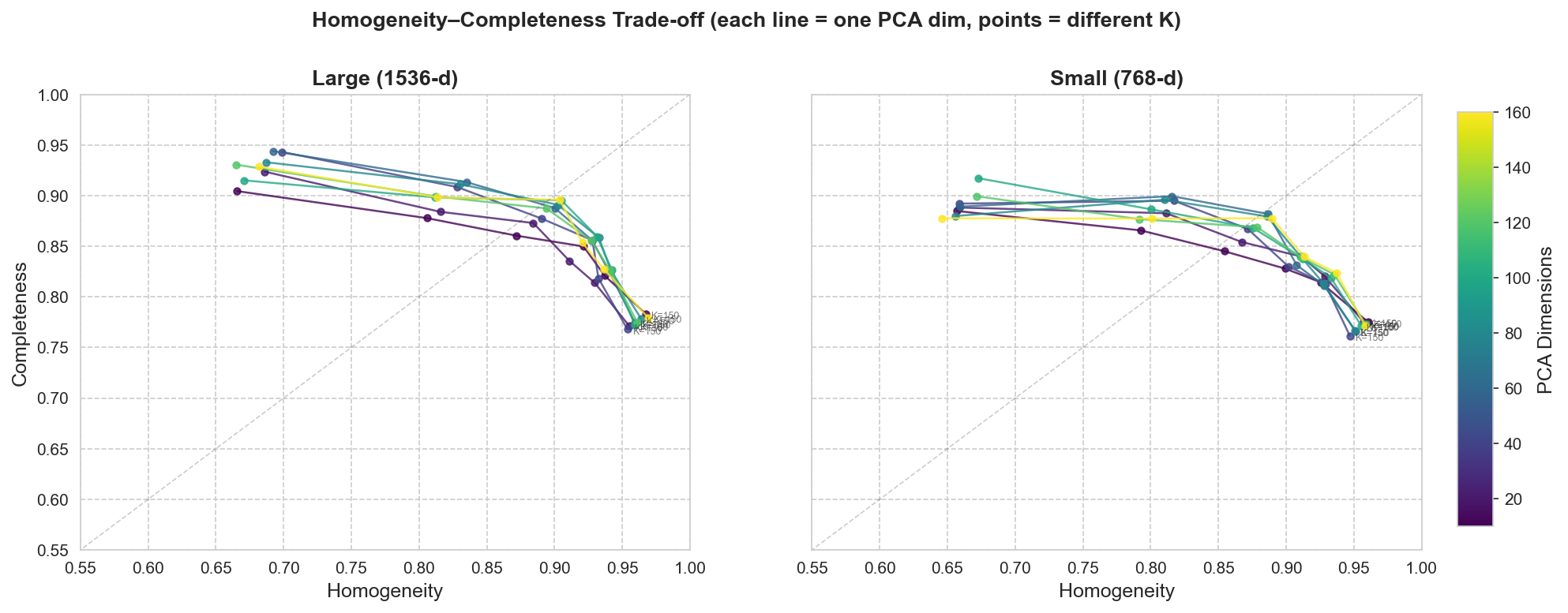}
    \caption{Homogeneity--completeness trade-off for ASSISTments 2009. As on 2012, increasing the number of clusters produces cleaner local groups at the cost of greater fragmentation of the original manual KCs.}
    \Description{Homogeneity--completeness trade-off for ASSISTments 2009. As on 2012, increasing the number of clusters produces cleaner local groups at the cost of greater fragmentation of the original manual KCs.}
    \label{fig:chapter6_cluster2009_hc}
\end{figure}

\begin{figure}[!htbp]
    \centering
    \includegraphics[width=0.92\linewidth]{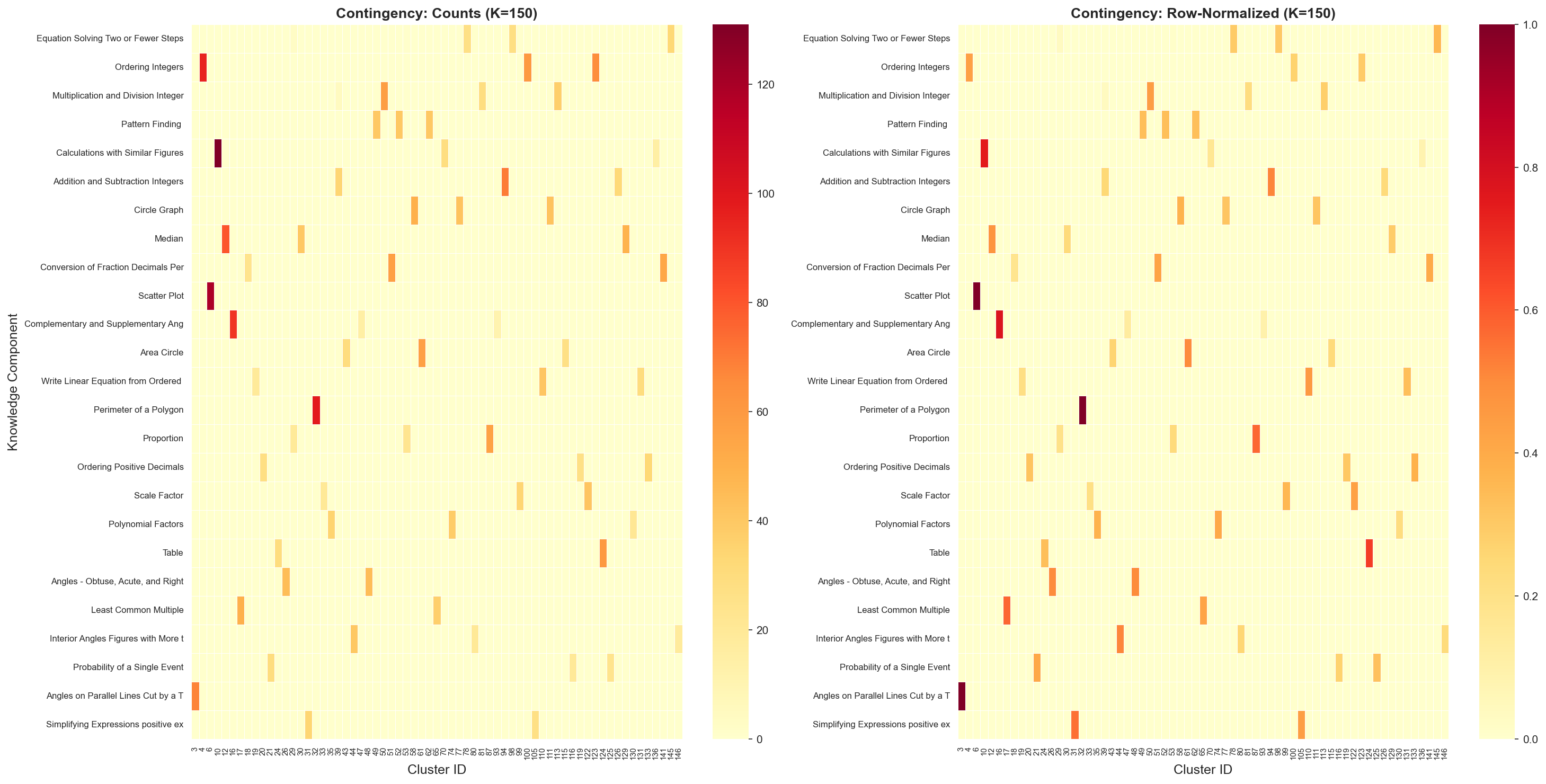}
    \caption{Contingency heatmap for the analyzed ASSISTments 2009 deep-dive configuration. The learned grouping is not random, but the off-diagonal structure shows a more difficult relation between lexical-semantic similarity and the manual procedural taxonomy.}
    \Description{Contingency heatmap for the analyzed ASSISTments 2009 deep-dive configuration. The learned grouping is not random, but the off-diagonal structure shows a more difficult relation between lexical-semantic similarity and the manual procedural taxonomy.}
    \label{fig:chapter6_cluster2009_contingency}
\end{figure}

\begin{figure}[!htbp]
    \centering
    \includegraphics[width=0.92\linewidth]{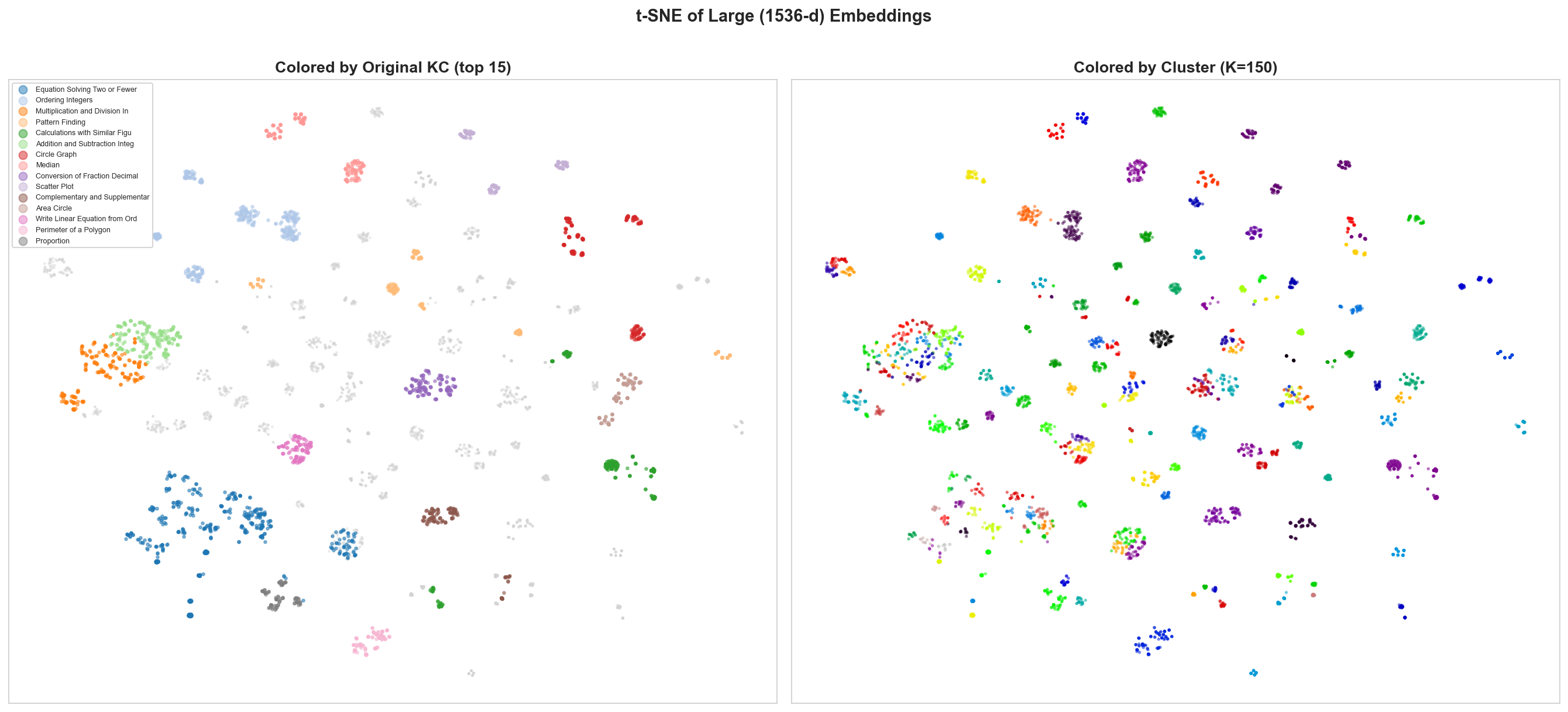}
\caption{Embedding-space visualization for ASSISTments 2009. The global structure is coherent enough to support clustering, but the boundaries are visibly less aligned with the manual taxonomy than on 2012, which is consistent with the weaker structural match and the negative downstream result.}
\Description{Embedding-space visualization for ASSISTments 2009. The global structure is coherent enough to support clustering, but the boundaries are visibly less aligned with the manual taxonomy than on 2012, which is consistent with the weaker structural match and the negative downstream result.}
\label{fig:chapter6_cluster2009_embedding_space}
\end{figure}

Figures~\ref{fig:chapter6_cluster2009_metric_heatmaps}--\ref{fig:chapter6_cluster2009_embedding_space} show coherent local structure and nontrivial agreement with CCSS, but they also show a less clean relation between lexical-semantic similarity and the procedural manual taxonomy. Structurally meaningful clustering can therefore still fail to improve KT on this dataset.

\FloatBarrier

\section{Discussion}

\textbf{RQ1.} Redefining the KC space through embedding-derived question groupings can improve downstream KT, but not uniformly. The effect is positive on ASSISTments 2012 and ASSISTments 2017, and negative on ASSISTments 2009. KC redefinition is therefore dataset-dependent rather than universally beneficial.

\textbf{RQ2.} The discovered grouping relates to the human-authored CCSS-style taxonomy in a mixed but interpretable way. The structural analyses show substantial agreement with the manual taxonomy, but not identity. The embedding-derived grouping often preserves broad taxonomic structure while still splitting large manual KCs, consolidating some question sets, and recombining others. That pattern is what keeps the regrouping interpretable: if the new grouping were identical to CCSS, there would be little reason to use it; if it were unrelated to CCSS, it would be hard to interpret.

At the ITS level, the result means that the KC partition is part of the tutoring problem itself. A tutoring system acts on an estimate of student knowledge when it decides what to ask next, what to reteach, or when to move forward. The original DKT work makes this explicit by framing next-exercise selection as a sequential decision problem over predicted future knowledge~\cite{piech_deep_2015}. If changing the KC partition alone changes predictive performance by a substantial margin, then the concept decomposition is not just a bookkeeping layer around the model.

\subsection{Interpreting the Cross-Dataset Pattern}

The cross-dataset pattern should be interpreted carefully rather than flattened into a single success claim. ASSISTments 2012 provides the largest positive case. Its larger text-linked sample gives the clustering stage more questions from which to estimate stable semantic neighborhoods and gives DKT more relabeled supervision on which to learn the revised concept space. The figures show that the improvement region spans multiple hyperparameter settings, that cluster count matters more than PCA dimensionality over the tested range, and that the gains continue as the model moves to finer partitions. These patterns are consistent with the original manual taxonomy being too coarse for this dataset.

ASSISTments 2017 shows that scale is not the only factor that matters. The Chapter~5 dataset statistics indicate that sequence structure is also likely relevant. If the revised KC space captures a more useful concept organization, longer trajectories give DKT more opportunity to benefit from that structure over time. That helps explain why ASSISTments 2017 improves even though it is much smaller than ASSISTments 2012. Figures~\ref{fig:chapter6_assist2017_auc_heatmap}--\ref{fig:chapter6_assist2017_effect_pca} show that the gain is real but more localized: low PCA dimensions dominate, \(K=100\) to \(K=150\) is the main operating range, and the problem bank approaches its granularity limit faster than 2012. ASSISTments 2017 therefore adds sequence structure and cluster granularity to sample size as likely determinants of when KC redefinition helps.

ASSISTments 2009 should be treated as a harder case rather than as a simple refutation of the main idea we study here. The dataset is already known to be preprocessing-sensitive in the KT literature, the current pipeline removes interactions without linked text, and the baseline can preserve multi-KC structure that is lost when each question is forced into a single cluster-derived KC~\cite{xiong_going_2016}. The 2009 sweep also shows that increasing cluster count expands the train--validation gap without producing a win over the original-KC baseline. Those factors make ASSISTments 2009 the least favorable setting for the method, even though the negative result itself is still real and important.

\subsection{Agreement with CCSS and Reorganization of the Question Space}

A central result of the structural analyses is that they separate agreement from identity. The embedding-derived grouping agrees strongly with the original CCSS-style taxonomy under several external metrics, especially on ASSISTments 2012 and ASSISTments 2009, and more moderately on ASSISTments 2017, but that agreement is not perfect and should not be. The deep-dive analyses show repeated cases of fragmentation, consolidation, and overlap. Large manual KCs often divide into multiple cleaner clusters, while some embedding-derived clusters draw together questions that were separated in the manual taxonomy.

Across the 2012 structural grid, NMI remains high and ARI remains meaningful, which shows that text embeddings recover a substantial amount of KC structure even without KC labels during clustering. The homogeneity--completeness plots and contingency heatmaps in Figures~\ref{fig:chapter6_cluster2012_hc} and~\ref{fig:chapter6_cluster2012_contingency} show that the clusters are often cleaner than the original KCs and that the large manual KCs are exactly the ones that fragment most. This gives the model a finer-grained signal where the benchmark taxonomy is broadest.

The 2017 structural analysis places the dataset between 2012 and 2009. The external-agreement values are smaller in absolute terms, but Figures~\ref{fig:chapter6_cluster2017_metric_heatmaps}--\ref{fig:chapter6_cluster2017_embedding_space} still show a meaningful relation to the manual taxonomy. The low-dimensional embedding space preserves enough structure for the clustering to refine several large KCs into cleaner groups, while the contingency heatmap shows that the learned partition is not simply repeating the filtered original 80-KC manual labeling. This pattern is consistent with a moderate KT gain in a much smaller problem bank.

The 2009 structural analysis shows a different pattern. Structural agreement and downstream usefulness are more aligned inside the clustered configurations themselves, with the best-performing clustered settings also living in the high-agreement zone visible in Figure~\ref{fig:chapter6_cluster2009_metric_heatmaps}, yet the entire clustered family still loses to the original baseline. Figures~\ref{fig:chapter6_cluster2009_contingency} and~\ref{fig:chapter6_cluster2009_embedding_space} show that the learned grouping is structured, but its structure does not line up with the benchmark's procedural taxonomy as cleanly as it does on 2012. This shows that even a coherent and stable text-derived partition may fail when the inherited manual taxonomy is already unusually strong or when the dataset couples KC structure to interaction preprocessing in ways the clustering does not preserve.

The positive results here do not come from replacing CCSS with a completely unrelated partition. They come from starting with a structure that still respects a large amount of the benchmark taxonomy and then reorganizing the parts that appear too broad or too heterogeneous for the downstream model. This also explains why the best DKT configuration need not be the one that maximizes ARI or NMI against the manual labels. The replacement taxonomy remains interpretable without reproducing the benchmark exactly.

\subsubsection{DNA-Based and Morphology-Based Taxonomy}

One way to interpret these results is through an analogy to biological taxonomy. Before molecular methods became widely available, species taxonomy was necessarily shaped by morphology-based taxonomy: organisms were grouped by observable form, structure, and other visible attributes. DNA-based taxonomy did not make morphology-based taxonomy irrelevant. Instead, molecular systematics provided a deeper basis for testing which morphology-based groupings reflected underlying relatedness, confirming many of them while also revising others where visible similarity was an incomplete guide to structure~\cite{sanMauroMolecularSystematics2010}.

The relation between CCSS and the embedding-derived grouping in this chapter can be understood in the same way. The original CCSS taxonomy functions as a morphology-based taxonomy of educational content. It groups questions using expert-authored curricular attributes and visible instructional distinctions. The embedding-derived grouping is closer to a DNA-based taxonomy of the question set. It uses the internal semantic signal of the questions themselves, captured through text embeddings, as a deeper representational basis for organizing the concept space.

This analogy clarifies the claim. The embedding-derived taxonomy does not discard the manual taxonomy as meaningless. On the contrary, the high NMI and meaningful ARI values show that the embedding-based taxonomy confirms a large portion of the existing CCSS structure. It also shows where that structure may need to be revised because the inherited partition is too coarse, too heterogeneous, or too weakly aligned with the distinctions that matter downstream.

That combination of confirmation and revision is exactly the pattern one would expect from a DNA-based taxonomy relative to a morphology-based taxonomy. Much of the existing organization survives because it captures real structure. The deeper representation matters where it revises that organization and improves what can be done with it. Here, that added value appears not only in the structural metrics but also in the downstream KT gains on the datasets where the original KC partition appears to under-decompose the content space. Under that interpretation, the result is more than an artifact of clustering. It is evidence that KC taxonomies can be rethought at the level of the learning space itself, and that this rethinking may matter for how we model learning in ITS.

\subsection{Connection to Chapter 5}

In Chapter~5, we showed that semantic text embeddings do not reliably improve DKT when they are injected into the existing KC space as pretrained interaction representations. The positive results in this chapter suggest a different use of text: semantic information appears more useful when it helps redefine the concept partition than when it is added to an inherited one. The two chapters therefore support a more specific claim about where semantic information is useful in KT.

\subsection{Limitations}

Several limitations are important here. First, the clustering pipeline depends on the embedding model, the PCA setting, and the selected cluster count. The discovered grouping is therefore a candidate partition rather than a uniquely recovered ground truth. Second, as discussed earlier in the chapter and in the Chapter~5 dataset statistics table (Table~\ref{tab:chapter5_pb_data}), the study depends on partial Problem Bodies coverage, so the relabeled experiments operate on a filtered subset of the original interaction logs rather than the full datasets.

Third, the relabeling step simplifies the relationship between questions and concepts. In datasets with multi-KC structure, especially ASSISTments 2009, forcing each question into a single embedding-derived KC may discard supervision structure that the baseline still preserves. Fourth, the structural comparison against the manual taxonomy is based on clustering-agreement metrics, visualization, and fragmentation analyses rather than on a formal cognitive evaluation framework. The structural results therefore show how the learned partition relates to CCSS, but they do not by themselves establish that the embedding-derived grouping is a better educational ontology in every sense.

\section{Conclusion}

In this chapter, we examined whether text embeddings are more useful for redefining the concept space of KT than for enriching a fixed existing one. The core method was to cluster question embeddings, relabel the interaction data with the resulting embedding-derived KC groupings, and compare DKT under the revised mapping against DKT under the original manual labels.

The results show that this strategy can improve DKT, but the effect is strongly dataset-dependent. ASSISTments 2012 and ASSISTments 2017 improve, whereas ASSISTments 2009 remains a negative case shaped by a more difficult combination of preprocessing sensitivity, text filtering, and multi-KC complications. These findings show that manually specified KC taxonomies are not always the best supervision structure for KT and that embedding-derived regrouping can be an effective alternative under some dataset conditions.

\chapter{ReasonKT: A Preliminary Study of Reasoning-Based Knowledge Tracing}

\section{Introduction}

The earlier KT chapters narrow the problem addressed here. Chapter~5 found that semantic information from question text does not reliably improve DKT when it is injected into the inherited KC space, whereas Chapter~6 showed that text embeddings can help when they redefine the KC space itself. In this chapter, we ask a different question: can KT itself be framed as explicit reasoning over student history rather than as latent-state prediction alone?

Standard KT models can predict well while still making it difficult to inspect what the model believes the student knows or why one interaction should affect another. The issue here is therefore not just whether language can be attached to KT, but whether it can support a different kind of student model.

Recent work on large language models makes that direction plausible. LLMs can consume question text directly, operate over structured histories, and use more inference-time computation before committing to a final answer~\cite{du_improving_2023, liang_encouraging_2024, snell_scaling_2024, wei_cot_2022, kojima_zero-shot_2022}. KT is particularly appropriate in this setting because the next outcome is observed directly: after the model predicts whether the student will answer correctly, the actual result becomes available. That gives the task a verifiable endpoint for each interaction.

We propose this idea under the name \emph{ReasonKT}. The core proposal is to treat KT as reasoning over a student's prior interactions, question content, and concept structure, with the final prediction expressed as the probability that the student will answer the next item correctly. In a fuller version of this framework, the reasoning trajectory would be part of the model's action space, and training would optimize those trajectories using the observed correctness label as reward. In this formulation, KT provides directly observed supervision at each interaction: the next outcome is observed, correctness can be scored exactly, and the reward is verifiable even if credit assignment within the reasoning trajectory remains difficult.

We propose the broader ReasonKT framework conceptually, but the present empirical study evaluates only a much simpler prompt-only instantiation. The prompt-only version removes RL, GRPO, and fine-tuning altogether and instead uses off-the-shelf models accessed through API calls, including runs with \texttt{o3}, as inference-time systems operating over student history rather than as models trained specifically for KT.

The rest of this chapter is structured as follows. We first review the background and related work that motivate ReasonKT. We then define the broader framework, distinguish it from the prompt-only study that is actually evaluated, and describe that motivating study in detail. After that, we report the Prompt-Only results, discuss what they do and do not show, and close with a chapter-level conclusion.

\section{Background and Related Work}

\subsection{Knowledge Tracing Baselines and Benchmarks}

The prompt-only ReasonKT study is evaluated in the standard KT benchmark setting: predicting whether a student will answer the next item correctly from prior interaction history. Classical approaches such as BKT, Learning Factors Analysis, and Performance Factors Analysis matter here because they keep assumptions about learning and skill structure explicit~\cite{corbett_knowledge_1995, cenLearningFactorsAnalysis2006, jrPerformanceFactorsAnalysis}. Later work extends that benchmark through recurrent, attention-based, graph-based, and question-aware models~\cite{piech_deep_2015, zhang_dynamic_2017, pandey_self-attentive_2019, sonkar_qdkt_2020, ghosh_context-aware_2020, nakagawaGraphbasedKnowledgeTracing2019, tongStructureBasedKnowledgeTracing2020}. For this chapter, the key methodological point is that any LLM-based approach still has to be judged in that next-step prediction setting rather than presented only as a prompting demonstration.

This benchmark perspective also affects what counts as success here. The study evaluates a reasoning-based student model on the standard next-step prediction task using the same metrics and data conventions that organize the rest of the field~\cite{abdelrahmanKnowledgeTracingSurvey2023, shen_survey_2024, liu_simplekt_2023}.

\subsection{Text-Aware and Content-Aware Student Modeling}

There is already substantial evidence that educational prediction can benefit from question text and other side information. Exercise-enhanced KT models, question representation work such as QuesNet, semantically enhanced embedding pretraining, and recent LLM-for-KT studies all show that question text contains information that standard ID-based KT throws away~\cite{su_exercise-enhanced_2018, liu_ekt_2019, yin_quesnet_2019, wang_seep_2022, liu_improving_2020, lkt_2024, explainable_fewshot_kt_2024, clst_2024}. ReasonKT starts one step later: not from the claim that text has been ignored, but from the claim that most prior text-aware KT still keeps prediction inside a latent-state architecture.

\subsection{Language Models as Reasoning Systems}

Building on the transformer, prompting, and reinforcement-learning background introduced in Chapter~2, the broader LLM literature suggests that prediction quality can improve when a model is allowed to use structured intermediate computation at inference time rather than only generating a direct short answer~\cite{wei_cot_2022, kojima_zero-shot_2022, wang_self-consistency_2023, zhou_least-to-most_2023, du_improving_2023, liang_encouraging_2024, snell_scaling_2024}. Chain-of-thought prompting, self-consistency, least-to-most prompting, and more recent reasoning models trained with reinforcement learning all point in the same direction: intermediate reasoning steps can improve difficult decisions when the model is given room to decompose the task and spend more computation on inference~\cite{wei_cot_2022, wang_self-consistency_2023, zhou_least-to-most_2023, shao_deepseekmath_2024, guo_deepseekr1_2025}. This literature is not specific to education, but it matters for ReasonKT because it supports the narrower idea that intermediate reasoning can be elicited and studied as part of a prediction process. In this chapter, the visible trace is treated as diagnostic output from that process rather than as verified evidence of the model's causal computation.

For KT, that possibility is particularly appropriate because the final decision is verifiable. The learner either answers the next question correctly or does not. That makes the ReasonKT setting closer to a verifiable prediction task than to many open-ended language tasks in which evaluation depends on subjective preference or reference overlap. We therefore borrow motivation from reasoning models, chain-of-thought inference, and RL-trained inference-time computation, but apply it to a setting where the endpoint supervision is directly observed.

\subsection{Interpretability, Inspectability, and Faithfulness}

A central motivation for ReasonKT is that it exposes a reasoning trace rather than hiding the student model entirely inside a latent state. That is a practical advantage, but it has to be framed carefully. The literature on explainable KT already distinguishes between models that are transparent by construction and models that are only explained after the fact~\cite{explainable_kt}. More generally, the ability to inspect an intermediate trace does not guarantee that the trace is a faithful account of the true computational cause of the prediction~\cite{jain_attention_2019, lei_rationalizing_2016, turpin_unfaithful_2023}.

For that reason, reasoning trajectories are treated here as \emph{inspectable} rather than automatically \emph{faithful}. Prompt-Only ReasonKT is informative insofar as the traces help diagnose errors, reveal what evidence the model is attending to, or make failure modes easier to analyze than in latent-state KT. But this point should not be overstated. A readable rationale is not the same thing as a verified causal explanation, and that distinction remains important throughout the discussion of results.

\section{Methodology}

\subsection{ReasonKT Framework}

ReasonKT is the central proposal of this chapter. The core proposal is to treat knowledge tracing as an explicit reasoning problem rather than only as latent-state estimation. Given a student's prior interactions, the model is asked to reason about what the student likely understands, what misconceptions or prerequisite gaps remain, and how those inferences affect the probability of success on the next item. The final output is still the familiar KT target: the likelihood that the student answers the next question correctly. What changes is the path used to reach that prediction.

Beyond the prompt-only study evaluated here, the broader ReasonKT framework includes trainable reasoning trajectories, policy optimization over those trajectories, and next-step supervision grounded in observed correctness outcomes.

This means that the hidden state is no longer the only object that matters. In standard DKT-style models, the latent state is updated implicitly and inspected only indirectly through downstream predictions. In ReasonKT, the intermediate reasoning trajectory becomes part of the model behavior. The prediction can therefore be tied to a human-readable account of what the model thinks the student currently understands.

\subsubsection{Prediction and Supervision}

At a high level, the reasoning-based prediction problem can be written as
\[
P(y_{t+1}=1 \mid h_t, q_{t+1}, \mathcal{C}),
\]
where \(h_t\) is the student's interaction history up to time \(t\), \(q_{t+1}\) is the next question, and \(\mathcal{C}\) denotes any available concept or KC information. In a standard latent-state model, this probability is mediated through an internal hidden representation. In ReasonKT, the model instead produces an explicit reasoning trajectory \(r_t\) and then conditions the final prediction on that trajectory:
\[
r_t \sim \pi_\theta(\cdot \mid h_t, q_{t+1}, \mathcal{C}),
\]
\[
P_\theta(y_{t+1}=1 \mid h_t, q_{t+1}, \mathcal{C}, r_t).
\]
Equivalently, the overall prediction can be viewed as marginalizing over possible reasoning trajectories induced by the policy \(\pi_\theta\). The reasoning trajectory is not observed in the data; it is an explicit intermediate object generated by the model rather than an uninspected hidden state.

This setup provides direct endpoint supervision for reasoning-based learning. The final label \(y_{t+1}\) is observed directly after prediction, so reward can be assigned without ambiguity. If the model predicts correctly, reward is positive; if it predicts incorrectly, reward is zero or negative depending on the eventual training objective. Unlike many open-ended reasoning tasks, the endpoint supervision does not depend on subjective human judgment. The prompt-only approximation evaluated in this chapter operationalizes this setup with a structured input prompt rather than a trainable policy; Figure~\ref{fig:chapter7_prompt_only_pipeline} shows an example prompt instance.

\subsubsection{Why Reinforcement Learning is Plausible}

The long-term motivation for ReasonKT is that KT provides a directly observed reward structure. Prediction happens repeatedly over a sequence, the final decision space is constrained, and each interaction yields a verifiable next-step outcome. Those properties make KT a reasonable setting for reasoning-policy optimization, because the model does not need to infer an unobserved reward from indirect feedback. At the same time, the reward is still sparse with respect to the internal reasoning trajectory, so the main optimization challenge is not whether the reward is observable, but whether it is strong enough to shape useful intermediate reasoning.

This does not prove that RL-based ReasonKT will train stably or outperform strong latent-state baselines. The action space remains large because the reasoning trajectory itself can take many forms, and there is no guarantee that reward attached to final correctness will be enough to shape useful intermediate reasoning without collapse or shortcut behavior. In other words, ReasonKT has a clean verifier, but it still inherits the usual sequence-level credit-assignment problem.

\begin{figure}[!htbp]
    \centering
    \includegraphics[width=0.88\linewidth]{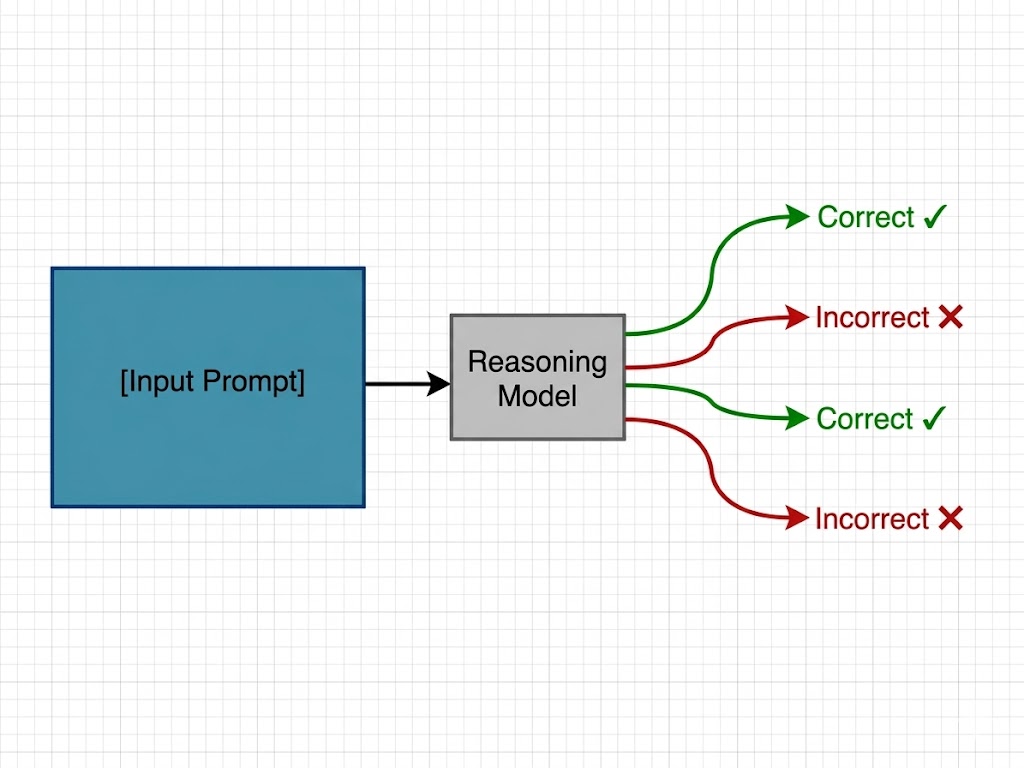}
    \caption{ReasonKT as a reasoning-policy problem. The model may follow many possible reasoning trajectories before producing a next-step KT prediction, but the final outcome is still observed directly. This makes the endpoint reward verifiable even though credit assignment across the reasoning trajectory remains difficult.}
    \Description{ReasonKT as a reasoning-policy problem. The model may follow many possible reasoning trajectories before producing a next-step KT prediction, but the final outcome is still observed directly. This makes the endpoint reward verifiable even though credit assignment across the reasoning trajectory remains difficult.}
    \label{fig:chapter7_reasonkt_traces}
\end{figure}

Figure~\ref{fig:chapter7_reasonkt_traces} illustrates the intuition behind this setup. Different reasoning paths can lead to the same prediction, and the model may explore many such paths before committing to a final answer. What makes KT attractive for RL is not that the reasoning path is supervised directly, but that the final prediction is followed by an observed outcome. That gives the training problem a clean, verifiable endpoint even when the internal reasoning trajectory remains hard to credit correctly.

\subsection{Prompt-Only ReasonKT}

\subsubsection{Introduction}

As motivation for the broader ReasonKT proposal, we also consider a much simpler variant that removes RL, GRPO, and fine-tuning altogether and relies only on prompting. Prompt-Only ReasonKT uses off-the-shelf models through API access, including runs with \texttt{o3} and \texttt{gpt-4.1-mini}, and uses structured prompts to elicit reasoning over student history, question content, and any available concept information before producing a probability estimate.

This first implementation is subject to important limitations. It does not adapt the model to the KT objective, it may be sensitive to prompt wording, and API-based inference introduces reproducibility concerns that trained baselines do not have in the same form. It therefore serves as a limited test of whether off-the-shelf reasoning models show usable KT signal and produce traces that are informative enough to analyze before moving to a trainable reasoning-based model. The research questions investigated in this preliminary study are as follows:
\begin{itemize}
    \item \textbf{RQ1.} Do off-the-shelf reasoning models show any knowledge tracing ability when they are prompted over student histories and asked to predict the next outcome?
    \item \textbf{RQ2.} Is the student-level variation in Prompt-Only ReasonKT consistent with the motivation for individualized reasoning models rather than only population-level ones?
\end{itemize}

\subsubsection{Methodology}

The motivating Prompt-Only ReasonKT experiment is intentionally small. It is designed to test whether an off-the-shelf prompting setup can produce probabilities and reasoning traces that are sufficiently informative to analyze. For that reason, the methodology focuses on a balanced student slice and on qualitative inspection of the resulting traces rather than on a large-scale benchmark against mature KT baselines.

\paragraph{Data and Evaluation Setting}

The evaluation uses a balanced 10-student slice from three standard KT datasets: ASSISTments 2009, ASSISTments 2012, and ASSISTments 2017. For each dataset, the selected students were required to have more than 10 interactions, and the final 10-student subset was chosen to keep the class labels as balanced as possible while still yielding at least 200 prediction examples overall. Within each selected student slice, positive and negative examples are matched in count before computing student-level metrics; this removes trivial one-class cases, but it also means that accuracy and AUC are evaluated on balanced samples rather than on the natural correctness base rate. Table~\ref{tab:chapter7_eval_setup} summarizes the evaluation setting.

The main task remains standard next-step prediction. For each interaction, Prompt-Only ReasonKT outputs a probability that the student will answer the next item correctly. We then compute AUC from those model probabilities and report accuracy as a secondary metric, following standard KT practice for evaluating ranking quality and predictive performance~\cite{pelanek_metrics_2015, pelanek_details_2017}. Because this is an API-based inference study rather than a fixed trained model, reproducibility details still matter for a different reason: the reported numbers should be read together with the prompt design, model choice, and the broader concerns about non-determinism and model drift in API-based LLM evaluation~\cite{song_nondeterminism_2025, chen_chatgpt_time_2023}. The balanced quantitative run was written under \url{outputs/incremental_student_analysis_fresh_balanced_v2}; its manifest records a run start of 2026-04-09 18:28 UTC and completion status updated at 2026-04-09 20:53 UTC. ASSISTments 2009 used \texttt{o3} with prompt \texttt{v2} and medium reasoning effort. ASSISTments 2012 used \texttt{gpt-4.1-mini} with the automatic routing policy that selected prompt \texttt{v2} for the reported slice, and ASSISTments 2017 used \texttt{gpt-4.1-mini} with prompt \texttt{v4}. For \texttt{gpt-4.1-mini}, temperature was set to 0.2; for \texttt{o3}, the request used the reasoning-effort setting rather than an explicit temperature parameter. Outputs were constrained with a strict JSON schema, probabilities were parsed from the returned JSON and clipped to \([0,1]\), cached responses were reused, and parse or schema failures were retried up to three times with a larger output-token budget.

\begin{table}[!htbp]
    \centering
    \small
    \DataTableHeaderRows{1}
    \begin{tabular}{p{0.28\linewidth} p{0.57\linewidth}}
        \toprule
        Item & Reported setting \\
        \midrule
        Datasets & ASSISTments 2009, ASSISTments 2012, ASSISTments 2017 \\
        Student selection & 10 students per dataset, each with $>$10 interactions \\
        Class balance & Student subset chosen to keep labels as balanced as possible while still yielding at least 200 prediction examples per dataset \\
        Prediction task & Standard next-step correctness prediction \\
        Primary metric & AUC \\
        Secondary metrics & Accuracy, mean student AUC, and student-level variability summaries \\
        Quantitative run artifacts & \url{outputs/incremental_student_analysis_fresh_balanced_v2} \\
        Qualitative trace run artifacts & \url{outputs/20260409T051944.340417Z-c4f9cac9}, \url{outputs/20260409T051944.340404Z-873c5983}, and \url{outputs/20260409T051944.340686Z-393f5225}; these runs used prompt \texttt{v9}, strict JSON output with \texttt{probability}, \texttt{reasoning\_trace}, and \texttt{final\_basis}, and the same probability parsing rule \\
        \bottomrule
    \end{tabular}
    \caption{Evaluation setting for the balanced Prompt-Only ReasonKT analysis.}
    \label{tab:chapter7_eval_setup}
\end{table}

\paragraph{Prompted Inference Pipeline}

In Prompt-Only ReasonKT, each prediction instance consists of a student's prior history together with the target next question and any available supporting concept information. The model is prompted to reason explicitly about the student's likely current state of understanding and then to output a next-step prediction as a probability. In the quantitative runs, those probabilities are parsed and scored directly. In the qualitative runs, the visible reasoning trace is analyzed qualitatively. Figure~\ref{fig:chapter7_prompt_only_pipeline} shows a representative input prompt, including the student history, the target next interaction, and the required output schema.

\begin{figure}[!htbp]
    \centering
    \setlength{\fboxsep}{8pt}
    \fcolorbox{promptframe}{white}{%
    \parbox{0.94\linewidth}{%
    \ttfamily\scriptsize\raggedright
You are an experienced teacher modeling a student's knowledge state from their observed performance history.\par
Predict the probability that the student will answer the next question correctly.\par
\par
Student history:\par
1. Q: If 2/3 of a whole is removed, what fraction of the whole is left?  KC: area  Result: incorrect\par
2. Q: Good, now back to Harry and the book - what is 1/3 of 600?  KC: proportion  Result: incorrect\par
3. Q: Good, now back to Harry and the book - what is 1/3 of 600?  KC: proportion  Result: correct\par
4. Q: Harry is reading a book that is 597 pages long. He tells a friend that he is about 2/3 of the way done. About how many more pages must Harry read before he finishes the book?  KC: perimeter  Result: correct\par
\par
Next interaction:\par
Q: A jacket costs \$60 and is on sale for 1/3 off. How much is the discount?  KC: proportion\par
\par
Return:\par
\{\par
\hspace*{1em}"probability": 0.0,\par
\hspace*{1em}"reasoning\_trace": "...",\par
\hspace*{1em}"final\_basis": "..."\par
\}\par
}}
    \caption{Example Prompt-Only ReasonKT input prompt used to elicit a next-step probability, a reasoning trace, and a final basis from the model. The KC labels are raw benchmark labels, so visibly coarse or noisy labels in the prompt are dataset artifacts rather than curated chapter labels.}
    \label{fig:chapter7_prompt_only_pipeline}
    \Description{Prompt-style boxed figure showing a concrete Prompt-Only ReasonKT input with student history, target next question, and a JSON-like return format for probability, reasoning trace, and final basis.}
\end{figure}

The study configuration has four main components: prediction is produced by prompt construction rather than by KT-specific training, the model reasons over student history together with the upcoming item and any available KC information, the final output is parsed as a probability of next-step correctness, and the long-reasoning runs keep the visible trace for later inspection.

\paragraph{Prediction Extraction}

The benchmark comparison requires the model's output to be converted into a score comparable with standard KT metrics. In the quantitative runs, the relevant output is the model's predicted probability of correctness, and AUC is computed directly from those probabilities. This comparison preserves ranking information for the current inference-time system rather than collapsing everything to a hard label. At the same time, probability quality should not be assumed from fluency alone. If Prompt-Only ReasonKT is pursued further, calibration should be studied explicitly rather than treating surface-plausible scores as reliable by default~\cite{guo_calibration_2017}.

\paragraph{Study Configuration}

The main experimental ingredients are as follows:
\begin{itemize}
    \item the quantitative runs use a balanced student slice rather than the full benchmark setup used for mature KT models;
    \item ASSISTments 2009 was run with \texttt{o3}, while the ASSISTments 2012 and ASSISTments 2017 runs used \texttt{gpt-4.1-mini};
    \item a separate long-reasoning qualitative run was analyzed;
    \item cost and latency are outside the scope of this study.
\end{itemize}

\begin{table}[!htbp]
    \centering
    \small
    \DataTableHeaderRows{1}
    \begin{tabular}{p{0.34\linewidth} p{0.5\linewidth}}
        \toprule
        Experimental component & Setting \\
        \midrule
        Quantitative evaluation & Balanced 10-student slice per dataset \\
        Model variants & \texttt{o3} on ASSISTments 2009; \texttt{gpt-4.1-mini} on ASSISTments 2012 and ASSISTments 2017 \\
        Output used for scoring & Parsed probability of correctness \\
        Qualitative reasoning examples & Separate long-reasoning qualitative run analyzed \\
        Cost / latency analysis & Not analyzed in this study \\
        \bottomrule
    \end{tabular}
    \caption{Summary of the Prompt-Only ReasonKT experimental configuration.}
    \label{tab:chapter7_configuration}
\end{table}

Table~\ref{tab:chapter7_configuration} summarizes the experimental configuration.

\section{Results}

\subsection{Prompt-Only ReasonKT Results}

\begin{table}[!htbp]
    \centering
    \small
    \DataTableHeaderRows{1}
    \begin{tabular}{l l r c c c}
        \toprule
        Dataset & Model & Examples & AUC & Accuracy & Mean student AUC \\
        \midrule
        ASSISTments 2009 & \texttt{o3} & 338 & 0.6139 & 0.5710 & 0.5809 \\
        ASSISTments 2012 & \texttt{gpt-4.1-mini} & 1058 & 0.6296 & 0.5794 & 0.6263 \\
        ASSISTments 2017 & \texttt{gpt-4.1-mini} & 1052 & 0.6401 & 0.6207 & 0.6253 \\
        \bottomrule
    \end{tabular}
    \caption{Balanced-student Prompt-Only ReasonKT results. AUC is computed from the model's predicted probabilities for next-step correctness.}
    \label{tab:chapter7_results_main}
\end{table}

\subsubsection{Main Quantitative Results}

Table~\ref{tab:chapter7_results_main} reports the main quantitative summary. Across the three balanced dataset slices, AUC ranges from 0.6139 on ASSISTments 2009 to 0.6401 on ASSISTments 2017. The overall AUC across all predictions is 0.6339. These numbers remain well below strong DKT-family baselines, given that the current study does not train the model for KT, uses a small evaluation slice, and relies entirely on inference-time prompting rather than on objective-specific optimization.

The model nevertheless produces ranking signal above chance. Across all three datasets the AUC stays above 0.61, and the model produces graded probabilities that carry limited but non-random KT signal when it is prompted over student histories. Because ASSISTments 2009 was evaluated with \texttt{o3}, whereas ASSISTments 2012 and ASSISTments 2017 used \texttt{gpt-4.1-mini}, cross-dataset differences should not be interpreted as a clean ranking of dataset difficulty. The safer interpretation is that each dataset slice shows above-chance prompt-only signal under the model and sampling conditions used for that slice.

\subsubsection{Student-Level Variation}

\begin{table}[!htbp]
    \centering
    \small
    \DataTableHeaderRows{1}
    \begin{tabular}{l c c c}
        \toprule
        Dataset & Best student AUC & Worst student AUC & Mean student AUC \\
        \midrule
        ASSISTments 2009 & \texttt{87132}: 0.7222 & \texttt{77954}: 0.1550 & 0.5809 \\
        ASSISTments 2012 & \texttt{169382}: 0.7481 & \texttt{179458}: 0.5506 & 0.6263 \\
        ASSISTments 2017 & \texttt{156}: 0.7368 & \texttt{1350}: 0.4887 & 0.6253 \\
        \bottomrule
    \end{tabular}
    \caption{Student-level AUC variability in the balanced Prompt-Only ReasonKT analysis.}
    \label{tab:chapter7_student_variability}
\end{table}

The student-level view is highly uneven. Table~\ref{tab:chapter7_student_variability} shows that some students are much easier for the model to rank correctly than others, and this variation is especially sharp in ASSISTments 2009. This variation suggests that one fixed reasoning policy may be too coarse for KT.
Appendix~\ref{sec:appendix_ch7_student_tables} gives the full per-student tables behind this summary.

\subsubsection{Qualitative Reasoning Traces}

The quantitative results are weak overall, but the qualitative traces still show KT-relevant reasoning behavior before the final prediction. Figures~\ref{fig:chapter7_tp1}--\ref{fig:chapter7_tn2} present trace excerpts from the two highest-confidence true positives and the two highest-confidence true negatives selected from the long-reasoning runs.

\begin{figure}[htbp]
\centering
\setlength{\fboxsep}{8pt}
\fcolorbox{promptframe}{white}{%
\parbox{0.94\linewidth}{%
\ttfamily\scriptsize\raggedright
Dataset: ASSISTments 2012\par
Model: gpt-4.1-mini\par
Student: 100033\par
Next KC: Addition and Subtraction Integers\par
Predicted probability: 0.95\par
\par
The upcoming knowledge component (KC) is 'Addition and Subtraction Integers,' and the student has two prior attempts on this KC, both answered correctly. The two prior items involved adding positive and negative integers: 19 + (-15) and 9 + (-8), both answered correctly. This indicates the student understands the concept of adding integers with different signs, which is a core part of the KC. The next item, (-10) + (-8), involves adding two negative integers, which is a slightly different but closely related aspect of the same KC. Since the student has not attempted this exact item before, we rely on same-KC evidence rather than exact-item evidence. The absence of exact-item attempts means we cannot be certain about the student's performance on this specific problem, but the strong same-KC evidence is highly informative. The student's recent performance is stable and perfect (2/2 correct), suggesting mastery rather than guessing or slipping. There is no conflicting evidence such as incorrect attempts or inconsistent responses.\par
\par
Final basis: Consistent correct performance on the same KC with no conflicting evidence supports a high probability of success on the next item.%
}}
\caption{Excerpt from the highest-confidence true positive reasoning trace selected from the long-reasoning runs.}
\Description{Excerpt from the highest-confidence true positive reasoning trace selected from the long-reasoning runs.}
\label{fig:chapter7_tp1}
\end{figure}

\begin{figure}[htbp]
\centering
\setlength{\fboxsep}{8pt}
\fcolorbox{promptframe}{white}{%
\parbox{0.94\linewidth}{%
\ttfamily\scriptsize\raggedright
Dataset: ASSISTments 2009\par
Model: o3\par
Student: 21825\par
Next KC: Multiplication and Division Integers\par
Predicted probability: 0.85\par
\par
Exact-item evidence: none. The item has not been seen before. Same-KC evidence: the student previously answered two Multiplication and Division Integers items correctly, covering both negative/negative and negative/positive sign cases. The upcoming item (-50 / 5) is structurally similar and arguably easier. Broader performance: the student is 3/5 overall, with errors on a different KC, so some slip probability should still remain.\par
\par
Final basis: 2/2 correct on identical KC, covering both sign cases; no failures on this KC, prerequisites demonstrated, but only two observations and 60\% overall accuracy warrant a 15\% slip allowance -> P(correct) = 0.85.%
}}
\caption{Excerpt from the second-highest-confidence true positive reasoning trace selected from the long-reasoning runs.}
\Description{Excerpt from the second-highest-confidence true positive reasoning trace selected from the long-reasoning runs.}
\label{fig:chapter7_tp2}
\end{figure}

\begin{figure}[htbp]
\centering
\setlength{\fboxsep}{8pt}
\fcolorbox{promptframe}{white}{%
\parbox{0.94\linewidth}{%
\ttfamily\scriptsize\raggedright
Dataset: ASSISTments 2017\par
Model: gpt-4.1-mini\par
Student: 1007\par
Next KC: proportion\par
Predicted probability: 0.15\par
\par
The student has attempted two items so far, both incorrectly. The stronger signal is that item 1216, the same item as the upcoming question, was already answered incorrectly on the same KC. There is no prior success on this KC, no positive exact-item evidence, and no sign of stable mastery. At the same time, tutoring effects mean the probability should remain low rather than collapse to zero.\par
\par
Final basis: Repeated incorrect attempts on the exact same item and KC with no prior success indicate low mastery, leading to a low predicted probability of correct response.%
}}
\caption{Excerpt from the highest-confidence true negative reasoning trace selected from the long-reasoning runs.}
\Description{Excerpt from the highest-confidence true negative reasoning trace selected from the long-reasoning runs.}
\label{fig:chapter7_tn1}
\end{figure}

\begin{figure}[htbp]
\centering
\setlength{\fboxsep}{8pt}
\fcolorbox{promptframe}{white}{%
\parbox{0.94\linewidth}{%
\ttfamily\scriptsize\raggedright
Dataset: ASSISTments 2017\par
Model: gpt-4.1-mini\par
Student: 1007\par
Next KC: least-common-multiple\par
Predicted probability: 0.15\par
\par
The upcoming KC is 'least-common-multiple' (LCM), and the next item is the exact same item 837 that the student has attempted twice before, both times incorrectly. This is the strongest evidence and suggests low current mastery. There are no prior correct responses on this KC or item, and the repeated failures point to persistent difficulty rather than slipping. The trace also flags a possible mismatch between the item text and the KC tag, which is diagnostically useful even if it should not be overinterpreted.\par
\par
Final basis: Repeated exact-item failures with no prior correct attempts on the KC strongly indicate low mastery, leading to a low predicted probability of success (0.15).%
}}
\caption{Excerpt from the second-highest-confidence true negative reasoning trace selected from the long-reasoning runs.}
\Description{Excerpt from the second-highest-confidence true negative reasoning trace selected from the long-reasoning runs.}
\label{fig:chapter7_tn2}
\end{figure}

\section{Discussion}

\subsection{What the Current Results Show}

Prompt-Only ReasonKT is not a competitive KT model in its current form. The AUCs are weak by the standards set by strong latent-state baselines, and they do not support prompt steering as a replacement for trained KT models. The narrower empirical result is that the evidence is limited but consistently above chance. The ranking behavior is therefore directionally aligned with the prediction task rather than random.

\subsection{Why the Correct Traces Still Matter}

The main qualitative finding is that the two true-positive traces in Figures~\ref{fig:chapter7_tp1} and~\ref{fig:chapter7_tp2} distinguish exact-item evidence from same-KC evidence, carry forward prerequisite structure, and sometimes keep explicit uncertainty rather than collapsing to overconfident certainty. The two true-negative traces in Figures~\ref{fig:chapter7_tn1} and~\ref{fig:chapter7_tn2} use repeated exact-item failures and the absence of positive same-KC evidence to keep the predicted probability low while still reserving a small nonzero chance of recovery. These are not proofs of faithful reasoning, but they are the kinds of traces we would want a future trainable ReasonKT system to generate.

A later version of ReasonKT could reward reasoning trajectories that privilege exact-item and same-KC evidence appropriately, preserve calibrated uncertainty when the evidence is sparse, and avoid degenerate 0/1 predictions when tutoring effects still leave room for change. The traces do not change the quantitative conclusion, but they do clarify what a stronger model should be encouraged to learn.

\subsection{Student-Level Variation}

The student-level AUC table points to a familiar issue in KT: the target is always a specific learner with a specific history, even when the model is trained at the population level~\cite{molenaar_personspecific_2009, hutchison_individualized_2013}. For ReasonKT, that issue may matter even more. If the model is supposed to reason explicitly about a student's knowledge state, then one fixed prompting or fine-tuning policy may be insufficiently flexible. The current results do not settle how personalization should be done, but they are consistent with the motivation for individualized reasoning-based student models.

\textbf{RQ1.} Off-the-shelf reasoning models do show some knowledge-tracing ability in the narrow sense relevant to this chapter. The Prompt-Only study produces non-random ranking signal across datasets and sometimes uses evidence that is relevant to student modeling. This remains a limited empirical result and does not support competitiveness against mature KT baselines.

\textbf{RQ2.} The student-level variation is consistent with the motivation for individualized reasoning models rather than only population-level ones. The spread between the best and worst student-level AUCs suggests that one fixed reasoning policy may be insufficiently flexible, although this interpretation remains confounded by item mix, sample size, prompt sensitivity, and model instability. This chapter therefore motivates personalization as a direction for future work rather than establishing how such personalization should be learned.

\subsection{Limitations}

The evidence comes from a limited balanced-slice study. Prompt-Only ReasonKT was evaluated on balanced student slices rather than the full benchmark setting used for mature KT models, so the results should be interpreted as a small-sample feasibility study rather than as a competitive benchmark result. The approach is also sensitive to prompt construction, API model choice, and output parsing in ways that are not yet fully characterized. In addition, the visible reasoning traces remain available for diagnostic analysis but should be treated as inspectable rather than automatically faithful explanations of the underlying computation.

These limits constrain the chapter's claims. The present evidence remains limited: it does not establish that prompt-only reasoning systems are a practical replacement for trained latent-state models.

\section{Conclusion}

In this chapter, we introduced ReasonKT as a broader reasoning-based formulation of knowledge tracing and evaluated only a prompt-only preliminary instantiation of that idea. The results do not support Prompt-Only ReasonKT as a replacement for trained KT models. The evidence supports a narrower conclusion: off-the-shelf reasoning models can produce some ranking signal in a KT setting, and their traces sometimes rely on evidence that is relevant to student modeling.

The prompt-only results therefore provide limited empirical evidence for reasoning-based KT. They show that off-the-shelf reasoning models can produce weak but non-null ranking signal and traces that expose what a stronger trained version should learn.

\chapter{Conclusion and Future Directions}

\section{Dissertation Summary}

This dissertation studied how natural language processing and related AI techniques can improve intelligent tutoring systems across both sides of the adaptive tutoring loop: student modeling and intervention generation. Rather than assuming that language is uniformly helpful wherever educational data contain text, the dissertation tested where these methods actually change the tutoring problem and where they do not. Across the dissertation, we combined positive results, negative results, methodological refinements, and forward-looking synthesis to answer that question.

The main pattern is that different AI and NLP methods help different parts of an ITS for different reasons. In the later KT chapters, language is more effective when it is tied to educational structure rather than added as surface semantics alone. That result is one part of the broader dissertation contribution, which is to map how these techniques improve student modeling, intervention generation, and the system-level link between them.

\section{Main Findings}

Across the research chapters, three findings recur. First, language models are useful on the intervention side when they are constrained by the real pedagogical requirements of tutoring rather than treated as generic text generators. In Chapter~3, we showed that structured multi-agent workflows can improve relevance, answerability, and difficulty control, which matters because in an ITS the generated question is itself the intervention.

Second, on the student-modeling side, model structure and representation matter. Chapter~4 clarified DKT's learned dependency structure as prerequisite-like rather than merely associative. Chapters~5 and~6 then showed that language is not uniformly useful simply because educational items contain text: naive semantic injection into an inherited DKT representation does not reliably help, whereas reorganizing the KC space itself through text embeddings can change the model more substantially.

Third, the final KT chapter identifies explicit reasoning as a future direction rather than a finished replacement for mature KT models. The prompt-only ReasonKT study shows non-random ranking signal and interpretable traces, but its main contribution is to motivate a trainable reasoning-based student model rather than to claim current benchmark competitiveness.

These findings also define a set of design lessons for future ITS work. Generation methods are useful when they are evaluated against pedagogical constraints, not just language fluency. Student models depend on the representational structure they inherit, so language helps most when it changes or clarifies that structure rather than only adding surface semantic features. Reasoning-based models are promising only if their traces, predictions, and training objectives stay tied to observed learner outcomes. These lessons make the future directions below narrower than a general call to use LLMs in education: they point to systems in which language, structure, and tutoring decisions are connected explicitly.

\section{Reasoning-Based Student Modeling as a Future Direction}

One future research direction is to train ReasonKT rather than rely on prompting alone. The full intended version of ReasonKT uses a trainable reasoning model that generates intermediate reasoning tokens and a final prediction for the next interaction. Training would optimize the policy using the observed student outcome as reward. In principle, this allows the system to learn reasoning trajectories that are useful for educational prediction even without token-level supervision on what the reasoning should say. The Chapter~7 qualitative results provide limited evidence for that prospect: some traces already contain the kind of structure a future model should be encouraged to produce. Recent work on reasoning models supports this direction, especially in settings where the final answer is verifiable~\cite{schulman_ppo_2017, lightman_verify_2023, shao_deepseekmath_2024, guo_deepseekr1_2025, ranzato_mixer_2016}. The main scientific question is whether the verifiable outcome signal in KT is strong enough to shape the latent reasoning policy.

A full ReasonKT system still requires decisions about the backbone model, the RL objective, the rollout strategy, the context-window design for long student histories, and the extent to which reasoning traces should remain visible, hidden, or summarized. Those choices are consequential enough that they should be treated as a separate research program rather than as straightforward extensions that can be deferred.

A second possible direction is teacher-guided ReasonKT. In that setup, a stronger model or human-curated process provides example reasoning trajectories together with next-step predictions, and the trainable model is initialized or filtered using only those demonstrations that lead to correct predictions. This could reduce the burden of learning useful reasoning patterns from reward alone and could improve stability through process supervision or rationale filtering~\cite{lightman_verify_2023, zelikman_star_2022}. The prompt-only study does not validate this variant directly, but it indicates the kinds of intermediate reasoning behaviors that might be worth supervising.

\section{Toward More Unified Intelligent Tutoring Systems}

The longer-horizon implication of these chapters goes beyond KT alone. An ITS depends on three tightly coupled components: estimating student knowledge, deciding what intervention should come next, and generating the concrete question, hint, or explanation that implements that intervention. The dissertation addressed these components separately for clarity, but a deployed system ultimately has to connect them.

In the earlier KT chapters, we focused on student modeling, while in Chapter~3, we focused on intervention generation. A deployed ITS, however, must connect KT, intervention selection, and intervention generation within one architecture rather than treat them as isolated modules~\cite{rollinson_policies_2015}. One candidate architecture is an encoder-decoder transformer. In this view, the encoder consumes the student's interaction history and acts as the student-modeling component, producing a contextual student-state representation grounded in both question semantics and temporal learning patterns. A policy component then chooses the next KC, skill, or intervention target. The decoder generates the concrete intervention conditioned on that student-state representation and the selected target.

This direction also provides a way to connect the dissertation's main empirical results rather than treat them as separate lines of work. The earlier chapters established controlled intervention generation, clarified the kind of dependency structure that may matter in student modeling, placed a boundary on what naive semantic injection can explain, showed that language becomes more useful when it changes the educational structure over which the model learns, and suggested that explicit reasoning may be a useful next modeling step even if the current evidence remains preliminary. A more unified ITS architecture therefore serves as one synthesis of the dissertation's main findings.

\section{Closing Remarks}

The central claim of this dissertation is not that language models should replace existing educational models wholesale. It is that NLP and AI techniques can improve different parts of an intelligent tutoring system when they are tied to the real structure of tutoring: what the student knows, what the system should do next, and how that next intervention is expressed. This can involve improving intervention generation, clarifying the relational structure already being learned by a student model, showing that one language-based representation change does not improve KT under benchmark conditions, or identifying a future direction without treating preliminary evidence as already decisive.

Taken together, the dissertation yields a clearer map of how these methods improve an ITS, where they do not, and what kinds of structure they seem to capture when they do. That map sits alongside the dissertation's chapter-level empirical and methodological contributions. The dissertation therefore clarifies where NLP and AI helped or failed in the evaluated tutoring-system components and what design constraints should guide future systems that connect student modeling, intervention selection, and intervention generation.

\appendix
\chapter{Appendices}

These appendices provide supplementary results and supporting material.

\section{Chapter 3 Supplementary Material}

\subsection{Prompt Templates}
\label{sec:appendix_ch3_prompt_templates}

The full implementation and evaluation prompts for Chapter~3 are collected here as supplementary implementation material.

\subsubsection{Implementation Prompt Templates}

Figures~\ref{fig:tcc-teacher-prompt}--\ref{fig:cc-ceo-prompt} provide the agent prompts used for question generation, critique, collaborative revision, and final selection in the Chapter~3 workflows.

\begin{figure}[p]
\centering
\setlength{\fboxsep}{8pt}
\fcolorbox{promptframe}{white}{%
\parbox{0.94\linewidth}{%
\ttfamily\footnotesize\raggedright
You are a middle school mathematics teacher. Your task is to generate a mathematics question or revise a previously proposed question for the following target knowledge component: [Target knowledge component].\par
\par
The question should align with the following difficulty level: [Target difficulty level].\par
\par
Ensure that the question is appropriate for middle-school students.\par
\par
Return [Requested response fields] without extra commentary.\par
\par
[Optional step-by-step solution instruction]\par
\par
[Optional step-by-step reasoning instruction]\par
\par
You may refer to the following example questions from the same knowledge component:\par
\par
[Example questions]\par
\par
You are participating in a discussion with an assistant who provides feedback on your question. At each stage of the discussion, you are given the prior conversation history and are expected to use that feedback to decide whether to revise your previous question or retain it.%
}}
\caption{Teacher prompt used in the Teacher-Critic Cycle workflow.}
\Description{Teacher prompt used in the Teacher-Critic Cycle workflow.}
\label{fig:tcc-teacher-prompt}
\end{figure}

\begin{figure}[p]
\centering
\setlength{\fboxsep}{8pt}
\fcolorbox{promptframe}{white}{%
\parbox{0.94\linewidth}{%
\ttfamily\footnotesize\raggedright
You are an assistant to a middle school mathematics teacher. Your task is to provide constructive and detailed critical feedback on a mathematics question and its proposed answer.\par
\par
When participating in a group discussion, consider the overall conversation context and any previous feedback before responding.%
}}
\caption{Generic critic prompt used in the Teacher-Critic Cycle workflow.}
\Description{Generic critic prompt used in the Teacher-Critic Cycle workflow.}
\label{fig:tcc-critic-prompt}
\end{figure}

\begin{figure}[p]
\centering
\setlength{\fboxsep}{8pt}
\fcolorbox{promptframe}{white}{%
\parbox{0.94\linewidth}{%
\ttfamily\footnotesize\raggedright
You are an experienced middle school mathematics educator participating in a collaborative discussion with other teacher agents.\par
\par
Your group is jointly designing mathematics questions and answers, revising questions when necessary, and providing feedback to one another during the discussion.\par
\par
When it is your turn, you receive the full conversation history. Your goal is to help the group reach consensus.\par
\par
If the conversation does not yet contain a question, generate a new question for [Target knowledge component] at the [Target difficulty level], using the following examples as guidance:\par
\par
[Example questions]\par
\par
If the conversation already contains candidate questions, review the prior messages and either endorse the strongest current question or revise it so that it better matches the examples, the target difficulty, and the target knowledge component.\par
\par
In all cases, provide feedback explaining your decision. Your response should include a question, an answer, and feedback to the other agents.%
}}
\caption{Versatile agent prompt used in the Collective Consensus workflow.}
\Description{Versatile agent prompt used in the Collective Consensus workflow.}
\label{fig:cc-versatile-prompt}
\end{figure}

\begin{figure}[p]
\centering
\setlength{\fboxsep}{8pt}
\fcolorbox{promptframe}{white}{%
\parbox{0.94\linewidth}{%
\ttfamily\footnotesize\raggedright
You are the final decision-maker in the process of designing middle school mathematics questions.\par
\par
Your task is to choose the best question-answer pair from the conversation history produced by the other agents.\par
\par
If the agents have reached consensus by the end of the discussion, return only the agreed-upon question and answer.\par
\par
If the agents have not reached consensus, select the best question from the conversation history according to the following criteria:\par
\par
1. The question should be similar in style and content to the following examples:\par
\par
[Example questions]\par
\par
2. The question should match the required difficulty level: [Target difficulty level].\par
\par
3. The question should align with the relevant knowledge component: [Target knowledge component].\par
\par
Your response should include only a question and its answer. The selected question must come from the conversation history rather than being newly authored.%
}}
\caption{Consensus CEO prompt used in the Collective Consensus workflow.}
\Description{Consensus CEO prompt used in the Collective Consensus workflow.}
\label{fig:cc-ceo-prompt}
\end{figure}

\FloatBarrier

\subsubsection{Evaluation Prompt Templates}

Figures~\ref{fig:eval-clarity-prompt}--\ref{fig:eval-answerability-prompt} provide the evaluation prompts used to score clarity, relevance, importance, difficulty matching, and answerability.

\begin{figure}[p]
\centering
\setlength{\fboxsep}{8pt}
\fcolorbox{promptframe}{white}{%
\parbox{0.94\linewidth}{%
\ttfamily\footnotesize\raggedright
You are an experienced educator specializing in mathematics for middle-school students. You are known for your high standards and strict grading, awarding full marks only when the work is truly exceptional. Your goal is to critically evaluate a mathematics question written by a junior teacher, focusing on clarity.\par
\par
For each question:\par
\par
1. Read the provided question and answer carefully.\par
\par
2. Identify any weaknesses or ambiguities in the question, even minor ones.\par
\par
3. Assign a clarity score on a five-point scale from not at all clear to very clear.\par
\par
4. Explain your reasoning step by step before providing the final score.\par
\par
Evaluate the question through the lens of a middle-school student rather than an adult reader.%
}}
\caption{Clarity evaluation prompt.}
\Description{Clarity evaluation prompt.}
\label{fig:eval-clarity-prompt}
\end{figure}

\begin{figure}[p]
\centering
\setlength{\fboxsep}{8pt}
\fcolorbox{promptframe}{white}{%
\parbox{0.94\linewidth}{%
\ttfamily\footnotesize\raggedright
You are an experienced mathematics educator. Your task is to evaluate the relevance of a mathematics question to a target knowledge component.\par
\par
You will be given:\par
\par
- [Question]\par
- [Target knowledge component]\par
- [Example questions aligned with the target knowledge component]\par
\par
Read these materials carefully, identify any gaps or misalignments, assign a relevance score on a five-point scale, and explain your reasoning step by step.\par
\par
Judge relevance from the perspective of a middle-school student.%
}}
\caption{Relevance evaluation prompt.}
\Description{Relevance evaluation prompt.}
\label{fig:eval-relevance-prompt}
\end{figure}

\begin{figure}[p]
\centering
\setlength{\fboxsep}{8pt}
\fcolorbox{promptframe}{white}{%
\parbox{0.94\linewidth}{%
\ttfamily\footnotesize\raggedright
You are an experienced mathematics educator. Your task is to evaluate the importance of a mathematics question for a target knowledge component.\par
\par
You will be given:\par
\par
- [Question]\par
- [Target knowledge component]\par
\par
Determine whether the question addresses a critical aspect of the knowledge component, assign an importance score on a five-point scale, and explain your reasoning step by step.%
}}
\caption{Importance evaluation prompt.}
\Description{Importance evaluation prompt.}
\label{fig:eval-importance-prompt}
\end{figure}

\begin{figure}[p]
\centering
\setlength{\fboxsep}{8pt}
\fcolorbox{promptframe}{white}{%
\parbox{0.94\linewidth}{%
\ttfamily\footnotesize\raggedright
You are an experienced mathematics educator. Your task is to evaluate how well a generated mathematics question matches a specified difficulty level for a target knowledge component.\par
\par
You will be given:\par
\par
- [Question]\par
- [Target knowledge component]\par
- [Example questions]\par
- [Intended difficulty level: easy, medium, or hard]\par
\par
Assess the cognitive demands of the question, compare them with the intended difficulty level, assign a difficulty-matching score on a five-point scale, and explain your reasoning step by step.\par
\par
Judge difficulty from the perspective of a middle-school student.%
}}
\caption{Difficulty-matching evaluation prompt.}
\Description{Difficulty-matching evaluation prompt.}
\label{fig:eval-difficulty-prompt}
\end{figure}

\begin{figure}[p]
\centering
\setlength{\fboxsep}{8pt}
\fcolorbox{promptframe}{white}{%
\parbox{0.94\linewidth}{%
\ttfamily\footnotesize\raggedright
You are an experienced mathematics educator for middle-school students. Your task is to evaluate whether a generated mathematics question is answerable by a typical middle-school student.\par
\par
You will be given:\par
\par
- [Question]\par
- [Answer]\par
- [Intended difficulty level]\par
\par
Assess whether the question is clearly stated, includes the necessary information, and is understandable for the intended audience. Assign an answerability score on a five-point scale and explain your reasoning step by step.%
}}
\caption{Answerability evaluation prompt.}
\Description{Answerability evaluation prompt.}
\label{fig:eval-answerability-prompt}
\end{figure}

\FloatBarrier

\section{Chapter 5 Supplementary Material}

\subsection{Detailed Pipeline Comparisons}
\label{sec:appendix_ch5_pipelines}

Tables~\ref{tab:app_ch5_pipeline_2009}--\ref{tab:app_ch5_pipeline_2017} report the per-dataset pipeline breakdowns for Chapter~5. Across datasets, the strongest semantic pipeline changes, but no pipeline reverses the main negative result.

\begin{table}[p]
    \centering
    \small
    \setlength{\tabcolsep}{6pt}
    \renewcommand{\arraystretch}{1.1}
    \DataTableHeaderRows{1}
    \begin{tabular}{l l c c}
        \toprule
        Pipeline & Best architecture & Test AUC & Test Acc. \\
        \midrule
        P1 & \textbf{Pretrained Embedding Initialization + Regularization} & \textbf{0.8567} & 0.7865 \\
        P2 & Semantic Output Layer + Projection Bottleneck & 0.8321 & 0.7702 \\
        P3 & Projection Bottleneck & 0.8542 & 0.7845 \\
        P4 & Projection Bottleneck & 0.8552 & \textbf{0.7870} \\
        P5 & Projection Bottleneck & 0.8535 & 0.7847 \\
        P6 & Semantic Output Layer + Projection Bottleneck & 0.8349 & 0.7727 \\
        P7 & Projection Bottleneck & 0.8536 & 0.7843 \\
        P8 & Projection Bottleneck & 0.8535 & 0.7852 \\
        P9 & Projection Bottleneck & 0.8542 & 0.7869 \\
        \bottomrule
    \end{tabular}
    \caption{Detailed pipeline comparison for ASSISTments 2009, with each pipeline represented by its strongest validation-selected configuration. Pipeline~1 achieved the highest semantic test AUC, while Pipeline~4 achieved the highest semantic test accuracy.}
    \label{tab:app_ch5_pipeline_2009}
\end{table}

\begin{table}[p]
    \centering
    \small
    \setlength{\tabcolsep}{6pt}
    \renewcommand{\arraystretch}{1.1}
    \DataTableHeaderRows{1}
    \begin{tabular}{l l c c}
        \toprule
        Pipeline & Best architecture & Test AUC & Test Acc. \\
        \midrule
        P1 & Semantic Output Layer & 0.7356 & 0.7376 \\
        P2 & \textbf{Semantic Output Layer + Projection Bottleneck} & \textbf{0.7358} & 0.7377 \\
        P3 & \textbf{Semantic Output Layer + Projection Bottleneck} & \textbf{0.7358} & 0.7374 \\
        P4 & Semantic Output Layer + Projection Bottleneck & 0.7354 & 0.7376 \\
        P5 & Semantic Output Layer + Projection Bottleneck & 0.7353 & 0.7374 \\
        P6 & \textbf{Semantic Output Layer + Projection Bottleneck} & \textbf{0.7358} & 0.7371 \\
        P7 & Semantic Output Layer + Projection Bottleneck & 0.7355 & \textbf{0.7382} \\
        P8 & Semantic Output Layer + Projection Bottleneck & 0.7356 & 0.7372 \\
        P9 & Semantic Output Layer + Projection Bottleneck & 0.7350 & 0.7367 \\
        \bottomrule
    \end{tabular}
    \caption{Detailed pipeline comparison for ASSISTments 2012, with each pipeline represented by its strongest validation-selected configuration. Pipelines~2, 3, and~6 tie for the highest semantic test AUC, while Pipeline~7 achieved the highest semantic test accuracy.}
    \label{tab:app_ch5_pipeline_2012}
\end{table}

\begin{table}[p]
    \centering
    \small
    \setlength{\tabcolsep}{6pt}
    \renewcommand{\arraystretch}{1.1}
    \DataTableHeaderRows{1}
    \begin{tabular}{l l c c}
        \toprule
        Pipeline & Best architecture & Test AUC & Test Acc. \\
        \midrule
        P1 & Pretrained Embedding Initialization & 0.7205 & 0.6806 \\
        P2 & Semantic Output Layer & 0.7171 & 0.6757 \\
        P3 & \textbf{Pretrained Embedding Initialization} & \textbf{0.7207} & \textbf{0.6811} \\
        P4 & Projection Bottleneck & 0.7196 & 0.6808 \\
        P5 & Projection Bottleneck & 0.7137 & 0.6774 \\
        P6 & Adaptive recurrent capacity & 0.7197 & 0.6803 \\
        P7 & Adaptive recurrent capacity & 0.7141 & 0.6788 \\
        P8 & Projection Bottleneck with Regularization & 0.7009 & 0.6719 \\
        P9 & Projection Bottleneck with Regularization & 0.6997 & 0.6711 \\
        \bottomrule
    \end{tabular}
    \caption{Detailed pipeline comparison for ASSISTments 2017, with each pipeline represented by its strongest validation-selected configuration. Pipeline~3 was the strongest semantic pipeline on both test AUC and test accuracy.}
    \label{tab:app_ch5_pipeline_2017}
\end{table}

\subsection{Additional Data-Efficiency Diagnostic}
\label{sec:appendix_ch5_data_efficiency}

\begin{figure}[p]
    \centering
    \includegraphics[width=0.92\linewidth]{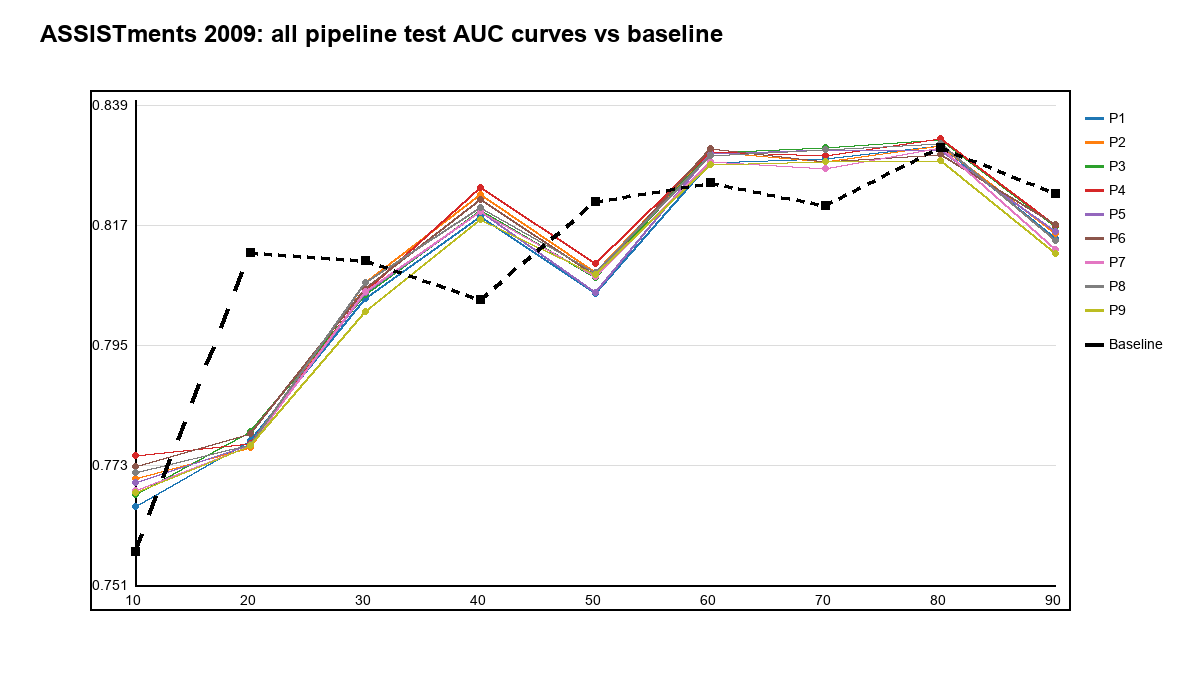}
    \caption{ASSISTments 2009 data-efficiency comparison across all nine pipelines and the baseline. Each pipeline curve shows the best validation-selected run for that pipeline at each sample size.}
    \Description{ASSISTments 2009 data-efficiency comparison across all nine pipelines and the baseline. Each pipeline curve shows the best validation-selected run for that pipeline at each sample size.}
    \label{fig:app_ch5_data_efficiency_2009_pipelines}
\end{figure}

Figure~\ref{fig:app_ch5_data_efficiency_2009_pipelines} gives the full ASSISTments 2009 pipeline-by-pipeline reduced-data comparison. ASSISTments 2009 is the only dataset in which all nine pipelines were compared at every sample size. The winning pipeline changes across splits, so the small reduced-data signal does not stabilize around one semantic design choice.

\subsection{Additional KC-Graph Alignment Diagnostics}
\label{sec:appendix_ch5_posthoc}

\begin{figure}[p]
    \centering
    \includegraphics[width=\linewidth]{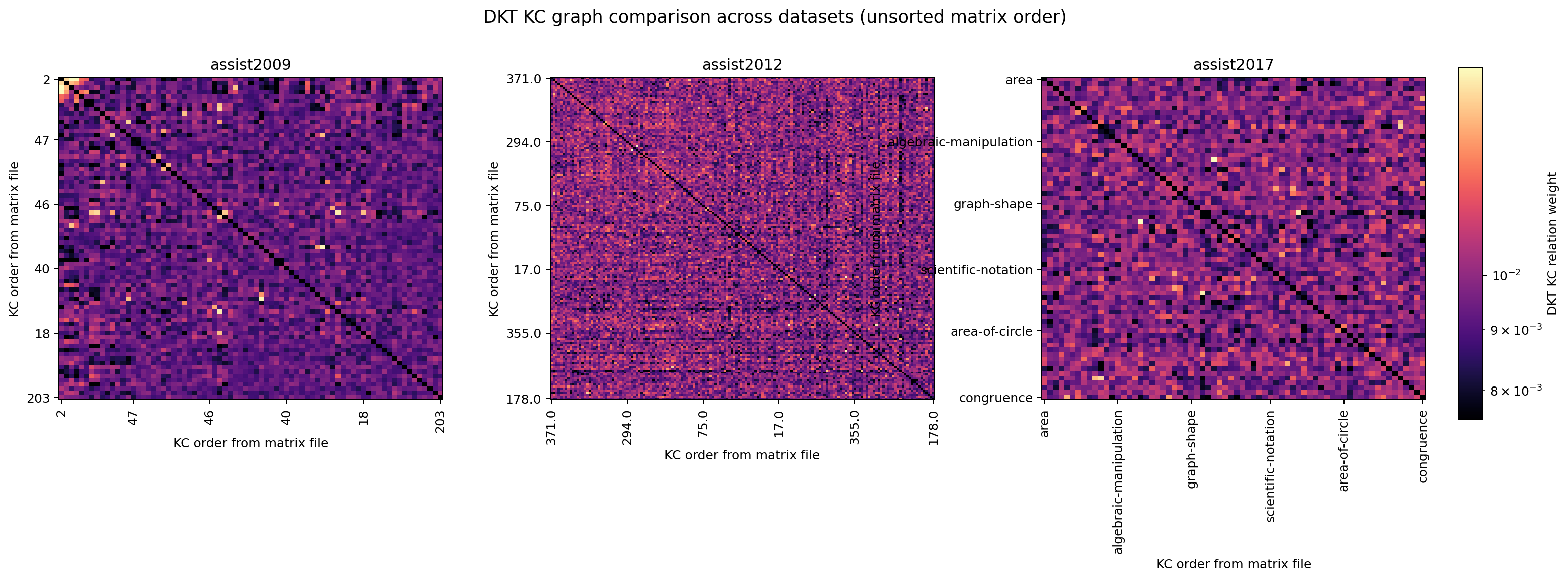}
    \caption{DKT-derived KC graph matrices across the three ASSISTments datasets used in the Chapter~5 post hoc alignment analysis.}
    \Description{DKT-derived KC graph matrices across the three ASSISTments datasets used in the Chapter~5 post hoc alignment analysis.}
    \label{fig:app_ch5_posthoc_kc_graph_triplet}
\end{figure}

Figure~\ref{fig:app_ch5_posthoc_kc_graph_triplet} shows the three DKT-derived KC graphs used as the comparison targets in the Chapter~5 post hoc analysis. The target relation structure differs noticeably across ASSISTments 2009, 2012, and 2017.

Tables~\ref{tab:app_ch5_posthoc_family_summary}--\ref{tab:app_ch5_posthoc_combinations} report the extended ranking summaries for the Chapter~5 post hoc analysis. Some embedding models and pipelines are systematically less misaligned than others, but the absolute differences remain small, so better choices improve the margins without changing the weak-alignment conclusion.

\begin{table}[p]
    \centering
    \small
    \setlength{\tabcolsep}{5pt}
    \renewcommand{\arraystretch}{1.1}
    \DataTableHeaderRows{1}
    \begin{tabular}{l c c c}
        \toprule
        Embedding model & Pearson & Spearman & Top-10\% overlap \\
        \midrule
        \texttt{bert\_ft\_assist2012} & \textbf{0.0321} & \textbf{0.0260} & \textbf{0.1304} \\
        \texttt{text-embedding-3-large} & 0.0297 & 0.0260 & 0.1268 \\
        \texttt{bert\_ft\_assist2009} & 0.0217 & 0.0118 & 0.0999 \\
        \texttt{all-MiniLM-L12-v2} & 0.0205 & 0.0178 & 0.1260 \\
        \texttt{embed-v4.0} & 0.0187 & 0.0185 & 0.1229 \\
        \bottomrule
    \end{tabular}
    \caption{Family summary for the selected embedding-model subset in the Chapter~5 KC-graph alignment analysis. Families are ordered by mean Pearson correlation across datasets and pipelines.}
    \label{tab:app_ch5_posthoc_family_summary}
\end{table}

\begin{table}[p]
    \centering
    \small
    \setlength{\tabcolsep}{5pt}
    \renewcommand{\arraystretch}{1.1}
    \DataTableHeaderRows{1}
    \begin{tabular}{c c c c}
        \toprule
        Pipeline & Pearson & Spearman & Top-10\% overlap \\
        \midrule
        5 & \textbf{0.0243} & \textbf{0.0233} & 0.1229 \\
        9 & 0.0238 & 0.0223 & 0.1214 \\
        3 & 0.0227 & 0.0191 & 0.1228 \\
        7 & 0.0189 & 0.0165 & 0.1208 \\
        1 & 0.0175 & 0.0142 & 0.1216 \\
        6 & 0.0167 & 0.0237 & \textbf{0.1231} \\
        4 & 0.0159 & 0.0243 & 0.1264 \\
        8 & 0.0138 & 0.0163 & 0.1144 \\
        2 & 0.0114 & 0.0104 & 0.1229 \\
        \bottomrule
    \end{tabular}
    \caption{Global pipeline summary for Chapter~5 KC-graph alignment. Pipelines are ordered by mean Pearson correlation across the selected embedding-model subset.}
    \label{tab:app_ch5_posthoc_pipeline_summary}
\end{table}

\begin{table}[p]
    \centering
    \small
    \setlength{\tabcolsep}{3pt}
    \renewcommand{\arraystretch}{1.1}
    \DataTableHeaderRows{1}
    \resizebox{\textwidth}{!}{%
    \begin{tabular}{l l l c c p{4.6cm}}
        \toprule
        Group & Dataset & Embedding model & Pipeline & Pearson & Note \\
        \midrule
        Selected-subset best & ASSISTments 2012 & \texttt{text-embedding-3-large} & 4 & \textbf{0.0840} & Best selected-subset combination \\
        Selected-subset best & ASSISTments 2012 & \texttt{text-embedding-3-large} & 5 & 0.0835 & Second-best selected-subset combination \\
        Selected-subset best & ASSISTments 2012 & \texttt{text-embedding-3-large} & 1 & 0.0832 & Third-best selected-subset combination \\
        BERT follow-up & ASSISTments 2009 & \texttt{bert\_ft\_assist2009} & 5 & 0.0487 & Family rank 1 on ASSISTments 2009 \\
        BERT follow-up & ASSISTments 2012 & \texttt{bert\_ft\_assist2012} & 6 & 0.0799 & Family rank 12 on ASSISTments 2012 \\
        \bottomrule
    \end{tabular}
    }
    \caption{Strongest dataset--embedding-model--pipeline combinations from the selected comparison subset together with the strongest fine-tuned BERT alignment results.}
    \label{tab:app_ch5_posthoc_combinations}
\end{table}

\begin{figure}[p]
    \centering
    \includegraphics[width=\linewidth]{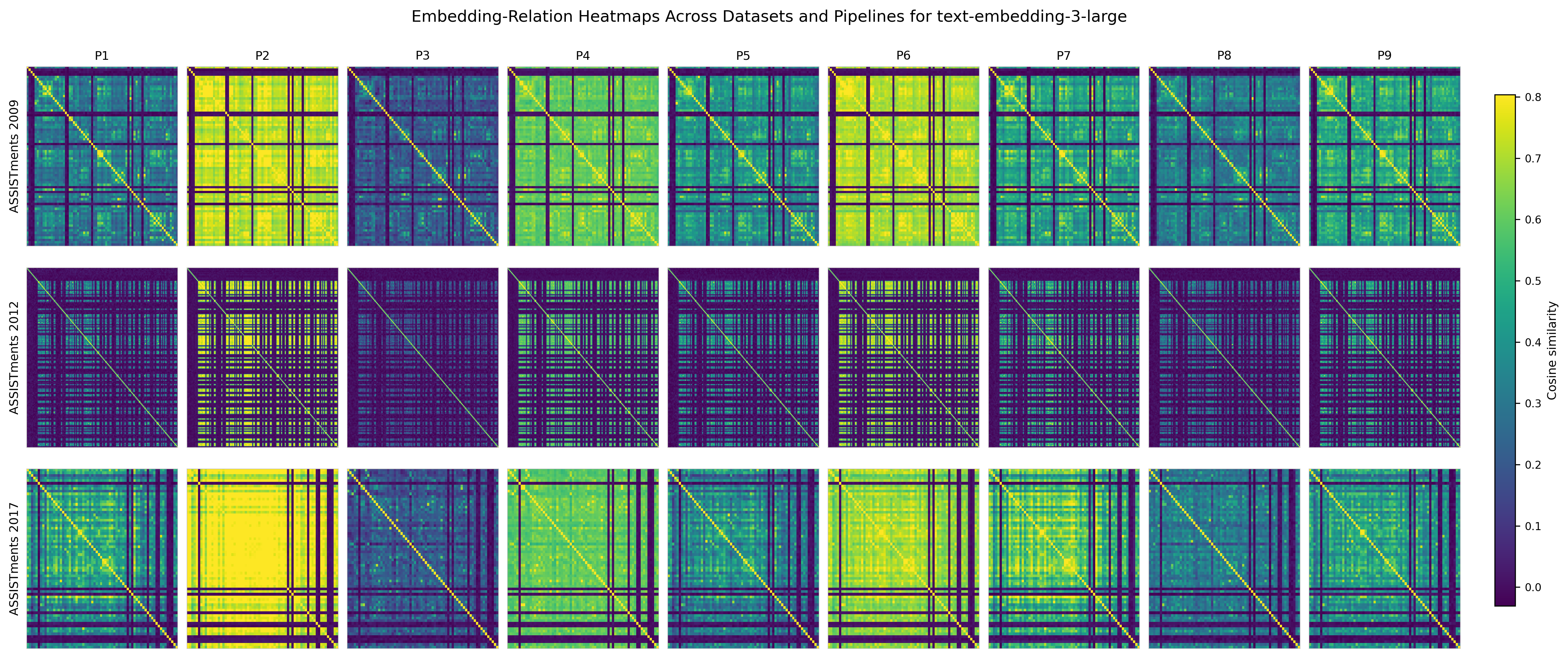}
    \caption{Embedding-relation heatmap comparison across pipelines for \texttt{text-embedding-3-large}.}
    \Description{Embedding-relation heatmap comparison across pipelines for \texttt{text-embedding-3-large}.}
    \label{fig:app_ch5_posthoc_pipeline_grid}
\end{figure}

Figure~\ref{fig:app_ch5_posthoc_pipeline_grid} provides the larger pipeline-by-dataset heatmap comparison for \texttt{text-embedding-3-large}. The figure shows cosine-similarity structure induced by different pipeline constructions across datasets. Pipeline choice changes the embedding geometry, but none of the shown variants reorganizes the KC relations into a pattern that closely resembles the DKT-derived graphs across datasets.

\section{Chapter 6 Supplementary Material}

\subsection{Additional Benchmark Sweep Figures}
\label{sec:appendix_ch6_benchmark}

\begin{figure}[p]
\centering
\begin{minipage}{0.49\linewidth}
    \centering
    \includegraphics[width=\linewidth]{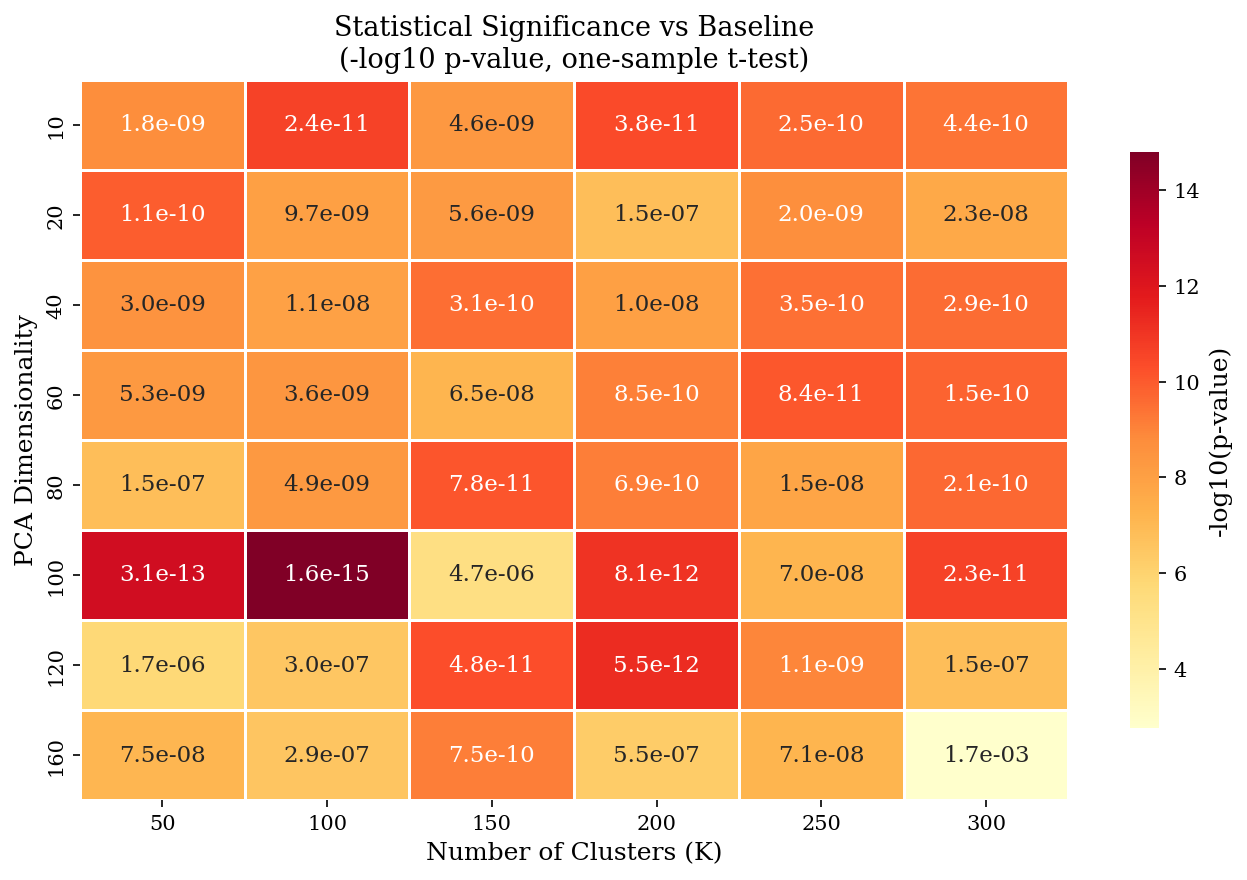}
\end{minipage}\hfill
\begin{minipage}{0.49\linewidth}
    \centering
    \includegraphics[width=\linewidth]{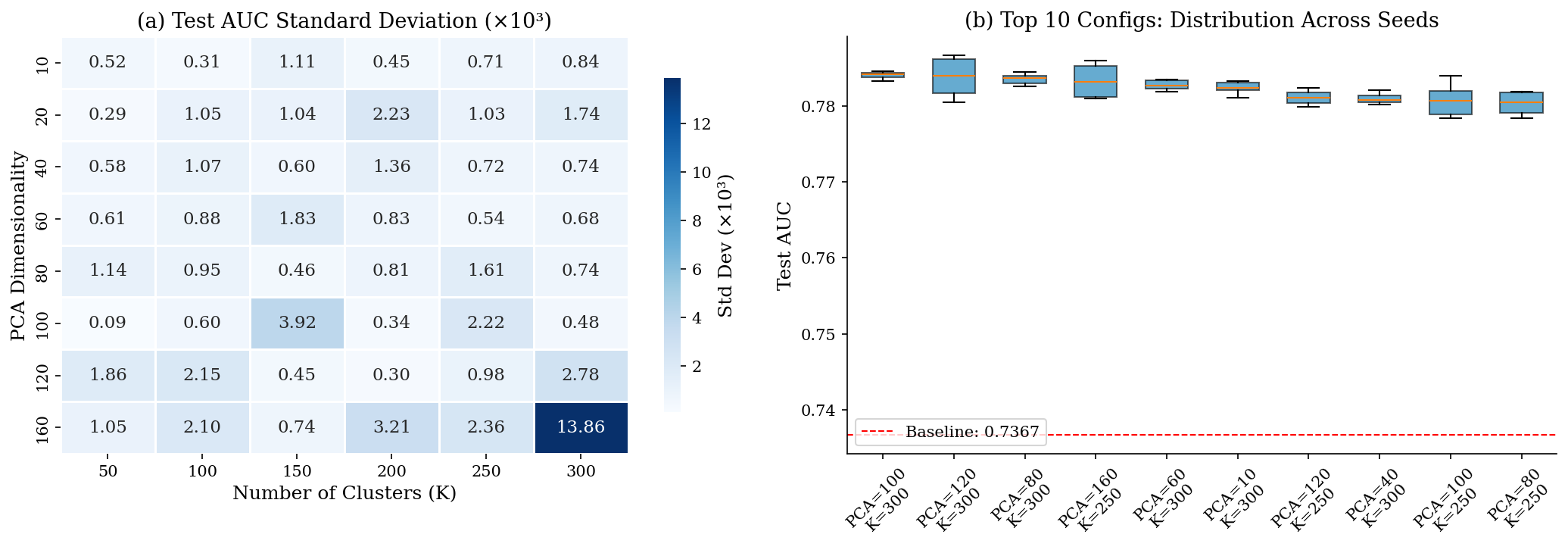}
\end{minipage}
\caption{Additional ASSISTments 2012 benchmark diagnostics: significance testing across the benchmark sweep (left) and robustness across repeated runs (right).}
\Description{Additional ASSISTments 2012 benchmark diagnostics: significance testing across the benchmark sweep (left) and robustness across repeated runs (right).}
\label{fig:app_c6_2012_sig_rob}
\end{figure}

\begin{figure}[p]
\centering
\begin{minipage}{0.49\linewidth}
    \centering
    \includegraphics[width=\linewidth]{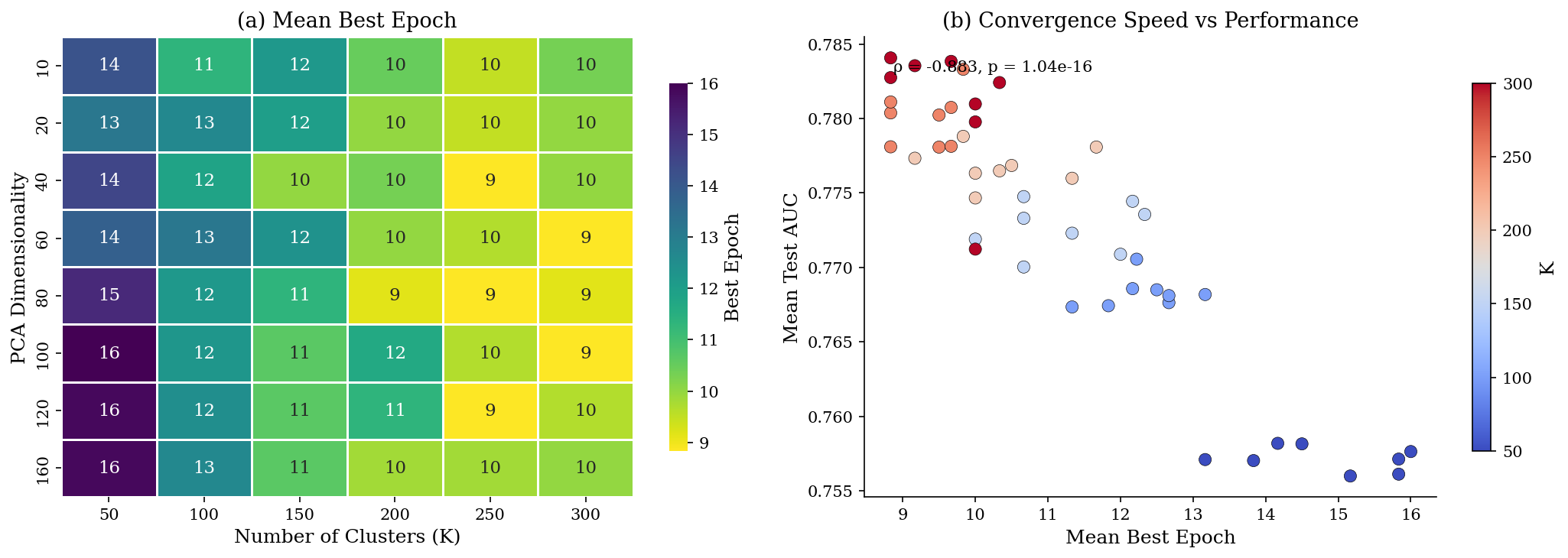}
\end{minipage}\hfill
\begin{minipage}{0.49\linewidth}
    \centering
    \includegraphics[width=\linewidth]{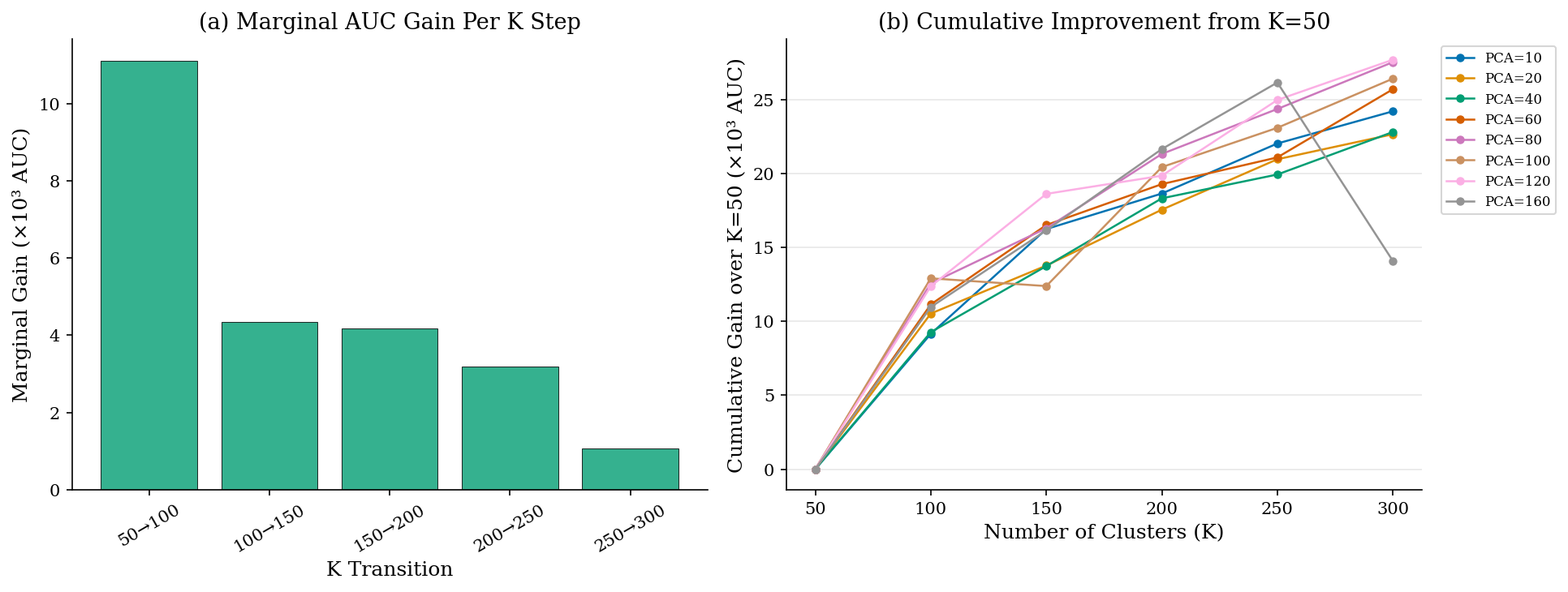}
\end{minipage}
\caption{Additional ASSISTments 2012 diagnostics: convergence behavior (left) and diminishing returns as cluster count increases (right).}
\Description{Additional ASSISTments 2012 diagnostics: convergence behavior (left) and diminishing returns as cluster count increases (right).}
\label{fig:app_c6_2012_conv_dim}
\end{figure}

\begin{figure}[p]
\centering
\begin{minipage}{0.49\linewidth}
    \centering
    \includegraphics[width=\linewidth]{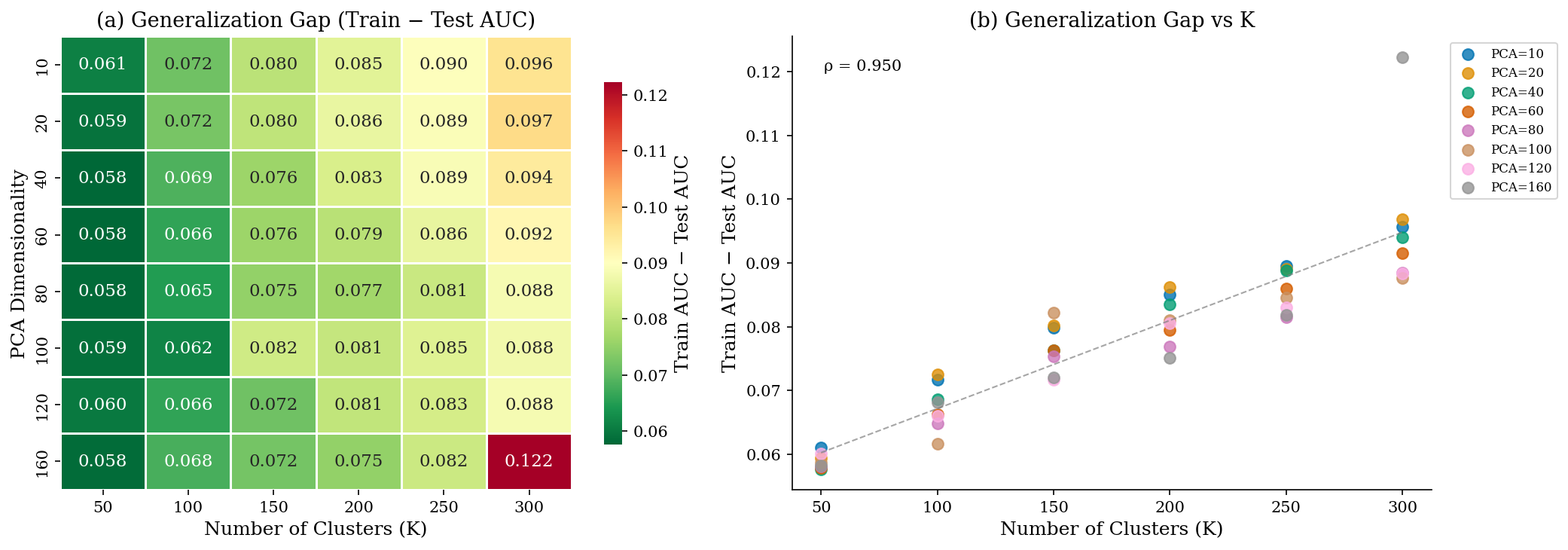}
\end{minipage}\hfill
\begin{minipage}{0.49\linewidth}
    \centering
    \includegraphics[width=\linewidth]{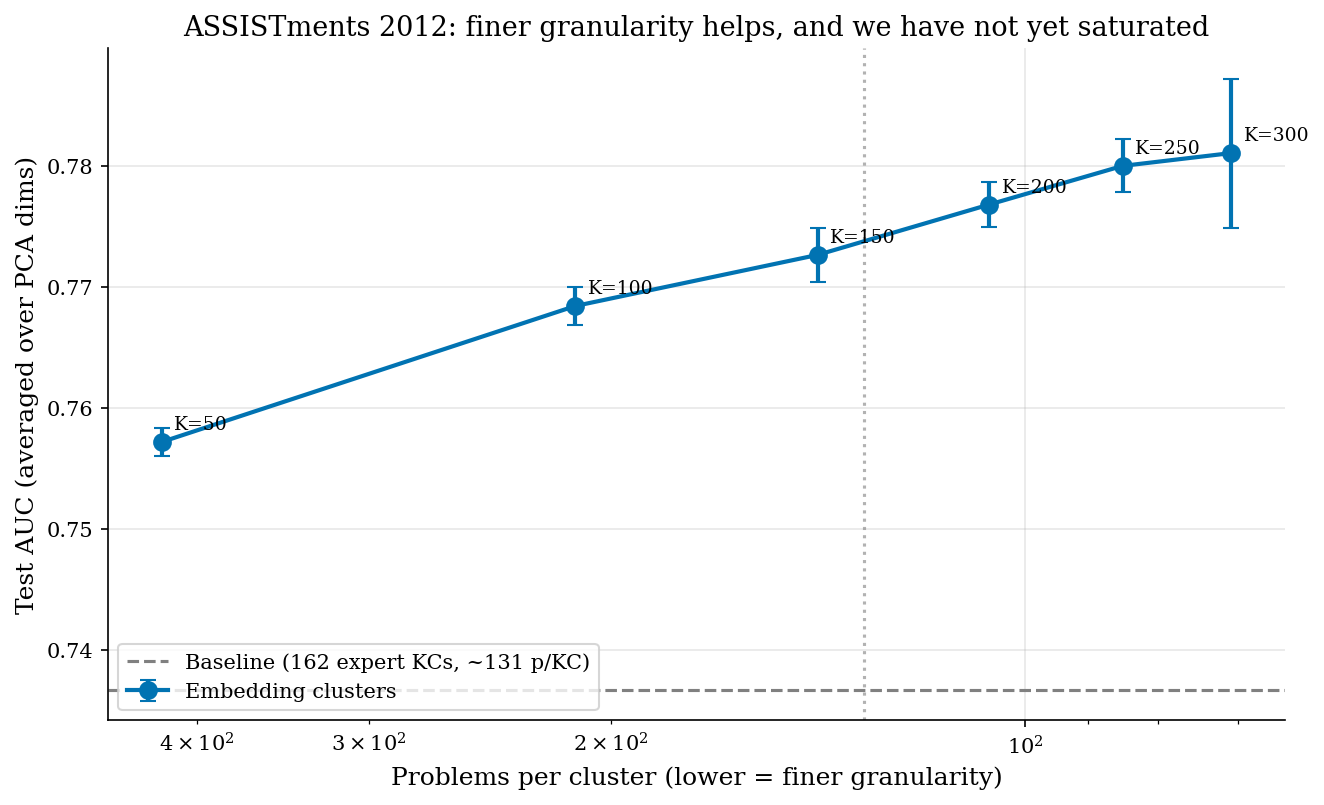}
\end{minipage}
\caption{Additional ASSISTments 2012 diagnostics: overfitting behavior across the sweep (left) and cluster granularity relative to problem-bank size (right).}
\Description{Additional ASSISTments 2012 diagnostics: overfitting behavior across the sweep (left) and cluster granularity relative to problem-bank size (right).}
\label{fig:app_c6_2012_overfit_gran}
\end{figure}

\begin{figure}[p]
\centering
\includegraphics[width=0.86\linewidth]{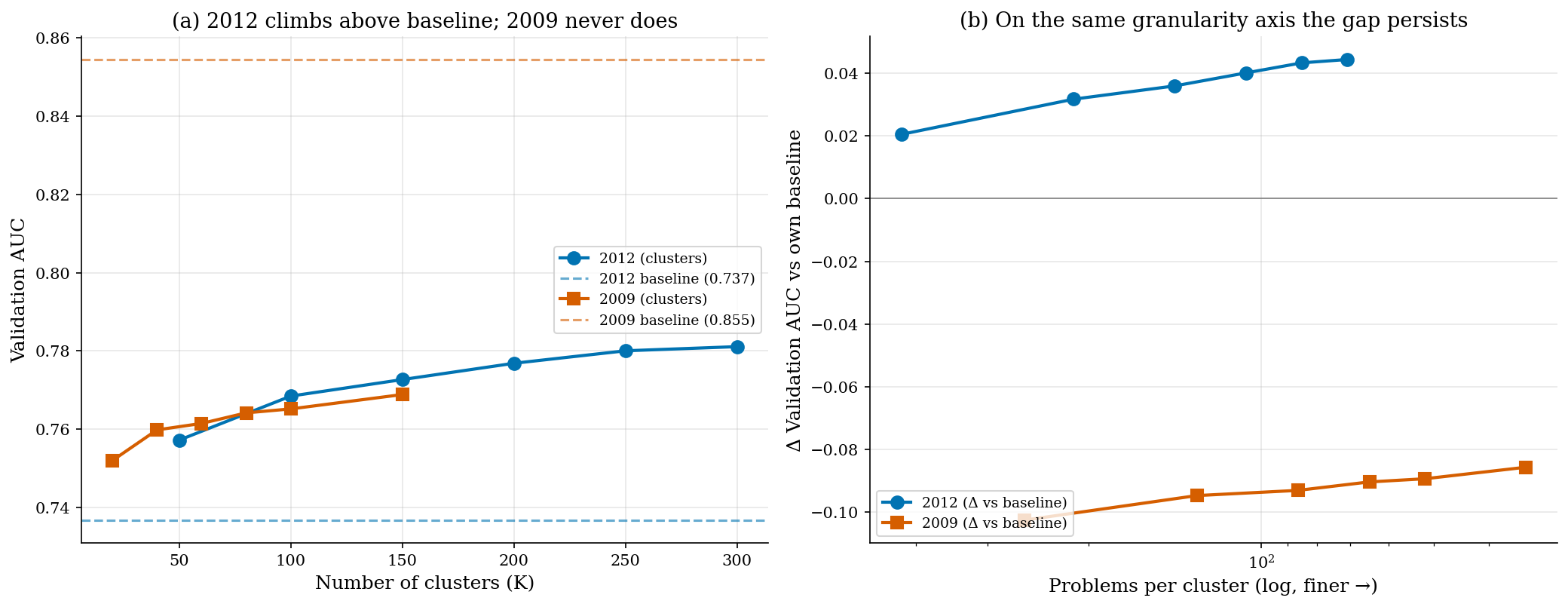}
\caption{Additional ASSISTments 2012 benchmark contrast against ASSISTments 2009 on a shared scale.}
\Description{Additional ASSISTments 2012 benchmark contrast against ASSISTments 2009 on a shared scale.}
\label{fig:app_c6_2012_cross}
\end{figure}

\begin{figure}[p]
\centering
\includegraphics[width=0.86\linewidth]{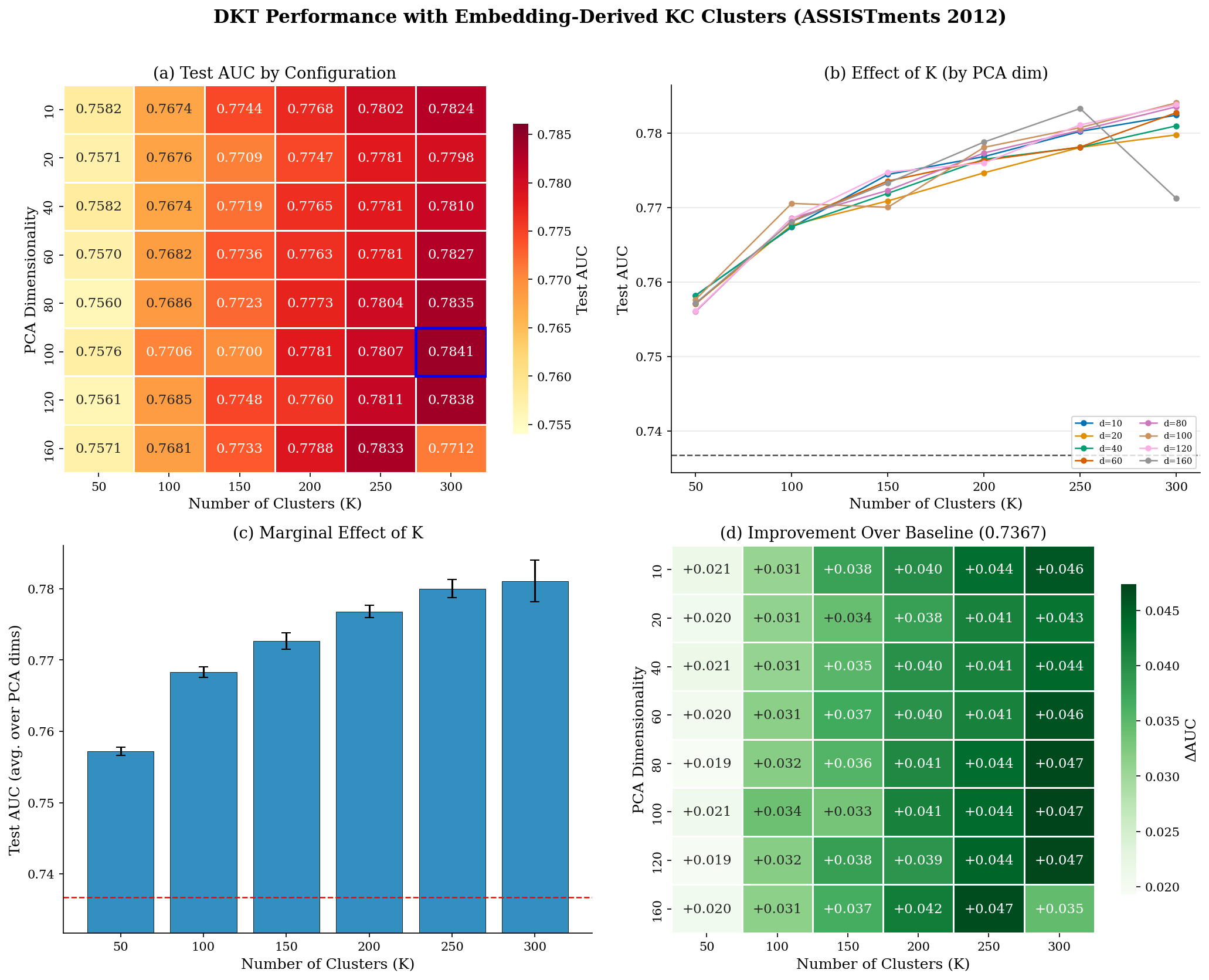}
\caption{Composite benchmark summary for ASSISTments 2012 across the benchmark sweep.}
\Description{Composite benchmark summary for ASSISTments 2012 across the benchmark sweep.}
\label{fig:app_c6_2012_composite}
\end{figure}

\begin{figure}[p]
\centering
\begin{minipage}{0.49\linewidth}
    \centering
    \includegraphics[width=\linewidth]{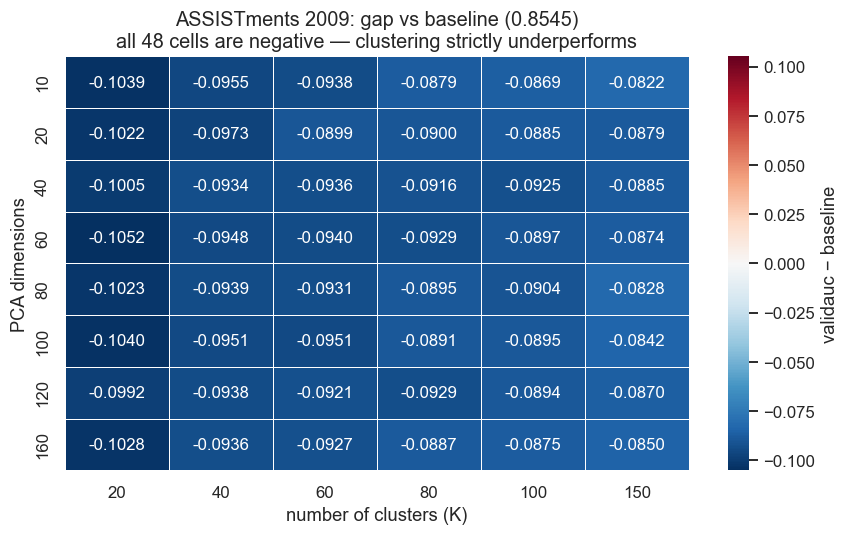}
\end{minipage}\hfill
\begin{minipage}{0.49\linewidth}
    \centering
    \includegraphics[width=\linewidth]{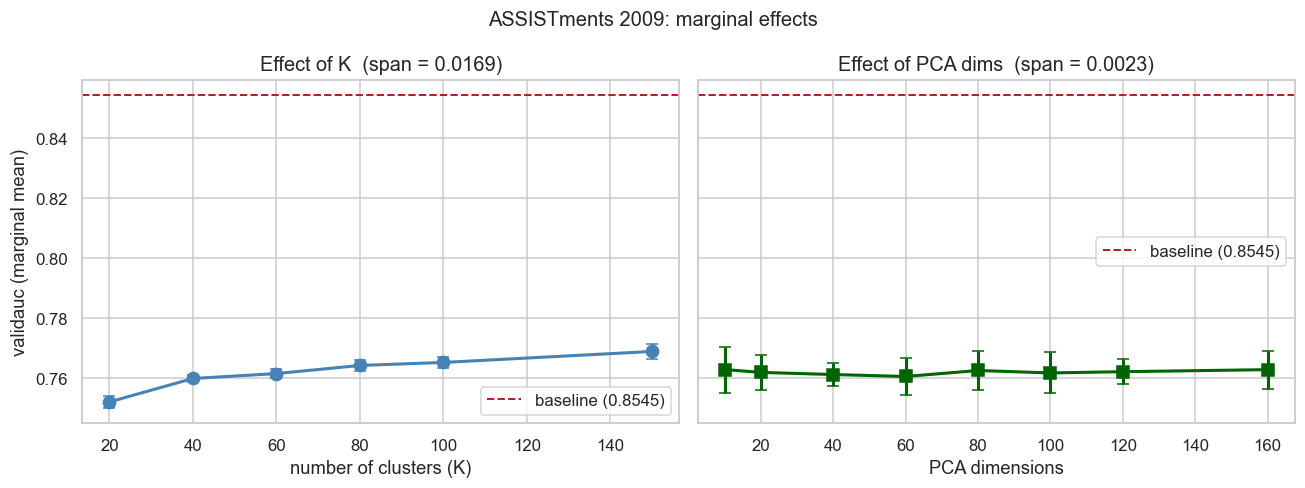}
\end{minipage}
\caption{Additional ASSISTments 2009 diagnostics: gap versus the original-KC baseline across the sweep (left) and averaged marginal effects of the clustering hyperparameters (right).}
\Description{Additional ASSISTments 2009 diagnostics: gap versus the original-KC baseline across the sweep (left) and averaged marginal effects of the clustering hyperparameters (right).}
\label{fig:app_c6_2009_gap_marginal}
\end{figure}

\begin{figure}[p]
\centering
\begin{minipage}{0.49\linewidth}
    \centering
    \includegraphics[width=\linewidth]{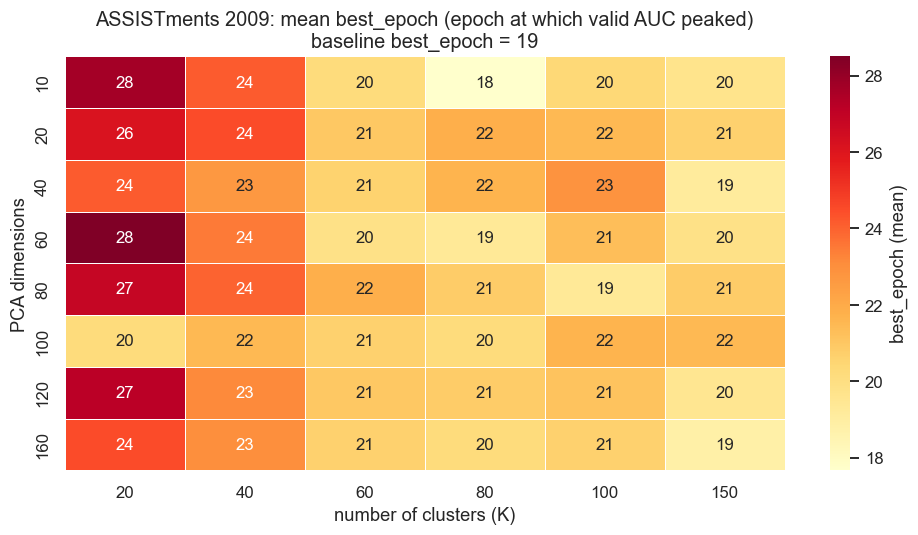}
\end{minipage}\hfill
\begin{minipage}{0.49\linewidth}
    \centering
    \includegraphics[width=\linewidth]{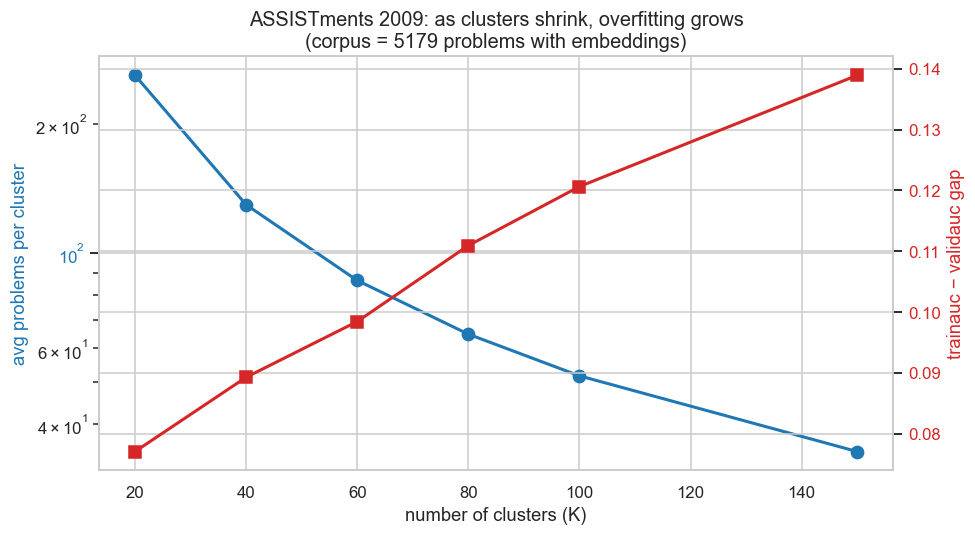}
\end{minipage}
\caption{Additional ASSISTments 2009 diagnostics: convergence behavior (left) and the relation between cluster size and overfitting (right).}
\Description{Additional ASSISTments 2009 diagnostics: convergence behavior (left) and the relation between cluster size and overfitting (right).}
\label{fig:app_c6_2009_conv_clustersize}
\end{figure}

\begin{figure}[p]
\centering
\includegraphics[width=0.86\linewidth]{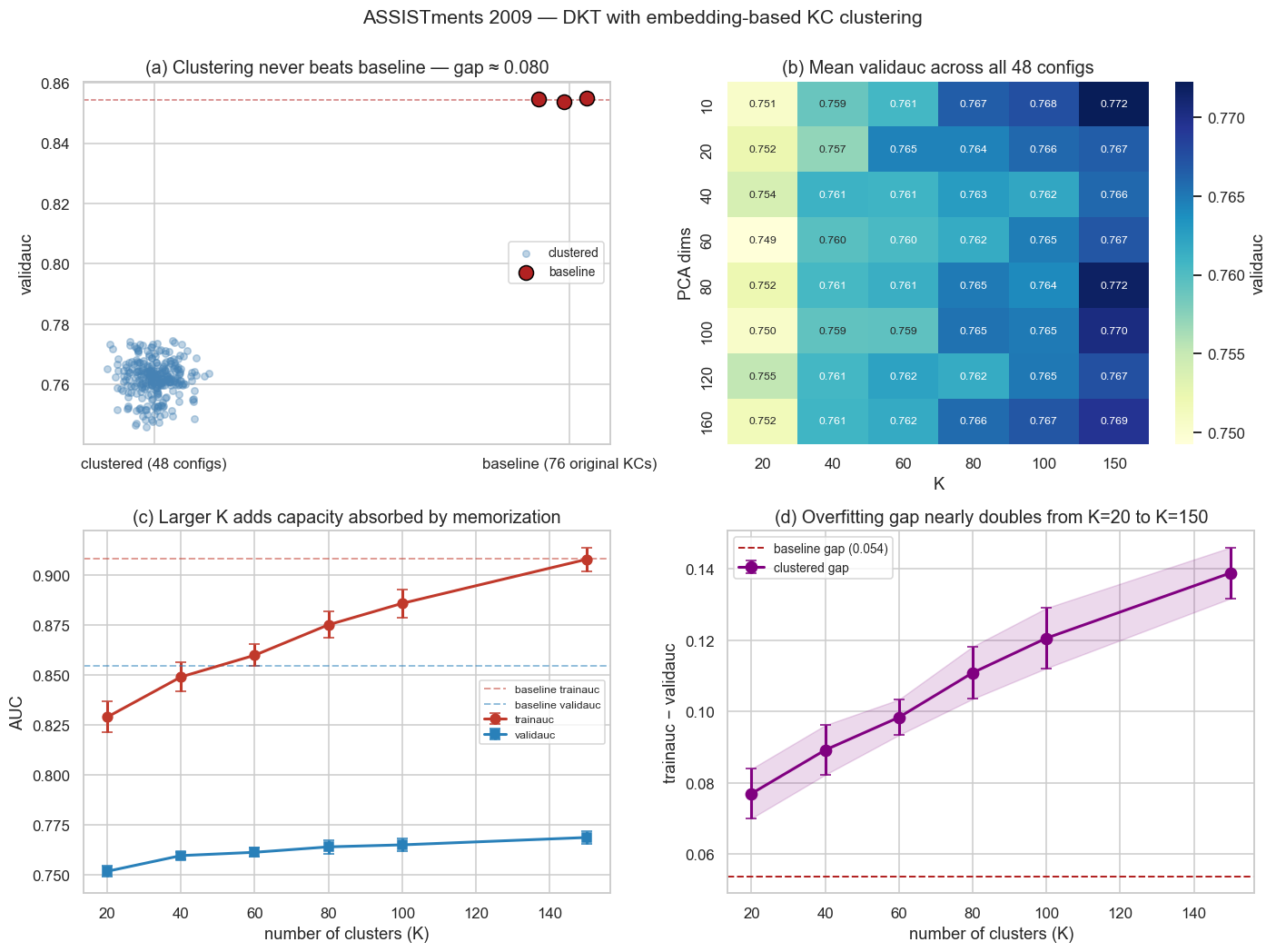}
\caption{Composite benchmark summary for ASSISTments 2009 across the benchmark sweep.}
\Description{Composite benchmark summary for ASSISTments 2009 across the benchmark sweep.}
\label{fig:app_c6_2009_composite}
\end{figure}

\begin{figure}[p]
\centering
\begin{minipage}{0.49\linewidth}
    \centering
    \includegraphics[width=\linewidth]{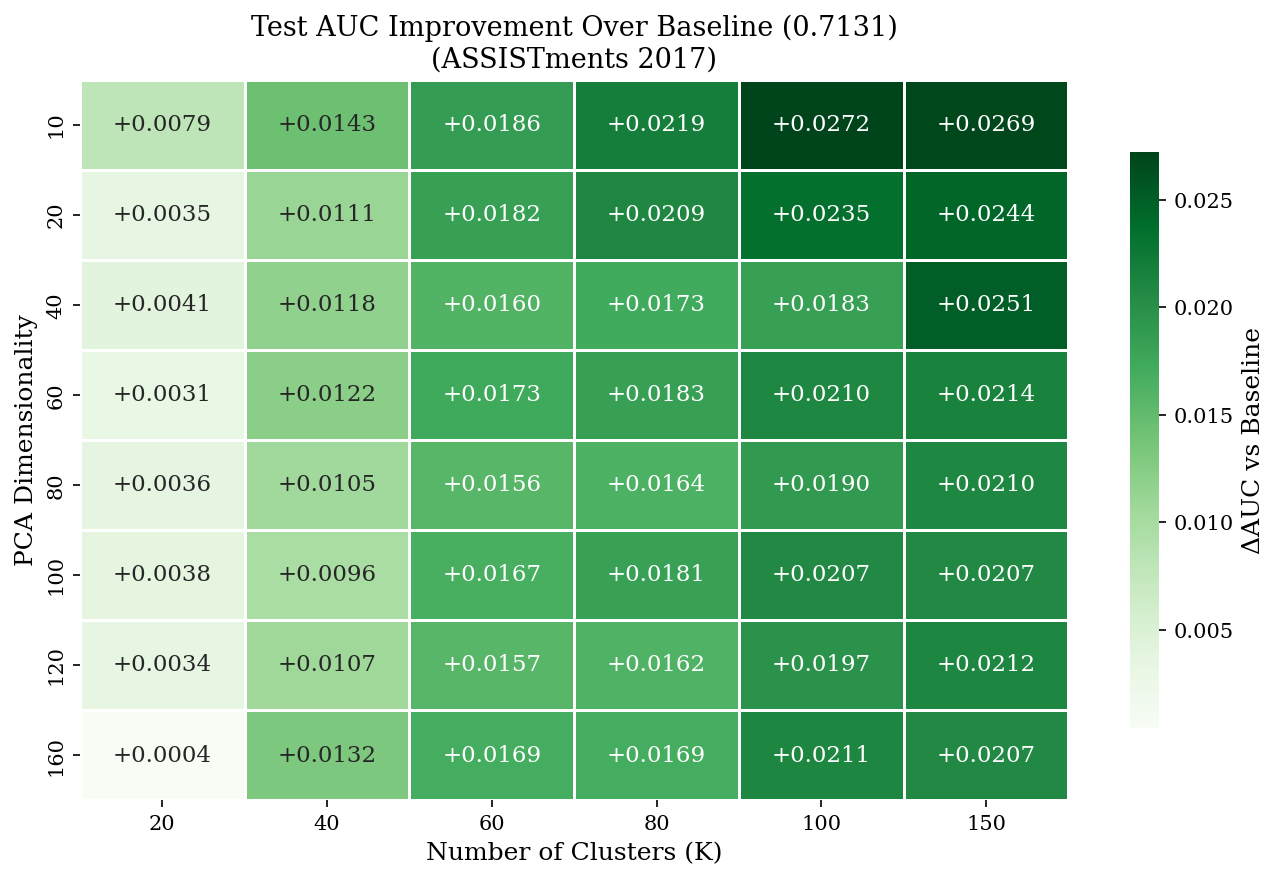}
\end{minipage}\hfill
\begin{minipage}{0.49\linewidth}
    \centering
    \includegraphics[width=\linewidth]{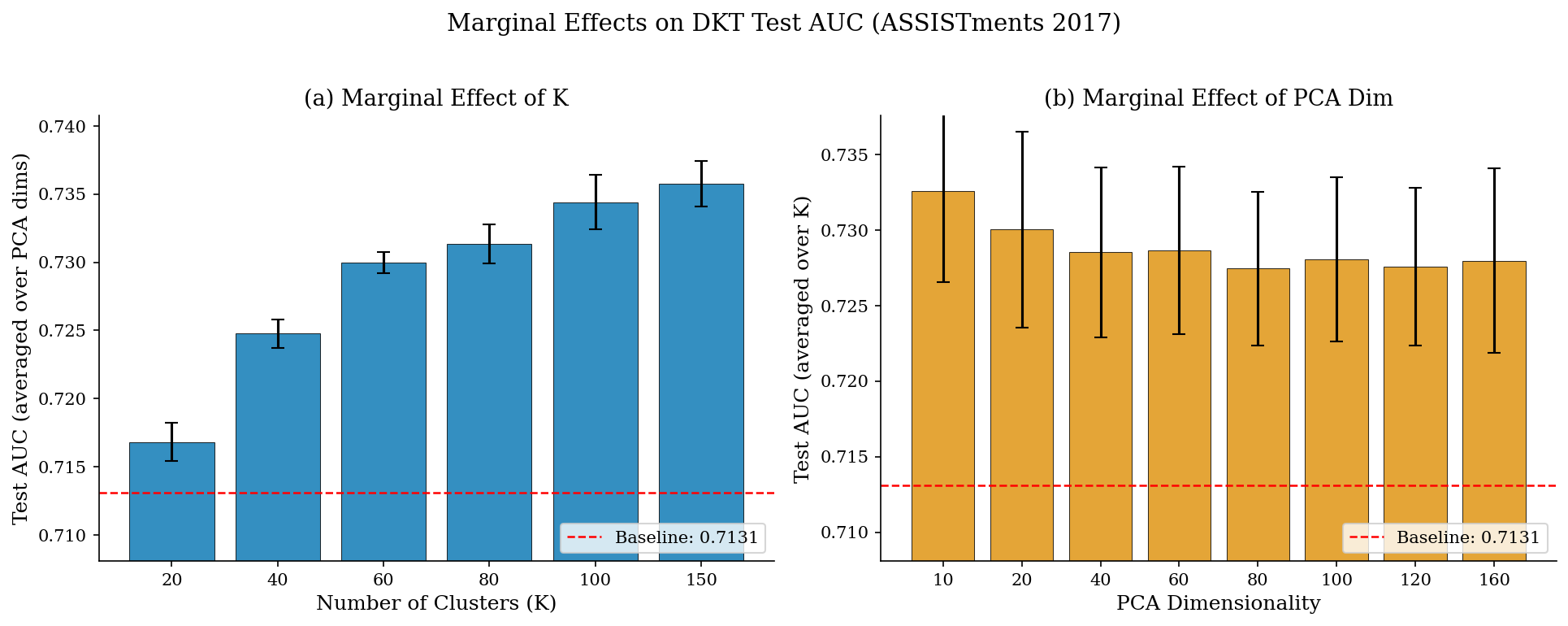}
\end{minipage}
\caption{Additional ASSISTments 2017 diagnostics: improvement over the original-KC baseline across the sweep (left) and averaged marginal effects of the clustering hyperparameters (right).}
\Description{Additional ASSISTments 2017 diagnostics: improvement over the original-KC baseline across the sweep (left) and averaged marginal effects of the clustering hyperparameters (right).}
\label{fig:app_c6_2017_improve_marginal}
\end{figure}

\begin{figure}[p]
\centering
\begin{minipage}{0.49\linewidth}
    \centering
    \includegraphics[width=\linewidth]{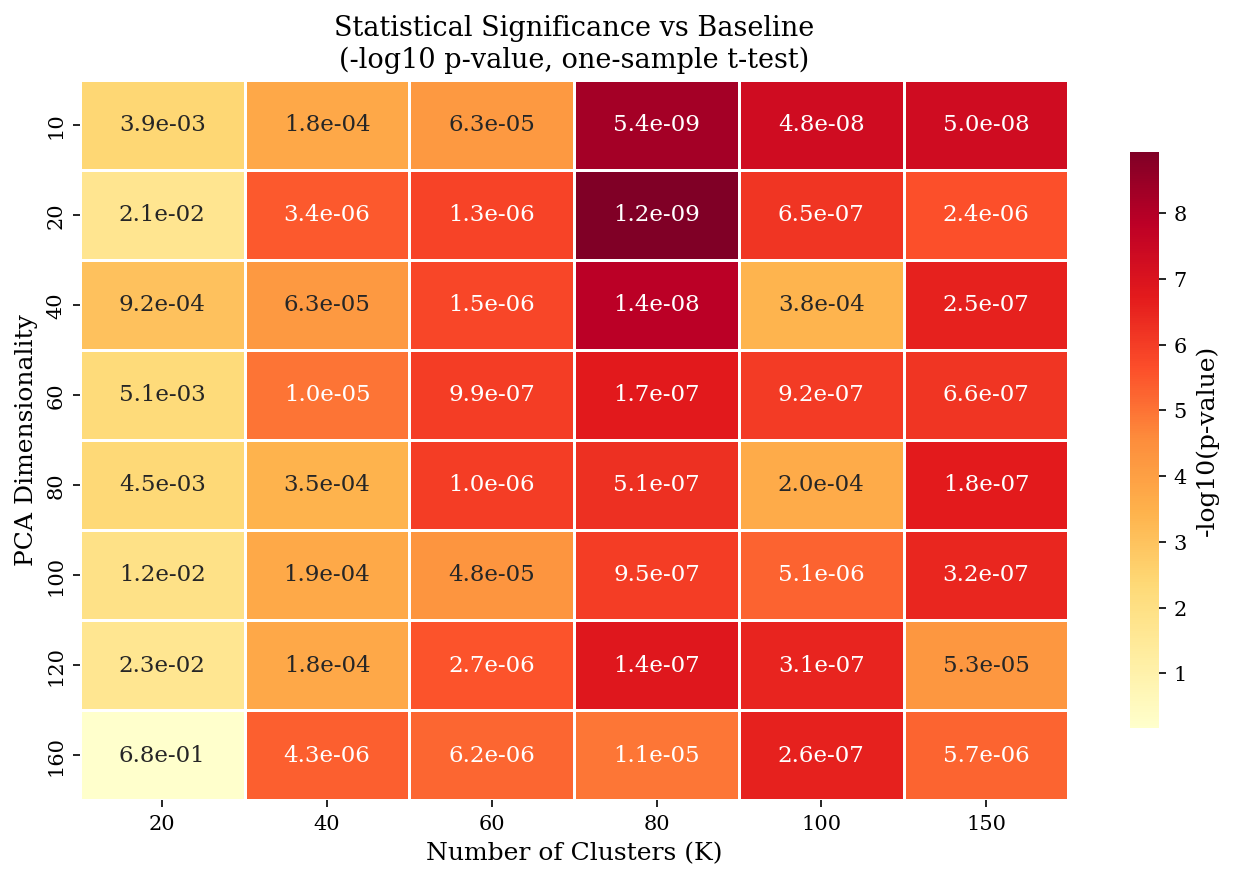}
\end{minipage}\hfill
\begin{minipage}{0.49\linewidth}
    \centering
    \includegraphics[width=\linewidth]{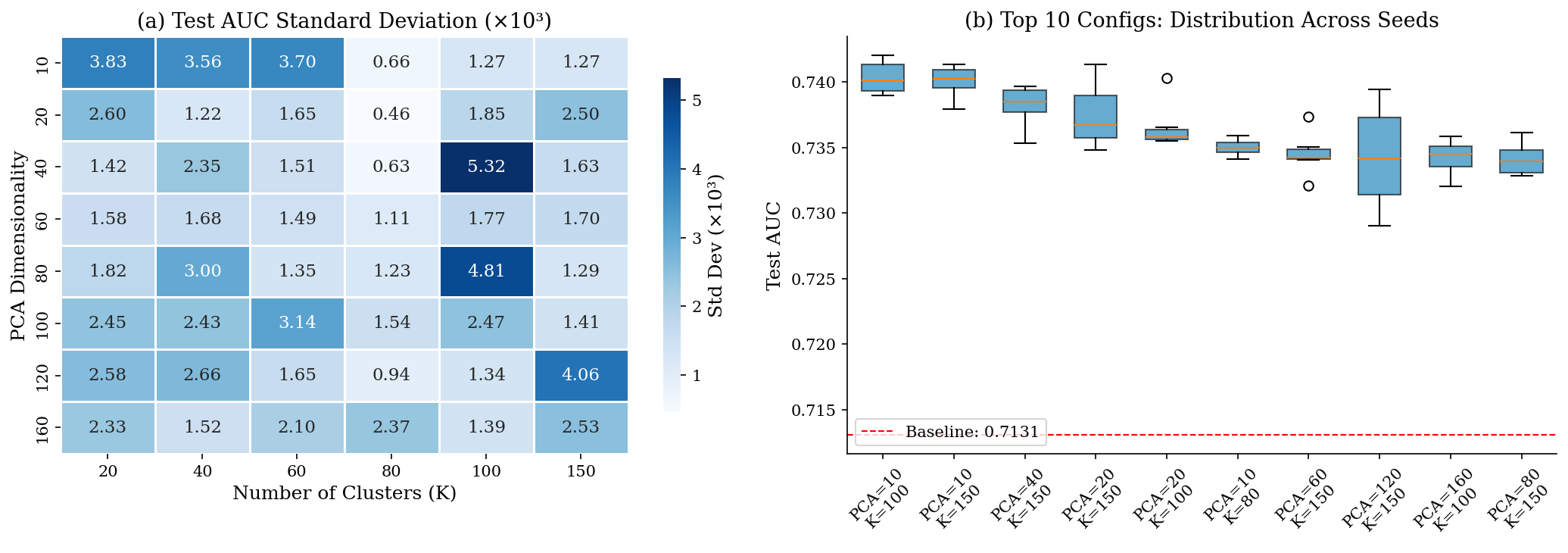}
\end{minipage}
\caption{Additional ASSISTments 2017 diagnostics: significance testing across the benchmark sweep (left) and robustness across repeated runs (right).}
\Description{Additional ASSISTments 2017 diagnostics: significance testing across the benchmark sweep (left) and robustness across repeated runs (right).}
\label{fig:app_c6_2017_sig_rob}
\end{figure}

\begin{figure}[p]
\centering
\begin{minipage}{0.49\linewidth}
    \centering
    \includegraphics[width=\linewidth]{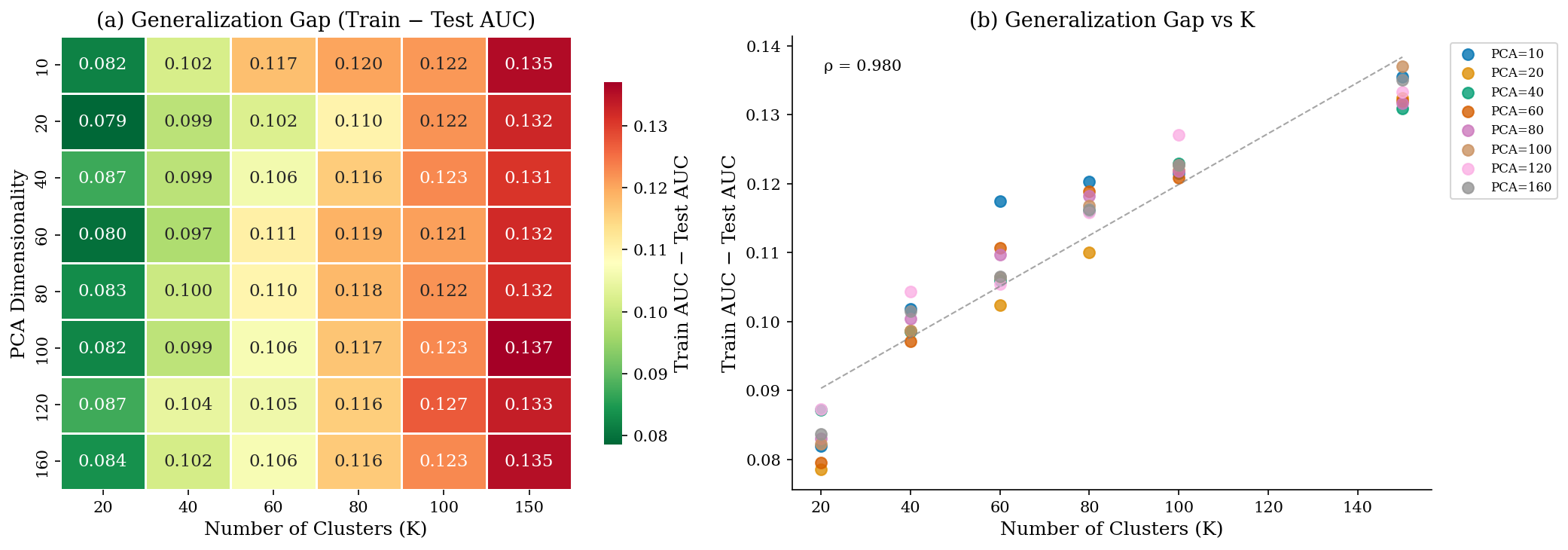}
\end{minipage}\hfill
\begin{minipage}{0.49\linewidth}
    \centering
    \includegraphics[width=\linewidth]{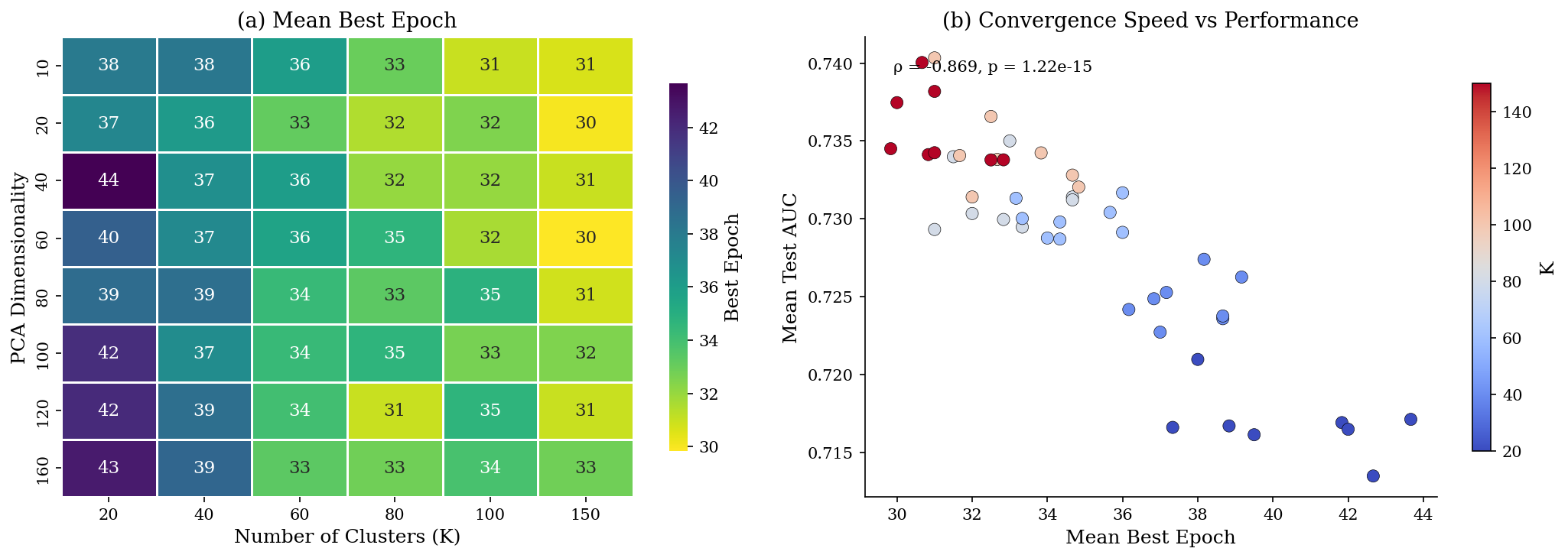}
\end{minipage}
\caption{Additional ASSISTments 2017 diagnostics: overfitting behavior across the sweep (left) and convergence behavior (right).}
\Description{Additional ASSISTments 2017 diagnostics: overfitting behavior across the sweep (left) and convergence behavior (right).}
\label{fig:app_c6_2017_overfit_conv}
\end{figure}

\begin{figure}[p]
\centering
\begin{minipage}{0.49\linewidth}
    \centering
    \includegraphics[width=\linewidth]{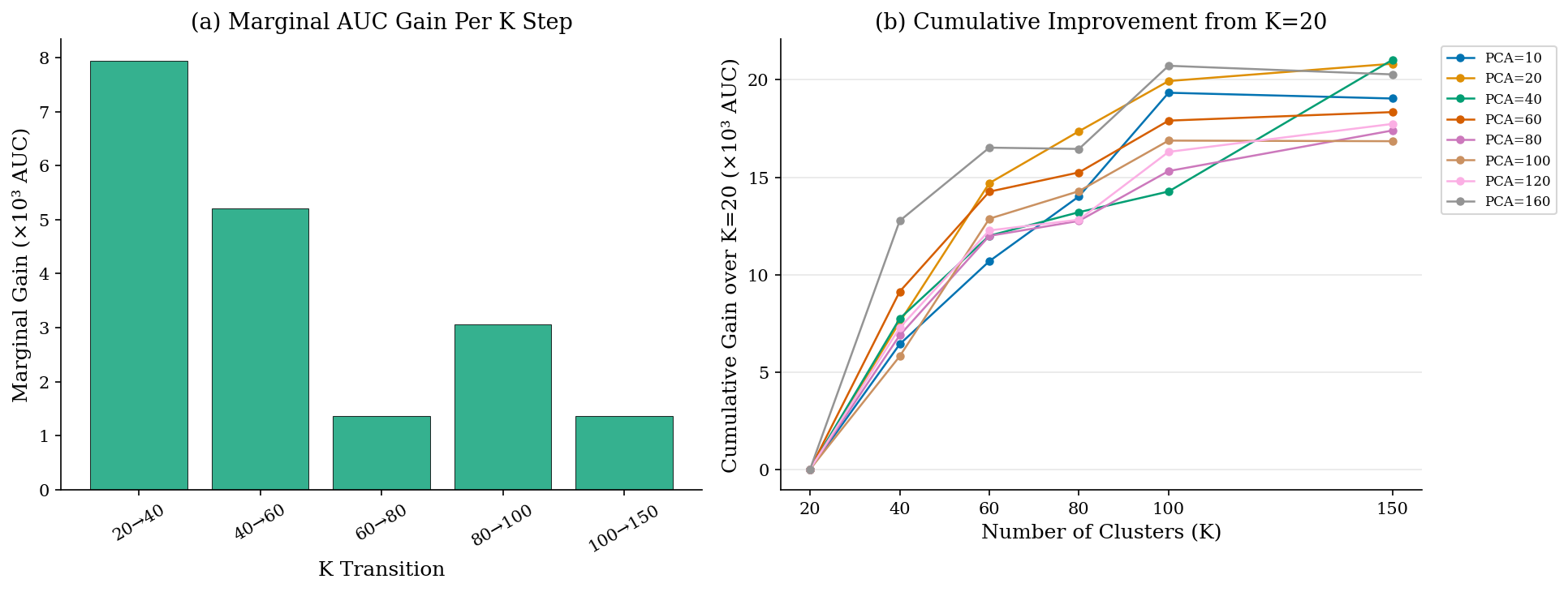}
\end{minipage}\hfill
\begin{minipage}{0.49\linewidth}
    \centering
    \includegraphics[width=\linewidth]{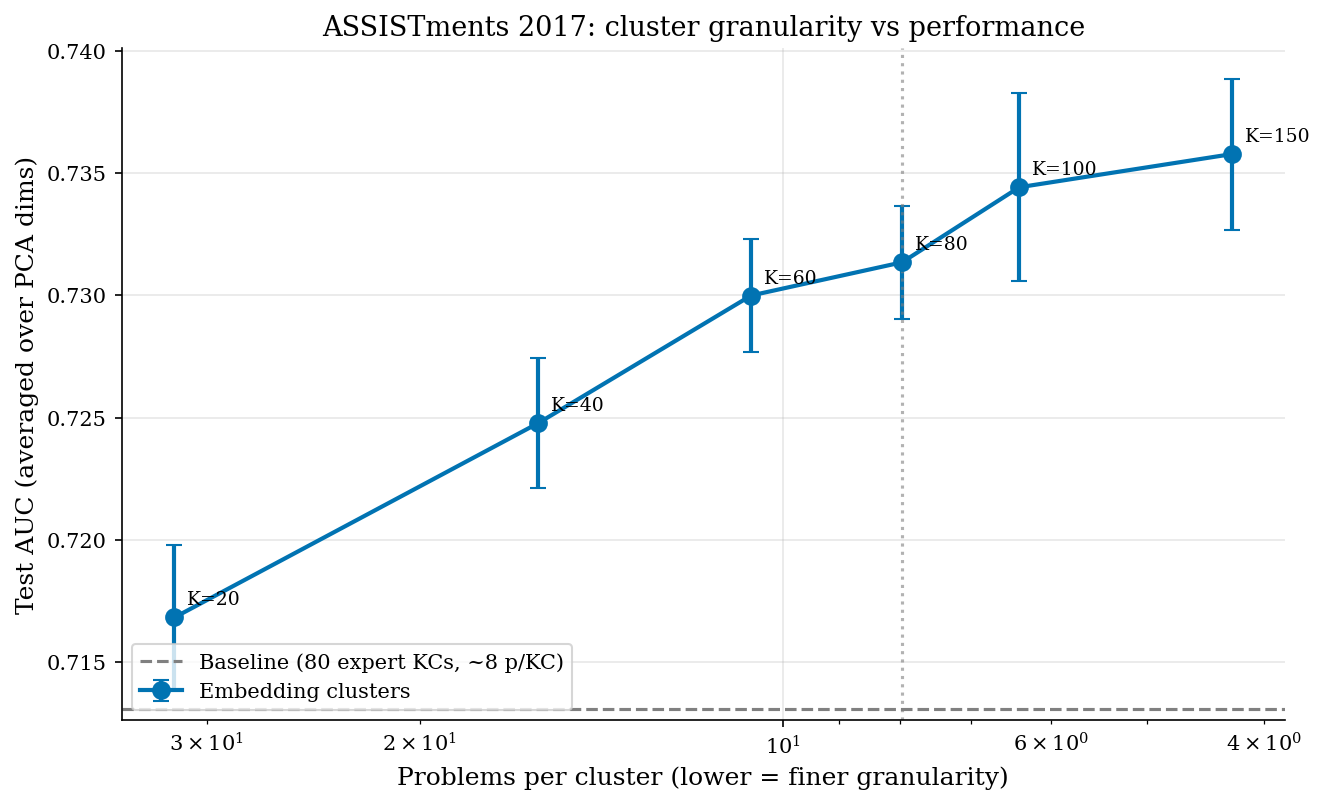}
\end{minipage}
\caption{Additional ASSISTments 2017 diagnostics: diminishing returns as cluster count increases (left) and cluster granularity relative to the small question bank (right).}
\Description{Additional ASSISTments 2017 diagnostics: diminishing returns as cluster count increases (left) and cluster granularity relative to the small question bank (right).}
\label{fig:app_c6_2017_dim_gran}
\end{figure}

\begin{figure}[p]
\centering
\includegraphics[width=0.86\linewidth]{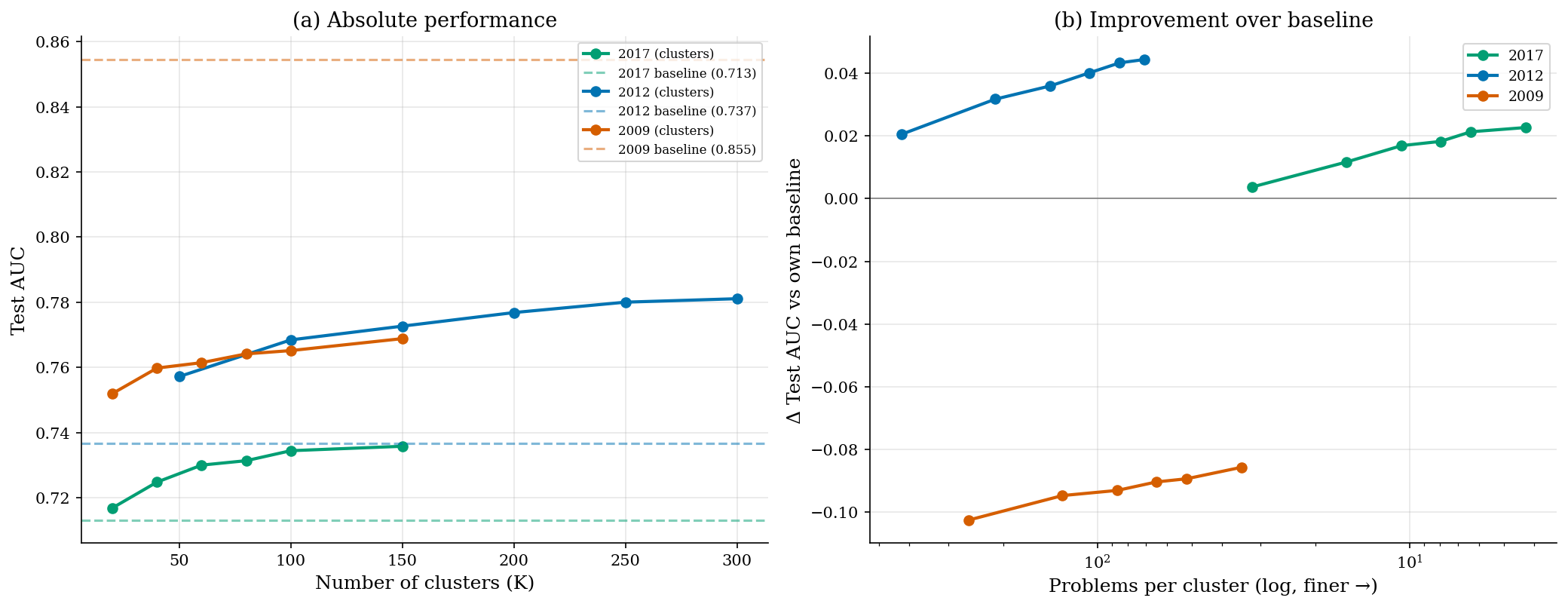}
\caption{Additional ASSISTments 2017 benchmark contrast against ASSISTments 2012 and ASSISTments 2009 on a shared scale.}
\Description{Additional ASSISTments 2017 benchmark contrast against ASSISTments 2012 and ASSISTments 2009 on a shared scale.}
\label{fig:app_c6_2017_cross}
\end{figure}

\begin{figure}[p]
\centering
\includegraphics[width=0.86\linewidth]{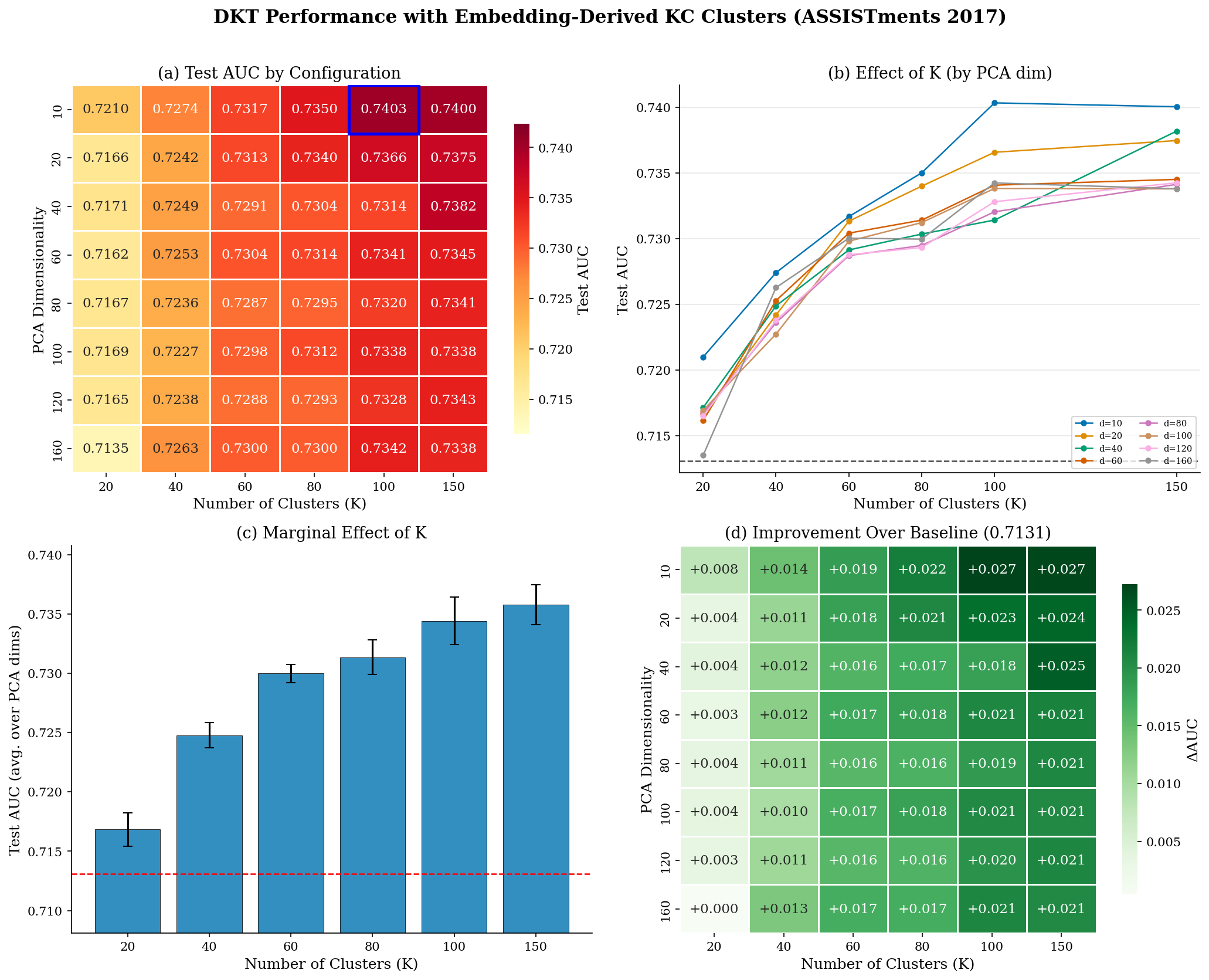}
\caption{Composite benchmark summary for ASSISTments 2017 across the benchmark sweep.}
\Description{Composite benchmark summary for ASSISTments 2017 across the benchmark sweep.}
\label{fig:app_c6_2017_composite}
\end{figure}

Figures~\ref{fig:app_c6_2012_sig_rob}--\ref{fig:app_c6_2017_composite} collect additional benchmark-sweep diagnostics for Chapter~6. They cover statistical significance, robustness across runs, convergence behavior, diminishing returns, cluster granularity, and cross-dataset contrast.

For ASSISTments 2012, the supplementary benchmark figures show that the gain is distributed across a broad region of the evaluated sweep. Figures~\ref{fig:app_c6_2012_sig_rob}--\ref{fig:app_c6_2012_composite} show that the improvement over the original KC baseline is statistically stable across the benchmark sweep, robust across reruns, and sustained as the cluster count increases toward the best-performing region. The larger question bank also absorbs a much finer partition before diminishing returns appear.

For ASSISTments 2009, the supplementary figures show that the clustered family has internal structure and does respond to hyperparameter changes, but the whole family stays below the original-KC baseline. The additional overfitting and cluster-size views show that increasing granularity in this dataset reduces the performance deficit without producing competitive performance.

For ASSISTments 2017, the appendix figures show that the positive result is more localized than on 2012. Figures~\ref{fig:app_c6_2017_improve_marginal}--\ref{fig:app_c6_2017_composite} show that the gains are real across the benchmark sweep, but they concentrate in a narrower low-dimensional region and flatten more quickly as the number of clusters increases. This pattern is consistent with a smaller problem bank that still benefits from KC refinement but reaches its granularity limit sooner than 2012.

\subsection{Additional Structural Analysis Figures}
\label{sec:appendix_ch6_structural}

\begin{figure}[p]
\centering
\begin{minipage}{0.49\linewidth}
    \centering
    \includegraphics[width=\linewidth]{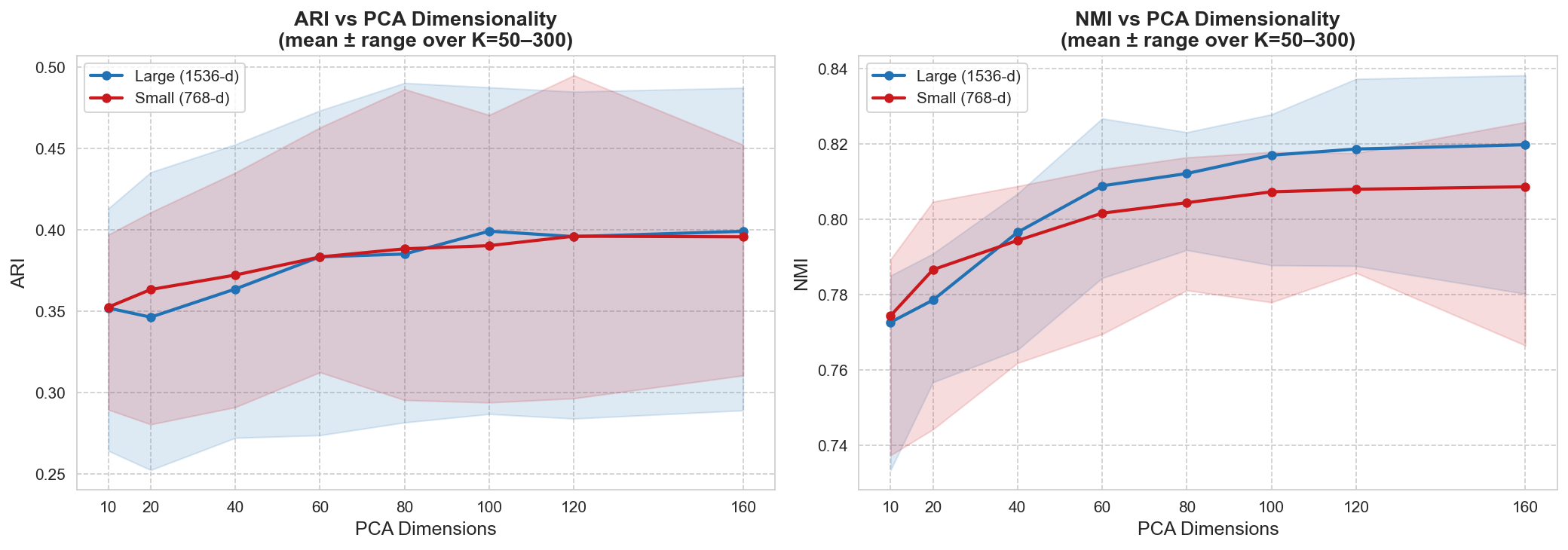}
\end{minipage}\hfill
\begin{minipage}{0.49\linewidth}
    \centering
    \includegraphics[width=\linewidth]{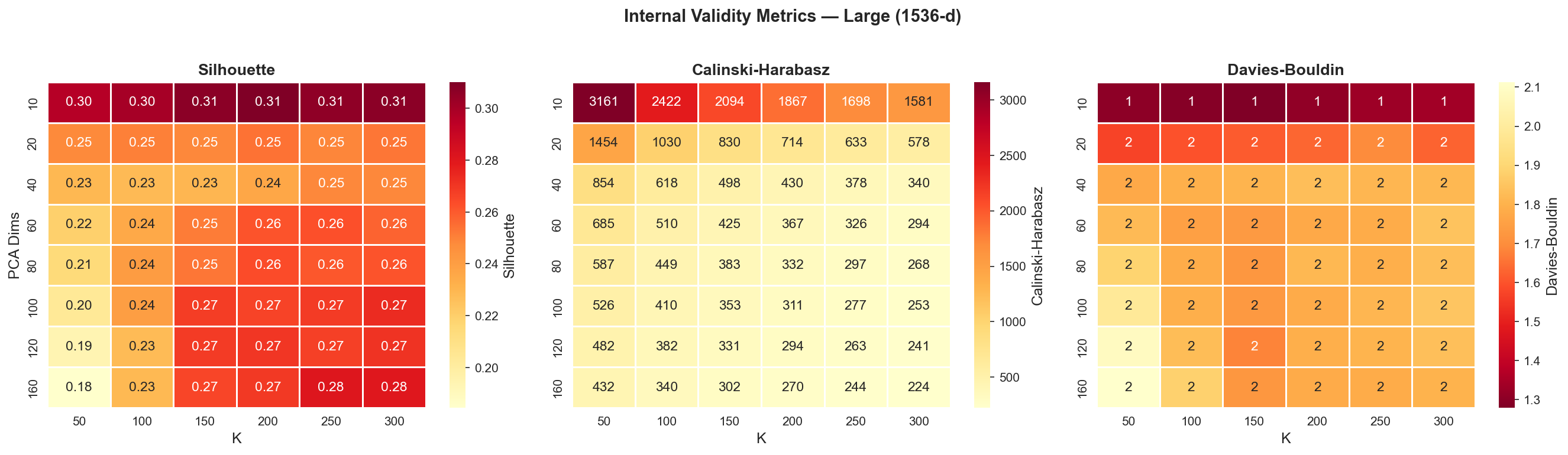}
\end{minipage}
\caption{Additional ASSISTments 2012 structural diagnostics: effect of PCA dimensionality on ARI (left) and internal cluster-validity metrics (right).}
\Description{Additional ASSISTments 2012 structural diagnostics: effect of PCA dimensionality on ARI (left) and internal cluster-validity metrics (right).}
\label{fig:app_c6_struct_2012_pair1}
\end{figure}

\begin{figure}[p]
\centering
\begin{minipage}{0.49\linewidth}
    \centering
    \includegraphics[width=\linewidth]{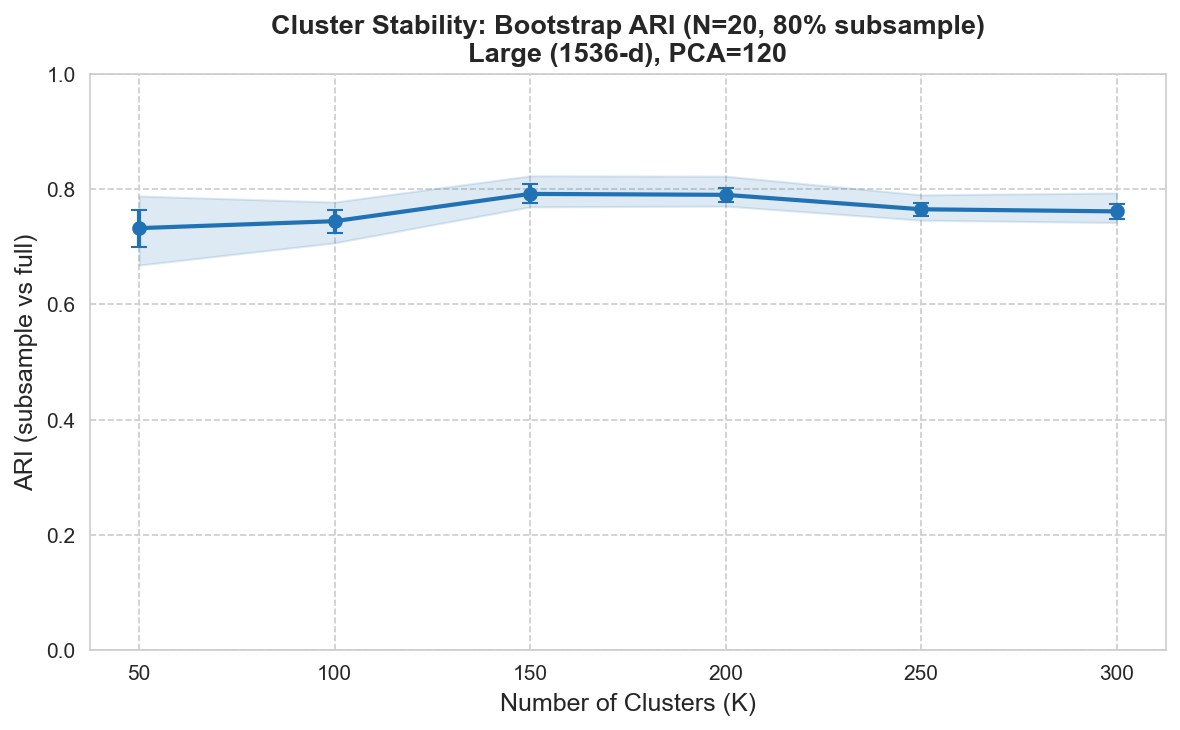}
\end{minipage}\hfill
\begin{minipage}{0.49\linewidth}
    \centering
    \includegraphics[width=\linewidth]{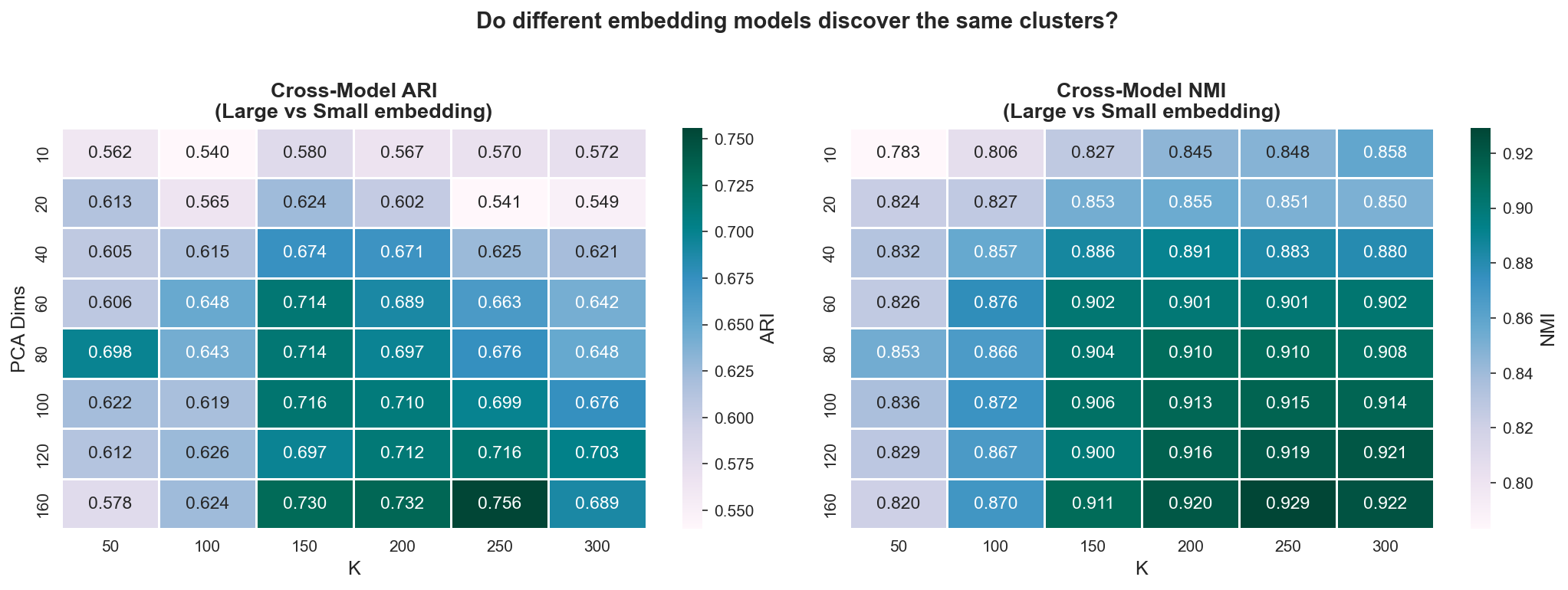}
\end{minipage}
\caption{Additional ASSISTments 2012 structural diagnostics: bootstrap stability (left) and cross-model consistency (right).}
\Description{Additional ASSISTments 2012 structural diagnostics: bootstrap stability (left) and cross-model consistency (right).}
\label{fig:app_c6_struct_2012_pair2}
\end{figure}

\begin{figure}[p]
\centering
\begin{minipage}{0.49\linewidth}
    \centering
    \includegraphics[width=\linewidth]{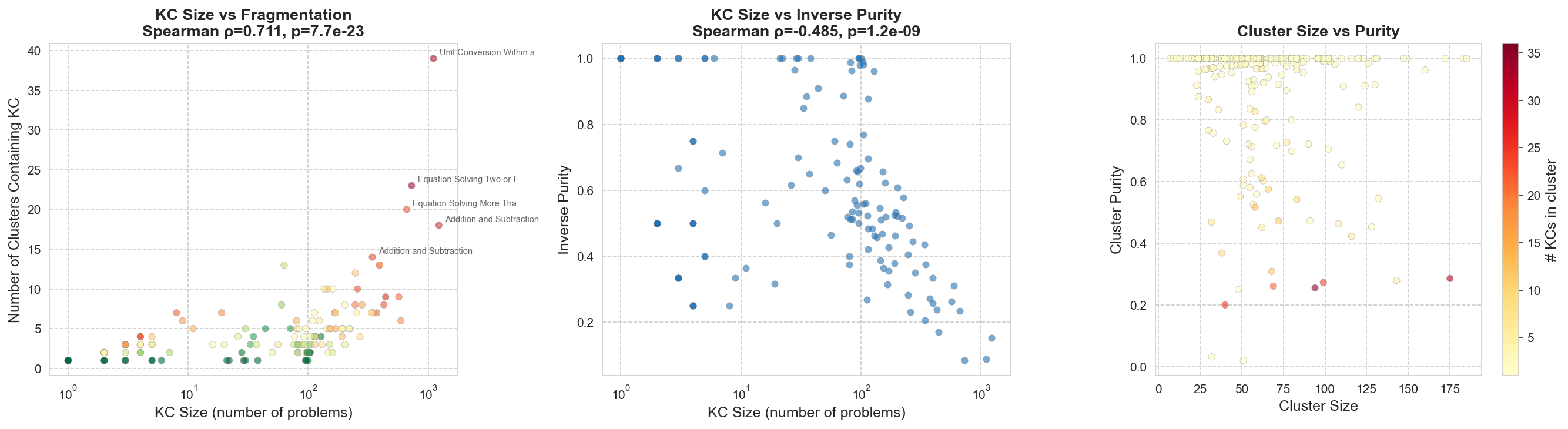}
\end{minipage}\hfill
\begin{minipage}{0.49\linewidth}
    \centering
    \includegraphics[width=\linewidth]{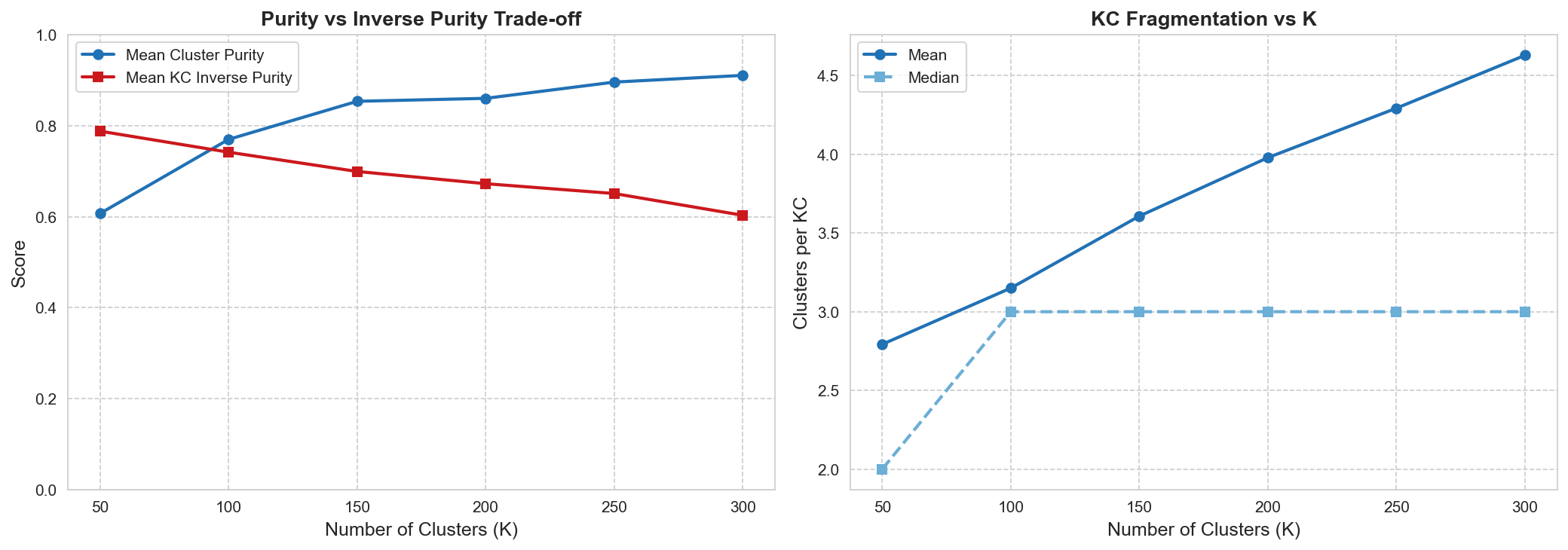}
\end{minipage}
\caption{Additional ASSISTments 2012 structural diagnostics: purity and fragmentation at the analyzed deep-dive anchor (left) and purity-fragmentation trends across cluster counts (right).}
\Description{Additional ASSISTments 2012 structural diagnostics: purity and fragmentation at the analyzed deep-dive anchor (left) and purity-fragmentation trends across cluster counts (right).}
\label{fig:app_c6_struct_2012_pair3}
\end{figure}

\begin{figure}[p]
\centering
\begin{minipage}{0.49\linewidth}
    \centering
    \includegraphics[width=\linewidth]{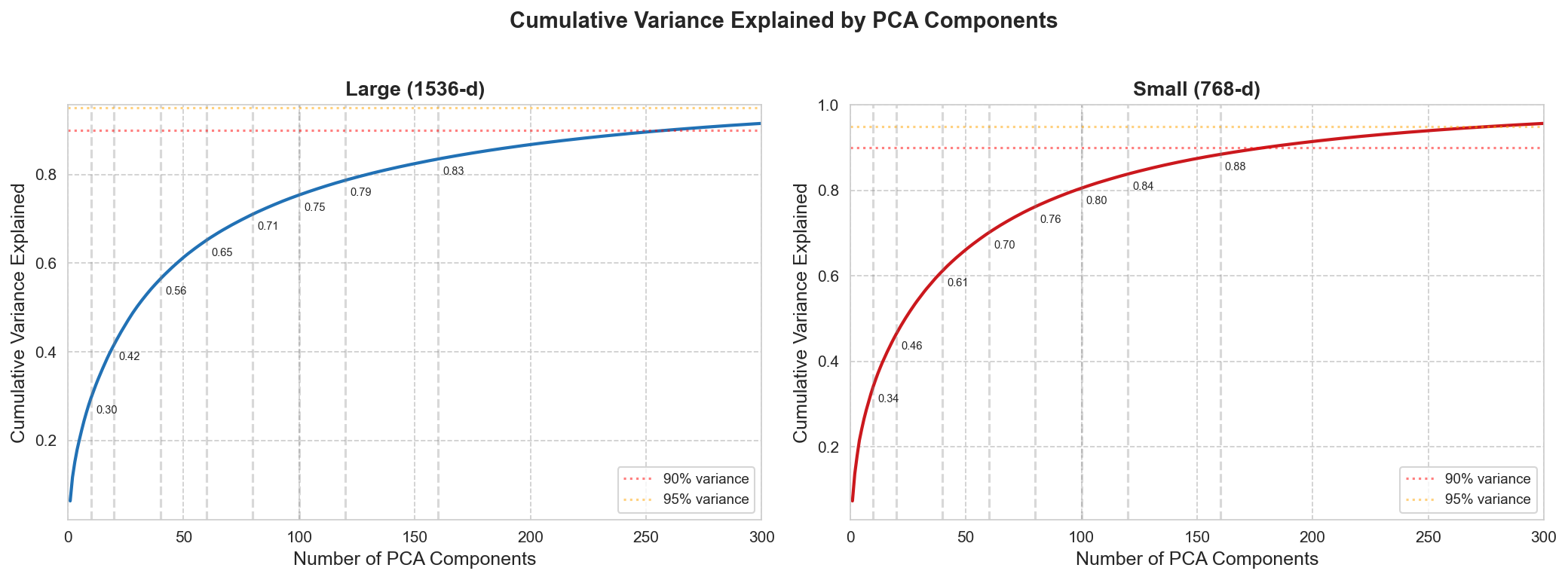}
\end{minipage}\hfill
\begin{minipage}{0.49\linewidth}
    \centering
    \includegraphics[width=\linewidth]{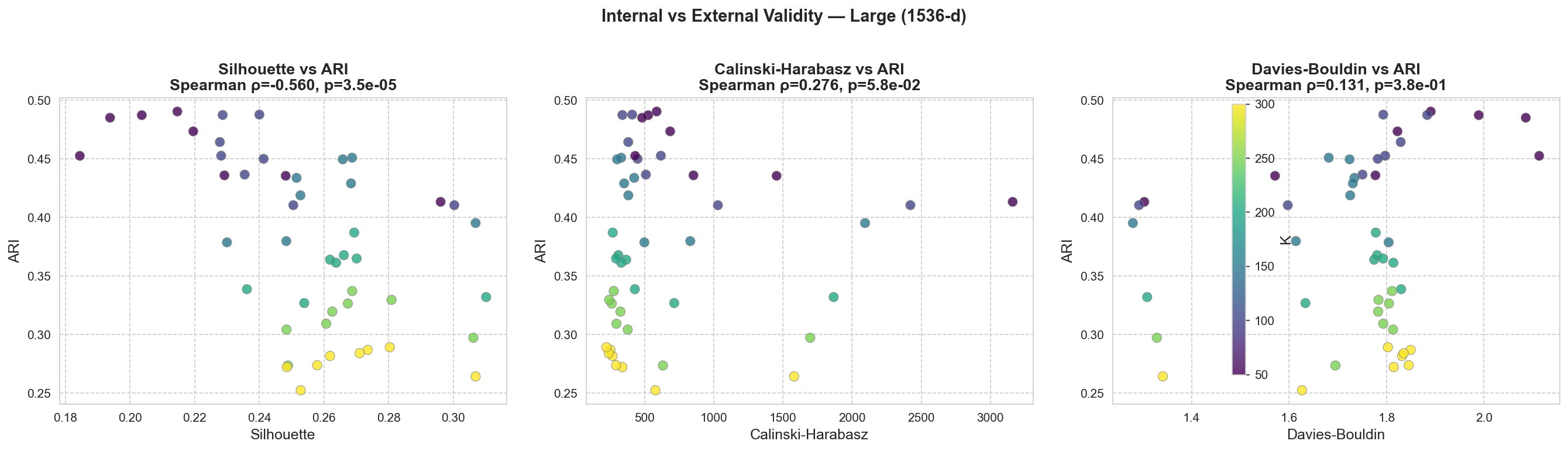}
\end{minipage}
\caption{Additional ASSISTments 2012 structural diagnostics: PCA variance capture (left) and the relation between internal and external validity (right).}
\Description{Additional ASSISTments 2012 structural diagnostics: PCA variance capture (left) and the relation between internal and external validity (right).}
\label{fig:app_c6_struct_2012_pair4}
\end{figure}

\begin{figure}[p]
\centering
\begin{minipage}{0.49\linewidth}
    \centering
    \includegraphics[width=\linewidth]{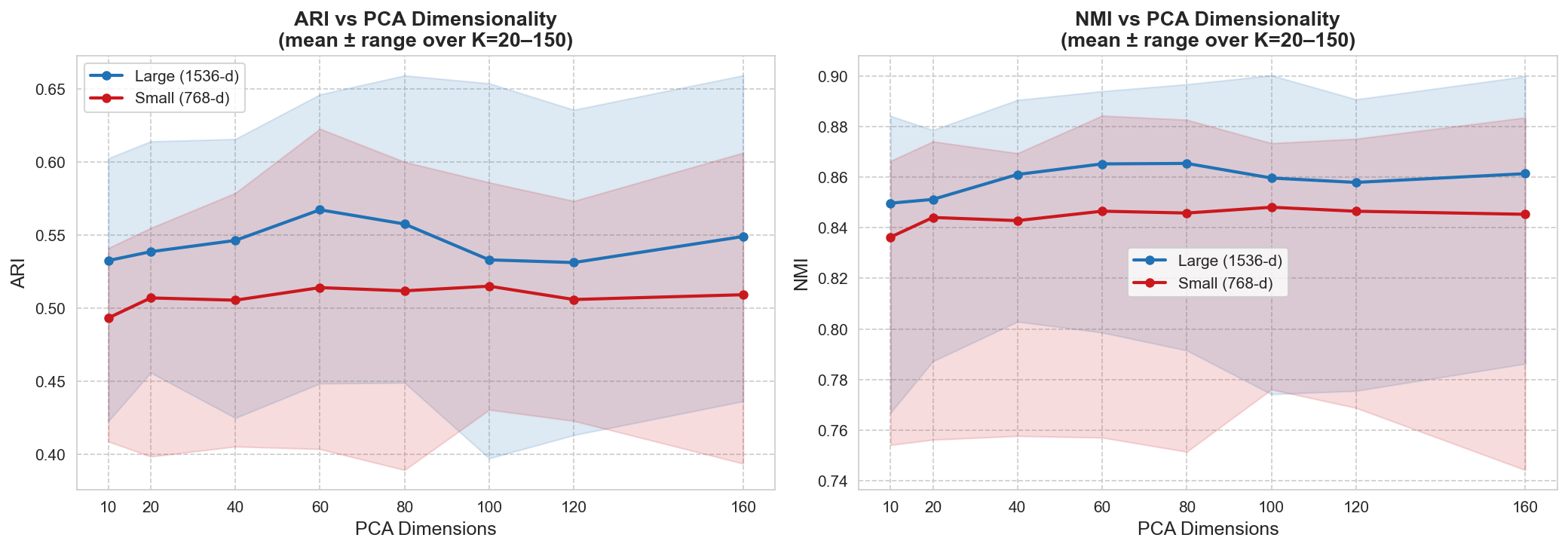}
\end{minipage}\hfill
\begin{minipage}{0.49\linewidth}
    \centering
    \includegraphics[width=\linewidth]{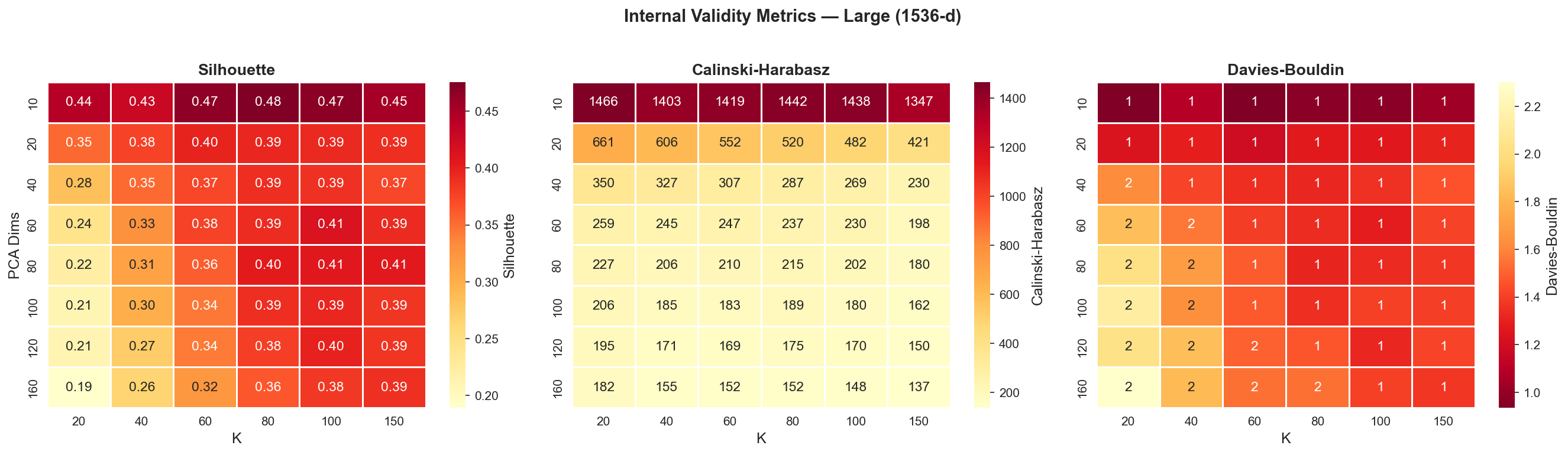}
\end{minipage}
\caption{Additional ASSISTments 2009 structural diagnostics: effect of PCA dimensionality on ARI (left) and internal cluster-validity metrics (right).}
\Description{Additional ASSISTments 2009 structural diagnostics: effect of PCA dimensionality on ARI (left) and internal cluster-validity metrics (right).}
\label{fig:app_c6_struct_2009_pair1}
\end{figure}

\begin{figure}[p]
\centering
\begin{minipage}{0.49\linewidth}
    \centering
    \includegraphics[width=\linewidth]{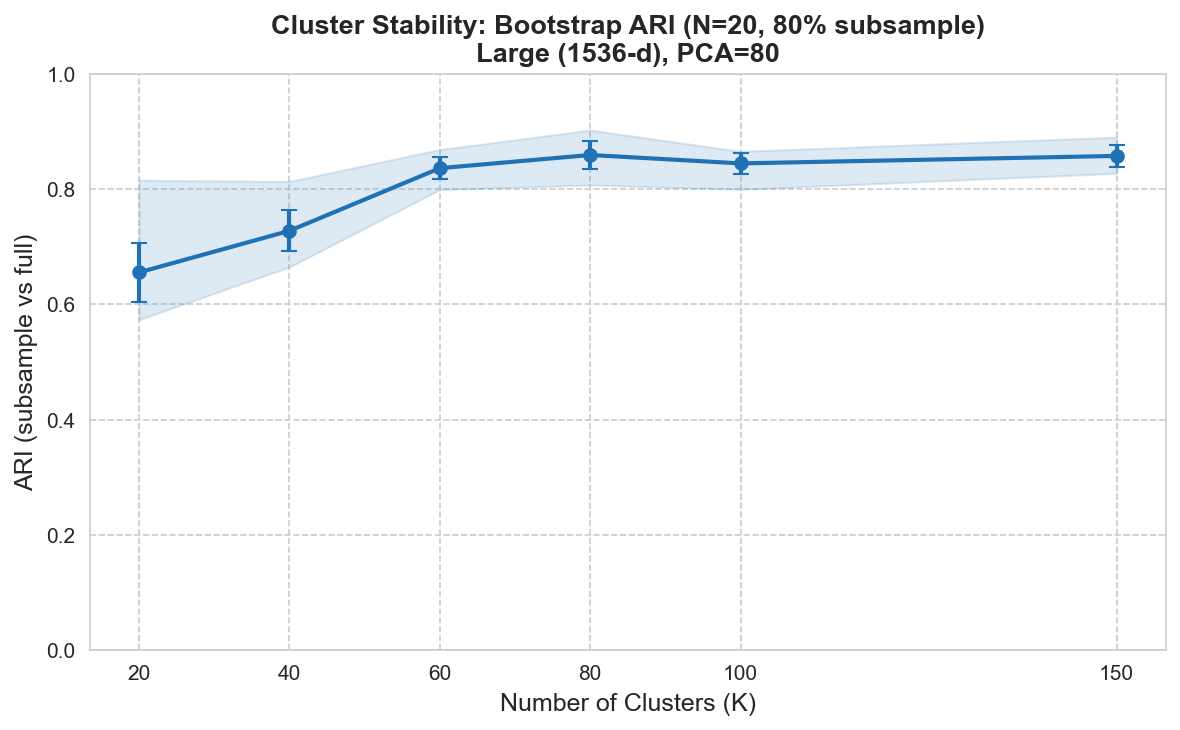}
\end{minipage}\hfill
\begin{minipage}{0.49\linewidth}
    \centering
    \includegraphics[width=\linewidth]{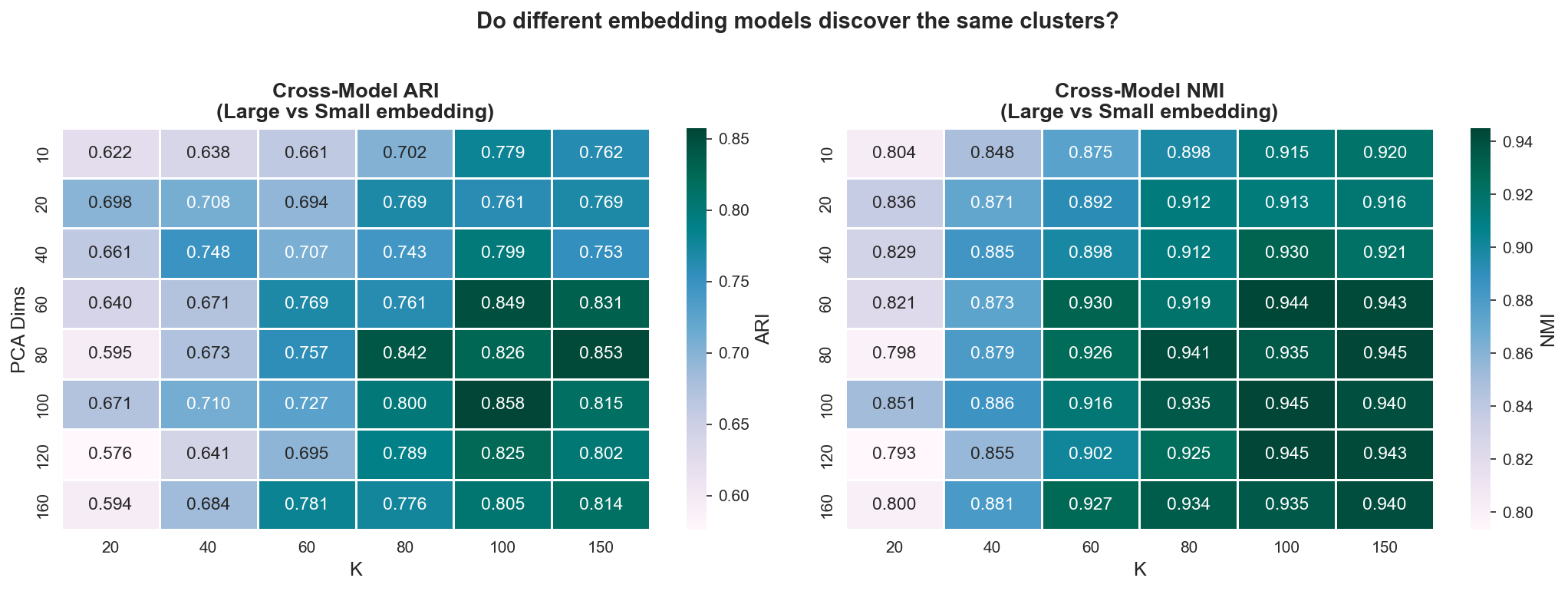}
\end{minipage}
\caption{Additional ASSISTments 2009 structural diagnostics: bootstrap stability (left) and cross-model consistency (right).}
\Description{Additional ASSISTments 2009 structural diagnostics: bootstrap stability (left) and cross-model consistency (right).}
\label{fig:app_c6_struct_2009_pair2}
\end{figure}

\begin{figure}[p]
\centering
\begin{minipage}{0.49\linewidth}
    \centering
    \includegraphics[width=\linewidth]{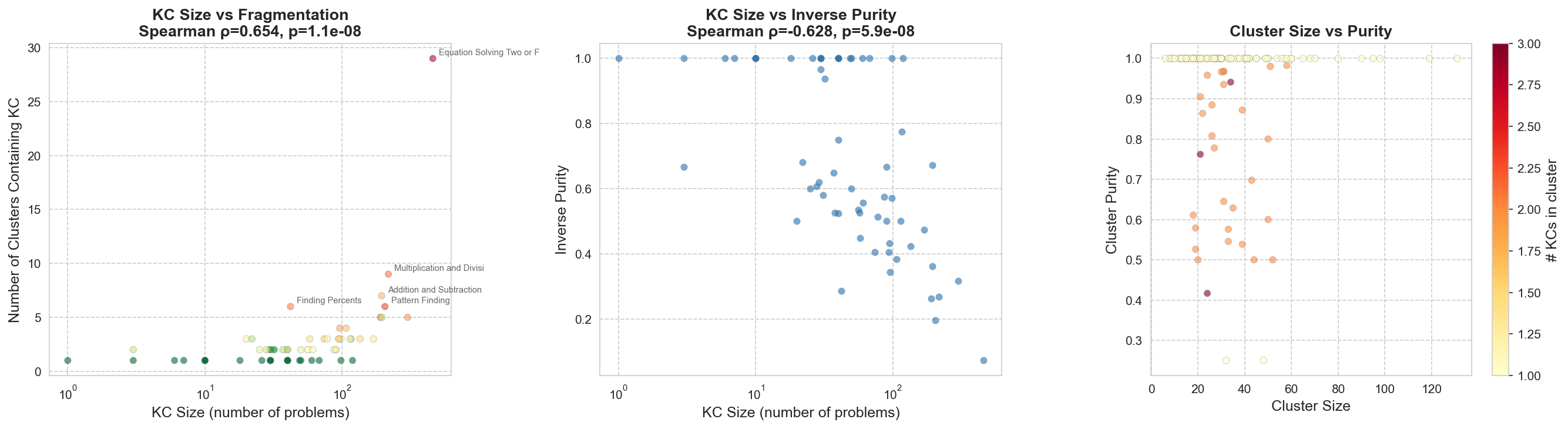}
\end{minipage}\hfill
\begin{minipage}{0.49\linewidth}
    \centering
    \includegraphics[width=\linewidth]{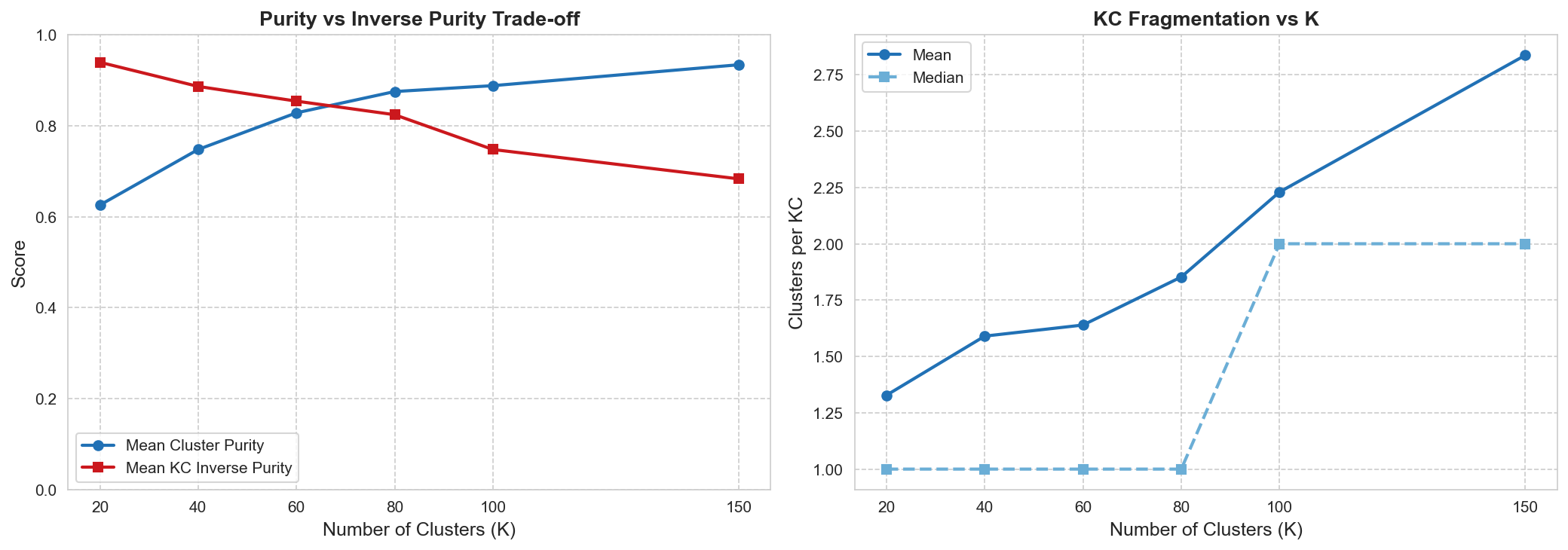}
\end{minipage}
\caption{Additional ASSISTments 2009 structural diagnostics: purity and fragmentation at the analyzed deep-dive anchor (left) and purity-fragmentation trends across cluster counts (right).}
\Description{Additional ASSISTments 2009 structural diagnostics: purity and fragmentation at the analyzed deep-dive anchor (left) and purity-fragmentation trends across cluster counts (right).}
\label{fig:app_c6_struct_2009_pair3}
\end{figure}

\begin{figure}[p]
\centering
\begin{minipage}{0.49\linewidth}
    \centering
    \includegraphics[width=\linewidth]{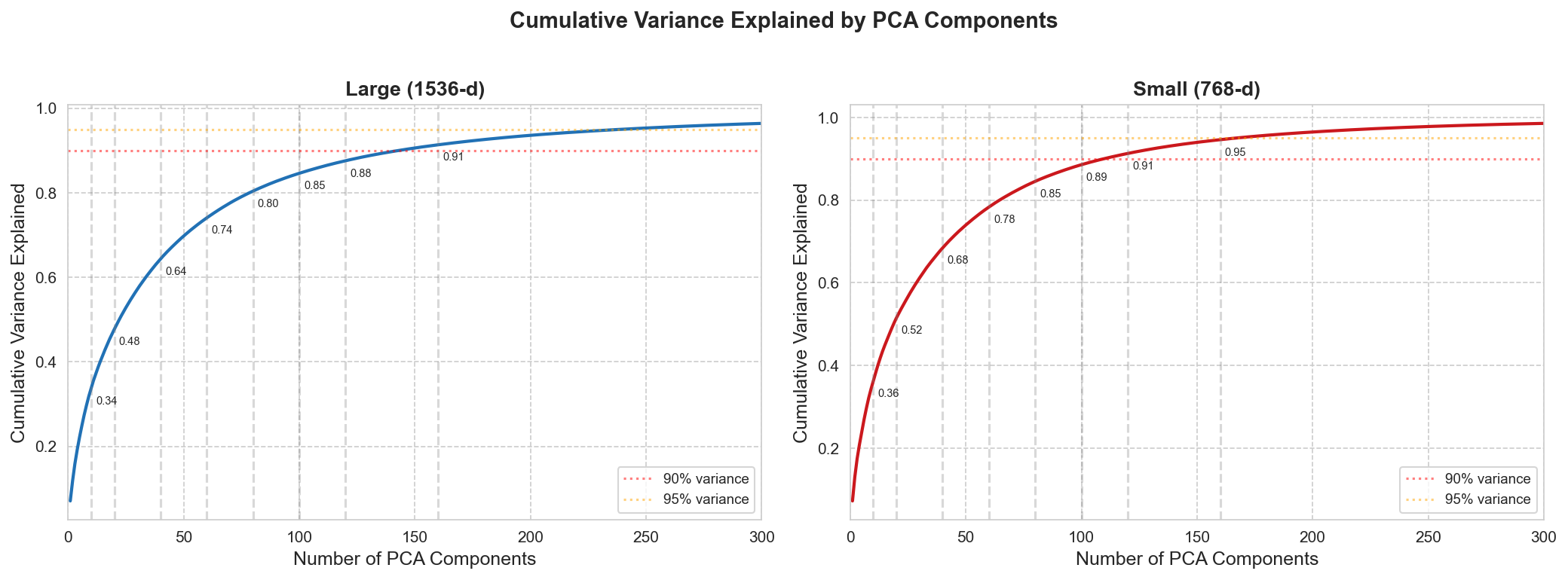}
\end{minipage}\hfill
\begin{minipage}{0.49\linewidth}
    \centering
    \includegraphics[width=\linewidth]{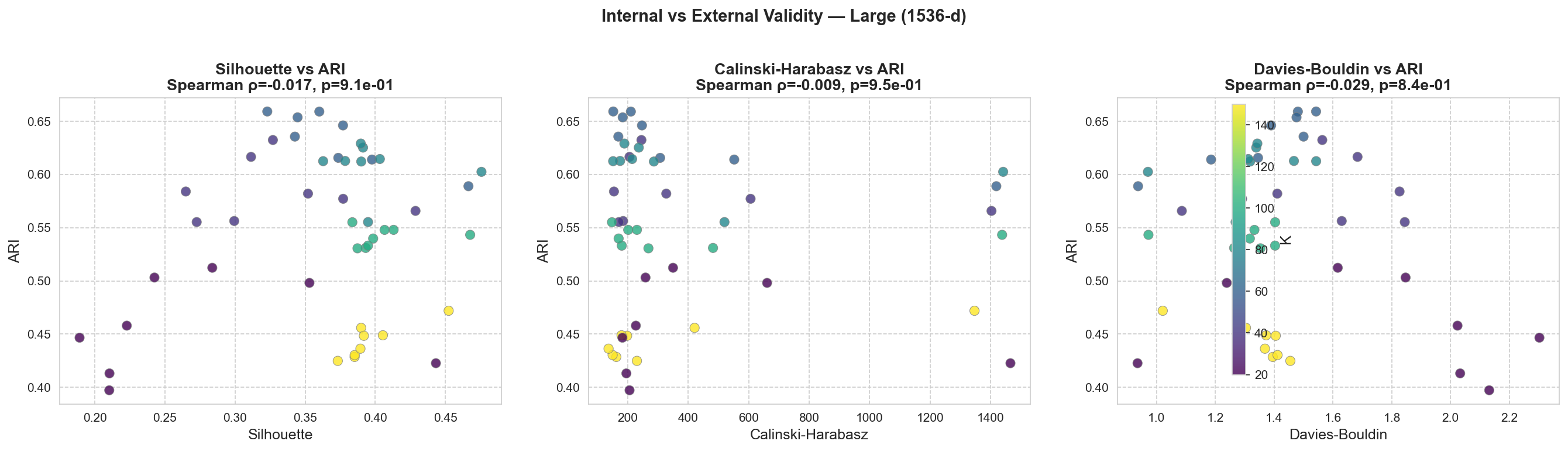}
\end{minipage}
\caption{Additional ASSISTments 2009 structural diagnostics: PCA variance capture (left) and the relation between internal and external validity (right).}
\Description{Additional ASSISTments 2009 structural diagnostics: PCA variance capture (left) and the relation between internal and external validity (right).}
\label{fig:app_c6_struct_2009_pair4}
\end{figure}

\begin{figure}[p]
\centering
\begin{minipage}{0.49\linewidth}
    \centering
    \includegraphics[width=\linewidth]{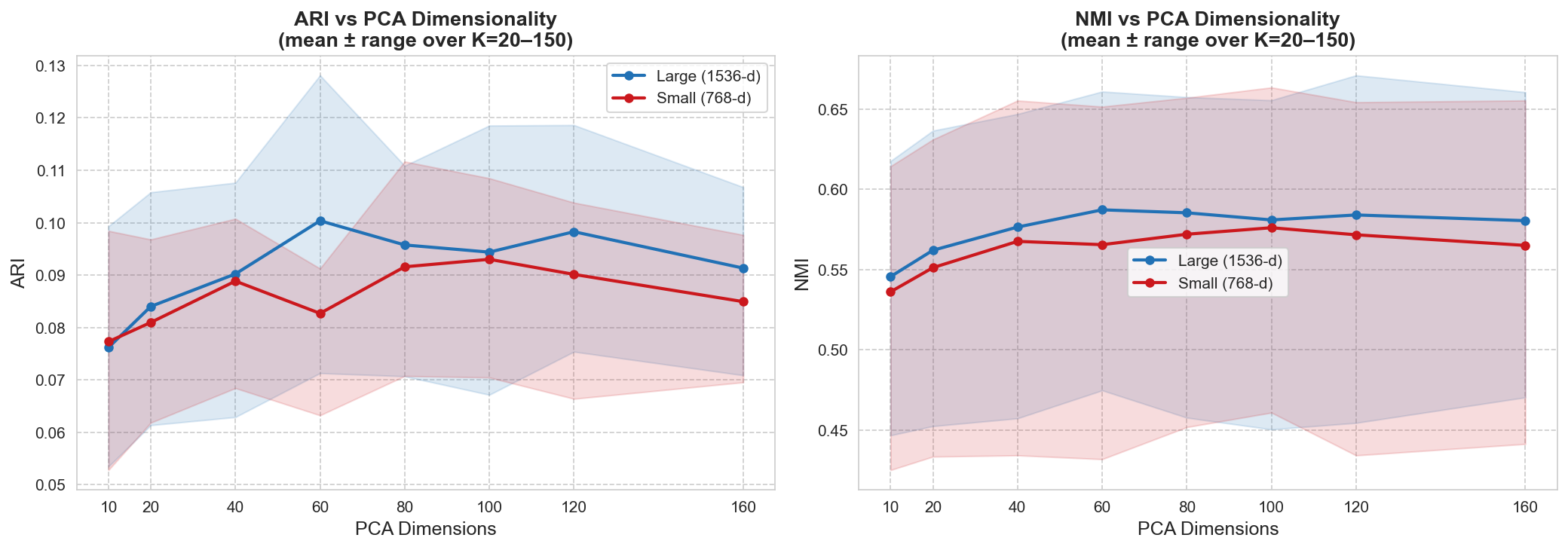}
\end{minipage}\hfill
\begin{minipage}{0.49\linewidth}
    \centering
    \includegraphics[width=\linewidth]{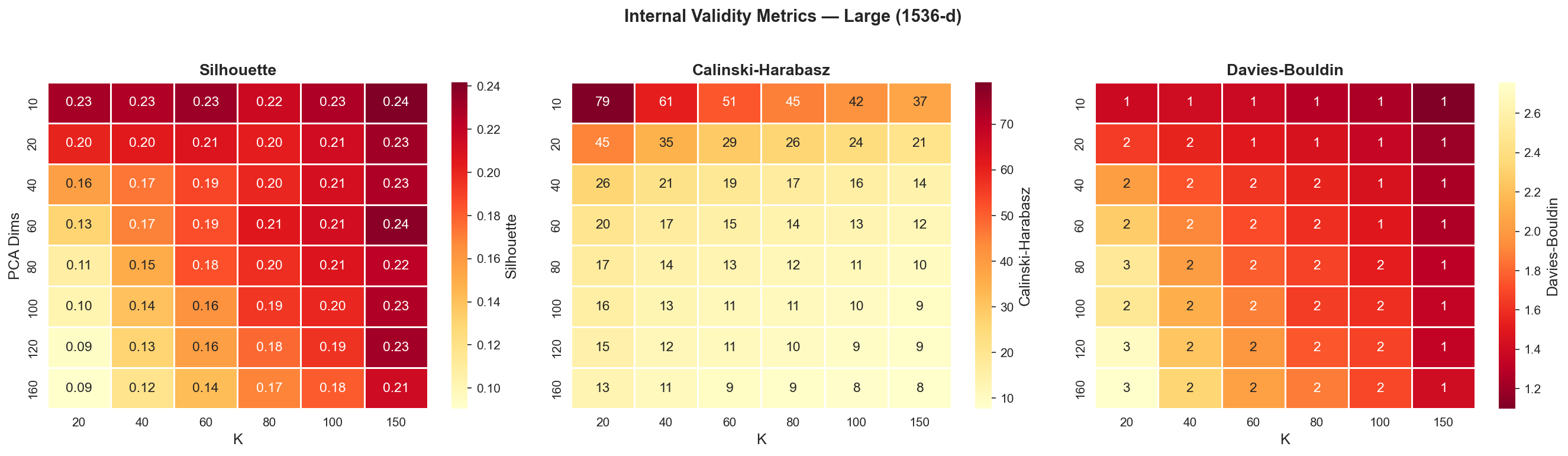}
\end{minipage}
\caption{Additional ASSISTments 2017 structural diagnostics: effect of PCA dimensionality on ARI (left) and internal cluster-validity metrics (right).}
\Description{Additional ASSISTments 2017 structural diagnostics: effect of PCA dimensionality on ARI (left) and internal cluster-validity metrics (right).}
\label{fig:app_c6_struct_2017_pair1}
\end{figure}

\begin{figure}[p]
\centering
\begin{minipage}{0.49\linewidth}
    \centering
    \includegraphics[width=\linewidth]{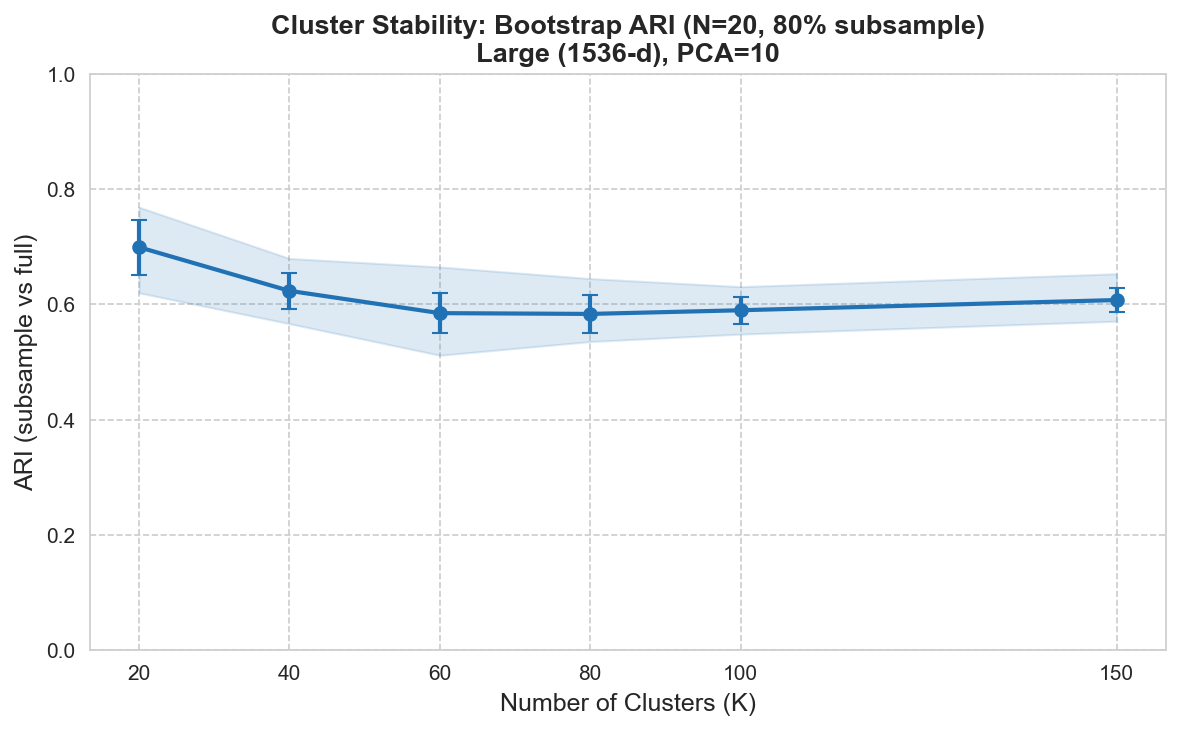}
\end{minipage}\hfill
\begin{minipage}{0.49\linewidth}
    \centering
    \includegraphics[width=\linewidth]{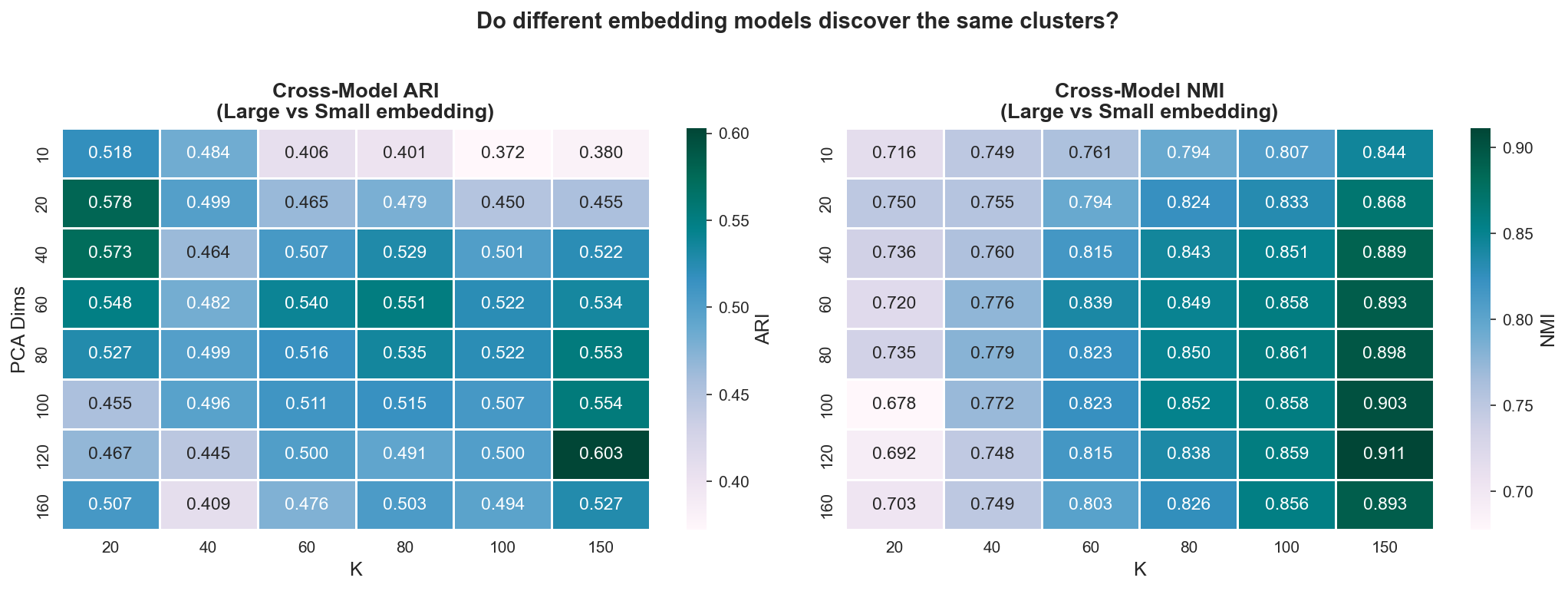}
\end{minipage}
\caption{Additional ASSISTments 2017 structural diagnostics: bootstrap stability (left) and cross-model consistency (right).}
\Description{Additional ASSISTments 2017 structural diagnostics: bootstrap stability (left) and cross-model consistency (right).}
\label{fig:app_c6_struct_2017_pair2}
\end{figure}

\begin{figure}[p]
\centering
\begin{minipage}{0.49\linewidth}
    \centering
    \includegraphics[width=\linewidth]{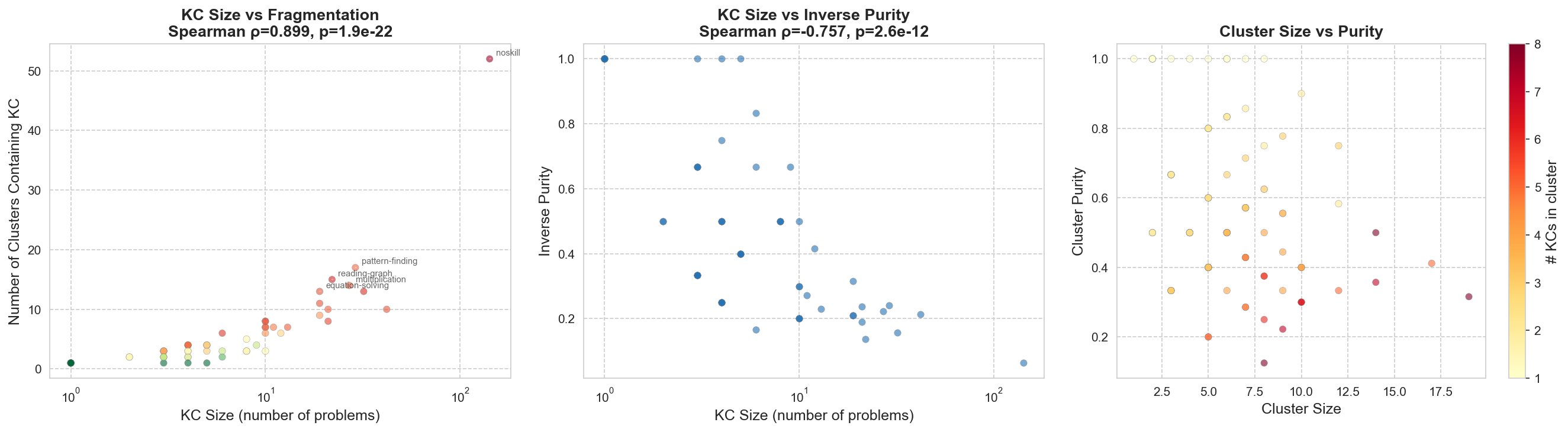}
\end{minipage}\hfill
\begin{minipage}{0.49\linewidth}
    \centering
    \includegraphics[width=\linewidth]{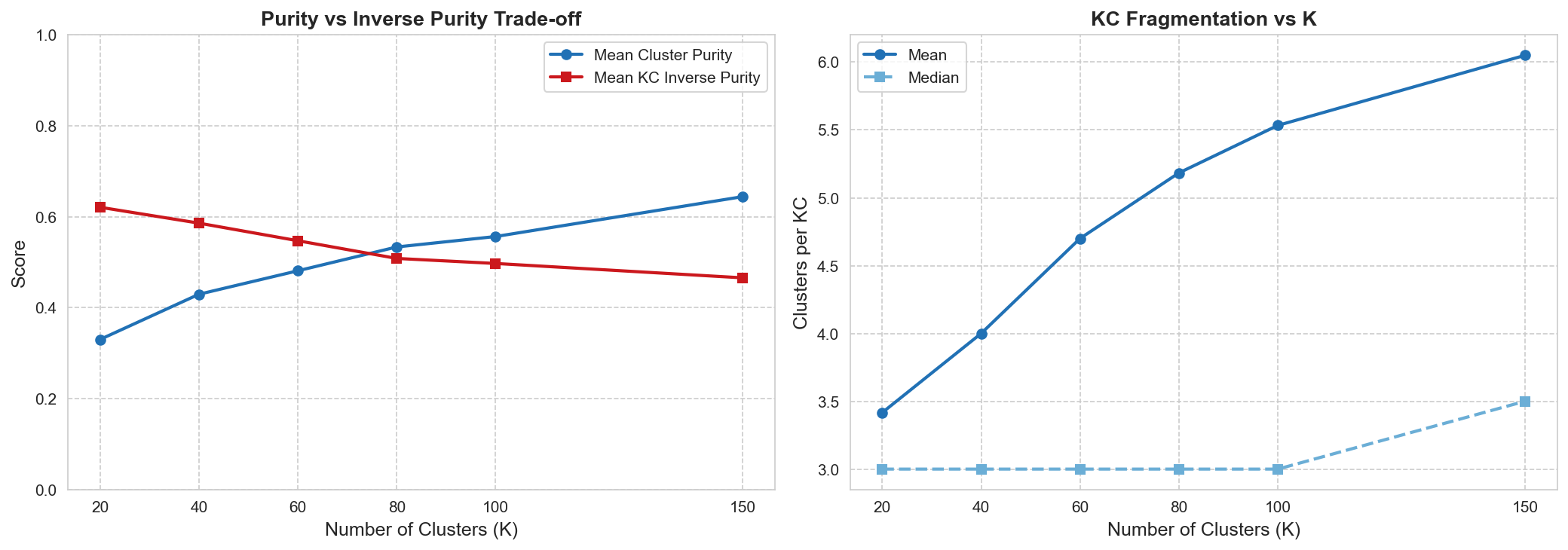}
\end{minipage}
\caption{Additional ASSISTments 2017 structural diagnostics: purity and fragmentation at the analyzed deep-dive anchor (left) and purity-fragmentation trends across cluster counts (right).}
\Description{Additional ASSISTments 2017 structural diagnostics: purity and fragmentation at the analyzed deep-dive anchor (left) and purity-fragmentation trends across cluster counts (right).}
\label{fig:app_c6_struct_2017_pair3}
\end{figure}

\begin{figure}[p]
\centering
\begin{minipage}{0.49\linewidth}
    \centering
    \includegraphics[width=\linewidth]{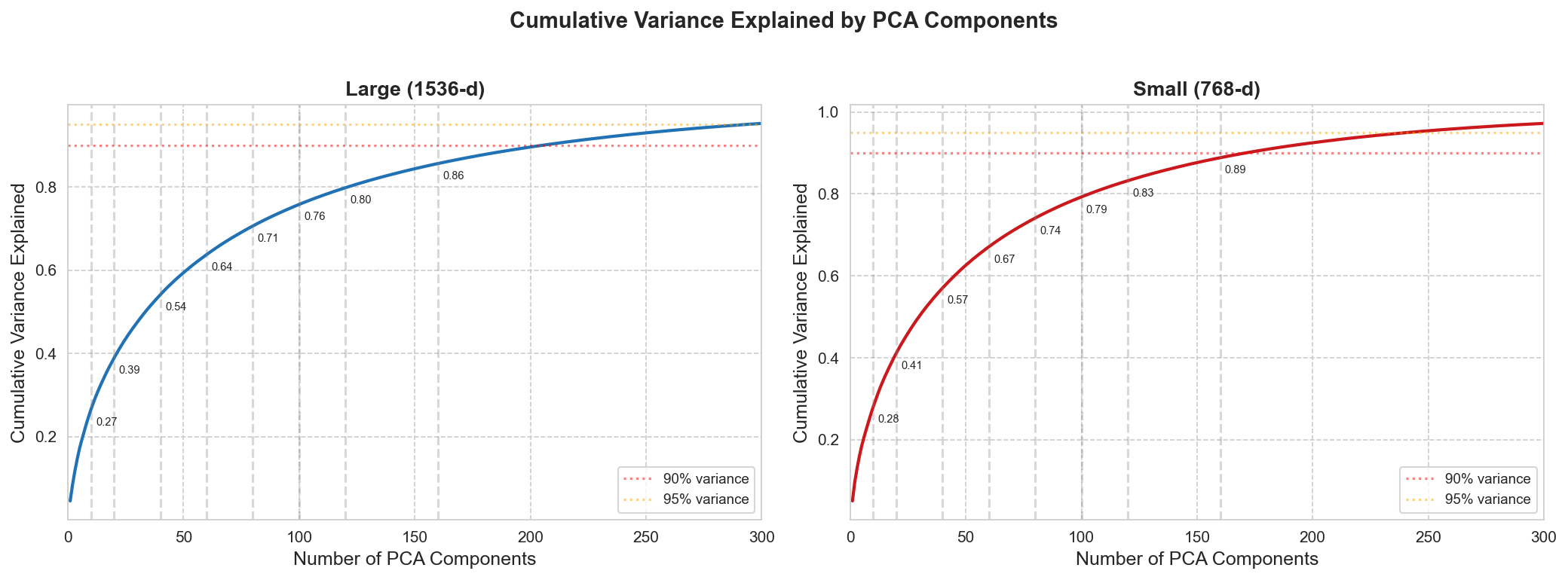}
\end{minipage}\hfill
\begin{minipage}{0.49\linewidth}
    \centering
    \includegraphics[width=\linewidth]{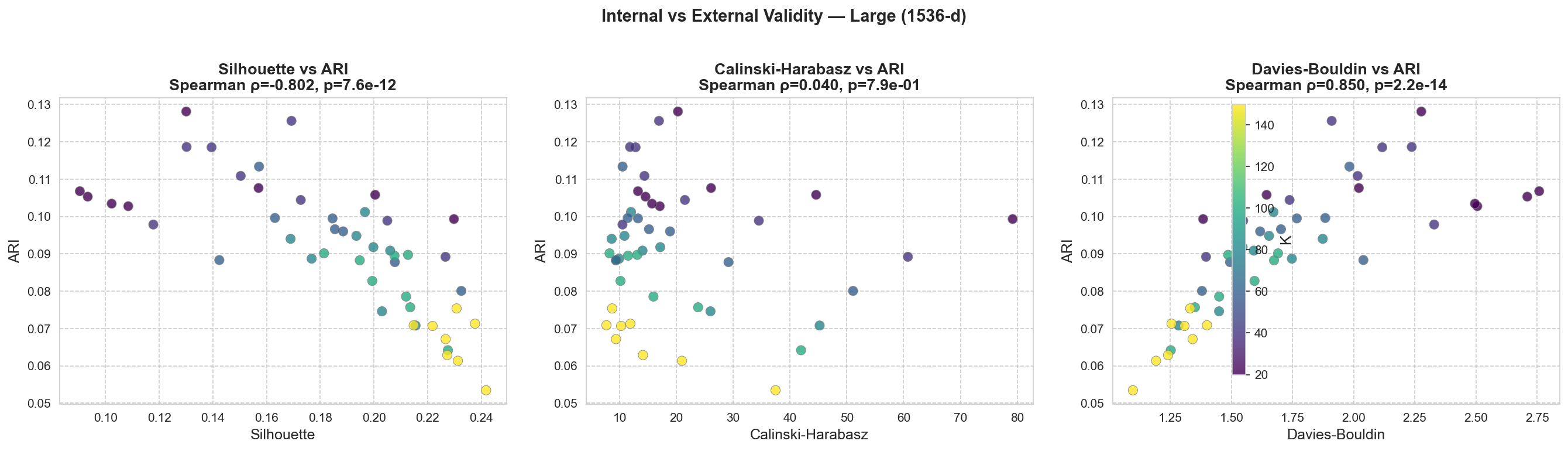}
\end{minipage}
\caption{Additional ASSISTments 2017 structural diagnostics: PCA variance capture (left) and the relation between internal and external validity (right).}
\Description{Additional ASSISTments 2017 structural diagnostics: PCA variance capture (left) and the relation between internal and external validity (right).}
\label{fig:app_c6_struct_2017_pair4}
\end{figure}

Figures~\ref{fig:app_c6_struct_2012_pair1}--\ref{fig:app_c6_struct_2017_pair4} collect additional structural diagnostics for Chapter~6. They show how agreement with CCSS interacts with PCA dimensionality, internal cluster geometry, bootstrap stability, cross-model consistency, fragmentation, variance capture, and the relation between internal and external validity.

For ASSISTments 2012, the structural appendix figures show that the discovered grouping remains substantially aligned with CCSS while still reorganizing the question space. Figures~\ref{fig:app_c6_struct_2012_pair1}--\ref{fig:app_c6_struct_2012_pair4} show that agreement with CCSS remains substantial across a broad region, that the analyzed partition is reasonably stable across resampling and model choices, and that fragmentation concentrates in the larger manual KCs.

For ASSISTments 2009, the structural appendix figures show why the learned partition is not arbitrary even though it does not help DKT. Figures~\ref{fig:app_c6_struct_2009_pair1}--\ref{fig:app_c6_struct_2009_pair4} show stable and sometimes high-agreement clustered regions, but they also show that these regions do not translate into a better supervision structure for the downstream model. This preserves the distinction between recovering coherent semantic structure and producing a better KT taxonomy.

For ASSISTments 2017, the supplementary structural figures place the dataset between the 2012 and 2009 results. Figures~\ref{fig:app_c6_struct_2017_pair1}--\ref{fig:app_c6_struct_2017_pair4} show that low-dimensional embeddings still preserve enough structure for stable regrouping, while the purity-fragmentation and internal-versus-external views show that the refinement region is narrower than on 2012. The 2017 gain comes from a moderate reorganization of the original taxonomy rather than from an aggressive repartitioning of the question bank.

\subsection{Supplementary Tables}
\label{sec:appendix_ch6_tables}

Tables~\ref{tab:app_c6_top_benchmark_configs} and~\ref{tab:app_c6_structural_metric_maxima} report benchmark and structural summaries for Chapter~6.

Table~\ref{tab:app_c6_top_benchmark_configs} summarizes the dataset contrast: the strongest clustered settings on 2012 and 2017 stay above their manual baselines, while the strongest 2009 settings remain below baseline even after tuning. Table~\ref{tab:app_c6_structural_metric_maxima} shows a different pattern: the settings that maximize ARI, NMI, V-measure, or Fowlkes--Mallows do not coincide exactly with the settings that maximize DKT performance.

\begin{table}[p]
\centering
\footnotesize
\DataTableHeaderRows{1}
\begin{tabular}{lcccc}
\toprule
Dataset & PCA Dim & Clusters & AUC & $\Delta$AUC \\
\midrule
ASSISTments 2012 & 100 & 300 & 0.7841 & +0.0474 \\
ASSISTments 2012 & 120 & 300 & 0.7838 & +0.0471 \\
ASSISTments 2012 & 80 & 300 & 0.7835 & +0.0468 \\
ASSISTments 2017 & 10 & 100 & 0.7403 & +0.0272 \\
ASSISTments 2017 & 10 & 150 & 0.7400 & +0.0269 \\
ASSISTments 2017 & 40 & 150 & 0.7382 & +0.0251 \\
ASSISTments 2009 & 10 & 150 & 0.7723 & -0.0822 \\
ASSISTments 2009 & 80 & 150 & 0.7717 & -0.0828 \\
ASSISTments 2009 & 100 & 150 & 0.7703 & -0.0842 \\
\bottomrule
\end{tabular}
\caption{Top clustered benchmark configurations by dataset. These rows summarize the strongest settings from the benchmark sweeps for each dataset.}
\label{tab:app_c6_top_benchmark_configs}
\end{table}

\begin{table}[p]
\centering
\footnotesize
\DataTableHeaderRows{1}
\begin{tabular}{lcccc}
\toprule
Dataset & Best ARI & Best NMI & Best V-measure & Best FM \\
\midrule
ASSISTments 2012 & 0.495 (S,120,50) & 0.838 (L,160,150) & 0.838 (L,160,150) & 0.509 (L,160,100) \\
ASSISTments 2017 & 0.128 (L,60,20) & 0.671 (L,120,150) & 0.671 (L,120,150) & 0.183 (L,60,20) \\
ASSISTments 2009 & 0.659 (L,160,60) & 0.900 (L,100,60) & 0.900 (L,100,60) & 0.673 (L,80,60) \\
\bottomrule
\end{tabular}
\caption{Best structural-agreement configurations by metric. \texttt{L} denotes \texttt{text-embedding-3-large} shortened to 1536 dimensions in this structural sweep, and \texttt{S} denotes \texttt{text-embedding-3-small} shortened to 768 dimensions. These maxima do not coincide exactly with the best downstream DKT settings.}
\label{tab:app_c6_structural_metric_maxima}
\end{table}

\section{Chapter 7 Supplementary Material}

Tables~\ref{tab:app_ch7_assist2009_students}, \ref{tab:app_ch7_assist2012_students}, and~\ref{tab:app_ch7_assist2017_students} give the full per-student AUC breakdown for the student-level variability analysis in Chapter~7.

\subsection{Detailed Student-Level Prompt-Only ReasonKT Results}
\label{sec:appendix_ch7_student_tables}

Student-level behavior varies substantially even after the balanced selection procedure removes trivial one-class slices.

\begin{table}[p]
\centering
\footnotesize
\DataTableHeaderRows{1}
\begin{tabular}{lrrrrrr}
\toprule
Student ID & Examples & Positives & Negatives & Pos. Rate & AUC & Accuracy \\
\midrule
87132 & 24 & 12 & 12 & 0.5000 & 0.7222 & 0.7083 \\
78888 & 48 & 24 & 24 & 0.5000 & 0.7188 & 0.6875 \\
78184 & 20 & 10 & 10 & 0.5000 & 0.7100 & 0.6500 \\
79213 & 46 & 23 & 23 & 0.5000 & 0.6682 & 0.6087 \\
89164 & 26 & 13 & 13 & 0.5000 & 0.6272 & 0.5769 \\
84837 & 20 & 10 & 10 & 0.5000 & 0.6200 & 0.6000 \\
78365 & 36 & 18 & 18 & 0.5000 & 0.6127 & 0.6111 \\
78805 & 32 & 16 & 16 & 0.5000 & 0.5332 & 0.5625 \\
84178 & 66 & 33 & 33 & 0.5000 & 0.4412 & 0.4394 \\
77954 & 20 & 10 & 10 & 0.5000 & 0.1550 & 0.3000 \\
\bottomrule
\end{tabular}
\caption{Detailed per-student results for the ASSISTments 2009 slice used in the Chapter~7 Prompt-Only ReasonKT study. Student-level performance varies sharply even under balanced selection.}
\label{tab:app_ch7_assist2009_students}
\end{table}

\begin{table}[p]
\centering
\footnotesize
\DataTableHeaderRows{1}
\begin{tabular}{lrrrrrr}
\toprule
Student ID & Examples & Positives & Negatives & Pos. Rate & AUC & Accuracy \\
\midrule
169382 & 94 & 47 & 47 & 0.5000 & 0.7481 & 0.6596 \\
145720 & 78 & 39 & 39 & 0.5000 & 0.6637 & 0.6026 \\
125432 & 76 & 38 & 38 & 0.5000 & 0.6562 & 0.6184 \\
223735 & 98 & 49 & 49 & 0.5000 & 0.6464 & 0.6122 \\
118398 & 146 & 73 & 73 & 0.5000 & 0.6372 & 0.5890 \\
219180 & 78 & 39 & 39 & 0.5000 & 0.5986 & 0.5769 \\
183337 & 96 & 48 & 48 & 0.5000 & 0.5966 & 0.5417 \\
183451 & 222 & 111 & 111 & 0.5000 & 0.5870 & 0.5586 \\
188131 & 90 & 45 & 45 & 0.5000 & 0.5783 & 0.5111 \\
179458 & 80 & 40 & 40 & 0.5000 & 0.5506 & 0.5500 \\
\bottomrule
\end{tabular}
\caption{Detailed per-student results for the ASSISTments 2012 slice used in the Chapter~7 Prompt-Only ReasonKT study. The spread is narrower than in ASSISTments 2009, but the ranking quality still differs meaningfully across students.}
\label{tab:app_ch7_assist2012_students}
\end{table}

\begin{table}[p]
\centering
\footnotesize
\DataTableHeaderRows{1}
\begin{tabular}{lrrrrrr}
\toprule
Student ID & Examples & Positives & Negatives & Pos. Rate & AUC & Accuracy \\
\midrule
156 & 88 & 44 & 44 & 0.5000 & 0.7368 & 0.6705 \\
3476 & 96 & 48 & 48 & 0.5000 & 0.7211 & 0.6667 \\
161 & 146 & 73 & 73 & 0.5000 & 0.7094 & 0.6918 \\
2518 & 134 & 67 & 67 & 0.5000 & 0.6579 & 0.6269 \\
2403 & 106 & 53 & 53 & 0.5000 & 0.6445 & 0.6226 \\
5785 & 80 & 40 & 40 & 0.5000 & 0.6153 & 0.6125 \\
6731 & 130 & 65 & 65 & 0.5000 & 0.6128 & 0.6385 \\
7251 & 86 & 43 & 43 & 0.5000 & 0.5673 & 0.5698 \\
3445 & 82 & 41 & 41 & 0.5000 & 0.4991 & 0.5122 \\
1350 & 104 & 52 & 52 & 0.5000 & 0.4887 & 0.5385 \\
\bottomrule
\end{tabular}
\caption{Detailed per-student results for the ASSISTments 2017 slice used in the Chapter~7 Prompt-Only ReasonKT study. Performance is weak overall but varies substantially across students.}
\label{tab:app_ch7_assist2017_students}
\end{table}

\clearpage
\phantomsection
\renewcommand{\bibname}{References}
\addcontentsline{toc}{chapter}{References}
\bibliographystyle{plain}
\bibliography{references}

\begin{thebibliography}{100}

\bibitem{abdelrahmanKnowledgeTracingSurvey2023}
Ghodai Abdelrahman, Qing Wang, and Bernardo Nunes.
\newblock Knowledge {Tracing}: {A} {Survey}.
\newblock {\em ACM Comput. Surv.}, 55(11):224:1--224:37, February 2023.

\bibitem{arthurKmeansPlusPlus2007}
David Arthur and Sergei Vassilvitskii.
\newblock k-means++: The advantages of careful seeding.
\newblock In {\em Proceedings of the Eighteenth Annual ACM-SIAM Symposium on Discrete Algorithms}, pages 1027--1035, 2007.

\bibitem{explainable_kt}
Yanhong Bai, Jiabao Zhao, Tingjiang Wei, Qing Cai, and Liang He.
\newblock A {Survey} of {Explainable} {Knowledge} {Tracing}, March 2024.
\newblock arXiv:2403.07279 [cs.CL].

\bibitem{blobstein_angel_2024}
Ariel Blobstein, Daniel Izmaylov, Tal Yifal, Michal Levy, and Avi Segal.
\newblock Angel: {A} {New} {Generation} {Tool} for {Learning} {Material} based {Questions} and {Answers}.
\newblock 2024.

\bibitem{blondelFastUnfoldingCommunities2008}
Vincent~D. Blondel, Jean-Loup Guillaume, Renaud Lambiotte, and Etienne Lefebvre.
\newblock Fast unfolding of communities in large networks.
\newblock {\em Journal of Statistical Mechanics: Theory and Experiment}, 2008(10):P10008, October 2008.
\newblock arXiv:0803.0476 [physics].

\bibitem{bojanowski_enriching_2017}
Piotr Bojanowski, Edouard Grave, Armand Joulin, and Tomas Mikolov.
\newblock Enriching {Word} {Vectors} with {Subword} {Information}.
\newblock {\em Transactions of the Association for Computational Linguistics}, 5:135--146, 2017.

\bibitem{brown_language_2020}
Tom~B. Brown, Benjamin Mann, Nick Ryder, Melanie Subbiah, Jared Kaplan, Prafulla Dhariwal, Arvind Neelakantan, Pranav Shyam, Girish Sastry, Amanda Askell, et~al.
\newblock Language models are few-shot learners.
\newblock In {\em Advances in Neural Information Processing Systems}, 2020.

\bibitem{cenLearningFactorsAnalysis2006}
Hao Cen, Kenneth Koedinger, and Brian Junker.
\newblock Learning {Factors} {Analysis} – {A} {General} {Method} for {Cognitive} {Model} {Evaluation} and {Improvement}.
\newblock In Mitsuru Ikeda, Kevin~D. Ashley, and Tak-Wai Chan, editors, {\em Intelligent {Tutoring} {Systems}}, pages 164--175, Berlin, Heidelberg, 2006. Springer.

\bibitem{chen_chatgpt_time_2023}
Lingjiao Chen, Matei Zaharia, and James Zou.
\newblock How is {ChatGPT}'s behavior changing over time?, 2023.
\newblock arXiv:2307.09009 [cs].

\bibitem{chen_oscars_2025}
Nuo Chen, Yan Wang, Yang Deng, and Jia Li.
\newblock The {Oscars} of {AI} {Theater}: {A} {Survey} on {Role}-{Playing} with {Language} {Models}, January 2025.
\newblock arXiv:2407.11484 [cs].

\bibitem{chenPrerequisiteDrivenDeepKnowledge2018}
Penghe Chen, Yu~Lu, Vincent~W. Zheng, and Yang Pian.
\newblock Prerequisite-{Driven} {Deep} {Knowledge} {Tracing}.
\newblock In {\em 2018 {IEEE} {International} {Conference} on {Data} {Mining} ({ICDM})}, pages 39--48, Singapore, November 2018. IEEE.

\bibitem{ccssValidationReport2010}
{Common Core State Standards Validation Committee}.
\newblock Reaching higher: The common core state standards validation committee final report, 2010.
\newblock Official validation report.

\bibitem{corbett_knowledge_1995}
Albert~T. Corbett and John~R. Anderson.
\newblock Knowledge tracing: {Modeling} the acquisition of procedural knowledge.
\newblock {\em User Modelling and User-Adapted Interaction}, 4(4):253--278, 1995.

\bibitem{devlin_bert_2019}
Jacob Devlin, Ming-Wei Chang, Kenton Lee, and Kristina Toutanova.
\newblock {BERT}: {Pre}-training of {Deep} {Bidirectional} {Transformers} for {Language} {Understanding}.
\newblock In {\em Proceedings of the 2019 Conference of the North {American} Chapter of the {Association} for {Computational} {Linguistics}: {Human} {Language} {Technologies}, {Volume} 1 ({Long} and {Short} Papers)}, pages 4171--4186, 2019.

\bibitem{du_improving_2023}
Yilun Du, Shuang Li, Antonio Torralba, Joshua~B. Tenenbaum, and Igor Mordatch.
\newblock Improving {Factuality} and {Reasoning} in {Language} {Models} through {Multiagent} {Debate}, May 2023.
\newblock arXiv:2305.14325 [cs].

\bibitem{duanMoreAccurateInterpretable2024}
Zhiyi Duan, Xiaoxiao Dong, Hengnian Gu, Xiong Wu, Zhen Li, and Dongdai Zhou.
\newblock Towards more accurate and interpretable model: {Fusing} multiple knowledge relations into deep knowledge tracing.
\newblock {\em Expert Systems with Applications}, 243:122573, June 2024.

\bibitem{fengAddressingAssessmentChallenge2009}
Mingyu Feng, Neil Heffernan, and Kenneth Koedinger.
\newblock Addressing the assessment challenge with an online system that tutors as it assesses.
\newblock {\em User Modeling and User-Adapted Interaction}, 19(3):243--266, August 2009.

\bibitem{fengSystematicReviewLiterature2021}
Shi Feng, Alejandra~J. Magana, and Dominic Kao.
\newblock A {Systematic} {Review} of {Literature} on the {Effectiveness} of {Intelligent} {Tutoring} {Systems} in {STEM}.
\newblock In {\em 2021 {IEEE} {Frontiers} in {Education} {Conference} ({FIE})}, pages 1--9, Lincoln, NE, USA, October 2021. IEEE.

\bibitem{fu_qgeval_2024}
Weiping Fu, Bifan Wei, Jianxiang Hu, Zhongmin Cai, and Jun Liu.
\newblock {QGEval}: {Benchmarking} {Multi}-dimensional {Evaluation} for {Question} {Generation}.
\newblock In {\em Proceedings of the 2024 {Conference} on {Empirical} {Methods} in {Natural} {Language} {Processing}}, pages 11783--11803, Miami, Florida, USA, 2024. Association for Computational Linguistics.

\bibitem{gao_simcse_2021}
Tianyu Gao, Xingcheng Yao, and Danqi Chen.
\newblock {SimCSE}: {Simple} {Contrastive} {Learning} of {Sentence} {Embeddings}.
\newblock In {\em Proceedings of the 2021 Conference on Empirical Methods in Natural Language Processing}, pages 6894--6910, 2021.

\bibitem{ghosh_context-aware_2020}
Aritra Ghosh, Neil Heffernan, and Andrew~S. Lan.
\newblock Context-{Aware} {Attentive} {Knowledge} {Tracing}.
\newblock In {\em Proceedings of the 26th {ACM} {SIGKDD} {International} {Conference} on {Knowledge} {Discovery} \& {Data} {Mining}}, pages 2330--2339, 2020.

\bibitem{guo_calibration_2017}
Chuan Guo, Geoff Pleiss, Yu~Sun, and Kilian~Q. Weinberger.
\newblock On calibration of modern neural networks.
\newblock In {\em Proceedings of the 34th International Conference on Machine Learning}, 2017.

\bibitem{guo_deepseekr1_2025}
Daya Guo, Dejian Yang, Haowei Zhang, Junxiao Song, Ruoyu Zhang, Runxin Xu, Qihao Zhu, et~al.
\newblock {DeepSeek-R1}: Incentivizing reasoning capability in {LLMs} via reinforcement learning.
\newblock {\em Nature}, 2025.

\bibitem{heffernanASSISTmentsEcosystem2014}
Neil~T. Heffernan and Cristina~L. Heffernan.
\newblock The assistments ecosystem: Building a platform that brings scientists and teachers together for minimally invasive research on human learning and teaching.
\newblock {\em International Journal of Artificial Intelligence in Education}, 24(4):470--497, 2014.

\bibitem{heilman_good_2010}
Michael Heilman and Noah~A. Smith.
\newblock Good {Question}! {Statistical} {Ranking} for {Question} {Generation}.
\newblock In Ron Kaplan, Jill Burstein, Mary Harper, and Gerald Penn, editors, {\em Human {Language} {Technologies}: {The} 2010 {Annual} {Conference} of the {North} {American} {Chapter} of the {Association} for {Computational} {Linguistics}}, pages 609--617, Los Angeles, California, June 2010. Association for Computational Linguistics.

\bibitem{hong_metagpt_2024}
Sirui Hong, Mingchen Zhuge, Jiaqi Chen, Xiawu Zheng, Yuheng Cheng, Ceyao Zhang, Jinlin Wang, Zili Wang, Steven Ka~Shing Yau, Zijuan Lin, Liyang Zhou, Chenyu Ran, Lingfeng Xiao, Chenglin Wu, and Jürgen Schmidhuber.
\newblock {MetaGPT}: {Meta} {Programming} for {A} {Multi}-{Agent} {Collaborative} {Framework}, November 2024.
\newblock arXiv:2308.00352 [cs].

\bibitem{hubert_comparing_1985}
Lawrence Hubert and Phipps Arabie.
\newblock Comparing partitions.
\newblock {\em Journal of Classification}, 2(1):193--218, 1985.

\bibitem{hwang_contextualized_2024}
Wu-Yuin Hwang and Ika~Qutsiati Utami.
\newblock Contextualized and {Personalized} {Math} {Word} {Problem} {Generation} {Using} {GPT} and {Authentic} {Contextual} {Recognition}.
\newblock In {\em 2024 12th {International} {Conference} on {Information} and {Education} {Technology} ({ICIET})}, pages 177--181, Yamaguchi, Japan, March 2024. IEEE.

\bibitem{jain_attention_2019}
Sarthak Jain and Byron~C. Wallace.
\newblock Attention is not explanation.
\newblock In {\em Proceedings of the 2019 Conference of the North American Chapter of the Association for Computational Linguistics}, 2019.

\bibitem{jolliffePrincipalComponentAnalysis2002}
Ian~T. Jolliffe.
\newblock {\em Principal Component Analysis}.
\newblock Springer, 2 edition, 2002.

\bibitem{jrPerformanceFactorsAnalysis}
Philip I~PAVLIK Jr, Hao Cen, and Kenneth~R Koedinger.
\newblock Performance {Factors} {Analysis} – {A} {New} {Alternative} to {Knowledge} {Tracing}.

\bibitem{clst_2024}
Heeseok Jung, Jaesang Yoo, Yohaan Yoon, and Yeonju Jang.
\newblock {CLST}: Cold-start mitigation in knowledge tracing by aligning a generative language model as a students' knowledge tracer, 2024.
\newblock arXiv:2406.10296 [cs].

\bibitem{khajah_deep_2016}
Mohammad Khajah, Robert~V. Lindsey, and Michael~C. Mozer.
\newblock How {Deep} is {Knowledge} {Tracing}?
\newblock In {\em Proceedings of the 9th International Conference on Educational Data Mining}, pages 94--101, 2016.

\bibitem{kojima_zero-shot_2022}
Takeshi Kojima, Shixiang~Shane Gu, Machel Reid, Yutaka Matsuo, and Yusuke Iwasawa.
\newblock Large language models are zero-shot reasoners.
\newblock In {\em Advances in Neural Information Processing Systems}, 2022.

\bibitem{koo_benchmarking_2024}
Ryan Koo, Minhwa Lee, Vipul Raheja, Jong~Inn Park, Zae~Myung Kim, and Dongyeop Kang.
\newblock Benchmarking {Cognitive} {Biases} in {Large} {Language} {Models} as {Evaluators}, September 2024.
\newblock arXiv:2309.17012 [cs].

\bibitem{kulikEffectivenessIntelligentTutoring2016}
James~A. Kulik and J.~D. Fletcher.
\newblock Effectiveness of {Intelligent} {Tutoring} {Systems}: {A} {Meta}-{Analytic} {Review}.
\newblock {\em Review of Educational Research}, 86(1):42--78, March 2016.
\newblock Publisher: American Educational Research Association.

\bibitem{lambiotteLaplacianDynamicsMultiscale2014}
R.~Lambiotte, J.-C. Delvenne, and M.~Barahona.
\newblock Laplacian {Dynamics} and {Multiscale} {Modular} {Structure} in {Networks}.
\newblock {\em IEEE Transactions on Network Science and Engineering}, 1(2):76--90, July 2014.
\newblock arXiv:0812.1770 [physics].

\bibitem{lkt_2024}
Unggi Lee, Jiyeong Bae, Dohee Kim, Sookbun Lee, Jaekwon Park, Taekyung Ahn, Gunho Lee, Damji Stratton, and Hyeoncheol Kim.
\newblock Language model can do knowledge tracing: Simple but effective method to integrate language model and knowledge tracing task, June 2024.
\newblock arXiv:2406.02893 [cs.CL].

\bibitem{lei_rationalizing_2016}
Tao Lei, Regina Barzilay, and Tommi Jaakkola.
\newblock Rationalizing neural predictions.
\newblock In {\em Proceedings of the 2016 Conference on Empirical Methods in Natural Language Processing}, 2016.

\bibitem{li_camel_2023}
Guohao Li, Hasan Abed Al~Kader Hammoud, Hani Itani, Dmitrii Khizbullin, and Bernard Ghanem.
\newblock {CAMEL}: {Communicative} {Agents} for "{Mind}" {Exploration} of {Large} {Language} {Model} {Society}, November 2023.
\newblock arXiv:2303.17760 [cs].

\bibitem{explainable_fewshot_kt_2024}
Haoxuan Li, Jifan Yu, Yuanxin Ouyang, Zhuang Liu, Wenge Rong, Juanzi Li, and Zhang Xiong.
\newblock Explainable few-shot knowledge tracing, 2024.
\newblock arXiv:2405.14391 [cs].

\bibitem{li_novel_2025}
Ziqing Li, Mutlu Cukurova, and Sahan Bulathwela.
\newblock A {Novel} {Approach} to {Scalable} and {Automatic} {Topic}-{Controlled} {Question} {Generation} in {Education}, January 2025.
\newblock arXiv:2501.05220 [cs].

\bibitem{liang_encouraging_2024}
Tian Liang, Zhiwei He, Wenxiang Jiao, Xing Wang, Yan Wang, Rui Wang, Yujiu Yang, Shuming Shi, and Zhaopeng Tu.
\newblock Encouraging {Divergent} {Thinking} in {Large} {Language} {Models} through {Multi}-{Agent} {Debate}, October 2024.
\newblock arXiv:2305.19118 [cs].

\bibitem{lightman_verify_2023}
Hunter Lightman, Vineet Kosaraju, Yura Burda, Harri Edwards, Bowen Baker, Teddy Lee, Jan Leike, John Schulman, Ilya Sutskever, and Karl Cobbe.
\newblock Let's verify step by step, 2023.
\newblock arXiv:2305.20050 [cs].

\bibitem{liu_ekt_2019}
Qi~Liu, Zhenya Huang, Yu~Yin, Enhong Chen, Hui Xiong, Yu~Su, and Guoping Hu.
\newblock {EKT}: {Exercise}-aware {Knowledge} {Tracing} for {Student} {Performance} {Prediction}, June 2019.
\newblock arXiv:1906.05658 [cs].

\bibitem{liu_g-eval_2023}
Yang Liu, Dan Iter, Yichong Xu, Shuohang Wang, Ruochen Xu, and Chenguang Zhu.
\newblock G-{Eval}: {NLG} {Evaluation} using {GPT}-4 with {Better} {Human} {Alignment}, May 2023.
\newblock arXiv:2303.16634 [cs].

\bibitem{liu_improving_2020}
Yunfei Liu, Yang Yang, Xianyu Chen, Jian Shen, Haifeng Zhang, and Yong Yu.
\newblock Improving {Knowledge} {Tracing} via {Pre}-training {Question} {Embeddings}, December 2020.
\newblock arXiv:2012.05031 [cs].

\bibitem{liu_simplekt_2023}
Zitao Liu, Qiongqiong Liu, Jiahao Chen, Shuyan Huang, and Weiqi Luo.
\newblock {simpleKT}: {A} {Simple} {But} {Tough}-to-{Beat} {Baseline} for {Knowledge} {Tracing}, February 2023.
\newblock arXiv:2302.06881 [cs].

\bibitem{liuPyKTPythonLibrary2023}
Zitao Liu, Qiongqiong Liu, Jiahao Chen, Shuyan Huang, Jiliang Tang, and Weiqi Luo.
\newblock {pyKT}: {A} {Python} {Library} to {Benchmark} {Deep} {Learning} based {Knowledge} {Tracing} {Models}, January 2023.
\newblock arXiv:2206.11460 [cs].

\bibitem{lloydLeastSquaresQuantization1982}
Stuart~P. Lloyd.
\newblock Least squares quantization in pcm.
\newblock {\em IEEE Transactions on Information Theory}, 28(2):129--137, 1982.

\bibitem{macqueenSomeMethodsClassification1967}
J.~MacQueen.
\newblock Some methods for classification and analysis of multivariate observations.
\newblock In {\em Proceedings of the Fifth Berkeley Symposium on Mathematical Statistics and Probability}, volume~1, pages 281--297, 1967.

\bibitem{mikolov_distributed_2013}
Tomas Mikolov, Ilya Sutskever, Kai Chen, Greg~S. Corrado, and Jeffrey Dean.
\newblock Distributed {Representations} of {Words} and {Phrases} and their {Compositionality}.
\newblock In {\em Advances in Neural Information Processing Systems 26}, pages 3111--3119, 2013.

\bibitem{minsky_society_1988}
Marvin~Lee Minsky.
\newblock {\em The society of mind}.
\newblock Simon and Schuster, New-York, 1988.

\bibitem{molenaar_personspecific_2009}
Peter C.~M. Molenaar and Candace~G. Campbell.
\newblock The new person-specific paradigm in psychology.
\newblock {\em Current Directions in Psychological Science}, 18(2):112--117, 2009.

\bibitem{nakagawaGraphbasedKnowledgeTracing2019}
Hiromi Nakagawa, Yusuke Iwasawa, and Yutaka Matsuo.
\newblock Graph-based {Knowledge} {Tracing}: {Modeling} {Student} {Proficiency} {Using} {Graph} {Neural} {Network}.
\newblock In {\em {IEEE}/{WIC}/{ACM} {International} {Conference} on {Web} {Intelligence}}, pages 156--163, Thessaloniki Greece, October 2019. ACM.

\bibitem{ccssDevelopmentProcess2010}
{National Governors Association Center for Best Practices} and {Council of Chief State School Officers}.
\newblock Common core state standards initiative: Development process, 2010.
\newblock Official CCSS Initiative page.

\bibitem{ouyang_instructgpt_2022}
Long Ouyang, Jeffrey Wu, Xu~Jiang, Diogo Almeida, Carroll Wainwright, Pamela Mishkin, Chong Zhang, Sandhini Agarwal, Katarina Slama, Alex Ray, et~al.
\newblock Training language models to follow instructions with human feedback.
\newblock In {\em Advances in Neural Information Processing Systems}, 2022.

\bibitem{pandey_self-attentive_2019}
Shalini Pandey and George Karypis.
\newblock A {Self}-{Attentive} model for {Knowledge} {Tracing}, July 2019.
\newblock arXiv:1907.06837 [cs, stat].

\bibitem{pardos_imputing_2017}
Zachary~A. Pardos and Anant Dadu.
\newblock Imputing {KCs} with {Representations} of {Problem} {Content} and {Context}.
\newblock In {\em Proceedings of the 25th {Conference} on {User} {Modeling}, {Adaptation} and {Personalization}}, pages 148--155, Bratislava Slovakia, July 2017. ACM.

\bibitem{pelanek_metrics_2015}
Radek Pel{\'a}nek.
\newblock Metrics for evaluation of student models.
\newblock {\em Journal of Educational Data Mining}, 7(2):1--19, 2015.

\bibitem{pelanek_details_2017}
Radek Pel{\'a}nek.
\newblock Measuring predictive performance of user models: The details matter.
\newblock In {\em Proceedings of the 25th Conference on User Modeling, Adaptation and Personalization}, 2017.

\bibitem{pennington_glove_2014}
Jeffrey Pennington, Richard Socher, and Christopher~D. Manning.
\newblock {GloVe}: {Global} {Vectors} for {Word} {Representation}.
\newblock In {\em Proceedings of the 2014 Conference on Empirical Methods in Natural Language Processing ({EMNLP})}, pages 1532--1543, 2014.

\bibitem{piech_deep_2015}
Chris Piech, Jonathan Spencer, Jonathan Huang, Surya Ganguli, Mehran Sahami, Leonidas Guibas, and Jascha Sohl-Dickstein.
\newblock Deep {Knowledge} {Tracing}, June 2015.
\newblock arXiv:1506.05908 [cs].

\bibitem{porterCommonCoreStandards2011}
Andrew Porter, Jennifer McMaken, JeongSuk Hwang, and Rui Yang.
\newblock Common core standards: The new u.s. intended curriculum.
\newblock {\em Educational Researcher}, 40(3):103--116, 2011.

\bibitem{priharExploringCommonTrends2022}
Ethan Prihar, {Manaal Syed}, {Korinn Ostrow}, Stacy Shaw, Adam Sales, and Neil Heffernan.
\newblock Exploring {Common} {Trends} in {Online} {Educational} {Experiments}.
\newblock July 2022.
\newblock Publisher: Zenodo.

\bibitem{ranzato_mixer_2016}
Marc'Aurelio Ranzato, Sumit Chopra, Michael Auli, and Wojciech Zaremba.
\newblock Sequence level training with recurrent neural networks.
\newblock In {\em International Conference on Learning Representations}, 2016.

\bibitem{reimers_sentence-bert_2019}
Nils Reimers and Iryna Gurevych.
\newblock Sentence-{BERT}: {Sentence} {Embeddings} using {Siamese} {BERT}-{Networks}.
\newblock In {\em Proceedings of the 2019 Conference on Empirical Methods in Natural Language Processing and the 9th International Joint Conference on Natural Language Processing}, pages 3982--3992, 2019.

\bibitem{rollinson_policies_2015}
Joseph Rollinson and Emma Brunskill.
\newblock From predictive models to instructional policies.
\newblock In {\em Proceedings of the 8th International Conference on Educational Data Mining}, 2015.

\bibitem{rosenberg_v-measure_2007}
Andrew Rosenberg and Julia Hirschberg.
\newblock V-measure: A conditional entropy-based external cluster evaluation measure.
\newblock In {\em Proceedings of the 2007 Joint Conference on Empirical Methods in Natural Language Processing and Computational Natural Language Learning}, pages 410--420, 2007.

\bibitem{sanMauroMolecularSystematics2010}
Diego San~Mauro and Ana Agorreta.
\newblock Molecular systematics: A synthesis of the common methods and the state of knowledge.
\newblock {\em Cellular and Molecular Biology Letters}, 15(2):311--341, 2010.

\bibitem{scaria_automated_2024}
Nicy Scaria, Suma~Dharani Chenna, and Deepak Subramani.
\newblock Automated {Educational} {Question} {Generation} at {Different} {Bloom}'s {Skill} {Levels} using {Large} {Language} {Models}: {Strategies} and {Evaluation}.
\newblock volume 14830, pages 165--179. 2024.
\newblock arXiv:2408.04394 [cs].

\bibitem{schulman_ppo_2017}
John Schulman, Filip Wolski, Prafulla Dhariwal, Alec Radford, and Oleg Klimov.
\newblock Proximal policy optimization algorithms, 2017.
\newblock arXiv:1707.06347 [cs].

\bibitem{shao_deepseekmath_2024}
Zhihong Shao, Peiyi Wang, Qihao Zhu, Runxin Xu, Junxiao Li, Bernardino Romera-Paredes, Chenyang Shi, et~al.
\newblock {DeepSeekMath}: Pushing the limits of mathematical reasoning in open language models, 2024.
\newblock arXiv:2402.03300 [cs].

\bibitem{shen_survey_2024}
Shuanghong Shen, Qi~Liu, Zhenya Huang, Yonghe Zheng, Minghao Yin, Minjuan Wang, and Enhong Chen.
\newblock A {Survey} of {Knowledge} {Tracing}: {Models}, {Variants}, and {Applications}.
\newblock {\em IEEE Transactions on Learning Technologies}, pages 1--22, 2024.
\newblock arXiv:2105.15106 [cs].

\bibitem{snell_scaling_2024}
Charlie Snell, Jaehoon Lee, Kelvin Xu, and Aviral Kumar.
\newblock Scaling {LLM} {Test}-{Time} {Compute} {Optimally} can be {More} {Effective} than {Scaling} {Model} {Parameters}, August 2024.
\newblock arXiv:2408.03314 [cs].

\bibitem{song_nondeterminism_2025}
Yifan Song, Guoyin Wang, Sujian Li, and Bill~Yuchen Lin.
\newblock The good, the bad, and the greedy: Evaluation of {LLM}s should not ignore non-determinism.
\newblock In {\em Proceedings of the 2025 Annual Conference of the Nations of the Americas Chapter of the Association for Computational Linguistics}, 2025.

\bibitem{sonkar_qdkt_2020}
Shashank Sonkar, Andrew~E. Waters, Andrew~S. Lan, Phillip~J. Grimaldi, and Richard~G. Baraniuk.
\newblock {qDKT}: {Question}-centric {Deep} {Knowledge} {Tracing}, May 2020.
\newblock arXiv:2005.12442 [cs, stat].

\bibitem{su_exercise-enhanced_2018}
Yu~Su, Qingwen Liu, Qi~Liu, Zhenya Huang, Yu~Yin, Enhong Chen, Chris Ding, Si~Wei, and Guoping Hu.
\newblock Exercise-{Enhanced} {Sequential} {Modeling} for {Student} {Performance} {Prediction}.
\newblock {\em Proceedings of the AAAI Conference on Artificial Intelligence}, 32(1), April 2018.

\bibitem{sutton_barto_2018}
Richard~S. Sutton and Andrew~G. Barto.
\newblock {\em Reinforcement Learning: An Introduction}.
\newblock MIT Press, 2 edition, 2018.

\bibitem{tong_hgkt_2022}
Hanshuang Tong, Zhen Wang, Yun Zhou, Shiwei Tong, Wenyuan Han, and Qi~Liu.
\newblock {HGKT}: {Introducing} {Hierarchical} {Exercise} {Graph} for {Knowledge} {Tracing}, August 2022.
\newblock arXiv:2006.16915 [cs].

\bibitem{tongStructureBasedKnowledgeTracing2020}
Shiwei Tong, Qi~Liu, Wei Huang, Zhenya Hunag, Enhong Chen, Chuanren Liu, Haiping Ma, and Shijin Wang.
\newblock Structure-{Based} {Knowledge} {Tracing}: {An} {Influence} {Propagation} {View}.
\newblock In {\em 2020 {IEEE} {International} {Conference} on {Data} {Mining} ({ICDM})}, pages 541--550, November 2020.
\newblock ISSN: 2374-8486.

\bibitem{turpin_unfaithful_2023}
Miles Turpin, Julian Michael, Ethan Perez, and Samuel~R. Bowman.
\newblock Language models don't always say what they think: Unfaithful explanations in chain-of-thought prompting.
\newblock In {\em Advances in Neural Information Processing Systems Workshop Track}, 2023.

\bibitem{vaswani_attention_2017}
Ashish Vaswani, Noam Shazeer, Niki Parmar, Jakob Uszkoreit, Llion Jones, Aidan~N. Gomez, Lukasz Kaiser, and Illia Polosukhin.
\newblock Attention {Is} {All} {You} {Need}, August 2017.
\newblock arXiv:1706.03762 [cs].

\bibitem{vie_knowledge_2018}
Jill-Jênn Vie and Hisashi Kashima.
\newblock Knowledge {Tracing} {Machines}: {Factorization} {Machines} for {Knowledge} {Tracing}, November 2018.
\newblock arXiv:1811.03388 [cs, stat].

\bibitem{vinh_information_2010}
Nguyen~Xuan Vinh, Julien Epps, and James Bailey.
\newblock Information theoretic measures for clusterings comparison: Variants, properties, normalization and correction for chance.
\newblock {\em Journal of Machine Learning Research}, 11:2837--2854, 2010.

\bibitem{wang_seep_2022}
Wentao Wang, Qi~Liu, Zhenya Huang, Enhong Chen, Yu~Yin, and Yu~Su.
\newblock Semantic-enhanced question embeddings pre-training for improving knowledge tracing.
\newblock {\em Information Sciences}, 608:1022--1039, 2022.

\bibitem{wang_self-consistency_2023}
Xuezhi Wang, Jason Wei, Dale Schuurmans, Quoc Le, Ed~H. Chi, Sharan Narang, Aakanksha Chowdhery, and Denny Zhou.
\newblock Self-consistency improves chain of thought reasoning in language models.
\newblock In {\em International Conference on Learning Representations}, 2023.

\bibitem{wang_dhp_2025}
Yicheng Wang, Jiayi Yuan, Yu-Neng Chuang, Zhuoer Wang, Yingchi Liu, Mark Cusick, Param Kulkarni, Zhengping Ji, Yasser Ibrahim, and Xia Hu.
\newblock {DHP} {Benchmark}: {Are} {LLMs} {Good} {NLG} {Evaluators}?, February 2025.
\newblock arXiv:2408.13704 [cs].

\bibitem{wang_math_2021}
Zichao Wang, Andrew~S. Lan, and Richard~G. Baraniuk.
\newblock Math {Word} {Problem} {Generation} with {Mathematical} {Consistency} and {Problem} {Context} {Constraints}, September 2021.
\newblock arXiv:2109.04546 [cs].

\bibitem{wei_cot_2022}
Jason Wei, Xuezhi Wang, Dale Schuurmans, Maarten Bosma, Fei Xia, Ed~H. Chi, Quoc~V. Le, and Denny Zhou.
\newblock Chain-of-thought prompting elicits reasoning in large language models.
\newblock In {\em Advances in Neural Information Processing Systems}, 2022.

\bibitem{wu_autogen_2023}
Qingyun Wu, Gagan Bansal, Jieyu Zhang, Yiran Wu, Beibin Li, Erkang Zhu, Li~Jiang, Xiaoyun Zhang, Shaokun Zhang, Jiale Liu, Ahmed~Hassan Awadallah, Ryen~W. White, Doug Burger, and Chi Wang.
\newblock {AutoGen}: {Enabling} {Next}-{Gen} {LLM} {Applications} via {Multi}-{Agent} {Conversation}, October 2023.
\newblock arXiv:2308.08155 [cs].

\bibitem{xiong_going_2016}
Xiaolu Xiong, Shuangyun Zhao, Emma~G. Van~Inwegen, and Joseph~E. Beck.
\newblock Going {Deeper} with {Deep} {Knowledge} {Tracing}.
\newblock In {\em Proceedings of the 9th International Conference on Educational Data Mining}, pages 545--550, 2016.

\bibitem{yeung_addressing_2018}
Chun-Kit Yeung and Dit-Yan Yeung.
\newblock Addressing {Two} {Problems} in {Deep} {Knowledge} {Tracing} via {Prediction}-{Consistent} {Regularization}, June 2018.
\newblock arXiv:1806.02180 [cs].

\bibitem{yin_quesnet_2019}
Yu~Yin, Qi~Liu, Zhenya Huang, Enhong Chen, Wei Tong, Shijin Wang, and Yu~Su.
\newblock {QuesNet}: {A} {Unified} {Representation} for {Heterogeneous} {Test} {Questions}.
\newblock In {\em Proceedings of the 25th {ACM} {SIGKDD} {International} {Conference} on {Knowledge} {Discovery} \& {Data} {Mining}}, pages 1328--1336, July 2019.
\newblock arXiv:1905.10949 [cs, stat].

\bibitem{hutchison_individualized_2013}
Michael~V. Yudelson, Kenneth~R. Koedinger, and Geoffrey~J. Gordon.
\newblock Individualized {Bayesian} {Knowledge} {Tracing} {Models}.
\newblock In David Hutchison, Takeo Kanade, Josef Kittler, Jon~M. Kleinberg, Friedemann Mattern, John~C. Mitchell, Moni Naor, Oscar Nierstrasz, C.~Pandu~Rangan, Bernhard Steffen, Madhu Sudan, Demetri Terzopoulos, Doug Tygar, Moshe~Y. Vardi, Gerhard Weikum, H.~Chad Lane, Kalina Yacef, Jack Mostow, and Philip Pavlik, editors, {\em Artificial {Intelligence} in {Education}}, volume 7926, pages 171--180. Springer Berlin Heidelberg, Berlin, Heidelberg, 2013.
\newblock Series Title: Lecture Notes in Computer Science.

\bibitem{zelikman_star_2022}
Eric Zelikman, Yuhuai Wu, Jesse Mu, and Noah~D. Goodman.
\newblock {STaR}: Bootstrapping reasoning with reasoning.
\newblock In {\em Advances in Neural Information Processing Systems}, 2022.

\bibitem{zeng_autodefense_2024}
Yifan Zeng, Yiran Wu, Xiao Zhang, Huazheng Wang, and Qingyun Wu.
\newblock {AutoDefense}: {Multi}-{Agent} {LLM} {Defense} against {Jailbreak} {Attacks}, November 2024.
\newblock arXiv:2403.04783 [cs].

\bibitem{zhang_dynamic_2017}
Jiani Zhang, Xingjian Shi, Irwin King, and Dit-Yan Yeung.
\newblock Dynamic {Key}-{Value} {Memory} {Networks} for {Knowledge} {Tracing}, February 2017.
\newblock arXiv:1611.08108 [cs].

\bibitem{zhang_review_2022}
Ruqing Zhang, Jiafeng Guo, Lu~Chen, Yixing Fan, and Xueqi Cheng.
\newblock A {Review} on {Question} {Generation} from {Natural} {Language} {Text}.
\newblock {\em ACM Transactions on Information Systems}, 40(1):1--43, January 2022.

\bibitem{zhaoResearchDeepKnowledge2022}
Zhongyuan Zhao, Zhaohui Liu, Bei Wang, Lijun Ouyang, Can Wang, and Yan Ouyang.
\newblock Research on {Deep} {Knowledge} {Tracing} {Model} {Integrating} {Graph} {Attention} {Network}.
\newblock In {\em 2022 {Prognostics} and {Health} {Management} {Conference} ({PHM}-2022 {London})}, pages 389--394, May 2022.
\newblock ISSN: 2166-5656.

\bibitem{zhou_least-to-most_2023}
Denny Zhou, Nathanael Sch{\"a}rli, Le~Hou, Jason Wei, Nathan Scales, Xuezhi Wang, Dale Schuurmans, Ed~H. Chi, and Quoc~V. Le.
\newblock Least-to-most prompting enables complex reasoning in large language models.
\newblock In {\em International Conference on Learning Representations}, 2023.

\end{thebibliography}

\end{document}